\documentclass[final]{uqthesis}

\usepackage{cite}				 %Allows abbreviated numerical citations.
\usepackage{pdfpages}			 %Allows you to include full-page pdfs.
\usepackage{wrapfig}			 %Lets you wrap text around figures.
\usepackage{bm} 				 %Bolded maths characters.
\usepackage{upgreek}			 %Upright Greek characters.
\usepackage{dsfont}				 %Double-struck fonts.
\usepackage{simplewick}			 %For typesetting Wick contractions.
\usepackage{mathtools}		     %Can be used to fine-tune the maths presentation.	
\usepackage{framed}			     %For boxed text.
\usepackage{microtype}			 %pdfLaTeX will fix your kerning.
\usepackage{marvosym}			 %Include symbols (like the Euro symbol, etc.).
\usepackage{color}				 %Nice for scalable pdf graphics using InkScape.
\usepackage{transparent}	     %Nice for scalable pdf graphics using InkScape.
\usepackage{placeins}			 %Lets you put in a \FloatBarrier to stop figures floating past this command.
\usepackage{mdframed,mdwlist}    %Use these for nice lists (less white space).
\usepackage{graphicx}            %Enhanced support for graphics.
\usepackage{float}               %Improved interface for floating objects. 
\usepackage{longtable}           %Allow tables to flow over page boundaries.
\usepackage{mathdots}            %Changed the basic LaTeX and plain TeX commands.
\usepackage{eucal}               %Font shape definitions to use the Euler script symbols in math mode.
\usepackage{array}               %Extending the array and tabular environments.
\usepackage{stmaryrd}            %The StMary’s Road symbol font.
\usepackage{amsthm}              %St Mary Road symbols for theoretical computer science. 
\usepackage{pifont}              %Access to PostScript standard Symbol and Dingbats fonts.
\usepackage{lipsum}              %Easy access to the Lorem Ipsum dummy text.
\usepackage{enumerate}           %Enumerate with redefinable labels. 
\usepackage[all]{xy}             %This is a special package for drawing diagrams.
\usepackage{amsmath}             %ATypesetting theorems (AMS style).
\usepackage{amssymb}             %Provided an extended symbol collection.
\usepackage[utf8]{inputenc}      %Allowed all displayable utf8 characters to be available as input.
\usepackage{blindtext}           %Produced 'blind' text for testing.
\usepackage{tikz}                %To create graphic elements.
\usepackage[figuresright]{rotating}	%Allows large tables to be rotated to landscape.
\usetikzlibrary{shapes.geometric, arrows}
\usepackage{algorithm}
\usepackage[noend]{algorithmic}
\usepackage{amsmath}
\usepackage{amsfonts}     
\newcommand{\SUB}[1]{\ENSURE \hspace{-0.15in} \textbf{#1}}

\usepackage{tabularx}
\newcommand{\dubbelop}{$^{\blacktriangle}$}
\newcommand{\dubbelneer}{$^{\blacktriangledown}$}
\newcommand{\enkelneer}{$^{\triangledown}$}
\newcommand{\enkelop}{$^{\vartriangle}$}
\usepackage{booktabs}

\usepackage{graphicx}
\usepackage{multirow}
\usepackage{subcaption}

\usepackage{url}

\usepackage{color}

\usepackage{soul}

\usepackage{tikz}
\usepackage{xcolor}
\newcommand*\circled[1]{\tikz[baseline=(char.base)]{
		\node[shape=circle,fill,inner sep=0.6pt] (char) {\bfseries\small\textcolor{white}{#1}};}}

\usepackage{amsthm}
\newtheorem{definition}{Definition}[section]  % This numbers definitions within sections

\usepackage{enumitem}

\title{Effective and Secure Federated Online Learning to Rank}

\author{Shuyi Wang}
\currentdegrees{M.S.}

\orcid{\href{add your ORCID url here}{0000-0002-4467-5574}}

\date{2024}
\submittedfor{Doctor}

\school{School of Electrical Engineering and Computer Science}

\begin{document}

\frontmatter
% Assemble title page
\maketitle
\clearpage

% ***************************************************
% Preface
%****************************************************
\section{Abstract}
\normalfont
%Open abstract.tex to edit
% ***************************************************
% Abstract
% ***************************************************
% TO PRODUCE A STAND-ALONE PDF OF YOUR ABSTRACT, uncomment this section and the \end{document} at the end of the file by removing the % from the start of each line.

%\documentclass[12pt, a4paper]{memoir}

%\input{LaTexPackages.tex}

%\begin{document}

%\begin{center}
	%\textbf{\large Your title goes here}

	%\textbf{Abstract}

	%Your Name, The University of Queensland, 20??
%\end{center}

% ********* Enter your text below this line: ********

%\noindent
%The abstract should outline the main approach and findings of the thesis and normally must be between 300 and 800 words.

Online Learning to Rank (OLTR) optimises ranking models using implicit user feedback, such as clicks. Unlike traditional Learning to Rank (LTR) methods that rely on a static set of training data with relevance judgements to learn a ranking model, OLTR methods update the model continually as new data arrives. Thus, it addresses several drawbacks such as the high cost of human annotations, potential misalignment between user preferences and human judgments, and the rapid changes in user query intents. However, OLTR methods typically require the collection of searchable data, user queries, and clicks, which poses privacy concerns for users.

Federated Online Learning to Rank (FOLTR) integrates OLTR within a Federated Learning (FL) framework to enhance privacy by not sharing raw data. While promising, FOLTR methods currently lag behind traditional centralised OLTR due to challenges in ranking effectiveness, robustness with respect to data distribution across clients, susceptibility to attacks, and the ability to unlearn client interactions and data. This thesis comprehensively investigates these challenges and proposes effective and secure FOLTR methods. 

First, we address the effectiveness of FOLTR by identifying the limitations of existing work and proposing a new method, termed Federated Pairwise Differentiable Gradient Descent (FPDGD). By adapting Pairwise Differentiable Gradient Descent (PDGD) to the Federated Averaging framework, we significantly improve upon the previous method, as demonstrated by empirical evaluations.

Next, we examine the robustness of FOLTR under non independent and identically distributed (non-IID) data across clients. We identify four data distribution scenarios that can lead to non-IID issues and evaluate their impact on ranking performance. Additionally, we assess common federated learning approaches to mitigate these non-IID challenges in FOLTR.

We also investigate data and model poisoning attack strategies and their impact on FOLTR effectiveness. We explore robust aggregation rules for federated learning to counter these attacks, providing insights into the effectiveness of attack and defense methods in FOLTR systems and identifying key factors influencing their success.

Finally, we propose an efficient unlearning method to remove a client's contributions from the FOLTR system using historical local updates. To verify the unlearning process, we introduce a novel evaluation approach based on poisoning attacks.

In summary, this thesis presents a comprehensive study on Federated Online Learning to Rank, addressing its effectiveness, robustness, security, and unlearning capabilities, thereby expanding the landscape of FOLTR.

% ***************************************************

%\end{document}

\clearpage
% ***************************************************
\section*{Declaration by author}
%DO NOT EDIT.
% ***************************************************
% Declaration by Author
% ***************************************************
% This is the DECLARATION BY AUTHOR
% All candidates to reproduce this section in their thesis verbatim
% DO NOT EDIT!
%
\begin{instructional}
\textit{(All candidates to reproduce this section in their thesis verbatim)\\}
\end{instructional}

\noindent
This thesis is composed of my original work, and contains no material previously published or written by another person except where due reference has been made in the text. I have clearly stated the contribution by others to jointly-authored works that I have included in my thesis.\\

\noindent
I have clearly stated the contribution of others to my thesis as a whole, including statistical assistance, survey design, data analysis, significant technical procedures, professional editorial advice, financial support and any other original research work used or reported in my thesis. The content of my thesis is the result of work I have carried out since the commencement of my higher degree by research candidature and does not include a substantial part of work that has been submitted to qualify for the award of any other degree or diploma in any university or other tertiary institution. I have clearly stated which parts of my thesis, if any, have been submitted to qualify for another award.\\

\noindent
I acknowledge that an electronic copy of my thesis must be lodged with the University Library and, subject to the policy and procedures of The University of Queensland, the thesis be made available for research and study in accordance with the Copyright Act 1968 unless a period of embargo has been approved by the Dean of the Graduate School. \\

\noindent
I acknowledge that copyright of all material contained in my thesis resides with the copyright holder(s) of that material. Where appropriate I have obtained copyright permission from the copyright holder to reproduce material in this thesis and have sought permission from co-authors for any jointly authored works included in the thesis.

\clearpage
%YOU MUST EDIT THIS DOCUMENT.
% ***************************************************
% PRELIMINARY PAGES
% ***************************************************
% The instructions contained within this part of the thesis template need to be suppressed from the final thesis. There are instructions on how to do this in the MainThesis.tex file.

% To ensure your work is not suppressed with the instructions please add your text only where instructed.

%***Publications included in this thesis***
\section*{Publications included in this thesis}

% ********* Enter your text below this line: ********
\begin{enumerate}
	
	\item \cite{wang2021federated} \textbf{Shuyi Wang}, Shengyao Zhuang, and Guido Zuccon, \href{https://doi.org/10.1007/978-3-030-72240-1_10}{Federated Online Learning to Rank with Evolution Strategies: A Reproducibility Study}, \textit{In European Conference on Information Retrieval (ECIR)}, pages 134-149, 2021
	
	\item \cite{wang2021effective} \textbf{Shuyi Wang}, Bing Liu, Shengyao Zhuang, and Guido Zuccon, \href{https://doi.org/10.1145/3471158.3472236}{Effective and Privacy-preserving Federated Online Learning to Rank}, \textit{In Proceedings of the 2021 ACM SIGIR International Conference on Theory of Information Retrieval (ICTIR)}, pages 3–12, 2021
	
	\item \cite{wang2022non} \textbf{Shuyi Wang}, and 
	Guido Zuccon, \href{https://doi.org/10.1145/3477495.3531709}{Is Non-IID Data a Threat in Federated Online Learning to Rank?}, \textit{In Proceedings of the 45th International ACM SIGIR Conference on Research and Development in Information Retrieval (SIGIR)}, pages 2801–2813, 2022
		
	\item \cite{wang2023analysis} \textbf{Shuyi Wang}, and Guido Zuccon, \href{https://doi.org/10.1145/3578337.3605117}{An Analysis of Untargeted Poisoning Attack and Defense Methods for Federated Online Learning to Rank Systems}, \textit{In Proceedings of the 2023 ACM SIGIR International Conference on Theory of Information Retrieval (ICTIR)}, pages 215–224, 2023
	
	\item \cite{DBLP:conf/ecir/WangLZ24} \textbf{Shuyi Wang}, Bing Liu, and Guido Zuccon, \href{https://doi.org/10.1007/978-3-031-56063-7\_7}{How to Forget Clients in Federated Online Learning to Rank?}, \textit{In European Conference on Information Retrieval (ECIR)}, pages 105--121, 2024
	
\end{enumerate}
% ***************************************************

%***Submitted manuscripts included in this thesis***
\section*{Submitted manuscripts included in this thesis}

% ********* Enter your text below this line: ********

No manuscripts submitted for publication.
% ***************************************************

%***Other publications during candidature***
\section*{Other publications during candidature}

% ********* Enter your text below this line: ********

No other publications.
% ***************************************************

%***Contributions by others to the thesis***
\section*{Contributions by others to the thesis}

% ********* Enter your text below this line: ********

The works included in this thesis were done under the supervision of Prof. Guido Zuccon and A/Prof. Bevan Koopman.

% ***************************************************

%***Statement of parts of the thesis submitted to qualify for the award of another degree***
\section*{Statement of parts of the thesis submitted to qualify for the award of another degree}

% ********* Enter your text below this line: ********

No works submitted towards another degree have been included in this thesis.

% ***************************************************

%***Research involving human or animal subjects***
\section*{Research involving human or animal subjects}

% ********* Enter your text below this line: ********

No animal or human subjects were involved in this research.

% ***************************************************

%***Acknowledgements***
\clearpage
\section*{Acknowledgments}

% ********* Enter your text below this line: ********
I would like to take this opportunity to express my gratitude to everyone who has contributed to my research and personal growth during my PhD journey.

First and foremost, I would like to express my deepest gratitude to my advisor, Prof. Guido Zuccon. His unwavering support, insightful guidance, and continuous encouragement have been invaluable throughout my PhD. I am deeply impressed by his profound knowledge, dedication to research, and enthusiasm for work. I would like to thank him for providing detailed feedback on my research, encouraging me when I encountered bottlenecks or faced paper rejections, offering generous support for my academic travels, and providing invaluable suggestions on career development. My co-advisor, A/Prof. Bevan Koopman, has also been a tremendous support, offering valuable feedback and engaging insightful discussions about my research. I have greatly benefited from his numerous talks on research methods and academic writing during our group meetings. Their expertise and dedication have profoundly shaped my research, and I am deeply grateful for their mentorship.

My special thanks to Prof. Shane Culpepper, Prof. Gabriele Tolomei and Dr. Harrie Oosterhuis for being my PhD committee. Thank you for your valuable feedback and discussion on improving this thesis.

I would like to extend my sincere thanks to my colleagues and friends in the ielab group: Ahmed, Anton, Daniel, Hang, Harry, Ismail, Joel, Jonathan, Katya, Kim, Linh, Shengyao, Shuai, Shuoqi, Sitthi, Suresh, Watheq, Xinyu. Their constructive feedback and discussions have made my research experience both enjoyable and enriching. I have truly enjoyed our lunch talks, coffee breaks, and group activities on the weekends. I am deeply impressed by their intelligence, hard work, and kindness. I want to thank friends that I met at UQ: Kexuan, Can, Shijie, Tianyu, Pengfei. I enjoyed our board games, road trips, and hanging out at weekends and holidays. Thank you for generously offering support to me and sharing your thoughts on academic life and career development. I wish you all a bright future in your continuing research and career. Further thanks to friends from Europe: Sophia, Gineke, Wojciech, Ferdinand and Maik. It was nice to meet your in Brisbane and thank you for the friendly company in the office. I would also like to thank the numerous others who have inspired, encouraged, and helped me along the way. Your kindness and support have been a source of strength and motivation for me.

I am deeply grateful to my parents and family for their unconditional love and support. Thank you for your understanding, encouragement, and for being there through the ups and downs of this journey. Your patience and sacrifices have made this achievement possible. To my parents, your guidance and values have shaped who I am today, and I am forever indebted to you for your endless support.

Lastly, I would like to acknowledge the love and support provided by Bing. Your encouragement and patience have been vital to me, and without you, this journey would not have been possible.

Thank you all for being a part of this journey.
% ***************************************************

%***Financial Support***
\clearpage
\section*{Financial support}

% ********* Enter your text below this line: ********

This research was supported by a scholarship from China Scholarship Council and a Google PhD Fellowship.

% ***************************************************

%***Order of remaining thesis content***
\clearpage

%***Dedication***
%If you wish to add a dedication (if appropriate), do so here.
%If not, comment out from here...
	\rmfamily
	\normalfont
    
    	\begin{vplace}[1]
		\begin{center}
        To my family for their love and support
		\end{center}
	\end{vplace}

%... to here.

\clearpage
\pagestyle{headings}

%***Table of Contents***
%These generate the table of contents, list of figures, and list of tables from items tagged with a \label{} command.
\tableofcontents
	\clearpage
\listoffigures
	\clearpage
\listoftables
\newpage
%*************************************
% List of abbreviations
%*************************************
% You can make a list of abbreviations here.
%
% There are LaTeX packages available to take care of these things, but you will
% need to manually add these to the template at this stage (support may be added
% in future releases).

%CHOOSE AN APPROPRIATE TITLE.
%\chapter[List of abbreviations]{List of abbreviations}
\chapter[List of Abbreviations and Symbols]{List of Abbreviations and Symbols}

%If the auto-sizing of the tables annoys you, consider the tabularx package.

%List of abbreviations.
\begin{center}
	\small
	\begin{longtable}{ll}
	\toprule
	Abbreviations & {} \\
	\bottomrule
	LTR	& Learning to Rank \\
	CLTR & Counterfactual Learning to Rank \\
	OLTR&Online Learning to Rank \\
	ULTR&Unbiased Learning to Rank \\
	PDGD&Pairwise Differentiable Gradient Descent \\
	FL&Federated Learning \\
	FedAvg&Federated Averaging \\
	FOLTR&Federated Online Learning to Rank \\
	FOLtR-ES&Federated OLTR with Evolutionary Strategies method \\
	FPDGD&Federated Pairwise Differentiable Gradient Descent \\
	non-IID&non independent and identically distributed \\
	SERPs&search engine result pages \\
	CCM&Cascade Click Model \\
	nDCG&normalised Discounted Cumulative Gain \\
	PBM&Position-Based Model \\
	PLMs&Pretrained Language Models \\
	\hline
	\end{longtable}
\end{center}

%*************************************
% List of symbols
%*************************************
%List of symbols. REMOVE IF NOT NEEDED.
%\begin{center}
%	\small
%	\begin{longtable}{ll}
%	\toprule
%	Symbols & {} \\
%	\bottomrule
%	$\hat{\rho}$		& Density operator \\
%	\etc{}					& \etc{} \\
%	\hline
%	\end{longtable}
%\end{center}

%***End of list of symbols and abbreviations*** %List of symbols. REMOVE IF NOT NEEDED.

%***End of front matter***

% ***************************************************
% Thesis Content
%****************************************************
\mainmatter

%Each chapter is a separate .tex file. Use \input to load them here, as demonstrated below for Chapter 1 and Chapter 2.
%We recommend keeping each in a separate subfolder, with its accompanying figures, etc. This is how the template is currently structured.
%If you wish to divide your thesis into parts (each containing multiple chapters), us the \part{} command.

%CHAPTER 1
\chapter[Introduction]{Introduction}
\label{Chap:Intro}

% ***************************************************

\section{Background}
%A little background, What is this thesis about, and Why it is important

% Ranking pipeline
%Ranking is a core task in Information Retrieval~\footnote{The broader research area of Information Retrieval encompasses various domain-specific applications and multimedia documents, such as legal search~\cite{locke2022case}, email search~\cite{kim2017understanding}, image search~\cite{datta2008image}, music search~\cite{orio2006music}, etc. In this thesis, we foucs on the general context of information retrieval known as document retrieval~\cite{cao2006adapting}. Examples of document include web pages, email, books, text messages, etc.}. In a typical search engine pipeline, after the user inputs a query, the search system first efficiently gather relevant documents - a process termed as first-stage retrieval. Subsequently, re-ranking of the retrieved documents is conducted to place the most relevant documents at the top. This second-stage re-ranking employs more complex models to refine the initial results. This thesis is pertained with the construction of such second-stage re-rankers.

%cascaded pipeline
%
%OLTR involves continuously updating the ranking model as new data become available. This is particularly useful in dynamic environments where user preferences or document collections change over time.
%
%The online aspect allows the model to adapt to changes in user behavior or document collections in real-time.

% introduce ranking and learning to rank

% Background - Ranking; Learning to rank.
Ranking is a core task in Information Retrieval~\footnote{The broader research area of Information Retrieval encompasses various domain-specific applications and multimedia documents, such as legal search~\cite{locke2022case}, email search~\cite{kim2017understanding}, image search~\cite{datta2008image}, music search~\cite{orio2006music}, etc. In this thesis, we focus on the general context of information retrieval known as document retrieval~\cite{cao2006adapting}. Examples of document include web pages, email, books, text messages, etc.}. It involves ordering documents in a way that places the most relevant results at the top of the list in response to a user need, typically formulated by means of a search query. The effectiveness of ranking algorithms directly impacts the user experience and satisfaction of search engines and various information retrieval applications. Consequently, significant research efforts are dedicated to developing and optimizing ranking models, with a particular focus on Learning to Rank (LTR) methods that leverage machine learning techniques to improve ranking performance. These models score the relevance of each candidate document to a given query and rank them based on these scores. To explore the relevance between queries and documents, the data used for training typically consist of the multi-dimensional feature representations of each query-document pair along with corresponding relevance judgements. Various loss functions, such as \textit{pairwise}, \textit{pointwise}, and \textit{listwise}~\cite{liu2009learning}, can be utilized to learn the ranking models, aiming to achieve better ranking effectiveness.

%Learning to Rank (LTR) methods leverage supervised machine learning techniques to train ranking models. 

% Why not learning to rank? - limitations of LTR
Although learning to rank methods have achieved exceptional ranking performance with the help of supervision, the necessity of human labeling presents a significant bottleneck, as data annotation is extremely costly in the field of information retrieval~\cite{sanderson2010test} and other machine learning tasks. In addition to being expensive and time-consuming to generate, human annotations may not always align with real users' preferences~\cite{sanderson2010test}, and these labels sometimes fail to capture the evolving nature of user search intents~\cite{lefortier2014online,zhuang2021how}. Additionally, human labeling raises ethical issues when applied to private data, such as emails, as users' privacy can not be guaranteed if the annotated documents contain sensitive personal information~\cite{wang2016learning}.

% Two groups of methods have been proposed to address the limitation of LTR
To address the document annotation challenges in learning to rank, two groups of methods have been developed by substituting human judgements with real users' feedback. When users interact with a list of ranked documents, their actions (such as clicks) on the search engine result pages (SERPs) are recorded and used as relevance signals for training ranking models. One method is termed Counterfactual Learning to Rank (CLTR), which utilizes logged user interactions from previous ranking systems as the training set for a new ranking system, following the common offline training setting before deployment. The other method is called Online Learning to Rank (OLTR), which collects real-time user interactions with the ranking result pages and directly learns from them to update the ranking model. This procedure is conducted in an online manner, meaning the ranking system simultaneously handles real users' information needs and learns from user feedback to update itself.

% Previous focus of CLTR and OLTR methods -> neglect the privacy
Unlike previous LTR methods, CLTR and OLTR require real user interactions. Although inexpensive to obtain, user interactions contain a significant amount of noise and biases~\cite{joachims2017accurately}, making the translation of these interactions into relevance signals a not straightforward solution. Noise in user interactions arises from the randomness of user behaviour, i.e., users click for unexpected reasons or misclick some documents. In practice, averaging over many interactions can mitigate the influence of noisy interactions, making them less threatening to the system. However, addressing biases in user interactions is more complex, as they consistently stem from user behaviours under the way of how ranking lists are presented. One of the most studied biases is position bias, where users are more likely to examine and click on higher-ranked documents, not necessarily due to their relevance. Another is selection bias, which indicates that user interactions are limited to the presented results only. In the active research area of Unbiased Learning to Rank\footnote{Unbiased Learning to Rank (ULTR) broadly includes two groups of methods: Counterfactual Learning to Rank and Online Learning to Rank, both of which share the same goal of learning unbiased user preferences from biased user interactions~\cite{oosterhuis2020unbiased,ai2021unbiased}.}, more biases are being explored along with corresponding methods for remedying them. To handle the inherent biases in user interactions, CLTR methods build user models that explicitly account for biases to optimize ranking metrics unbiasedly using historical interactions. Differently, OLTR methods address biases by randomizing results through online interventions, i.e. actively controlling what is displayed to the user, and optimize ranking models using online interactions directly. OLTR involves continuously updating the ranking model as new data become available. The online aspect allows the model to adapt to changes in user preferences or document collections in real-time. This thesis focuses on study of OLTR methods.

\begin{figure}[t]
	\includegraphics[width=0.4\textwidth]{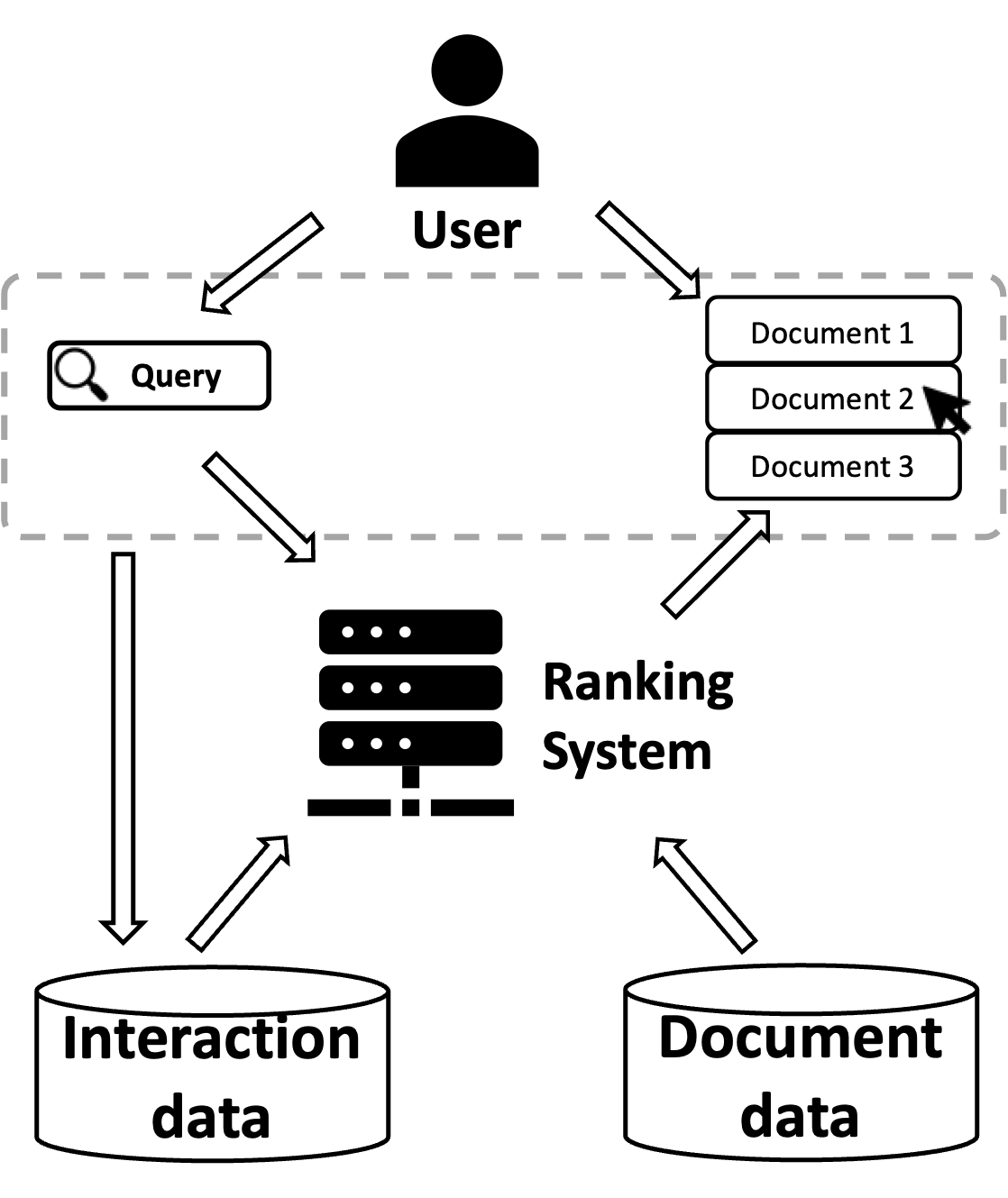}
	\centering
	\caption{Illustration of Online Learning to Rank (OLTR) system.}
	\label{fig:oltr_overview} 
\end{figure}

% Privacy is important but is being neglected in the previous study of CLTR and OLTR
One key issue that has been neglected in previous learning-from-user-interaction methods is that user interactions with ranking result pages always reflect user's preferences, posing a threat to their privacy. Figure~\ref{fig:oltr_overview} demonstrates an overview of an OLTR system where user interactions are gathered in an online manner. The ranking system is deployed on one or several centralised servers responsible for indexing the documents, producing the result lists, and collecting users' queries and search interactions to optimise the ranker. A drawback of this centralised paradigm is that user's privacy cannot be fully guaranteed. Firstly, the server maintains a record of the searchable documents in the form of an index. For web search where documents are crawled on the Internet, the threat to user privacy is limited as the documents are publicly accessible. However, it is not desirable when the documents to be searched are personal, such as in email search~\cite{kim2017understanding} or desktop search~\cite{cohen2008ranking}, or when the data contains private or valuable information, such as in medical record search~\cite{hanauer2006emerse}, or enterprise search~\cite{DBLP:conf/adc/Hawking04}. Another potential privacy risk is that the central server also observes other information that the users may wish to keep private, such as the queries they issue and their clicks on search results, which reflect personal information needs, search interests, and user behaviour. This information can disclose users' demographic attributes, personal habits, political views, etc~\cite{barbaro2006face,DBLP:conf/iir/Carpineto016,sousa2021privacy}. With growing awareness of protecting personal data and the enforcement of data protection regulations such as the General Data Protection Regulation (GDPR) in the European Union, and the California Consumer Privacy Act (CCPA) in California, privacy protection has become not only an ethical issue but also a legal requirement. Thus, the design of a responsible ranking system requires anonymising the user-generated data during search activities.

%Why Online Learning to Rank?
%Techniques like data encryption~\cite{gentry2009fully} or user anonymity~\cite{mivule2017data} can be used to protect user's interaction data for CLTR as it is fully implemented offline.
%However, solutions under the context of Online Learning to Rank is not that obvious and accessible.
In this thesis, we focus on addressing the aforementioned privacy issues in the context of Online Learning to Rank. As an online method, OLTR has been shown to be more adaptive to changes in user search intents~\cite{zhuang2021how} compared to CLTR which is fully implemented offline. However, the current implementation of OLTR systems poses threats to user privacy: the privatization of user queries and interactions cannot be guaranteed, nor can the confidentiality of documents containing personal information, such as those in desktop or email search scenarios. Therefore, there is a need for methods that ensure more reliable, effective, and privacy-preserving OLTR systems.

% [Writing outlines]

%What is FL?
%Background of FOLTR.
%What are the research gap for FOLTR. And uniqueness of FOLTR comparing to other ML tasks.
%What are the research questions under this thesis and why study this.

\section{Task Definition}

To address privacy issues in OLTR, this thesis focuses on a new training pipeline that transitions previous centralised OLTR approaches into a Federated Learning (FL) setting, termed Federated Online Learning to Rank (FOLTR)~\cite{kharitonov2019federated}).

%What is FL?
First proposed by McMahan et al.~\cite{mcmahan2017communication}, Federated Learning addresses data privacy issues in training machine learning models. In the traditional setting, known as Centralised Machine Learning, training data are typically gathered by a central server before the learning process. This approach poses a threat to user privacy, especially if the training data contain sensitive information such as personal health records, user profiles, or financial transactions. For example, multiple hospitals want to collaboratively improve their machine learning models for diagnosing certain diseases. The sharing of and gathering patient data to a central server will be infeasible due to the data privacy restriction. With stricter data regulations and increased user awareness of data protection, obtaining such private data has become increasingly difficult and infeasible. Additionally, as these data are usually distributed across multiple devices, gathering them centrally is costly due to limited communication bandwidth. However, machine learning models, particularly deep learning models, require large amounts of data to achieve high performance, especially in today's natural language processing and computer vision tasks. Federated Learning provides a solution: instead of sharing data, the training is moved to the device of each data owner (the "client"). Each client trains a local model using its local dataset. All local models have the same structure and only the information needed for collaborative training is sent to a central server. In the most commonly used federated learning algorithm, FedAvg~\cite{mcmahan2017communication}, model updates are communicated from each client to a central server, which takes the weighted average of the model updates from all clients to perform an update for the global model. The newly updated global model is then sent back to each client to replace the local model, and subsequent local training is based on this updated model. Federated Learning allows each client to benefit from the data of all participants through collaboration, without the threat of sharing local data.

% A brief introduction of a FOLTR system with a figure

\begin{figure}[t]
	\includegraphics[width=0.7\textwidth]{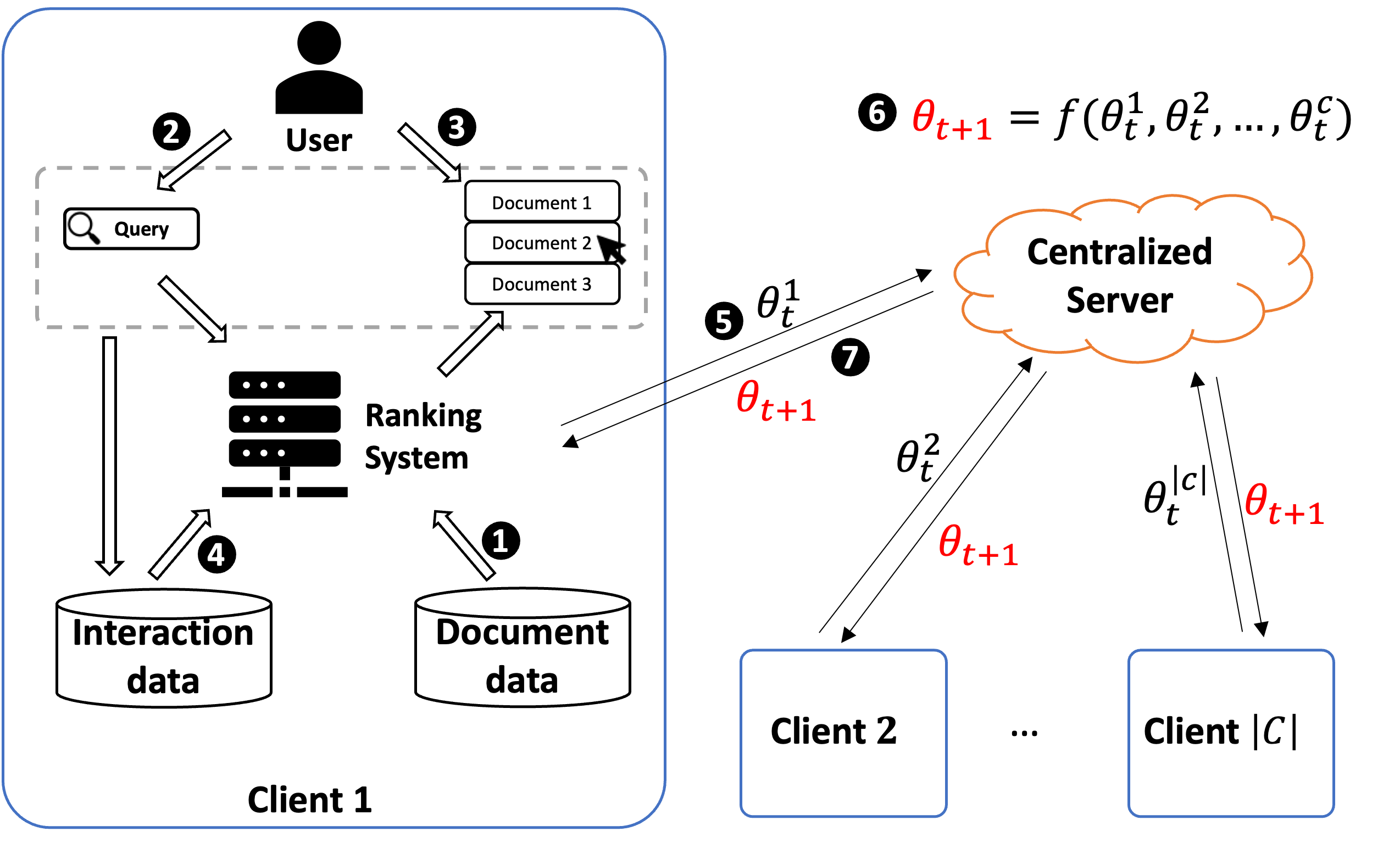}
	\centering
	\caption{Schematic representation of Federated Online Learning to Rank (FOLTR) setting.}
	\label{fig:foltr_overview} 
\end{figure}

In a typical Federated Online Learning to Rank (FOLTR)~\footnote{In general, FL has been approximately grouped into three types: horizontal FL, vertical FL and transfer FL~\cite{yang2019federated}. In this thesis, we only consider horizontal FL for our FOLTR framework.} setting, depicted in Figure~\ref{fig:foltr_overview}, the documents to be searched, along with user queries and interactions, are withheld from the central server, and so is also the responsibility of producing search results. Instead, search is performed within the user device \circled{2}, which also indexes the user's data \circled{1}, collects users interactions \circled{3}, and performs the online updates required by the OLTR method \circled{4}. Updates from several user devices are then shared with a central server \circled{5}. The central server is responsible for combining the ranker updates from different clients to produce a new version of the ranker \circled{6}, which is then distributed to the user devices \circled{7}. The advantage of a federated OLTR solution over a traditional OLTR approach is that neither user data nor user queries and interactions are seen by or shared with anyone, thus potentially preserving user privacy.

%Background of FOLTR.
%The first work in Federated Online Learning to Rank (shortened as "FOLtR-ES") was proposed by Kharitonov~\cite{kharitonov2019federated}, which transitions the centralised online learning to rank setting into a federated learning framework and uses evolution strategies~\cite{salimans2017evolution} as the optimisation method. User privacy is further protected through a privatization procedure based on the theory of differential privacy~\cite{dwork2014algorithmic}. Although this pioneering study demonstrates promising results on certain datasets, a follow-up study by Wang et al.~\cite{wang2021federated} highlights limitations of FOLtR-ES on large-scale datasets and under more commonly used evaluation metrics in the area of OLTR.

In this thesis, we conduct a comprehensive study of Federated Online Learning to Rank, focusing on addressing the limitation of previous methods and tackling unique challenges in this area.

\section{Research Questions}

%What are the research gap for FOLTR. And uniqueness of FOLTR comparing to other ML tasks.
Unlike most machine learning tasks, stochastic gradient descent (SGD) optimisation is not commonly used in Online Learning to Rank~\cite{yue2009interactively} due to two main reasons: 1) the algorithm learns directly from user interaction data in an online manner, and 2) pairwise, listwise and pointwise loss functions are commonly used in ranking tasks~\cite{liu2009learning}. However, Federated Learning typically relies on local SGD under an offline training setting with local datasets on device. This brings specific challenges for adopting FL for OLTR, demonstrated also by the existence of just a single FOLTR method prior to the work in this thesis.

%challenges in the area of FOLTR
%In particular, the existing studies on FOLTR are still facing several challenges:

\textbf{Challenge 1: Low Effectiveness in Balancing Ranking Performance and Privacy Protection.}
The only FOLTR method existing prior to the work of this thesis, FOLtR-ES~\cite{kharitonov2019federated}, demonstrated the feasibility of conducting OLTR within the FL framework. Empirical studies and theoretical analyses have shown its effectiveness in producing ranking lists and ensuring user privacy protection with the help of differential privacy. This has highlighted the potential of addressing privacy issues in OLTR using FL. However, due to the lack of some commonly-used evaluation protocols in OLTR, this method requires to be further explored and verified its effectiveness. As we later demonstrate in this thesis, several limitations of FOLtR-ES exist: it performs poorly on large-scale datasets and other common evaluation practices in OLTR, it does not effectively optimize neural ranking models, and its ranking performance lags behind the state-of-the-art OLTR in centralised setting. As the optimal goal of OLTR is to serve users' information needs, the ranking performance should be given equal importance to user privacy protection. One key aspect neglected in the previous FOLTR method is addressing the biases in user interactions. Without accounting for bias interactions, the exploration of model updates can only reach to a suboptimal space. Therefore, how to effectively address biases from user interactions in a federated setting and simultaneously protect user privacy remain the highest priority challenge.
%how to effectively address the noise and biases from user interactions in a federated setting, which, in return, achieve competitve performance compared to the state-of-the-art OLTR methods, and at the same time, further protect user's privacy, remain the challenge of first priority.

\textbf{Challenge 2: Uncertain Robustness under Heterogeneous Clients.}
In a federated setting, training data is distributed across local devices, leading to data heterogeneity or non independent and identically distributed (non-IID) data. Since local data is not shared between clients and the central server, ignoring non-IID issues can result in divergence among local model updates, negatively impacting the convergence of the global model during federated aggregation. While several methods have addressed FL under non-IID data and proposed solutions for tasks in computer vision and natural language processing~\cite{zhao2018federated,briggs2020federated,zhu2021federated,li2022federated}, there is a lack of focus on this issue in FOLTR. Given that users have different search intents and interaction behaviours, it further exacerbates the non-IID problem and complicates the development of robust and effective FOLTR systems. Due to the diversity of search activities and lack of non-IID benchmarks, the key factors that lead to non-IID issues and their impact on ranking performance remain unclear and require investigation.

\textbf{Challenge 3: Potential Security Risk Posed by Malicious Clients.}
Existing FOLTR systems are not necessarily secure: the federated mechanism provides malicious clients with opportunities to attack the effectiveness of the global ranker by deliberately manipulating their local data (data poisoning) or local model updates (model poisoning). For example, malicious clients can send arbitrary weights to the server, disrupting the convergence of the global ranker after aggregation. This kind of attack, termed untargeted poisoning attack, aims to compromise the integrity of the global model trained federatively~\cite{bagdasaryan2020backdoor}. This issue is critical for federated learning systems but has not yet been studied on FOLTR where the impact of a poisoned ranking model is further exacerbated due to its online updating and service. To understand the vulnerability of existing FOLTR methods to these attacks and the effectiveness of related defense mechanisms, two main challenges exist: first, how to design and implement existing data and model poisoning attacks to the context of FOLTR; second, how to design defense strategies and validate their utility under various settings of attack.

%However, the vulnerability of existing FOLTR methods to these attacks and the effectiveness of related defense mechanisms remain unknown. Thus, designing data poisoning and model poisoning strategies, understanding the potential security risk posed by these attacks, and developing effective defense strategies for FOLTR remain challenging and need to be addressed.

\textbf{Challenge 4: Addressing User's Right to Be Forgotten.}
Recent data protection legislations, such as the General Data Protection Regulation (GDPR)~\cite{voigt2017eu} and the California Consumer Privacy Act (CCPA)~\cite{harding2019understanding}, emphasize a user’s unconditional right to be forgotten. This means users can request that contributions made using their data be removed from learned models, a concept also known as machine unlearning~\cite{bourtoule2021machine} and federated unlearning~\cite{liu2021federaser}. A federated learning system should accommodate the possibility for clients to leave the federation and request the erasure of their data contributions. However, existing FOLTR systems lack an unlearning mechanism to forget certain users' contributions, making their design defective in this respect. Developing an effective and efficient unlearning mechanism presents two main challenges. The first challenge is how to efficiently unlearn without requiring an unreasonable amount of additional computation while ensuring the new ranker maintains comparable effectiveness to a model retrained from scratch. The second challenge is how to evaluate the effectiveness of an unlearning method. In a FOLTR system with many clients, the impact of removing a client's data on the ranker’s effectiveness may be marginal or even unnoticeable. A natural question from a user leaving the federation is: how can it be proven that the impact of their data on the ranker has been erased? Therefore, an adequate evaluation method is required to verify whether an unlearning process effectively achieves forgetting.

%What are the research questions under this thesis and why study this.

In this thesis, we address the unique research gaps in FOLTR by focusing on the following research questions.

\subsection{RQ1: How to design an effective and generalisable FOLTR method that can achieve competitive performance compared to state-of-the-art OLTR methods?}
% One sentence about the RQ.

% Further paragraphs to explain what this RQ is about.

As the service for addressing users' information needs, the ranking effectiveness of FOLTR should be a primary concern. It is currently unknown if existing methods can achieve performance on par with centralised OLTR methods. If not, the key question is how to enhance the ranking effectiveness for FOLTR, ensuring that the use of federated learning for user privacy protection does not significantly compromise ranking performance. Additionally, exploring methods for integrating other data protection techniques, such as differential privacy~\cite{dwork2014algorithmic}, into FOLTR could provide feasible and adaptive solutions for balancing privacy protection and model utility.
This RQ aligns with Challenge 1.

\subsubsection{RQ1.1: Does the only existing FOLTR method (FOLtR-ES) generalize to other commonly-used datasets and metrics in OLTR?}

With this RQ, we aim to conduct a comprehensive analysis of previous FOLTR method with a focus on ranking utility. Specifically, to compare the performance with centralised setting, we evaluate the existing FOLTR method on large-scale datasets and use commonly-used metrics against OLTR baselines trained in a centralised manner. A thorough understanding of the existing method, including their pros and cons, will inform the design of future methods in this field. This analysis will provide insights into the generalisability and effectiveness of FOLtR-ES across diverse scenarios.

\subsubsection{RQ1.2: How to integrate the state-of-the-art OLTR method, PDGD, with the widely-used, easy-to-implement FedAvg framework while further protecting user privacy?}

Unlike other applications of FL, OLTR methods handle noisy and biased user feedback in an online manner, and Stochastic Gradient Descent (SGD) is not commonly used for optimising ranking models in OLTR. In this RQ, we investigate whether the state-of-the-art OLTR method, Pairwise Differentiable Gradient Descent (PDGD)~\cite{oosterhuis2018differentiable}, can be integrated into the common FL framework to achieve a highly effective ranker. Simultaneously, methods for integrating other data protection techniques, such as differential privacy~\cite{dwork2014algorithmic}, will be explored to provide enhanced privacy guarantees. This research will focus on achieving a balance between ranking performance and privacy protection by adapting the OLTR method to fit the FL framework and enhancing it with additional privacy-preserving techniques.

\subsection{RQ2: Are FOLTR methods robust to non-IID data among clients?}
%\subsection{RQ2: Is non-IID data a threat in FOLTR?}

% One sentence about the RQ.
% Further paragraphs to explain what this RQ is about.

Within this RQ, we consider a scenario where the local training data are non independent and identically distributed (non-IID) across participating clients in FOLTR. This situation violates the common IID assumption in machine learning and can cause the global model to update in a biased direction~\cite{zhu2021federated}. Investigating this scenario is crucial because the learning of FOLTR relies on user-inputted queries, local documents, and relevance signals inferred from user interactions. Both search behaviours and interactions with search engines result pages can vary among users, naturally leading to non-IID issues for FOLTR. Understanding the robustness of FOLTR methods under these conditions will help in designing more resilient and effective ranking systems that can handle diverse user behaviours and interaction patterns.
This RQ is in line with Challenge 2.

\subsubsection{RQ2.1: What are the potential types of non-IID data in FOLTR and their impact on model performance?}

Since there has been no prior study addressing the non-IID challenge in FOLTR, the typical types of non-IID data specific to search scenarios remain unidentified, as do their potential impacts on ranking performance. In this RQ, we aim to identify and categorise types of non-IID data specific to search scenarios and understand which types cause significant degradation in ranking effectiveness compared to IID settings. By concretizing the types of non-IID data specific to search scenarios and analyzing their impacts, this RQ aims to uncover the potential challenges and degradations in ranking performance caused by non-IID issues in FOLTR systems.

\subsubsection{RQ2.2: Do existing methods that deal with non-IID data in FL for other tasks generalise to FOLTR?}

As demonstrated in this thesis, certain types of non-IID data negatively impact the ranking performance of FOLTR. The focus of this RQ is to investigate methods that can alleviate the impact of non-IID data. Specifically, we exploit existing methods proven effective in supervised federated learning under offline setting and migrate them into FOLTR, where training occurs online and relevance feedback is noisy and biased. By exploring and adapting existing methods for handling non-IID data in general FL to the context of FOLTR, this RQ aims to mitigate the negative impacts of non-IID data on ranking performance in FOLTR systems.

\subsection{RQ3: Are FOLTR methods secure in the face of potential risks brought by malicious participants?}

% One sentence about the RQ.

% Further paragraphs to explain what this RQ is about.

FOLTR relies on collaboration among participating clients. However, this exposes the system to potential risks from malicious participants who aim to obstruct the convergence of the global model and impair its performance by transmitting arbitrary or carefully crafted model weights. These are termed poisoning attacks with two main categories: \textit{data poisoning}, where the attack is to inject perturbation into the training data without any modification upon the training process itself, and \textit{model poisoning}, where the attackers directly manipulate the training process by altering the objective function, directly editing the model post-training, and so on. Given the lack of previous research on how these attacks affect FOLTR systems, within this RQ, we aim to provide a comprehensive understanding of potential risks brought by poisoning attacks and investigate methods for defending against them.
This RQ aligns with Challenge 3.

\subsubsection{RQ3.1: Can data poisoning and model poisoning strategies undermine ranking performance on FOLTR?}

This research question explores the potential risks associated with poisoning attacks in the context of FOLTR. Inspired by the established data and model poisoning strategies within general FL, we tailor and perform them under the specific nuances of the FOLTR environment. Furthermore, we plan to examine several crucial factors that influence the extent of ranking performance degradation. Understanding these factors is expected to inform the development of more robust defensive strategies against such attacks.

\subsubsection{RQ3.2: Can defense strategies mitigate the impact of malicious attacks in FOLTR?}

As we demonstrate in this thesis, poisoning attacks can cause degradation to the ranking effectiveness under FOLTR. Within this RQ, we exploit defense strategies to alleviate the impact from malicious attacks. Specifically, we employ widely recognized robust aggregation rules as defensive measures and assess their performance in counteracting both data poisoning and model poisoning attacks. The evaluation considers various attack configurations to understand how these defenses can best maintain the integrity and effectiveness of the ranking system. Furthermore, to assess the feasibility of deploying these defensive strategies in real-world settings, we examine scenarios without any attacks to verify if there is any decline in effectiveness when these defenses are implemented.

\subsection{RQ4: How to effectively and efficiently forget client data from the trained ranker when a client requests to exit the FOLTR system?}

% One sentence about the RQ.

% Further paragraphs to explain what this RQ is about.

Users of FOLTR have the right to discontinue their participation when they no longer require the service. In response to the requirements driven by users' privacy concerns and recent data regulations, the model should eliminate any knowledge derived from the data of users who opt out. A straightforward method involves retraining the model from scratch without the data from the exiting users, but this approach is computationally expensive and time-consuming. To overcome this, methods for Machine Unlearning and Federated Unlearning have been studied. This research question delves into these methods which are underexplored within the context of FOLTR. Specifically, it assesses both the effectiveness — vital as FOLTR addresses users' information needs in real-time — and the efficiency — crucial as FOLTR operates in an online environment and need to adapt swiftly to dynamic conditions.
This RQ aligns with Challenge 4.

\subsubsection{RQ4.1: How to evaluate the effectiveness of unlearning?}
%\subsubsection{RQ4.1: How can the effectiveness of unlearning be evaluated?}

Evaluating the effectiveness of unlearning presents significant challenges, as evidenced by existing literature. Several factors contribute to these challenges. First, due to the stochastic inherence of the learning processes of machine learning and federated learning, the learnt model is influenced by numerous configuration factors that are difficult to control. This variability can result in vastly different models even when trained on the same dataset, making it unclear whether changes in the model are due to effective unlearning strategies or other variables. Second, the leaving of a small number of users may not significantly impact model performance, as the effect of one user's data can be relatively minor. Consequently, traditional evaluation settings and metrics are inadequate for assessing unlearning effectiveness accurately. Within this RQ, we aim to highlight these issues in the FOLTR context and develop a suitable evaluation framework that adequately measures the effectiveness of unlearning strategies.

\subsubsection{RQ4.2: How to efficiently forget the client without requiring retraining from scratch?}

In addition to effectiveness, the efficiency of unlearning is another question worth studying. Specifically, unlearning methods need to demonstrate efficiency compared to the baseline that consists of retraining from scratch. If not, the necessity for unlearning diminishes, rendering the method ineffective. Within this RQ, we explore unlearning methods tailored to FOLTR and analyse the factors influencing both the effectiveness and efficiency of unlearning within the established evaluation framework.

\section{Contributions}

In this thesis, we thoroughly explore the aforementioned research questions with the aim of developing effective, robust, secure federated online learning to rank methods.

We commence by examining the effectiveness of FOLTR, identifying shortcomings in existing work, and proposing an effective method named FPDGD. Addressing a crucial gap in previous research, we tackle the bias stemming from user implicit feedback by adapting the Pairwise Differentiable Gradient Descent method to the Federated Averaging framework. 
To ensure a robust privacy guarantee, we introduce a noise-adding clipping technique grounded in the theory of differential privacy to be used in combination with FPDGD. 
Our empirical evaluations demonstrate that FPDGD significantly outperforms the sole existing federated OLTR method. Moreover, FPDGD exhibits greater robustness across various privacy guarantee requirements than the current method, rendering it more dependable for real-world applications.
%The following publications support these findings:
This work has resulted in two publications:

\begin{enumerate}
	
	\item Shuyi Wang, Shengyao Zhuang, and Guido Zuccon, \href{https://doi.org/10.1007/978-3-030-72240-1_10}{Federated Online Learning to Rank with Evolution Strategies: A Reproducibility Study}, \textit{In European Conference on Information Retrieval (ECIR)}, pages 134-149, 2021
	
	\item Shuyi Wang, Bing Liu, Shengyao Zhuang, and Guido Zuccon, \href{https://doi.org/10.1145/3471158.3472236}{Effective and Privacy-preserving Federated Online Learning to Rank}, \textit{In Proceedings of the 2021 ACM SIGIR International Conference on Theory of Information Retrieval (ICTIR)}, pages 3–12, 2021
\end{enumerate}

We assess the resilience of FOLTR under non independent and identically distributed (non-IID) data across participating clients. Initially, we enumerate four potential data distribution settings that may give rise to non-IID issues. Subsequently, we delve into the impact of each setting on ranking performance, pinpointing which data distribution may present challenges for FOLTR methods. This analysis constitutes an important contribution to the current landscape of FOLTR research as understanding the factors influencing performance, particularly the effects of non-IID data, is paramount for the deployment of FOLTR systems.
%Our findings are detailed in the following publication:
This work has resulted in the following publication:

\begin{enumerate}
	
	\item Shuyi Wang, and Guido Zuccon, \href{https://doi.org/10.1145/3477495.3531709}{Is Non-IID Data a Threat in Federated Online Learning to Rank?}, \textit{In Proceedings of the 45th International ACM SIGIR Conference on Research and Development in Information Retrieval (SIGIR)}, pages 2801–2813, 2022

\end{enumerate}

We examine both data and model poisoning attack strategies to showcase their impact on FOLTR search effectiveness. Additionally, we investigate the efficacy of robust aggregation rules in federated learning to counter these attacks. This exploration offers understanding of the effect of attack and defense methods for FOLTR systems, as well as identifying the key factors influencing their effectiveness. Our research contributes to a deeper understanding of the dynamics between attack and defense methods in FOLTR systems, providing insights for enhancing their resilience in real-world scenarios.
%The outcomes of our study are detailed in the following publication:
This work has resulted in the following publication:

\begin{enumerate}
	
	\item Shuyi Wang, and Guido Zuccon, \href{https://doi.org/10.1145/3578337.3605117}{An Analysis of Untargeted Poisoning Attack and Defense Methods for Federated Online Learning to Rank Systems}, \textit{In Proceedings of the 2023 ACM SIGIR International Conference on Theory of Information Retrieval (ICTIR)}, pages 215–224, 2023

\end{enumerate}

Lastly, we study an effective and efficient unlearning method designed to remove a client’s contribution without compromising the overall ranker effectiveness, and without necessitating a complete retraining of the global ranker. A pivotal challenge lies in assessing whether the model successfully unlearns the contributions from the client requesting removal (referred to as "the unlearned client"). To tackle this, we instruct the unlearned client to execute a poisoning attack through adding noise to its updates. Subsequently, we measure whether the impact of the attack diminishes after the unlearning process has been executed. Through empirical experiments, we showcase the effectiveness and efficiency of the unlearning strategy under various parameter settings combinations.
%Our contributions resulted in the following publication:
This work has resulted in the following publication:

\begin{enumerate}
	
	\item Shuyi Wang, Bing Liu, and Guido Zuccon, \href{https://doi.org/10.1007/978-3-031-56063-7\_7}{How to Forget Clients in Federated Online Learning to Rank?}, \textit{In European Conference on Information Retrieval (ECIR)}, pages 105--121, 2024
	
\end{enumerate}

In summary, this thesis delivers a comprehensive examination on Federated Online Learning to Rank, addressing various unexplored areas concerning its effectiveness, robustness, and security. By delving into these domains, the research broadens the landscape of FOLTR, offering insights and advancements to the field.

\section{Thesis Overview}

The thesis is structured as follows:

\textbf{Chapter~\ref{Chap:background}: Background and Literature Review.}
This chapter provides the background and literature review of Learning to Rank, Online Learning to Rank, Federated Learning and related methods pertinent to this thesis. Specifically, an explanation of the relevant terminology, methodologies, evaluation protocols, and datasets is provided, which lays the groundwork for subsequent chapters.

\textbf{Chapter~\ref{Chap:fpdgd}: Effective and Privacy-Preserving FOLTR.}
We conduct an analysis of the existing FOLTR method and introduce our method designed for effective and privacy-preserving FOLTR. 
(Supporting paper: \cite{wang2021federated,wang2021effective})

\textbf{Chapter~\ref{Chap:foltr-noniid}: FOLTR with non-IID Data.}
We present our work, which introduces a terminology of non-IID data in FOLTR and analyses its performance under various federated aggregation rules.
(Supporting paper: \cite{wang2022non})

\textbf{Chapter~\ref{Chap:foltr-poisoning}: FOLTR under Poisoning Attacks.}
We introduce our study, which investigates the potential risks posed by data and model poisoning attacks to FOLTR systems, along with the effectiveness of corresponding defense strategies.
(Supporting paper: \cite{wang2023analysis})

\textbf{Chapter~\ref{Chap:foltr-unlearning}: Unlearning for FOLTR.}
We introduce our method for effective and efficient unlearning in FOLTR, which is the first study of its kind.
(Supporting paper: \cite{DBLP:conf/ecir/WangLZ24})

\textbf{Chapter~\ref{Chap:Conclusion}: Conclusion.}
We conclude by providing a comprehensive summary of our findings, emphasizing our contributions to the field of FOLTR. We discuss avenues for further research, particularly in areas crucial for practical FL implementations for ranking tasks but still underexplored.

\chapter[Background and Literature Review]{Background and Literature Review}
\label{Chap:background}	%CREATE YOUR OWN LABEL.
\pagestyle{headings}

% ********* Enter your text below this line: ********
%Introduce the broad layout of the chapter.

% ***************************************************

%\section{Introduction}
%\label{Sec:label}	%CREATE YOUR OWN LABEL.
%
%% ********* Enter your text below this line: ********
%Add your text here. 
%
%% ***************************************************
%
%
%\subsection{Datasets}
%
%MQ2007, MQ2008~\cite{DBLP:journals/corr/QinL13}

This chapter provides an overview of Learning to Rank, Online Learning to Rank methods, Federated Learning and other related works that are involved in this thesis. Additionally, the common experimental settings, including datasets, user simulation, and evaluation metrics, will be introduced in detail. 
\label{sec:relatect works}

\section{Learning to Rank}

Learning to Rank (LTR) is the task to construct a ranking model with the aim to rank documents in a manner that optimally satisfies user queries~\cite{liu2009learning}. To achieve this, the ranking function is learned with the help of machine learning techniques using a set of labelled query-document pair examples, represented by predefined features~\cite{DBLP:journals/corr/QinL13}. Specifically, LTR has evolved through three primary approaches: the pointwise, pairwise, and listwise approaches. Pointwise approaches, such as those presented in the works of Gao et al.~\cite{gao2005linear}, treat ranking as a regression or classification problem on individual documents. Pairwise methods, exemplified by the RankNet model developed by Burges et al.~\cite{burges2005learning}, focus on comparing pairs of documents to determine their relative orders. Listwise approaches, with models like ListNet~\cite{cao2007learning} and LambdaRank~\cite{burges2006learning}, consider the entire list of documents when optimizing the ranking. Recent advancements in deep learning have further propelled LTR, integrating neural networks to enhance ranking accuracy and scalability. Despite these advancements, challenges such as data sparsity, scalability, and data annotation continue to drive research in this dynamic field.

A key limitation of LTR is the reliance on explicit relevance annotations, which require substantial effort and costs to collect~\cite{DBLP:journals/jmlr/ChapelleC11}. Editorial labels also pose ethical issue when labelling private data~\cite{wang2016learning} such as emails. In addition, user preferences may not agree with that of annotators~\cite{sanderson2010test} and these labels cannot reflect evolving user preferences and search intents~\cite{lefortier2014online}. To solve the data annotation issues for LTR, two group of methods have been proposed and investigated with the integration of user feedback. One method is termed Counterfactual Learning to Rank (CLTR)~\cite{agarwal2019general,jagerman2019model}, which improves ranking models using logged user interaction data from previous ranking systems, such as clicks, in a manner that accounts for potential biases and confounding factors by using techniques from causal inference and counterfactual reasoning. The other method is called Online Learning to Rank (OLTR), which collects real-time user interactions and directly learns from them to update the ranking model. This procedure is conducted in an online and interactive environment, meaning the ranking system simultaneously handles real users' information needs and learns from user feedback to update itself.

\section{Online Learning to Rank}

Online Learning to Rank (OLTR) differs from traditional Learning to Rank (LTR) methods in that it makes use of user online feedback (such as clicks) as the optimization goal in order to tackle some data-annotation drawbacks of LTR.

The Dueling Bandit Gradient Descent (DBGD) uses online evaluation to unbiasedly compare two or more rankers given a user interaction~\cite{yue2009interactively}. Subsequent work developed more reliable or more efficient online evaluation methods, including Probabilistic Interleaving (PIGD)~\cite{hofmann2011probabilistic}, and its extension, the Probabilistic Multileaving (PMGD)~\cite{oosterhuis2016probabilistic}, which compares multiple rankers at each interaction, resulting in the best DBGD-based algorithm. However, this method suffers from a high computational cost because it requires sampling ranking assignments to infer outcomes~\cite{zhuang2020counterfactual}. Further variations that reuse historical interaction data to accelerate the learning in DBGD have also been investigated~\cite{hofmann2013reusing}.

The Pairwise Differentiable Gradient Descent method (PDGD)~\cite{oosterhuis2018differentiable} constructs a pairwise gradient to update the ranking model according to the click feedback. Compared to previous OLTR methods, PDGD has showcased faster learning (higher online performance) and convergence to higher effectiveness (higher offline performance). Unlike other methods, PDGD can be applied to any differentiable ranking model and it is an unbiased method. Oosterhuis and de Rijke trained a linear and a neural ranker and both showed effective results~\cite{oosterhuis2018differentiable}.

Counterfactual Online Learning to Rank (COLTR)~\cite{zhuang2020counterfactual} uses the same DBGD framework for updating the ranker interactively. Instead of interleaving or multileaving, the key difference of COLTR from DBGD-based methods is the use of counterfactual evaluation from Counterfactual Learning to Rank (CLTR) to replace online evaluation thus COLTR uses clicks collected by the current ranker to evaluate candidate rankers.

Another group of methods focus on leveraging powerful offline learning learning rank solutions to the online setting. PairRank~\cite{jia2021pairrank} directly learns a pairwise logistic regression ranker online and explores the pairwise ranking space via a divide-and-conquer strategy based on the model’s uncertainty about the documents’ rankings. A follow-up work by Jia and Wang~\cite{jia2022learning} is proposed to extend PairRank with a multi-layer neural ranker. However, this method suffers from high computational complexity caused by the exploration in the pairwise document ranking space. To solve the computational efficiency bottleneck, Jia et al.~\cite{jia2022scalable} propose an efficient exploration strategy based on bootstrapping: an ensemble of ranking models trained with perturbed user click feedback.

\subsection{Evaluation Practice in OLTR}

It is common practice in OLTR to simulate user interactions with the search results using labelled offline learning to rank datasets~\cite{hofmann2013reusing,DBLP:conf/wsdm/SchuthOWR16,oosterhuis2018differentiable}. This is because no public dataset with LTR features and clicks is available; in addition OLTR methods directly manipulate the rankings that have to be shown to users, so even if a public dataset with LTR features and clicks was to be available, this could not be used for OLTR. This section introduces the common implementation and evaluation practice in OLTR, which this thesis also follows.

\subsubsection{Datasets}
\label{chap2-dataset}
In the study of OLTR, four learning-to-rank (LTR) datasets are commonly used: MQ2007~\cite{DBLP:journals/corr/QinL13}, MSLR-WEB10K~\cite{DBLP:journals/corr/QinL13}, Yahoo~\cite{DBLP:journals/jmlr/ChapelleC11}, and Istella-S~\cite{lucchese2016post}. Each dataset contains query ids and candidate document lists for each query, which is formalized as exclusive query-document pairs. Each query-document pair is represented by a multi-dimensional feature vector and annotated relevance label of the corresponding document. 

Among the selected datasets, MQ2007 is relatively small and has fewer assessed documents per query. The dataset, compiled using data from the TREC 2007 Million Query Track~\cite{allan2007million},  contains 1,700 queries, divided into 5 folds, with associated relevance assessments, expressed on a 3-level scale, from \emph{not relevant} (0) to \emph{very relevant} (2). Each query-document pair is represented by a 46-dimensional feature vector.

Provided by commercial search engines, the other three datasets are larger and more recent. MSLR-WEB10K, constructed from a retired labelling set of a Microsoft web search engine, has 10,000 queries and each query is associated with 125 documents on average, each represented with 136 features. Yahoo has 29,900 queries and each query-document pair has 700 features. Istella-S is the largest, with 33,018 queries, 220 features, and an average of 103 documents per query. These three commercial datasets are all annotated for relevance on a five-grade-scale: from \textit{not relevant} (0) to \textit{perfectly relevant} (4). Both MQ2007 and MSLR-WEB10K datasets have five data splits (stored separately in five data folders) while Yahoo and Istella-S contain only one.

\subsubsection{User simulations} 
\label{chap2-click}
In standard setup for user simulations in OLTR~\cite{hofmann2013reusing,DBLP:conf/wsdm/SchuthOWR16,oosterhuis2018differentiable}, user's queries are selected by randomly choosing from the dataset and user interactions on the search engine result pages (SERPs) are simulated accordingly. While there are many forms of user interactions, the most common and widely-used one is the click signal, which is also the only form of user interaction that is considered and simulated throughout this thesis.

In particular, user's clicks are simulated relying on the \textit{Cascade Click Model} (CCM) click model~\cite{guo2009efficient} based the relevance label and ranking position of the candidate documents. Specifically, for each query, the SERPs are commonly limited to $k$ documents. User clicks on the displayed ranking list are generated based on the CCM click model. Each user is assumed to inspect every document displayed in a SERP from top to bottom while clicks the document with click probability $P(click =1 | rel(d))$ conditioned with the actual relevance label. After a click happens, the user will stop the browsing session with stop probability $P(stop = 1 | click = 1, rel(d))$, or continue otherwise. In the aforementioned previous works, same as experiments in this thesis, three widely-used instantiations of CCM click model are considered: \textit{perfect, navigational, informational}. A \textit{perfect} user inspects every document in a SREP with high chance to click high relevant documents thus provides very reliable feedback. A \textit{navigational} user also searches for reasonably relevant document but has a higher chance of stopping browsing after one click. An \textit{informational} user typically clicks on many documents without a specific information preference thus provides the noisiest click feedback. The implementations of these click models are detailed in Table~\ref{tbl:chap2:click1} and Table~\ref{tbl:chap2:click2}.

\newcommand{\tc}[1]{\multicolumn{1}{c}{#1}}
\setlength{\tabcolsep}{3mm}

% Redefine X to center its content
\newcolumntype{C}{>{\centering\arraybackslash}X}

\begin{table}[ht!]
	\centering
	\caption{The three click model instantiations used for the MQ2007 dataset. $r = rel(d)$ represents the relevance label for the query-document pair.}\label{tbl:chap2:click1}
	\begin{tabularx}{\textwidth}{p{2.8cm}CCCCCC}
		\toprule
		& \multicolumn{3}{c}{$p(click=1|rel(d))$} & \multicolumn{3}{c}{$p(stop=1|click=1,rel(d))$} \\
		\cmidrule(r){2-4}  \cmidrule(){5-7}
		\textit{click model} & \tc{$r$ = 0} & \tc{$r$ = 1} &\tc{$r$ = 2} &  \tc{$r$ = 0} & \tc{$r$ = 1} & \tc{$r$ = 2} \\
		\midrule
		$perfect$ & 0.0 & 0.5 & 1.0& 0.0 & 0.0 & 0.0 \\
		$navigational$ & 0.05 & 0.5 & 0.95 & 0.2 & 0.5 & 0.9 \\
		$informational$ & 0.4 & 0.7 & 0.9 & 0.1 & 0.3 & 0.5\\
		\bottomrule
	\end{tabularx}
\end{table}

\begin{table}[ht!]
	\centering
	\small
	\caption{The three click model instantiations used for the datasets with five-grade-scale relevance annotation (MSLR-WEB10K, Yahoo!, and Istella-S datasets). $r = rel(d)$ represents the relevance label for the query-document pair.}\label{tbl:chap2:click2}
	\begin{tabularx}{\textwidth}{p{2cm}XXXXXXXXXX}
		\toprule
		& \multicolumn{5}{c}{$p(click=1|rel(d))$} & \multicolumn{5}{c}{$p(stop=1|click=1,rel(d))$} \\
		\cmidrule(r){2-6}  \cmidrule(){7-11}
		\textit{click model} & $r$ = 0 & $r$ = 1 &$r$ = 2 & $r$ = 3 & $r$ = 4 & $r$ = 0 & $r$ = 1 & $r$ = 2 & $r$ = 3 & $r$ = 4 \\
		\midrule
		$perfect$ & 0.0 & 0.2 & 0.4 & 0.8 & 1.0& 0.0 & 0.0 & 0.0 & 0.0 & 0.0\\
		$navigational$ & 0.05 & 0.3 & 0.5 & 0.7 & 0.95& 0.2 & 0.3 & 0.5 & 0.7 & 0.9\\
		$informational$ & 0.4 & 0.6 & 0.7 & 0.8 & 0.9& 0.1 & 0.2 & 0.3 & 0.4 & 0.5\\
		\bottomrule
	\end{tabularx}
\end{table}

%. Under SDBN, users are assumed to examine a SERP from top to bottom. Each document is inspected and clicked with click probability $P(click = 1 | R)$, conditioned on the relevance label $R = rel(d)$. After a click occurs, the user stops scanning with stopping probability $P(stop = 1 | R)$, or continues otherwise. It is common practice in OLTR to consider three instantiations of the SDBN: a \textit{perfect} user with very reliable feedback, a \textit{navigational} user searching for reasonably relevant documents, and an \textit{informational} user with the noisiest click feedback among the three instantiations. The values used for $P(click = 1 | R)$ and $P(stop = 1 | R)$ for SDBN in all considered datasets are reported in Table~\ref{tbl:sdbn}. In the study of click preference skewness (Section~\ref{noniid:cps}), we also adopt the \emph{Position-Based Model} (PBM)~\cite{craswell2008experimental}, which considers the probability that the user examines a document is highly relevant to the rank or position of the document on a SERP (search engine result page). The probability of examining typically decreases with the rank and the degree of decreasing is parameterized by $\eta$. The larger value of $\eta$ represents higher decreasing degree, which brings about higher position bias.

\subsubsection{Evaluation metric} 
\label{chap2-metric}
The evaluation practice from previous OLTR work consists of measuring the rankers' offline and online performance by using normalised Discounted Cumulative Gain (nDCG)~\cite{oosterhuis2018differentiable,jia2022learning}.

As each SERP is limited to $k$ documents, $nDCG@k$ is used for evaluating the overall offline effectiveness of the ranker. Effectiveness is measured by averaging the $nDCG$ scores of the ranker on the queries in the held-out test dataset with the actual relevance label. This performance value is recorded after each update of the ranker and the final convergence performance is also recorded.

To measure online performance, it is common practice to compute the cumulative discounted $nDCG@k$ for the rankings displayed during the training phase~\cite{hofmann2013fast}. For $t$-th query in a sequence of $T$ queries, with $R^t$ representing the ranking list displayed to the user, online performance is computed as:
\begin{equation}
	\label{eq:onlinemetric}
	Online\_Performance = \sum_{t = 1}^{T}nDCG(R^t)\cdot\gamma^{t-1} \text{.}
\end{equation}
This measure indicates the quality of the user experience during training; the discount factor $\gamma$ puts more importance to the interactions in the early phases of training. Following previous work~\cite{oosterhuis2018differentiable}, in this thesis, we set $\gamma = 0.9995$.

In the common evaluation practice in OLTR, each experiment is repeated multiple times, spread evenly over all dataset training folds. All evaluation results are averaged and statistical significant differences between system pairs are evaluated using two-tailed Student's t-test with Bonferroni correction.

\section{Federated Learning}

\subsection{Federated learning with privacy protection}
Federated Learning~\cite{mcmahan2017communication} considers a  machine learning setting where several data owners (clients) collaboratively train a model without sharing the data. To this aim, a coordinator (central server) is required and each participating client sends what is essential for the training (for example, the local gradient) to the central server. The Federated Averaging (FedAvg) algorithm~\cite{mcmahan2017communication} relies on local stochastic gradient descent (SGD), which is computed by each client, while the centralised server performs the global model update by weighted averaging the client's updates.

%\textcolor{red}{Some common methods used for privacy protection in Federated Learning include Multi-Party Computation (MPC)~\cite{}, Homomorphic Encryption (HE)~\cite{} and Differential Privacy (DP)~\cite{}. Secure MPC~\cite{} is a class of cryptography technique for many participants to jointly train a model in a peer-to-peer topology. DP~\cite{} protects privacy by adding random noise to the uploaded data, which has an impact on model accuracy.} 

Differential privacy~\cite{dwork2006calibrating,dwork2014algorithmic} is a method often used to ensure user privacy preservation in federated learning (the alternative is the encryption of the messages). We shall review the notion of  $\epsilon$-differential privacy in Section~\ref{dp}. Abadi et al.~\cite{abadi2016deep} have developed the algorithmic techniques for combining SGD-based deep learning methods with differential privacy. Differential privacy has also been added to FedAvg algorithm to provide user-level differential privacy guarantees for language models~\cite{mcmahan2018learning}, as for these models it has been shown that the gradients or shared model's weights can reveal sensitive information regarding the clients' data~\cite{fredrikson2015model, shokri2017membership, DBLP:conf/nips/GeipingBD020}. 
%Recently, a novel notion called \emph{federated f-differential privacy} is introduced by~\cite{zheng2021federated} based on the Gaussian differential privacy framework, which tackles record-level privacy guarantee in federated learning.

\subsection{Federated OLTR}
OLTR has been primarily investigated in a centralized paradigm, where a central server holds the data to be searched and collects the users' search interactions (e.g., queries, clicks). The training of the ranker is also conducted by the server. This centralized paradigm is not suited to a privacy-preserving situation where each client does not want to share the data on which it wants to search, along with queries and other search interactions.

The Federated OLTR with Evolutionary Strategies method (FOLtR-ES)~\cite{kharitonov2019federated} has been the first method that was proposed to address the above privacy-preserving issues in OLTR. This method extends the OLTR optimization scenario to the Federated SGD~\cite{mcmahan2017communication} and uses Evolution Strategies as the optimization method~\cite{salimans2017evolution}. To further protect the gradient updates from a malicious attacker, FOLtR-ES also includes a privatization procedure, which leverages the theory of $\epsilon$-local differential privacy~\cite{dwork2014algorithmic}. FOLtR-ES is shown to perform well on small-scale datasets (MQ2007) for the MaxRR metric, although performance degrades when privacy preservation is required~\cite{kharitonov2019federated}. However, the effectiveness of FOLtR-ES is not consistent across other, larger-scale datasets nor when typical OLTR metrics are considered (nDCG)~\cite{wang2021federated}.

%\textcolor{red}{
	%	Attacks to gradient updates: privacy~\cite{geiping2020inverting}, integrity~\cite{tolpegin2020data} 
	%}

%\textcolor{red}{
	%	Federated learning in search (Suggested in SIGIR review)%\cite{ahmad2018intent}, \cite{yu2018hide}, 
	%	%\cite{toledo2016lower}, 
	%	\cite{hartmann2019federated}, \cite{yao2021fedps}
	%}

\subsection{Federated Learning in other Information Retrieval Tasks}

%\textcolor{red}{todo: check section 2.2 in the non-iid paper (chapter 4)}

Aside from its usage in online learning to rank,  recent works have applied federated learning in other information retrieval contexts~\cite{pinelli2023flirt}. Zong et al.~\cite{zong2021fedcmr} provide a solution for cross-modal retrieval in a distributed data storage scenario, which uses federated learning to reduce the potential privacy risks and the high maintenance costs encountered when dealing with large amount of training data. Wang et al.~\cite{wang2021efficient} study learning to rank (but not in the online setting we consider in this thesis) in a cross-silo federated learning setting; this work is aimed at helping companies that have access to limited labelled data to collaboratively build a document retrieval system, in an efficient way.
Hartmann et al.~\cite{hartmann2019federated} use federated learning to improve the ranking of suggestions
in the Firefox URL bar, so that the training of the ranker on user interactions is performed in a privacy-preserving way; they show that this federated approach improves on the suggestions produced by the previously employed heuristics in Firefox.
Yang et al. ~\cite{yang2018applied} describe the use of federated learning for search query suggestions in the Google Virtual Keyboard (GBoard) product. Here, a baseline model is used to identify relevant query suggestions given a user query; candidate suggestions are then filtered using a triggering model learnt using federated learning. 
Aside from the previous examples, federated learning has also seen adoption in the area of personalised search~\cite{ghorab2013personalised}, which aims  to return search results that cater to the specific user's interests. Yao et al.~\cite{yao2021fedps} recently proposed a privacy protection enhanced personalized search framework which adapts federated learning to the state-of-the-art personalized search model.

%\textcolor{red}{Inspired by the recent success of federated learning, it has started being used in some related works for Information Retrieval tasks. Zong et al.~\cite{zong2021fedcmr} provide a solution for Cross-modal retrieval in the distributed data storage scenario, which uses federated learning to reduce the potential privacy risks and high maintenance costs for the large amount of training data. Wang et al.~\cite{wang2021efficient} study learning to rank (LTR) in a cross-silo federated learning (FL) setting to help enterprises with limited data to build the document retrieval systems collaboratively in an efficient way.
	%In personalized search which aims to return the ranking results that cater to user's interest, feature-based~\cite{carman2010towards, bennett2012modeling, harvey2013building} and deep learning-based~\cite{song2014adapting, ge2018personalizing, yao2020rlper} methods are widely developed. However, the privacy issue in personalized search is continuously overlooked, especially for user's query logs which are collected by the central server for personalized ranking model training. To tackle this issue, Yao et al.~\cite{yao2021fedps} recently propose a privacy protection enhanced personalized search framework which adapts federated learning to the state-of-the-art personalized search model.}

\subsection{Federated Learning with non-IID data}
Zhu et al. have compiled a comprehensive survey on the impact of non-IID data on federated learning ~\cite{zhu2021federated}, also reviewing the current research on handling these challenges.  
Early work from Zhao et al.~\cite{zhao2018federated} shows a deterioration of the accuracy of federated learning if non-IID or heterogeneous data is present; they also provide a solution to this problem by creating a small subset of globally shared data between all clients (local devices).
Li et al.~\cite{li2020convergence} analyse the convergence of the federated learning algorithm \verb|FedAvg|~\cite{mcmahan2017communication} (which is a component of the FOLTR method we rely upon for investigation~\cite{wang2022non}) on non-IID data and empirically show that data heterogeneity slows down the convergence. This raised attention to the presence of non-IID data in federated learning. 

Generally speaking, existing approaches for handling non-IID issues in federated learning can be classified into three categories: data-based approaches, algorithm-based approaches, and system-based approaches~\cite{zhu2021federated}.
Data sharing~\cite{zhao2018federated} and data augmentation~\cite{duan2019astraea} are two kinds of typical data-based approaches. While they achieve state-of-the-art performance, they fundamentally conflict with the objective of federated learning: that of not sharing data across clients. This is because, for example, methods such as data sharing require a subset of private data to be shared across all clients. While proposals have been made to use synthetic, rather than real, data for the data sharing mechanism~\cite{wang2021non} it is unclear (1) what the effectiveness loss of the sharing of synthetic data in place of real data is, and (2) whether the sharing of synthetic data could still jeopardise privacy as this synthetic data is typically generated from real data, and thus analysis of the synthetic data may reveal key aspects of and information contained in the real data.
Algorithm-based approaches mainly focus on personalisation methods like local fine-tuning of a neural model~\cite{wang2019federated} and Personalized FedAvg (Per-FedAvg)~\cite{fallah2020personalized} -- which are both limited mainly to neural models -- or the casting of the federated learning process into a multi-task learning problem~\cite{smith2017federated}.
System based approaches adopt clustering~\cite{sattler2020clustered} and tree-based structure~\cite{ghosh2020efficient} to deal with non-IID data.
Limitations exist among all proposed approaches, and this is still a much unexplored line of research.

\subsection{Poisoning Attacks on Federated Learning}

Poisoning attacks on federated learning systems aim to compromise the integrity of the system's global model. Poisoning attacks can be grouped according to the goals of the attack into two categories: untargeted poisoning attacks, and targeted poisoning attacks (also known as backdoor attacks).

Targeted poisoning attacks aim to manipulate a global model according to the attacker's objectives, such as misclassifying a group of data with certain features to a label chosen by the attacker, while maintaining normal model effectiveness under other conditions. This is accomplished through backdoor attacks~\cite{bagdasaryan2020backdoor, bhagoji2019analyzing}, which are designed to allow the targeted manipulations to transpire stealthily and without detection. %, while still providing high effectiveness for users.

%Targeted poisoning attacks intend to force the global model to perform in an intended way which the adversary wish the model to perform under certain adversarial condition (for example, misclassify a certain group of data with specific features to an attacker-assigned label) while at the same time, maintain the overall performance of other conditions. To achieve so, backdoor attacks~\cite{bagdasaryan2020backdoor, bhagoji2019analyzing} are designed to promote the targeted attacking stealthily while providing high-quality performance to benign users.

In contrast, untargeted poisoning attacks (also known as Byzantine failures~\cite{lamport1982byzantine, blanchard2017machine, yin2018byzantine, fang2020local, shejwalkar2021manipulating}) aim to decrease the overall effectiveness of the global model indiscriminately for all users and data groups. Current untargeted poisoning methods can be divided into two categories: data poisoning and model poisoning. 
Label flipping~\cite{biggio2012poisoning} is a representative data poisoning method:  the labels of honest training data are changed without altering their features.
Model poisoning, on the other hand, directly affects the local model updates before they are sent to the centralized server. For example, Baruch et al.~\cite{baruch2019little} poisons the local model updates through the addition of noise computed from the variance between the before-attack model updates, while Fang et al.~\cite{fang2020local}'s attacks are optimized to undermine specific robust aggregation rules.

%Instead, for attackers who conduct untargeted poisoning (also known as Byzantine failures in some literatures~\cite{lamport1982byzantine, blanchard2017machine, yin2018byzantine, fang2020local, shejwalkar2021manipulating}), their aim is to degrade the overall performance (or effectiveness) of global model on the held-out test set. Current untargeted poisoning methods can be grouped as two types: data poisoning, model poisoning. 
%One typical data poisoning strategy is label flipping~\cite{biggio2012poisoning, tolpegin2020data}, in which the labels of honest training data of one class are flipped to another while the corresponding features maintain unchanged. Unlike data poisoning, model poisoning attacks directly compromise the local model updates before sending them to the centralised server. Baruch et al.~\cite{baruch2019little} directly poisons the local updates through adding noise computed from variance between before-attack model updates. Fang et al.~\cite{fang2020local} propose model poisoning strategies tailored to specific robust aggregation rules through formulating attacks as optimization problems. 

Among untargeted poisoning attacks on federated learning systems, model poisoning methods have been found to be the most successful~\cite{bagdasaryan2020backdoor}. In particular, data poisoning attacks have limited success when Byzantine-robust defense aggregation rules are in use~\cite{fang2020local}; we introduce these defense methods in Section~\ref{defense}.
Furthermore, most data poisoning attacks assume that the attacker has prior knowledge about the entire training dataset, which is often unrealistic in practice.

In this thesis, we focus on untargeted poisoning attacks, delving into the effectiveness of both data poisoning and model poisoning methods. These attack methods are studied within the framework of a FOLTR system based on FPDGD, with and without the integration of defense countermeasures. 
%Along with attack methods, we also study the effectiveness ofunder both defensive and non-defensive settings tailored to federated online learning to rank system. 

%Previous studies have shown the evidence that model poisoning is more effective than data poisoning in FL~\cite{bagdasaryan2020backdoor}. Also, most data poisoning attacks assume that the adversary has prior knowledge about the entire training dataset, which is unfeasible in FL. Existing work has also shown that data poisoning attacks have limited success to attack Federated Learning under Byzantine-robust defensive aggregation rules~\cite{fang2020local} which we introduce in Section~\ref{defense}.

\subsection{Machine Unlearning and Federated Unlearning}

Machine Unlearning pertains to the removal of any evidence of a chosen data point from the model, a process commonly known as selective amnesia. Except ensuring the removal of certain data points from the model being the primary objective, the unlearning procedure should also do not affect the model's effectiveness. Machine unlearning has been explored in both centralised~\cite{DBLP:conf/sp/CaoY15,bourtoule2021machine} and federated settings~\cite{liu2021federaser,halimi2022federated,wang2022federated,che2023fast}.

\subsubsection{Methods in Machine Unlearning}
Methods in machine unlearning can be broadly classified into two families: exact unlearning and approximate unlearning. Exact unlearning methods are designed to provide a theoretical guarantee that the methods can completely remove the influence of the data to be forgotten~\cite{DBLP:journals/ml/BaumhauerSZ22,DBLP:conf/icml/BrophyL21,DBLP:journals/cluster/ChenXXZ19,DBLP:conf/nips/GinartGVZ19,mahadevan2021certifiable}; but a limitation of these methods is that they can only be applied to simple machine learning models.
%Some works~\cite{DBLP:journals/ml/BaumhauerSZ22,DBLP:conf/icml/BrophyL21,DBLP:journals/cluster/ChenXXZ19,DBLP:conf/nips/GinartGVZ19} studied exact unlearning methods which are designed to theoretically guarantee that models can totally remove the influence of removed data but can only be applied to simple methods.
Approximate unlearning methods, on the other hand,  are characterized by higher efficiency, which is achieved by relying on specific assumptions regarding the accessibility of training information, and by permitting a certain amount of reduction in the model's effectiveness~\cite{DBLP:journals/access/AldaghriMB21,DBLP:conf/ccs/Chen000H022,chundawat2023zero}.
These methods can be used on more complex machine learning models (e.g., deep neural networks). 
%On the contrary, some other works~\cite{DBLP:journals/access/AldaghriMB21,DBLP:conf/ccs/Chen000H022} investigated approximate unlearning, which can achieve unlearning in more complex models like neural networks.
%
The existing unlearning methods for federated learning belong to the approximate unlearning family~\cite{liu2021federaser,DBLP:journals/corr/abs-2003-10933,DBLP:conf/infocom/LiuXYWL22}.
However, most of these methods are applied to classification tasks and no previous work considers either ranking or FOLTR systems.
%Our method is migrated from FedEraser~\cite{liu2021federaser}.

%\todo{split the section or not?}
\subsubsection{Evaluation of Machine Unlearning}
%\textbf{Evaluation of Machine Unlearning.}
A key challenge posed by the task of unlearning in the federated learning context is how to evaluate whether an unlearning method has successfully removed the contributions from the client that requested to leave the federation. This evaluation is at times termed as unlearning verification~\cite{nguyen2022survey}, which specifically aims to certify that the unlearned model is unrelated to the data that needed to be removed. 
Due to the stochastic nature of the training for many machine learning models, it is difficult to distinguish the individual clients' models and their unlearned counterparts after a certain group of data is removed. 

To evaluate the effectiveness of unlearning, a group of methods leverage membership inference attacks~\cite{liu2021federaser,chen2021machine,hu2022membership}, i.e. a kind of attack method that predicts whether a data item belongs to the training data of a certain model~\cite{shokri2017membership}: it is then straightforward to adapt this type of attacks to verify if the unlearned samples participated in the unlearning process. However, conducting successful membership inference attacks needs subtle design and training of the inference model. The effectiveness of such attacks on FOLTR is unknown as it is an unexplored area. Thus, we do not consider this type of verification in this thesis.

Another type of methods is inspired from the idea of poisoning attacks. These methods add arbitrary noise~\cite{yuan2023federated} or backdoor triggers~\cite{gao2024verifi,sommer2022athena,wu2022federated,halimi2022federated} to the data that is needed to be deleted, with the aim of manipulating the effectiveness of the trained model. After the unlearning of poisoned datasets has taken place, the impact from the arbitrary poisoning or backdoor triggers should be reduced in the unlearned model: the extent of this reduction (and thus model gains) determines the extent of the success of unlearning.

%
%\todo{Besides methods that studies conducting unlearning effectively, evaluation on the success of unlearning is also important. Due to the stochastic factors in the training of machine learning models, it is difficult to distinguish the model and its unlearned counterparts after a certain group of data is removed. The evaluation process is also termed as unlearning verification~\cite{nguyen2022survey}, with the aim of certifying that the unlearn models is unrelated to the data to be removed. To evaluate the effectiveness of unlearning, a group of methods leverage membership inference attacks~\cite{liu2021federaser, chen2021machine, hu2022membership}, i.e. a kind of attack methods that predict whether a data item belongs to the training data of a certain model~\cite{shokri2017membership}. Thus, it is straightforward to adapt this type of attacks to verify if the unlearned samples participate in the unlearning process. Another type of methods is inspired from the idea of poisoning attacks. With the aim of manipulating the performance of the trained model, arbitrary noise~\cite{yuan2023federated} or backdoor triggers~\cite{gao2022verifi, sommer2022athena, wu2022federated, halimi2022federated} are added to the data that is to be deleted. After unlearning the poisoned datasets, the impact from arbitrary poisoning or backdoor triggers should be reduced in the unlearned model.} 

\section{Chapter Summary}

In this chapter, we introduce the basic concepts and related works for this thesis. In particular, we give an overview of existing methods in Online Learning to Rank and the common evaluation practice used in OLTR. We introduce a group of related works in Federated Learning with focus on privacy-protection techniques, dealing with non-IID data, poisoning attacks and defense, and unlearning methods. Furthermore, we summarise federated learning applications in other Information Retrieval and Recommendation tasks. In the next chapter, we start introducing our main work, findings and contributions for this research.

% HOW TO ADD ADDITIONAL CHAPTERS
% Step One: Add a new folder called "ChapterX" (X being the chapter number).
% Step Two: Within the folder add a new .tex file by clicking the "New File" button in the Overleaf Menu. Rename the file to a title of your choice.
% Step Three: Copy the Chapter 2 headline and "\input" command located above and insert it below Chapter 2.
% Step Four: Rename the headline to your specific chapter number, change the input command to include the name of the folder you created and the name of the file you created.
% Repeat this process for every chapter.

%%%% Main Chapters

%CHAPTER 3
%If you are presenting work which has been previously published, acknowledge this here.
% ***************************************************
% How to introduce a previously published chapter
% ***************************************************
%This is an example of how you might introduce a chapter that has been published previously. 
\cleartoevenpage
\pagestyle{empty}	
%Use this command (above) to suppress the header from the preceding chapter.

\noindent
The following publication has been incorporated as Chapter~\ref{Chap:fpdgd}.

\noindent
1.~\cite{wang2021federated} \textbf{Shuyi Wang}, Shengyao Zhuang, and Guido Zuccon, \href{https://doi.org/10.1007/978-3-030-72240-1_10}{Federated Online Learning to Rank with Evolution Strategies: A Reproducibility Study}, \textit{In European Conference on Information Retrieval (ECIR)}, pages 134-149, 2021

\noindent
2.~\cite{wang2021effective} \textbf{Shuyi Wang}, Bing Liu, Shengyao Zhuang, and Guido Zuccon, \href{https://doi.org/10.1145/3471158.3472236}{Effective and Privacy-preserving Federated Online Learning to Rank}, \textit{In Proceedings of the 2021 ACM SIGIR International Conference on Theory of Information Retrieval (ICTIR)}, pages 3–12, 2021

%\begin{table}[h]
%	\begin{center}
%	\begin{tabular}{|c|l|l|}
%		\hline
%		Contributor & Statement of contribution & \% \\
%		\hline
%		\textbf{Your Name}				& writing of text 					& 70\\
%															& proof-reading							& 60 \\
%															& theoretical derivations 	& 70\\
%															& numerical calculations 		& 100\\
%															& preparation of figures 		& 80 \\
%															& initial concept						& 10 \\
%		\hline
%		Co-author 1								& writing of text 					& 20\\
%															& proof-reading							& 10 \\
%															& supervision, guidance 		& 20\\
%															& theoretical derivations 	& 10\\
%															& preparation of figures 		& 20 \\
%															& initial concept						& 10 \\
%		\hline
%		Final Author							& writing of text 					& 10\\
%															& proof-reading							& 30 \\
%															& supervision, guidance 		& 80 \\
%															& theoretical derivations 	& 20 \\
%															& preparation of figures 		& 10 \\
%															& initial concept						& 80 \\
%		\hline
%	\end{tabular}
%	\end{center}
%\end{table}
%
%If your task breakdown requires further clarification, do so here. Do not exceed a single page.

% ***************************************************
% Example of an internal chapter
% ***************************************************
%This is an internal chapter of the thesis.
%If you have a long title, you can supply an abbreviated version to print in the Table of Contents using the optional argument to the \chapter command.
\chapter[Effective and Privacy-Preserving FOLTR]{Effective and Privacy-Preserving Federated Online Learning to Rank}
\label{Chap:fpdgd}	%CREATE YOUR OWN LABEL.
\pagestyle{headings}

% ********* Enter your text below this line: ********

This chapter considers the problem of devising effective online learning to rank (OLTR) methods embedded in a federated system. Specifically, we investigate the following research questions:

\begin{enumerate}[label=\textbf{RQ1:}, leftmargin=*]
	\item \bfseries How to design an effective and generalisable FOLTR method that can achieve competitive performance compared to state-of-the-art OLTR methods? \itshape
	\begin{enumerate}[label=RQ1.\arabic*:, leftmargin=*]
		\item Does the existing FOLTR method (FOLtR-ES) generalize to other commonly-used datasets and metrics in OLTR?
		\item How to integrate the state-of-the-art OLTR method with the widely-used, easy-to-implement FedAvg framework while further protecting user privacy?
	\end{enumerate}
\end{enumerate}

We investigate this problem into two parts. The first part correspond to Section~\ref{rq1.1} where we conduct reproduction of the previous FOLTR method with the focus on understanding its limitation to answer RQ1.1. The second part corresponds to Section~\ref{rq1.2} where we propose an effective FOLTR method with further privacy-preserving plugin and term it as FPDGD. Following common experimental analysis in OLTR and FL, we demonstrate the effectiveness improvement of FPDGD compared to the FOLtR-ES baseline, which address RQ1.2.

\section{Limitations of FOLtR-ES}
\label{rq1.1}

	%\section{Introduction}

In the context of few attention on guaranteeing user's privacy in OLTR, Kharitonov proposed FOLtR-ES~\cite{kharitonov2019federated} as the first attempt in the area of FOLTR. The method relies on the common federated learning paradigm~\cite{yang2019federated}, in which data (collection, queries, interactions) is maintained at each client's side along with a copy of the ranker, and updates to the rankers that are learned from the interaction on the client side are shared to the central server, which is responsible for aggregating the update signal from clients and propagate the aggregated ranker update. In this specific case, all users observe and act on the same feature space; each user however retains control of their own data, which includes the collection, the queries and the interactions. FOLtR-ES uses evolutionary strategies akin to those in genetic algorithms to make client rankers explore the feature space, and a parametric privacy preserving mechanism to further anonymise the feedback signal that is shared by clients to the central server.  

However, the original research study that introduced this method only evaluated it on a small Learning to Rank (LTR) dataset and with no conformity with respect to current OLTR evaluation practice. It further did not explore specific parameters of the method, such as the number of clients involved in the federated learning process, and did not compare FOLtR-ES with the current state-of-the-art OLTR method.

To remedy the aforementioned gap, in this section, we replicate and then reproduce the experiments from the original work of Kharitonov~\cite{kharitonov2019federated}, investigating the effect different configurations of that federated OLTR method have on effectiveness and user experience, extending and generalising its evaluation to different settings commonly used in OLTR and to different collections. Specifically, we address the following research questions:

\begin{enumerate}[label=\textbf{RQ1.1.\arabic*:}, leftmargin=*]
	\item \textit{Does the performance of FOLtR-ES generalise beyond the MQ2007/2008 datasets?} The original method was only evaluated using MQ2007/2008~\cite{DBLP:journals/corr/QinL13}, while current OLTR practice is to use larger datasets that are feature richer and that contain typical web results.
	\item \textit{How does the number of clients involved in FOLtR-ES affect its performance?} FOLtR-ES was previously evaluated using a set number of clients involved in the federated OLTR process ($n=2,000$), and it was left unclear whether considering more or less client would impact performance.
	\item \textit{How does FOLtR-ES compare with current state-of-the-art OLTR methods?} Compared to OLTR methods, FOLtR-ES preserves user privacy, but it is unclear to what expense in terms of search performance: the original work compared FOLtR-ES to rankers in non-federated settings, but the rankers used in there were not the current state-of-the-art in OLTR.
	\item \textit{How does FOLtR-ES performance generalise to the evaluation settings commonly used for OLTR evaluation, i.e. measuring offline and online performance, with respect to nDCG and with relevance labels?} The original evaluation of FOLtR-ES considered an unusual setting for OLTR, consisting of using MaxRR~\cite{radlinski2008learning} as evaluation measure in place of nDCG, computed on simulated clicks instead of on relevance labels.
\end{enumerate}
	
\subsection{Federated OLTR with Evolution Strategies}\label{sec:method}

We provide a brief overview of the FOLtR-ES method, which extends online LTR to federated learning; this is done by exploiting evolution strategies optimization, a widely used paradigm in Reinforcement Learning. 
The FOLtR-ES method consists of three parts. First, it casts the ranking problem into the federated learning optimization setting. Second, it uses evolution strategies to estimate gradients of the rankers. Finally, it introduces a privatization procedure to further protect users' privacy.

\subsubsection{Federated Learning Optimization Setting}
The federated learning optimization setting consists in turn of several steps, and assumes the presence of a central server and a number of distributed clients. First, a client downloads the most recently updated ranker from the server. Afterwards, the client observes $B$ user interactions (search queries and examination of SERPs) which are served by the client's ranker. The performance metrics of these interactions are averaged by the client and a privatized message is sent to the centralized server. After receiving messages from $N$ clients, the server combines them to estimate a single gradient $g$ and performs an optimization step to update the current ranker. Finally, the clients download the newly updated ranker from the server.

\subsubsection{Gradient Estimation} \label{sec-gradient-est}
The method assumes that the ranker comes from a parametric family indexed by vector $\theta \in R^{n}$. Each time a user $u$ has an interaction $a$, the ranking quality is measured; this is denoted as $f$. The goal of optimization is to find the vector $\theta^*$ that can maximize the mean of the metric $f$ across all interactions $a$ from all users $u$:
\begin{equation}
	\theta^{*}=\arg \max _{\theta} F(\theta)=\arg \max _{\theta} \mathbb{E}_{u} \mathbb{E}_{a \mid u, \theta} f(a ; \theta, u) \label{eq-theta}
\end{equation}

Using Evolution Strategies (ES)~\cite{salimans2017evolution}, FOLtR-ES considers a population of parameter vectors which follow the distribution with a density function $p_{\phi}(\theta)$. The objective aims to find the distribution parameter $\phi$ that can maximize the expectation of the metric across the population:
\begin{equation}
	 \mathbb{E}_{\theta\sim p_{\phi}(\theta)}~[F(\theta)] \label{eq-expectation}
\end{equation}

The gradient $g$ of the expectation of the metric across the population (Equation~\ref{eq-expectation}) is obtained in a manner similar to REINFORCE~\cite{williams1992simple}:
\begin{equation}
	\begin{aligned}
		g &=\nabla_{\phi} \mathbb{E}_{\theta}[F(\theta)]=\nabla_{\phi} \int_{\theta} p_{\phi}(\theta) F(\theta) d \theta=\int_{\theta} F(\theta) \nabla_{\phi} p_{\phi}(\theta) d \theta=\\
		&=\int_{\theta} F(\theta) p_{\phi}(\theta)\left(\nabla_{\phi} \log p_{\phi}(\theta)\right) d \theta=\mathbb{E}_{\theta}\left[F(\theta) \cdot \nabla_{\phi} \log p_{\phi}(\theta)\right]
	\end{aligned}
\end{equation}

Following the Evolution Strategies method, FOLtR-ES instantiates the population distribution $p_{\phi}(\theta)$ as an isotropic multivariate Gaussian distribution with mean $\phi$ and fixed diagonal covariance matrix $\sigma^2I$. Thus a simple form of gradient estimation is denoted as:
\begin{equation}
	g=\mathbb{E}_{\theta \sim p_{\phi}(\theta)}\left[F(\theta) \cdot \frac{1}{\sigma^{2}}(\theta-\phi)\right]
\end{equation}

Based on the federated learning optimization setting, $\theta$ is sampled independently on the client side. Combined with the definition of $F(\theta)$ in Equation~\ref{eq-theta}, the gradient can be obtained as:
\begin{equation}
	g=\mathbb{E}_{u} \mathbb{E}_{\theta \sim p_{\phi}(\theta)}\left[\left(\mathbb{E}_{a \mid u, \theta} f(a ; \theta, u)\right) \cdot \frac{1}{\sigma^{2}}(\theta-\phi)\right] \label{eq-gradient}
\end{equation}

To obtain the estimate $\hat{g}$ of $g$ from Equation~\ref{eq-gradient}, $\hat{g} \approx g$, the following steps are followed: (i) each client $u$ randomly generates a pseudo-random seed $s$ and uses the seed to sample a perturbed model $\theta_{s} \sim \mathbb{N}\left(\phi, \sigma^{2} I\right)$, (ii) the average of metric $f$ over $B$ interactions is used to estimate the expected loss $\hat{f} \approx \mathbb{E}_{a \mid u, \theta_{s}} f(a;\theta_s, u) $ from Equation~\ref{eq-gradient}, (iii) each client communicates the message tuple $(s,\hat{f})$ to the server, (iv) the centralized server computes the estimate $\hat{g}$ of Equation~\ref{eq-gradient} according to all message sent from the $N$ clients.

To reduce the variance of the gradient estimates, means of antithetic variates are used in FOLtR-ES: this is a common ES trick~\cite{salimans2017evolution}. The algorithm of the gradient estimation follows the standard ES practice, except that the random seeds are sampled at the client side.

\subsubsection{Privatization Procedure}
To ensure that the clients' privacy is fully protected, in addition to the federated learning setting, FOLtR-ES also proposes a privatization procedure that introduces privatization noise in the communication between the clients and the server.

Assume that the metric used on the client side is discrete or can be discretized if continuous. Then, the metric takes a finite number ($n$) of values, $f_0, f_1, ..., f_{n-1}$. For each time the client experiences an interaction, the true value of the metric is denoted as $f_0$ and the remaining $n-1$ values are different from $f_0$. When the privatization procedure is used, the true metric value $f_0$ is sent with probability $p$. Otherwise, with probability $1-p$, a randomly selected value $\hat{f}$ out of the remaining $n-1$ values is sent. To ensure the same optimization goal described in Gradient Estimation of Section~\ref{sec-gradient-est}, FOLtR-ES assumes that the probability $p > 1/n$.

Unlike other federated learning methods, FOLtR-ES adopts a strict notion of $\epsilon$-local differential privacy~\cite{kharitonov2019federated}, in which the privacy is considered at the level of the client, rather than of the server. Through the privatization procedure, $\epsilon$-local differential privacy is achieved, and the upper bound of $\epsilon$ is:
\begin{equation}
	\epsilon \leq log\frac{p(n-1)}{1-p} 
\end{equation}

\noindent
This means that, thanks to the privatization scheme, at least $log[p(n-1)/(1-p)]$-local differential privacy can be guaranteed. At the same time, any $\epsilon$-local differential private mechanism also can obtain $\epsilon$-differential privacy~\cite{dwork2014algorithmic}.
	\subsection{Experimental Settings}

\subsubsection{Datasets}
The original work of Kharitonov~\cite{kharitonov2019federated} conducted experiments on the MQ2007 and MQ2008 learning to rank datasets~\cite{DBLP:journals/corr/QinL13}, which are arguably small and outdated. In our work, we instead consider more recent and larger datasets: MSLR-WEB10K~\cite{DBLP:journals/corr/QinL13} and Yahoo! Webscope~\cite{DBLP:journals/jmlr/ChapelleC11}, which are commonly-used in offline and online learning to rank~\cite{zhuang2020counterfactual,hofmann2013reusing,jagerman2019model,oosterhuis2018differentiable}. Compared to MQ2007/2008, both MSLR-WEB10K and Yahoo! contain many more queries and corresponding candidate documents. In addition, both datasets have much richer and numerous features. For direct comparison with the original FOLtR-ES work, we also use MQ2007/2008. Detailed descriptions about each dataset are discussed in Section~\ref{chap2-dataset}.

\subsubsection{Simulation}
Akin to previous work~\cite{DBLP:conf/wsdm/SchuthOWR16, hofmann2013reusing, kharitonov2019federated}, we simulate users and their reaction with the search results using labelled offline learning to rank datasets.

For the experiment, we follow the same method used by the original FOLtR-ES work. We sample $B$ queries for each client randomly and use the local perturbed model to rank documents. The length for each ranking list is limited to 10 documents. We simulate users' clicks using the Cascade Click Model (CCM) as discussed in Section~\ref{chap2-click}\footnote{For simulating clicks for the MQ2007/2008, we use the same parameter settings from Table 1 in the original FOLtR-ES paper~\cite{kharitonov2019federated} (shown in Table~\ref{tbl:chap2:click1}): these are partially different from those used for MSLR-WEB10K and Yahoo! because relevance labels in these datasets are five-graded, while they are three-graded in MQ2007/2008.
}. After simulating users clicks, we record the quality metric for each interaction and perform the privatization procedure with probability $p$. Next, we send the averaged metric and pseudo-random seed to optimize the centralized ranker. Finally, each client receives the updated ranker.

\subsubsection{Evaluation metric}
For direct comparison with the original FOLtR-ES work, we use the reciprocal rank of the highest clicked result in each interaction (MaxRR~\cite{radlinski2008learning}). This metric is computed on the clicks produced by the simulated users on the SERPs. 

The evaluation setting above is unusual for OLTR. In RQ1.1.4 of this section, we also consider the more commonly-used normalised Discounted Cumulative Gain (nDCG) metric, as FOLtR-ES is designed to allow optimization based on any absolute measures of ranking quality. We thus record the nDCG@10 values from the relevance labels of the SERP displayed to users during interactions. This is referred to as online nDCG and the scores represent users' satisfaction~\cite{hofmann2013reusing}. For online evaluation, we take the online $nDCG@10$ score averaged across all clients. We also record the nDCG@10 of the globally-updated ranker measured with a held-out test set: this is refer to as offline nDCG. We record the offline $nDCG@10$ score of the global ranker during each federated training update. Detailed implementation of these metrics are introduction in Section~\ref{chap2-metric}.

\subsubsection{FOLtR-ES and Comparison OLTR Methods}
In all experiments, we adopt the same models and optimization steps used by Kharitonov~\cite{kharitonov2019federated}, and rely on the well document implementation made publicly available by the author. The two ranking models used by FOLtR-ES are a linear ranker and a neural ranker with a single hidden layer of size 10. For optimization, we use Adam~\cite{kingma2014adam} with default parameters.

To study how well FOLtR-ES compares with current state-of-the-art OLTR (RQ1.1.3), we implemented the Pairwise Differentiable Gradient Descent (PDGD)~\cite{oosterhuis2018differentiable} (more details of this algorithm is introduced in Section~\ref{sec:PDGD}). Unlike many previous OLTR methods that are designed for linear models, PDGD also provides effective optimization for non-linear models such as neural rankers. During each interaction, a weighted differentiable pairwise loss is constructed in PDGD and the gradient is directly estimated by document pairs preferences inferred from user clicks. PDGD has been empirically found to be significantly better than traditional OLTR methods in terms of final convergence, learning speed and user experience during optimization, making PDGD the current state-of-the-art method for OLTR~\cite{oosterhuis2018differentiable,jagerman2019model,zhuang2020counterfactual}.

	\subsection{Results and Analysis}
\label{sec3.1:results}

\subsubsection{RQ1.1.1: Generalisation of FOLtR-ES performance beyond MQ2007/2008}

\begin{figure}[!htbp]
	\centering
	\begin{subfigure}{1\textwidth}
		\centering
		\includegraphics[width=1\textwidth]{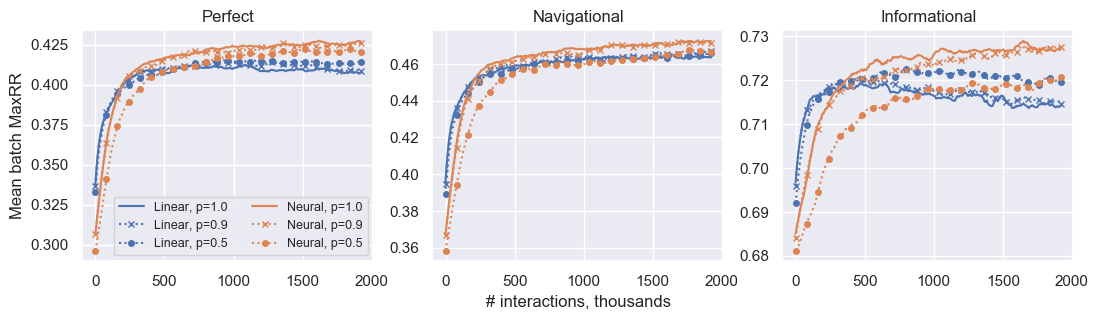}
		\caption{Mean batch MaxRR for MQ2007.}
		\label{fig:mq2007-rq1}
	\end{subfigure}
	\begin{subfigure}{1\textwidth}
		\centering
		\includegraphics[width=1\textwidth]{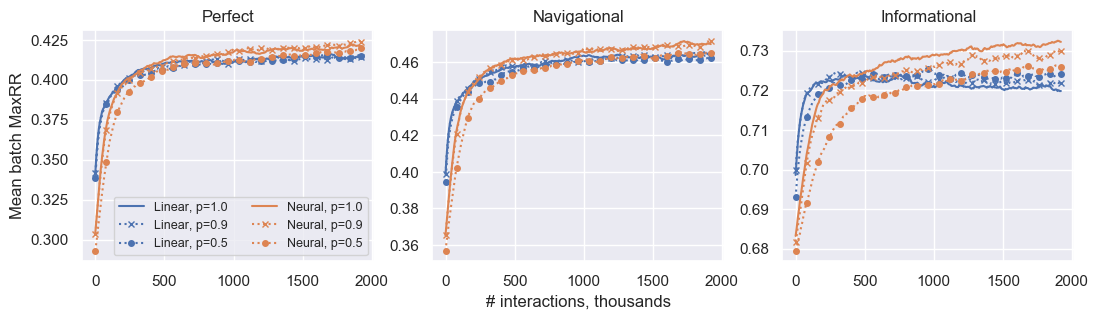}
		\caption{Mean batch MaxRR for MQ2008.}
		\label{fig:mq2008-rq1}
	\end{subfigure}
	\begin{subfigure}{1\textwidth}
		\centering
		\includegraphics[width=1\textwidth]{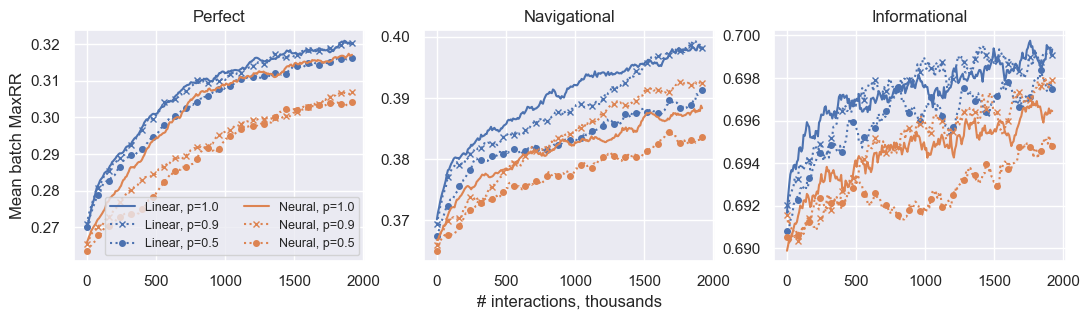}
		\caption{Mean batch MaxRR for MSLR-WEB10K.}
		\label{fig:mslr10k-rq1}
	\end{subfigure}
	\begin{subfigure}{1\textwidth}
		\centering
		\includegraphics[width=1\textwidth]{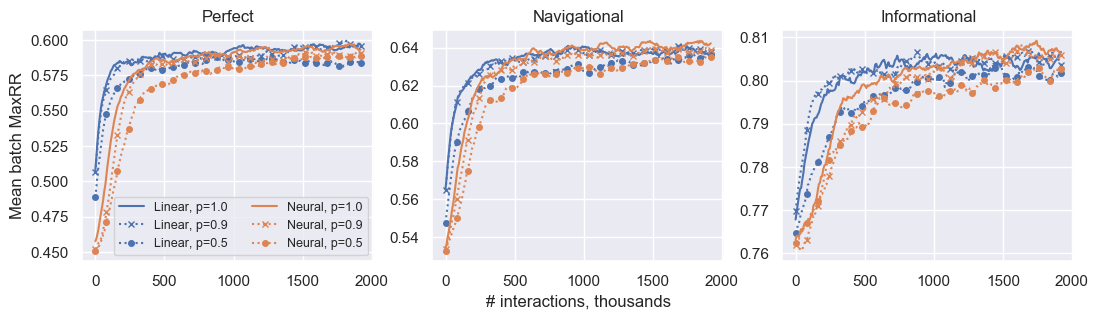}
		\caption{Mean batch MaxRR for Yahoo!.}
		\label{fig:yahoo-rq1}
	\end{subfigure}
	\caption{Results for RQ1.1.1: performance of FOLtR-ES across datasets under three different click models (averaged across all dataset splits). \label{fig:RQ1}}
\end{figure}

For answering RQ1.1.1 we replicate the results obtained by Kharitonov~\cite{kharitonov2019federated} on the MQ2007 and MQ2008 datasets; we then reproduce the experiment on MSLR-WEB10K and Yahoo! datasets, on which FOLtR-ES has not been yet investigated, and we compare the findings across datasets. For these experiments we use antithetic variates, set the number of interactions $B = 4$ and simulate 2,000 clients, use MaxRR as reward signal and for evaluation on clicked items. 

Figure~\ref{fig:mq2007-rq1} reports the results obtained by FOLtR-ES on the MQ2007 dataset with respect to the three click models considered, various settings for the privatization parameter $p$, and the two ranking models (linear and neural). Our results fully replicate those of Kharitonov~\cite{kharitonov2019federated} and indicate the following findings: (1) FOLtR-ES allows for the iterative learning of effective rankers; (2) high values of $p$ (lesser privacy) provide higher effectiveness; 
(3) the neural ranker is more effective than the linear ranker when $p \rightarrow 1$ (small to no privacy), while the linear model is equivalent, or better (for informational clicks) when $p=0.5$. Figure~\ref{fig:mq2008-rq1} reports similar results obtained from MQ2008 dataset.

However, not all these findings are applicable to the results obtained when considering MSLR-WEB10K and Yahoo!, which are displayed in Figures~\ref{fig:mslr10k-rq1} and~\ref{fig:yahoo-rq1}. In particular, we observe that (1) the results for MSLR-WEB10K (and to a lesser extent also for Yahoo!) obtained with the informational click model are very unstable, and, regardless of the click model, FOLtR-ES requires more data than with MQ2007/2008 to arrive at a stable performance, when it does; (2) the neural ranker is less effective than the linear ranker, especially on MSLR-WEB10K. We believe these findings are due to the fact that query-document pairs in MSLR-WEB10K and Yahoo! are represented by a larger number of features than in MQ2007/2008. Thus, more data is required for effective training, especially for the neural model; we also note that FOLtR-ES is largely affected by noisy clicks in MSLR-WEB10K.

\subsubsection{RQ1.1.2: Effect of number of clients on FOLtR-ES}
To answer RQ1.1.2 we vary the number of clients involved in FOLtR-ES; we investigate the values \{50, 1,000, 2,000\}. Kharitonov~\cite{kharitonov2019federated} used 2,000 in the original experiments, and the impact of the number of clients has not been studied. To be able to fairly compare results across number of clients, we fixed the total number of ranker updates to 2,000,000; we also set $B = 4$ and $p=0.9$. 

\begin{figure}[!htbp]
	\centering
	\begin{subfigure}{1\textwidth}
		\centering
		\includegraphics[width=1\textwidth]{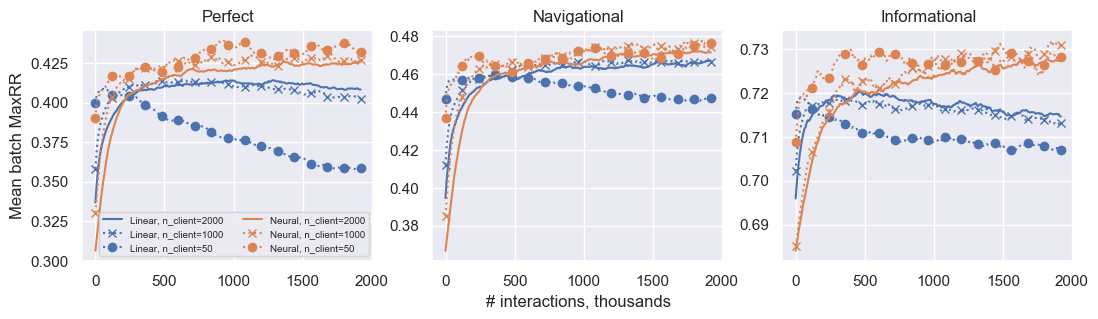}
		\caption{Mean batch MaxRR for MQ2007.}
		\label{fig:mq2007-rq2}
	\end{subfigure}
	\begin{subfigure}{1\textwidth}
		\centering
		\includegraphics[width=1\textwidth]{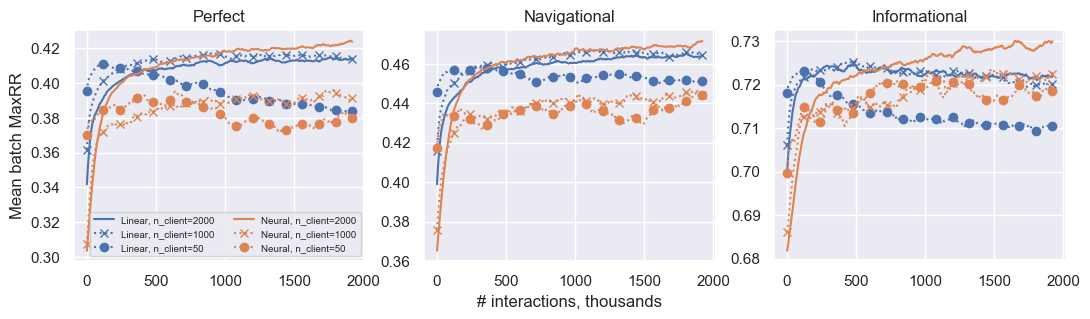}
		\caption{Mean batch MaxRR for MQ2008.}
		\label{fig:mq2008-rq2}
	\end{subfigure}
	\begin{subfigure}{1\textwidth}
		\centering
		\includegraphics[width=1\textwidth]{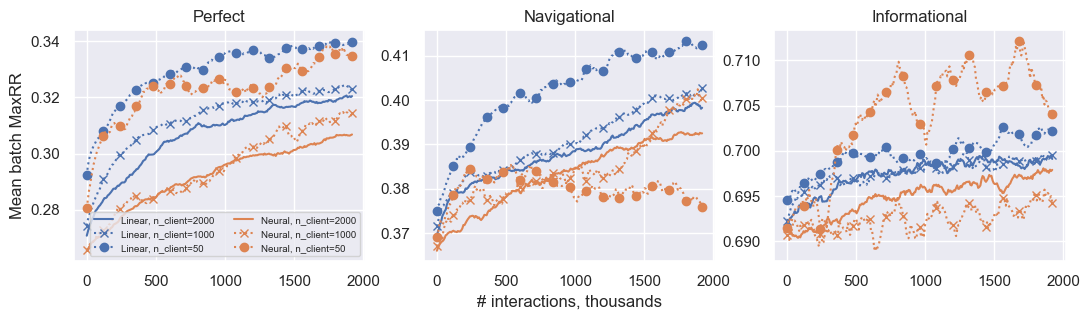}
		\caption{Mean batch MaxRR for MSLR-WEB10K.}
		\label{fig:mslr10k-rq2}
	\end{subfigure}
	\begin{subfigure}{1\textwidth}
		\centering
		\includegraphics[width=1\textwidth]{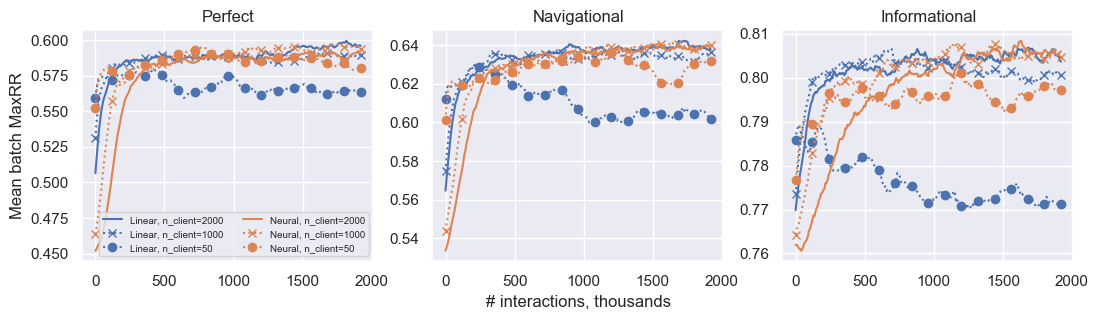}
		\caption{Mean batch MaxRR for Yahoo!.}
		\label{fig:yahoo-rq2}
	\end{subfigure}
	\caption{Results for RQ1.1.2: performance of FOLtR-ES with respect to number of clients (averaged across all dataset splits). \label{fig:RQ2}} 
\end{figure}

The results of these experiments are reported in Figure~\ref{fig:RQ2}, and they are mixed. For MQ2007, the number of clients have little effect on the neural ranker used in FOLtR-ES, although when informational clicks are provided this ranker is less stable, although often more effective, if very few clients (50) are used. Having just 50 clients, instead, severally hits the performance of the linear ranker, when compared with 1,000 or 2,000 clients. For MQ2008, in general, with more clients, the ranker gains higher performance. A degradation trend is also found with only 50 clients for linear ranker on MQ2008. The findings on MSLR-WEB10K, however, are different. In this dataset, a smaller number of clients (50), is generally better than larger numbers, both for linear and neural ranker. An exception to this is when considering navigational clicks: in this case the linear ranker obtains by far the best performance with a small number of clients, but the neural ranker obtains the worst performance. For Yahoo! dataset, the number of clients has little impact on the ranking performance with exceptions on linear ranker with 50 clients. Under the noisiest informational clicks, the growing trend of ranking effectiveness is less stable. This suggest that the number of clients greatly affects FOLtR-ES: but trends are not consistent across click types and datasets.

\begin{figure}[!htbp]
	\centering
	\begin{subfigure}{1\textwidth}
		\centering
		\includegraphics[width=1\textwidth]{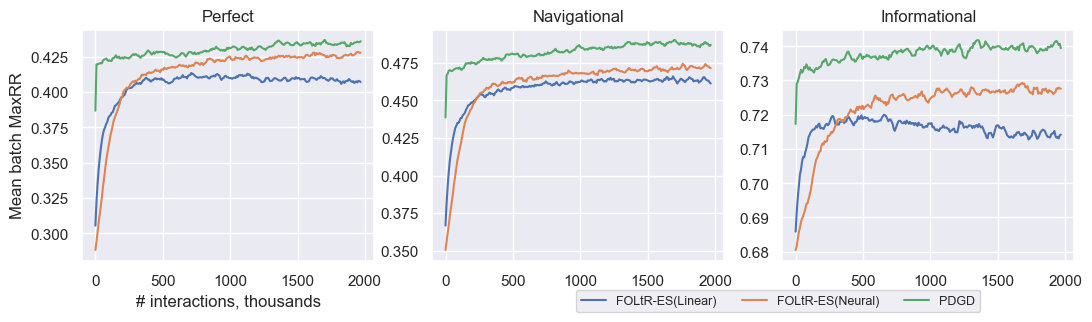}
		\caption{Mean batch MaxRR for MQ2007.}
		\label{fig:mq2007-rq3}
	\end{subfigure}
	\begin{subfigure}{1\textwidth}
		\centering
		\includegraphics[width=1\textwidth]{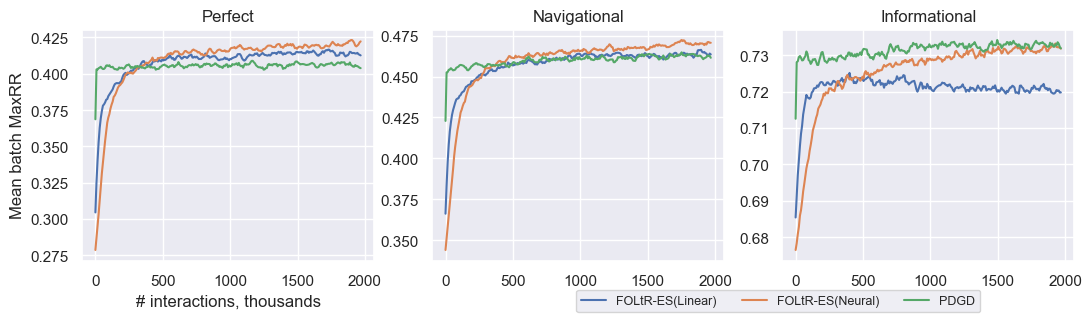}
		\caption{Mean batch MaxRR for MQ2008.}
		\label{fig:mq2008-rq3}
	\end{subfigure}
	\begin{subfigure}{1\textwidth}
		\centering
		\includegraphics[width=1\textwidth]{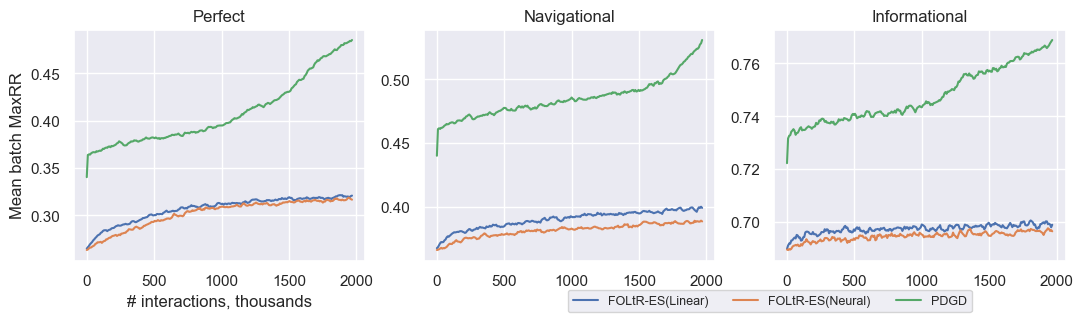}
		\caption{Mean batch MaxRR for MSLR-WEB10K.}
		\label{fig:mslr10k-rq3}
	\end{subfigure}
	\begin{subfigure}{1\textwidth}
		\centering
		\includegraphics[width=1\textwidth]{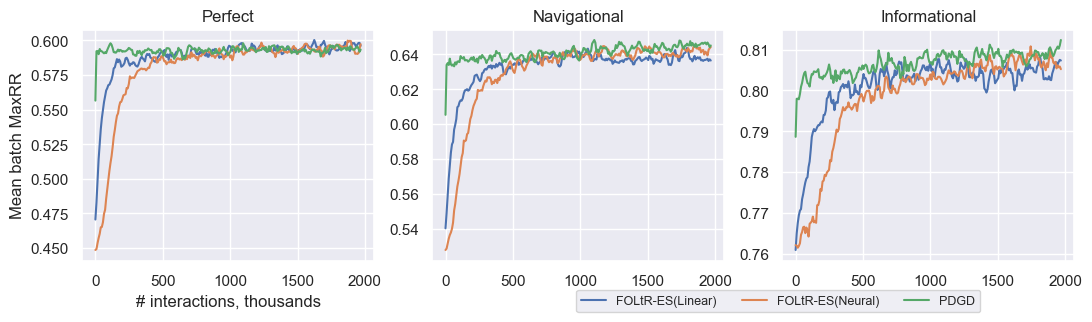}
		\caption{Mean batch MaxRR for Yahoo!.}
		\label{fig:yahoo-rq3}
	\end{subfigure}
	\caption{Results for RQ1.1.3: performance of FOLtR-ES and PDGD across datasets with privatization parameter $p=1$ and 2,000 clients (averaged across all dataset splits). \label{fig:RQ3}} 
\end{figure}

\subsubsection{RQ1.1.3: Comparing FOLtR-ES to state-of-the-art OLTR methods}

The original study of FOLtR-ES did not compare the method with non-federated OLTR approaches. To contextualise the performance of FOLtR-ES and to understand the trade-off between privacy and performance when designing FOLtR-ES, we compare this method with the current state-of-the-art OLTR method, the Pairwise Differentiable Gradient Descent (PDGD)~\cite{oosterhuis2018differentiable}. For fair comparison, we set the privatization parameter $p=1$ (lowest privacy) and the number of clients to 2,000. In addition note that in normal OLTR settings, rankers are updated after each user interaction: however in FOLtR-ES, rankers are updated in small batches. For fair comparison, we adapt PDGD to be updated in batch too. Instead of updating the ranker after each interaction (batch size 1), we accumulate gradients computed on the same batch size as for FOLtR-ES. Specifically, with 2000 clients for FOLtR-ES, the batch size of each update is 8,000 iterations (4 $\times$ 2,000). We then compute the updated gradients for PDGD on 8,000 interactions too. %Note, the number of updates after 2m user interaction now becomes 2m/8000 = 250.

Results are shown in Figure~\ref{fig:RQ3}: for MQ2008 and Yahoo! datasets, PDGD converges faster than FOLtR-ES, which is observed from the higher ranking performance in the early interaction steps. However, the performance gap between PDGD and FOLtR-ES is narrowed after more interactions. For MQ2007 and MSLR-WEB10K datasets, regardless of linear or neural ranker, FOLtR-ES is less effective than PDGD. The gap in performance is greater in larger dataset (i.e., MSLR-WEB10K) than in the smaller MQ2007. This gap becomes even bigger, especially for the first iterations, if the PDGD ranker was updated after each iteration (not shown here), rather than after a batch has been completed. This highlights that FOLtR-ES has the merit of being the first privacy preserving federated OLTR approach available; however, more work is needed to improve the performance of FOLTR based methods so as to close the gap between privacy-oriented approaches and centralise approaches that do not consider user privacy.

\subsubsection{RQ1.1.4: Extending FOLtR-ES evaluation to common OLTR practice}

\begin{figure}[!htbp]
	\centering
	\begin{subfigure}{1\textwidth}
		\centering
		\includegraphics[width=1\textwidth]{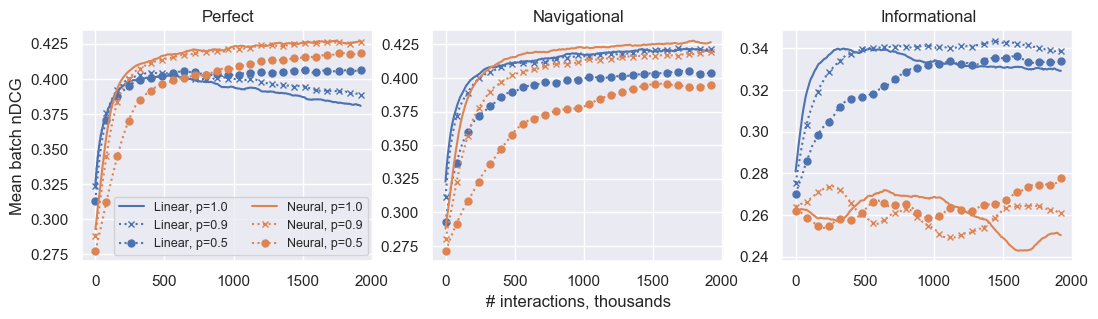}
		\caption{Mean batch nDCG@10 for MQ2007.}
		\label{fig:mq2007-rq4}
	\end{subfigure}
	\begin{subfigure}{1\textwidth}
		\centering
		\includegraphics[width=1\textwidth]{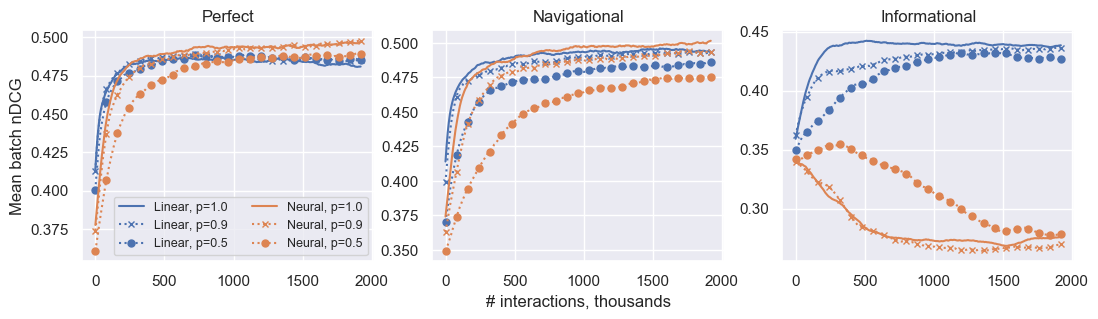}
		\caption{Mean batch nDCG@10 for MQ2008.}
		\label{fig:mq2008-rq4}
	\end{subfigure}
	\begin{subfigure}{1\textwidth}
		\centering
		\includegraphics[width=1\textwidth]{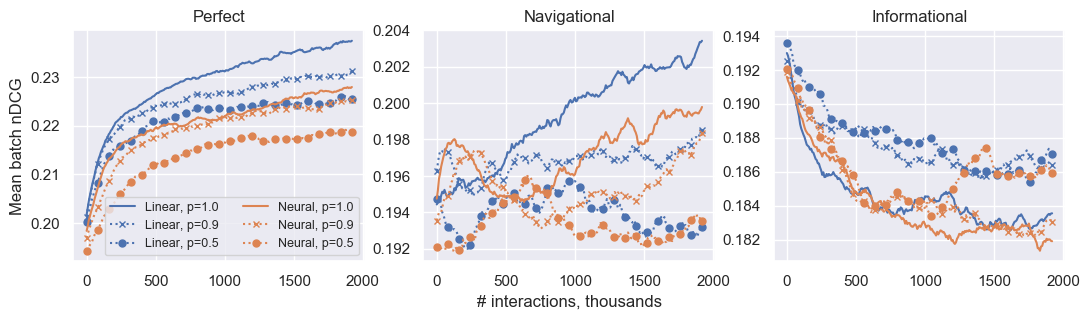}
		\caption{Mean batch nDCG@10 for MSLR-WEB10K.}
		\label{fig:mslr10k-rq4}
	\end{subfigure}
	\begin{subfigure}{1\textwidth}
		\centering
		\includegraphics[width=1\textwidth]{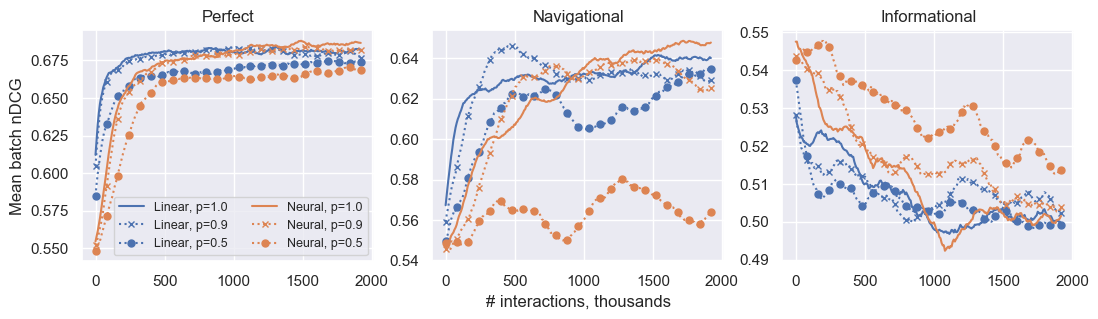}
		\caption{Mean batch nDCG@10 for Yahoo!.}
		\label{fig:yahoo-rq4}
	\end{subfigure}
	\caption{Results for RQ1.1.4: performance of FOLtR-ES in terms of online nDCG@10 computed using relevance labels and the SERPs used for obtaining user iterations (averaged across all dataset splits). \label{fig:RQ4}} 
\end{figure}

In the original work and in the sections above, FOLtR-ES was evaluated using MaxRR computed with respect to the clicks performed by the simulated users (click models). This is an unusual evaluation for OLTR because: (1) usually nDCG@10 is used in place of MaxRR as metric, (2) nDCG is computed with respect to relevance labels, and not clicks, and on a withheld portion of the dataset, not on the interactions observed -- this is used to produce learning curves and is referred to as offline nDCG, (3) in addition online nDCG is measured from the relevance labels in the SERPs from which clicks are obtained, and either displayed as learning curves or accumulated throughout the sessions -- these values represent how OLTR has affected user experience. We then consider this more common evaluation of OLTR next, where we set the number of clients to 2,000 and experiment with $p=\{0.5, 0.9, 1.0\}$.

\begin{figure}[!htbp]
	\centering
	\begin{subfigure}{1\textwidth}
		\centering
		\includegraphics[width=1\textwidth]{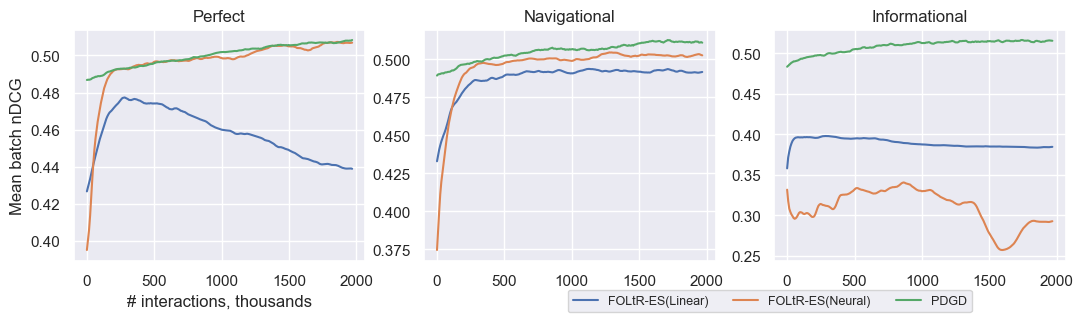}
		\caption{Mean batch nDCG@10 for MQ2007.}
		\label{fig:mq2007-rq4-offline}
	\end{subfigure}
	\begin{subfigure}{1\textwidth}
		\centering
		\includegraphics[width=1\textwidth]{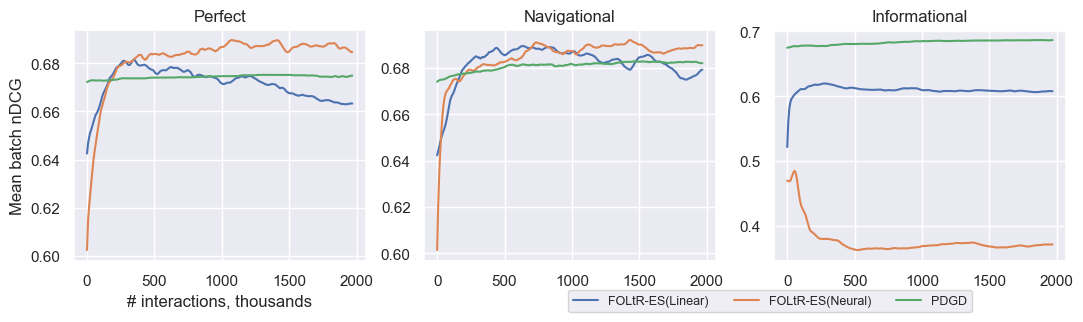}
		\caption{Mean batch nDCG@10 for MQ2008.}
		\label{fig:mq2008-rq4-offline}
	\end{subfigure}
	\begin{subfigure}{1\textwidth}
		\centering
		\includegraphics[width=1\textwidth]{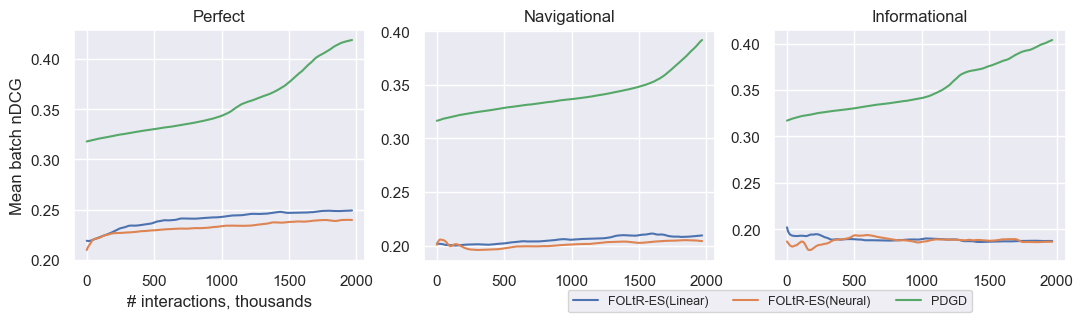}
		\caption{Mean batch nDCG@10 for MSLR-WEB10K.}
		\label{fig:mslr10k-rq4-offline}
	\end{subfigure}
	\begin{subfigure}{1\textwidth}
		\centering
		\includegraphics[width=1\textwidth]{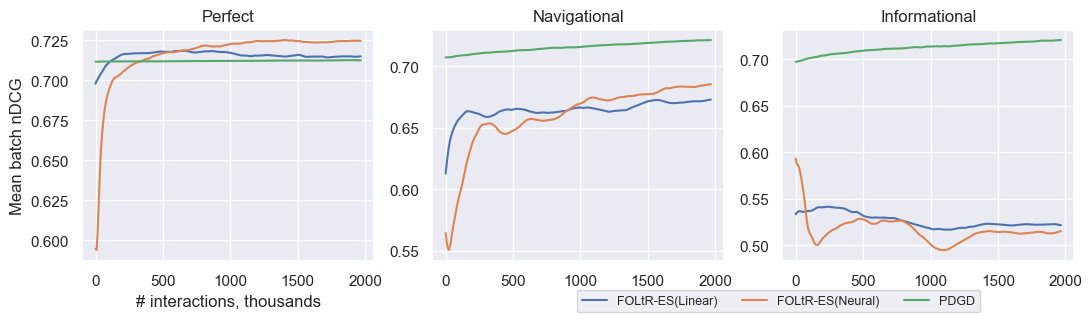}
		\caption{Mean batch nDCG@10 for Yahoo!.}
		\label{fig:yahoo-rq4-offline}
	\end{subfigure}
	\caption{Results for RQ1.1.4: performance of FOLtR-ES and PDGD in terms of offline nDCG@10 with privatization parameter $p=1$ and 2,000 clients (averaged across all dataset splits).\label{fig:RQ4-offline}} 
\end{figure}

Results are reported in Figure~\ref{fig:RQ4}. It is interesting to compare these plots with those in Figure~\ref{fig:RQ1}, that relate to the unusual (for OLTR) evaluation setting used in the original FOLtR-ES work. By comparing the figures, we note that for MQ2007/2008, FOLtR-ES can effectively learn rankers for perfect and navigational clicks. However, when the clicks become noisier (informational clicks), then FOLtR-ES learning is effective for the linear ranker but no learning occurs for the neural ranker: this is unlikely in the evaluation settings of the original work (Figure~\ref{fig:RQ1}). We note this finding repeating also for MSLR-WEB10K and Yahoo!, but this time this affects both linear and neural rankers; we also note that the online performance in MSLR-WEB10K on navigational clicks is also quite unstable and exhibits little learning for specific values of $p$ and ranker type. The online performance on MSLR-WEB10K for informational clicks (noisiest clicks) even exhibits a decreasing trend as iterations increase. Similar findings are also observed from results on Yahoo! dataset.

We further investigate the performance of FOLtR-ES with respect to offline nDCG@10. Results are shown in Figure~\ref{fig:RQ4-offline}, and are plotted along with the offline nDCG@10 of PDGD for additional context. Also the offline performance confirm that FOLtR-ES does not provide stable learning across click settings, datasets and ranker types. We also note that the performance of PDGD are sensibly higher than that of FOLtR-ES for MSLR-WEB10K dataset and for all datasets when informational clicks are considered.

These findings suggest that FOLtR-ES is yet far from being a solution that can be considered for use in practice, and more research is required for devising effective federated, privacy-aware OLTR techniques.
	\subsection{Summary}

In this section we considered the federated online learning to rank with evolutionary strategies (FOLtR-ES) method recently proposed by Kharitonov~\cite{kharitonov2019federated}, which represents the first method addressing privacy requirements in OLTR.

We set to explore four research questions related to FOLtR-ES. RQ1.1.1 aimed to investigate the generalisability of the original results obtained by FOLtR-ES on the MQ2007/2008 dataset to other datasets used in current OLTR practice. Our  experiments on MQ2007/2008 show consistent findings with that of Kharitonov~\cite{kharitonov2019federated}. However, when larger LTR datasets are considered, results change. In particular, the neural ranker used in FOLtR-ES is less effective than the linear ranker, especially on MSLR-WEB10K. %which shows FOLtR-ES needs more data to achieve effective training on those datasets with larger number of features.

RQ1.1.2 aimed to investigate the effect varying the number of clients involved in FOLtR-ES has on the effectiveness of the method. %We set the total times of interaction to a fix number and discover the ranking quality in terms of different number of clients in online simulation. 
Our experiments show mixed results with respect to the number of clients: the effect largely varies depending on dataset, ranker type and click settings. 
%that little clients number harm the performance in the linear ranker. But for the neural ranker, the difference is minor.

RQ1.1.3 aimed to compare FOLtR-ES with current OLTR state-of-the-art methods to understand the gap required to be paid for maintaining privacy with the help of federated learning. Our experiments show that FOLtR-ES lags behind the current OLTR state-of-the-art in terms of ranking performance: differences become more substantial when noisy clicks or larger datasets are considered. 

RQ1.1.4 aimed to investigate the generalisability of the original results obtained for FOLtR-ES to common evaluation practice in OLTR. Our experiments show that if the common evaluation settings used in OLTR are used to evaluate FOLtR-ES, then the method shows high variability in effectiveness across datasets, rankers and clicks types -- and overall that FOLtR-ES is unreliable on large datasets and noisy clicks. This finding suggests that more research and improvements are needed before a federated OLTR method, and FOLtR-ES in particular, can be used in practice. 

The results of our empirical investigation of FOLtR-ES help understanding the specific settings in which this technique works, and the trade-offs between user privacy and search performance (in terms of effectiveness and user experience). They also unveil that more work is require to devise effective federated methods for OLTR that can guarantee some degree of user privacy without sensibly compromising search performance.

Code, experiment scripts and experimental results are provided at \url{https://github.com/ielab/foltr}.

\section{Federated Pairwise Differentiable Gradient Descent (FPDGD)}
\label{rq1.2}

	%\section{Introduction}

In the previous section, we introduce FOLtR-ES and our further analysis of it. We find that the effectiveness of previous approach for federated OLTR (i.e., FOLtR-ES) exhibits large gaps in performance compared to current state-of-the-art, not federated, OLTR methods. It is this gap that we want to fill in this section. Specifically, in this section, we put forward the following contributions:

\begin{itemize}
	\item We cast the current state-of-the-art online LTR approach, the Pairwise Differentiable Gradient Descent (PDGD)~\cite{oosterhuis2018differentiable}, into the federated learning paradigm, specifically using the federated averaging algorithm. To do so, we devise a federated gradient update for PDGD. Our federated PDGD provides large, significant gains in effectiveness over the current federated OLTR method (FOLtR-ES~\cite{kharitonov2019federated,wang2021federated}), and its effectiveness generalises across datasets.
	
	\item We apply differential privacy on top of the proposed federated PDGD to secure the gradient updates performed in the federated environment against privacy attacks. We further investigate the trade-off between privacy preservation and ranker effectiveness.
\end{itemize}
	\subsection{Methodology}
\label{FPDGD}

In this section we propose a novel federated version of the current state-of-the-art OLTR method, the Pairwise Differentiable Gradient Descent (PDGD)~\cite{oosterhuis2018differentiable}, based on the Federated Averaging approach~\cite{mcmahan2017communication}. We then show how the gradient update of our federated OLTR method can be secured using differential privacy. First though, we provide a brief   introduction to the PDGD method.

\subsubsection{Pairwise Differentiable Gradient Descent}
\label{sec:PDGD}

Consider optimizing a ranking model $f_\theta(\textbf{d})$ which sorts documents by decreasing output scores, where $\textbf{d}$ represents the features of a query-document pair. The goal of an OLTR algorithm is to find the parameter vector $\theta$ for which the ranker displays an optimal ranking list.

Given this generic formulation of the ranking problem, PDGD applies a Plackett-Luce (PL) model to the ranking function $f_\theta(\cdot)$, resulting in a distribution over the document set $D$:
\begin{equation}
\label{eq:docprob}
P(d|D) = \frac{e^{f_\theta(d)}}{\sum_{d'\in{D}}e^{f_\theta(d')}} \text{.}
\end{equation}
A ranking $R$ of length $k$ is then created by sampling from the distribution $k$ times. With $d_i$ representing the document at position $i$, the probability of ranking $R$ is computed as:
\begin{equation}
\label{eq:listprob}
P(R|D) = \prod_{i=1}^{k} P(d_i | D \setminus \{d_1, ... , d_{i-1} \}) \text{.}
\end{equation}

After receiving the displayed list, the user may click on some of its documents. The method assumes that clicked documents represent a preference by the user over unclicked ones. For example, if $d_k$ is clicked while $d_l$ is not, the document preference inferred from this click event is $d_k >_c d_l$ and the probability that the preferred document $d_k$ is sampled before $d_l$ is increased. Based on this, PDGD adopts a pairwise optimization and estimates the gradient of the user preferences by the weighted sum:
\begin{equation}
\nabla f_\theta (\cdot) \approx \sum_{d_k >_c d_l}^{} \rho(d_k, d_l, R, D)[\nabla P(d_k > d_l)] \text{.}
\end{equation}
From~\cite{DBLP:conf/nips/SzorenyiBPH15}, the probability that the preferred document $d_k$ is sampled before $d_l$ is:
\begin{equation}
P(d_k > d_l) = \frac{P(d_k | D)}{P(d_k | D) + P(d_l | D)} = \frac{e^{f(d_k)}}{e^{f(d_k)} + e^{f(d_l)}} \text{.}
\end{equation}
Thus, the gradient estimation equals:
\begin{equation}
\label{eq:novelgradient}
\nabla f_\theta (\cdot) = \sum_{d_k >_c d_l}^{} \rho(d_k, d_l, R, D) \frac{e^{f_\theta (d_k)} e^{f_\theta (d_l)}}{(e^{f_\theta (d_k)} + e^{f_\theta (d_l)})^2} ({f_\theta}'(d_k) - {f_\theta}'(d_l)) \text{.}
\end{equation}

The function $\rho$ is applied to reweight the click preferences by the ratio between the occurring probabilities of $R$ or $R^*(d_k, d_l, R)$, which is the reversed pair ranking for $R$, i.e. the same ranking as $R$ except the position of $d_k$ and $d_l$ are swapped. This is done to reduce the position bias caused by the current ranking $R$. The reweighing function is defined as:
\begin{equation}
\label{eq:rho}
\rho(d_k, d_l, R, D) = \frac{P(R^*(d_k, d_l, R) | D)}{P(R | D) + P(R^*(d_k, d_l, R) | D)} \text{.}
\end{equation}

\begin{algorithm}[t]
	\caption{Pairwise Differentiable Gradient Descent(PDGD)~\cite{oosterhuis2018differentiable}} 
	\label{alg:pdgd:chap3}
	\begin{algorithmic}[1]
		\STATE \textbf{Input}: initial weights: $\mathbf{\theta}_1$; scoring function: $f$; learning rate $\eta$.  \label{line:novel:initmodel}
		\FOR{$t \leftarrow  1 \ldots \infty$ }
		\STATE $q_t \leftarrow \mathit{receive\_query}(t)$\hfill \textit{\small // obtain a query from a user} \label{line:novel:query}
		\STATE $D_t \leftarrow \mathit{preselect\_documents}(q_t)$\hfill \textit{\small // preselect documents for query} \label{line:novel:preselect}
		\STATE $\mathbf{R}_t \leftarrow \mathit{sample\_list}(f_{\theta_t}, D_t)$ \hfill \textit{\small // sample list according to Eq.~\ref{eq:docprob},\ref{eq:listprob}} \label{line:novel:samplelist}
		\STATE $\mathbf{c}_t \leftarrow \mathit{receive\_clicks}(\mathbf{R}_t)$ \hfill \textit{\small // show result list to the user} \label{line:novel:clicks}
		\STATE $\nabla f_{\theta_{t}} \leftarrow \mathbf{0}$ \hfill \textit{\small // initialize gradient} \label{line:novel:initgrad}
		\FOR{$d_k >_{\mathbf{c}} d_l \in \mathbf{c}_t$} \label{line:novel:prefinfer}
		\STATE $w \leftarrow \rho(d_k, d_l, R, D)$  \hfill \textit{\small // initialize pair weight (Eq.~\ref{eq:rho})} \label{line:novel:initpair}
		\STATE $w \leftarrow w 
		\frac{ 
			e^{f_{\theta_t}(\mathbf{d}_k)}e^{f_{\theta_t}(\mathbf{d}_l)}
		}{
			(e^{f_{\theta_t}(\mathbf{d}_k)} + e^{f_{\theta_t}(\mathbf{d}_l)}) ^ 2
		}$
		\hfill \textit{\small // pair gradient (Eq.~\ref{eq:novelgradient})} \label{line:novel:pairgrad}
		\STATE  $\nabla f_{\theta_{t}} \leftarrow \nabla f_{\theta_{t}} + w (f'_{\theta_t}(\mathbf{d}_k) - f'_{\theta_t}(\mathbf{d}_l))$
		\hfill \textit{\small // model gradient(Eq.~\ref{eq:novelgradient})} \label{line:novel:modelgrad}
		\ENDFOR
		\STATE $\theta_{t+1} \leftarrow \theta_{t} + \eta \nabla f_{\theta_{t}}$
		\hfill \textit{\small // update the ranking model} \label{line:novel:update}
		\ENDFOR
	\end{algorithmic}
\end{algorithm}

Algorithm~\ref{alg:pdgd:chap3} details the PDGD method.

\subsubsection{Federated Averaging for PDGD}\label{sec:fpdgd}

\begin{algorithm}[t]
	\caption{FederatedAveraging PDGD. \\
		$\bullet$ set of clients participating training: $C$, each client is indexed by $c$; \\
		$\bullet$ number of local interactions for client $c$: $n_c$ ($\sum_{c=1}^{|C|}n_c=n$) \\
		$\bullet$ local interaction set: $B$, model weights: $\theta$.}
	\label{alg:fpdgd:chap3}
	\begin{algorithmic}
		\SUB{Server executes:}
		\STATE initialize $\theta_0$; scoring function: $f$; learning rate: $\eta$
		\FOR{each round $t = 1, 2, \dots$}
			\FOR{each client $c \in C$ \textbf{in parallel}}
			\STATE $\theta_{t+1}^c, n_c \leftarrow \text{ClientUpdate}(c, \theta_t)$
			\ENDFOR
		\STATE $\theta_{t+1} \leftarrow \sum_{c=1}^{|C|} \frac{n_c}{n} \theta_{t+1}^c$
		\ENDFOR
		\STATE
		
		\SUB{ClientUpdate($c, \theta$):}\ \ \ // \emph{Run on client $c$}
		\FOR{each local update $i$ from $1$ to $B$}
		\STATE $\theta \leftarrow \theta + \eta\nabla f_\theta$    \hfill \textit{\small //PDGD update shown in Algorithm.~\ref{alg:pdgd:chap3}}
		\ENDFOR
		\STATE return ($\theta, n_c$) to server
	\end{algorithmic}	
\end{algorithm}

The Federated Averaging algorithm~\cite{mcmahan2017communication} is a popular approach for Federated Learning. It has the advantages of being able to handle large-scale data spread across numerous devices and reducing communication overhead by only sharing model updates instead of raw data. Next, we provide a working description of the algorithm.

Generally speaking, the optimization step of a generic machine learning algorithm consists of minimizing a loss function $f$: 
\begin{equation}
	\label{eq:problem}
	\min_{\theta \in {\ensuremath{\mathbb{R}}}^d} f(\theta)
	{\qquad \text{where} \qquad}
	f(\theta) {\overset{\text{def}}{=}} \frac{1}{n} \sum_{i=1}^n f_i(\theta).
\end{equation}
For each data example $(x_i, y_i)$, $f_i(\theta) = loss(x_i, y_i; \theta)$. 
Stochastic gradient descent (SGD) is commonly used to solve this optimization problem. By using SGD with a fixed learning rate $\eta$, the model is updated as: 
\begin{equation}
	\label{eq:sgd}
	\theta_{t+1} \leftarrow \theta_t - \eta \nabla f(\theta_t) \text{.}
\end{equation}

To perform the training process in federated manner, we assume that $|C|$ clients hold their own training dataset. For each client $c$ holding the local dataset $D_c$, the volume of the dataset is $n_c = |D_c|$. The loss function Eq.~\ref{eq:problem} can be re-written as:
\begin{equation}
	\label{eq:problem:distributed}
	f(\theta) = \sum_{c = 1}^{|C|} \frac{n_c}{n} F_c(\theta)
	\quad \text{where} \quad
	F_c(\theta) = \frac{1}{n_c} \sum_{i \in D_c} f_i(\theta).
\end{equation}

Based on Eq.~\ref{eq:problem:distributed}, $\nabla f(\theta_t) = \sum_{c = 1}^{|C|} \frac{n_c}{n} \nabla F_c(\theta)$. On each client $c$, the average gradient on the local data for the current model is denoted as $g_c = \nabla F_c(\theta_t)$. Then, the local model is updated using $\theta_{t+1}^c \leftarrow \theta_t - \eta g_c$ and the global model is updated by weighted averaging all local models: 
\begin{equation}
	\label{eq:weighted-averaging}
	\theta_{t+1} \leftarrow \sum_{c = 1}^{|C|} \frac{n_c}{n}\theta_{t+1}^c \text{.}
\end{equation}

 To do so, each client locally takes one step of gradient descent on the current model using its local data and the server then uses the weighted average of the local models as the global updated model. This process forms the FederatedSGD (FedSGD) algorithm~\cite{mcmahan2017communication}. Federated Averaging (FedAvg) extends FederatedSGD by adding more updating times to each client:  the local update $\theta_{t+1}^c \leftarrow \theta_t - \eta g_c$ is performed within a minibatch portion of the local dataset, before the averaging step.

In our approach (FPDGD), we implement PDGD using FedAvg. In Algorithm~\ref{alg:fpdgd:chap3}, each client considers $B$ interactions and updates the local ranker using PDGD gradients accordingly. After the local update is finished, each client returns the trained weights to the server. The server then aggregates those local messages by averaging the weights and sends back to the clients the latest ranker.

\subsubsection{Securing the Gradient: Differential Privacy for FPDGD}
\label{dp}

A key motivation for adopting a federated learning approach is to  protect users' private data: this is achieved by avoiding having to share data across clients or with a server -- instead, only gradient updates are shared. 
However, in the federated learning setting, the model parameters can be attacked to reveal sensitive information about the training data~\cite{fredrikson2015model, DBLP:conf/nips/GeipingBD020}.

To provide a strong privacy guarantee for the proposed federated PDGD, we introduce a noise-adding clipping technique~\cite{mcmahan2018learning, mugunthan2020privacyfl} which is based on the theory of differential privacy~\cite{dwork2006calibrating, dwork2014algorithmic}.

\begin{definition}($\epsilon$-Differential Privacy):
	A randomized mechanism $\mathcal{M} \colon \mathcal{D}
	\rightarrow \mathcal{R}$ with a domain $\mathcal{D}$ (e.g., possible training datasets) and range $\mathcal{R}$ (e.g., all possible trained models) satisfies $\epsilon$-differential privacy when there exists $\epsilon > 0$ such that:
	\begin{equation}
		P[{\mathcal{M}}(D_1)\in S]\leq e^{\epsilon}P[{\mathcal{M}}(D_2) \in S]
	\end{equation}
where $P[\cdot]$ is a probability, $D_1$ and $D_2$ are any two datesets differing for only one data instance, and $S$ denotes all subsets of possible outputs that $\mathcal{M}$ produces: $S \subseteq \mathcal{R}$. The parameter $\epsilon$ represents the privacy budget, where the lower the value of $\epsilon$, the higher the privacy guarantee.
\end{definition}

\begin{definition}(Global Sensitivity):
	For any real-valued query function $t \colon \mathcal{D} \rightarrow \mathcal{R}$, where $\mathcal{D}$ denotes the set of all possible datasets, the global sensitivity $\Delta$ of $t$ is defined as:
	\begin{equation}
	\Delta = 	\max_{D_1, D_2} |t(D_1) - t(D_2)|
	\end{equation}
for all $D_1, D_2 \in \mathcal{D}$.
\end{definition}

Since there is no apriori knowledge about the sensitivity of PDGD, we clip the weights outputed by PDGD to the bound $\frac{\Delta}{2}$, as in Eq.~\ref{eq:clipping}, so that the parameter updating can meet the global sensitivity $\Delta$. It should be noticed that the sensitivity $\Delta$ is used as hyper-parameter and will be determined through grid search for a given privacy level $\epsilon$.
\begin{equation}
	\theta = \theta \cdot \min(1, \frac{\Delta}{2 \cdot \parallel \theta \parallel}) \text{.}
	\label{eq:clipping}
\end{equation}

The Laplacian mechanism preserves $\epsilon$-differential privacy~\cite{dwork2006calibrating} through leveraging random noise $X$ sampled from a symmetric Laplacian distribution. The probability density function $f(x)$ of a zero-mean Laplacian distribution is:
\begin{equation}
	f(x) = \frac{1}{2\lambda} e^{-\frac{|x|}{\lambda}}
\end{equation}
where $\lambda$ is the scale parameter of the Laplacian distribution. Given $\Delta$ of the query function $t$, and the privacy loss parameter $\epsilon$, the Laplacian mechanism $\mathcal{M}$ uses random noise $X$ drawn from the Laplacian distribution with scale $\lambda = \frac{\Delta}{\epsilon}$.

\begin{algorithm}[t]
	\caption{FederatedAveraging PDGD with differential privacy. \\
		$\bullet$ set of clients participating training: $C$, each client is indexed by $c$; \\
		$\bullet$ number of local interactions for client $c$: $n_c$ ($\sum_{c=1}^{|C|}n_c=n$) \\
		$\bullet$ local interaction set: $B$, model weights: $\theta$.}
	\label{alg:fpdgd-dp}
	\begin{algorithmic}
		\SUB{Server executes:}
		\STATE initialize $\theta_0$; scoring function: $f$; learning rate: $\eta$
		\FOR{each round $t = 1, 2, \dots$}
		\FOR{each client $c \in C$ \textbf{in parallel}}
		\STATE $\theta_{t+1}^c, n_c \leftarrow \text{ClientUpdate}(c, \theta_t)$
		\ENDFOR
		\STATE $\theta_{t+1} \leftarrow \sum_{c=1}^{|C|} \frac{n_c}{n} \theta_{t+1}^c$
		\ENDFOR
		\STATE
		
		\SUB{ClientUpdate($c, \theta$):}\ \ \ // \emph{Run on client $c$}
		\FOR{each local update $i$ from $1$ to $B$}
		\STATE $\theta \leftarrow \theta + \eta\nabla f_\theta$   \hfill \textit{\small //PDGD update shown in Algorithm.~\ref{alg:pdgd:chap3}}
		\ENDFOR
		\STATE $\theta \leftarrow \theta \cdot \min(1, \frac{\Delta}{2 \cdot \parallel \theta \parallel} )$ \hfill \textit{\small //clip the weights}
		\STATE return ($\theta + \gamma - \gamma',n_c$) to server\hfill \textit{\small //$\gamma$,$\gamma'$ gamma noise sampled from Eq.~\ref{eq:gamma}}
	\end{algorithmic}	
\end{algorithm}

In the federated learning setting, each client contributes a portion of differentially private noise so that the sum of all contributions results in a differentially private noise value. Previous work has used Gamma random variables to generated the Laplace Random variable~\cite{mugunthan2020privacyfl}. Specifically, a Laplace Random variable can be generated from the sum of $n$ Gamma random variables:
\begin{equation}
	L(\mu, \lambda) = \frac{\mu}{n} + \sum_{p = 1}^{n}\gamma_p - \gamma'_p
\end{equation}
where $\mu$ is the mean parameter and $\lambda$ is the scale parameter of the Laplacian mechanism. 

$\gamma_p$ and $\gamma'_p$ are Gamma random variables with probability density function defined as:
\begin{equation}
\label{eq:gamma}
	f(x) = \frac{(1/s)^{1/n}}{\Gamma(1/n)} x^{1/n-1} e^{-x/s}
\end{equation}
where $1/n$ is the shape parameter, $s$ is the scale parameter and $\Gamma(p) = \int_{0}^{\infty} x^{p-1} e^{-x}\,dx$. In our federated setting, $n$ equals to number of clients $|C|$ and $s$ is equivalent to $\lambda$ of the Laplacian distribution. 

We set $\mu = 0$ and make use of this differential privacy technique so that each client adds $\gamma_p - \gamma'_p$ noise to each gradient update. This process is shown in Algorithm~\ref{alg:fpdgd-dp}.

	\subsection{Experimental Setup}

We rely on the typical evaluation setup used by previous OLTR research~\cite{hofmann2013fast, oosterhuis2016probabilistic, oosterhuis2018differentiable} to empirically investigate the effectiveness of the proposed FPDGD approach, and how it compares to other methods. This consists of use standard learning to rank datasets, simulate user interactions with SERPs, i.e. simulate clicks, and measure both online and offline performance.

\subsubsection{Datasets}
We use the MQ2007~\cite{DBLP:journals/corr/QinL13} and MSLR-WEB10K~\cite{DBLP:journals/corr/QinL13} datasets. While other, larger datasets for learning to rank are currently available, our choice considered the trade-off between representativeness and the high computational costs associated with experimenting with the federated OLTR methods on larger datasets. MQ2007 is relatively small and has fewer assessed documents per query, and was chosen to allow direct comparison with previous work on federated OLTR~\cite{kharitonov2019federated}, which mainly used MQ2007, and because this dataset, being small, allows for computationally treatable experiments. The MSLR-WEB10K is larger and more recent than MQ2007, and it was chosen as results in Section~\ref{sec3.1:results} suggested findings for federated OLTR observed on MQ2007 may not generalise on MSLR-WEB10K~\cite{wang2021federated}.

\subsubsection{User simulations}
Both the querying and clicking behaviour of users are simulated in our experiments. 

For the querying behaviour, for each client participating in the federated OLTR, we sample $B$ queries randomly, in line with previous work on FOLTR~\cite{kharitonov2019federated, wang2021federated}. For each query, we use the local ranking model (i.e. that held by the client) to rank documents; we limit SERP to 10 documents. 

For the click behaviour, we rely on the Cascade Click Model (CCM) as described in Section~\ref{chap2-click}.

\subsubsection{Federated setup}
We simulate the federated OLTR scenario as follows. Each client holds a copy of the current ranker. For each client, we consider $|B|$ user queries along with the respective interactions, during which the local ranker is optimised using the simulated user clicks. After $|B|$ interactions have occurred, the client sends the local message (updated weights) to the central server. The central server optimises the global ranker by aggregating the local messages and sends the newly-updated ranker back to each client. 

In our experiments for MQ2007, unless otherwise specified, we simulate $|C| = 1,000$ clients with each client performing $|B| = 4$ interactions (queries) locally to contribute to each global model update. We restrict the total interaction budget to 4 million queries, which results in 1,000 global updates. For MSLR-WEB10K dataset, unless otherwise specified, we also simulate $|C| = 1,000$ clients but with only $|B| = 2$ local interactions and we set the total interaction budget to 400,000 queries; thus resulting in 200 global updates.

\subsubsection{Baselines}
To investigate the impact of the federated averaging process on PDGD, we compare FPDGD against the original, centralised, PDGD model~\cite{oosterhuis2018differentiable}. We follow the original paper and set the learning rate $\eta = 0.1$. For both methods, we train a linear model as the ranker. In federated OLTR, global rankers are updated in batches, that is, the central server updates rankers after each client has executed $B$ searches and the local ranker updates have been sent to the server. However, in centralised OLTR settings, rankers are updated after each user interaction (batch size = 1). For fair comparison, we adapt PDGD to also be updated in batch by accumulating gradients updates. 

To date, FOLtR-ES is the only federated OLTR algorithm proposed in the literature~\cite{kharitonov2019federated}, and it constitutes a natural baseline for comparison with FPDGD, thus allowing to compare different federated OLTR methods.
For FOLtR-ES, we train a linear model to make fair comparison with our method trained on linear ranker. We set the learning rate $\eta = 0.001$ and choose the reciprocal rank of the highest clicked result (MaxRR) in an interaction as the optimization metric in FOLtR-ES.

\subsubsection{Evaluation measures}
To compare the ranking performance, we rely on the evaluation practice from previous OLTR work, which consists of measuring the ranker's offline and online performance using \textit{nDCG@k} as discussed in Section~\ref{chap2-metric}. We consider $k = 10$ as the number of documents in a ranked list for a query. For offline evaluation, we record the offline $nDCG@10$ score of the global ranker during each federated training update. For online evaluation, we take the online $nDCG@10$ score averaged across all clients.

Each experiment is repeated 25 times, spread evenly over all dataset training folds. All evaluation results are averaged and statistical significant differences between system pairs are evaluated using two-tailed Student's t-test with Bonferroni correction.

\subsubsection{Choosing privacy parameters}
The differential privacy process in FPDGD is controlled by the parameter $\epsilon$ as described in Section~\ref{dp}. To make comparison with the baseline FOLtR-ES method fair, we utilize the FOLtR-ES's privacy parameter $p$ to fix an upper bound on the value of $\epsilon$. In FOLtR-ES,  the higher the $p$ value, the lower the privacy. In our experiments, we set $p \in \{0.25, 0.5, 0.9, 1.0\}$; these settings correspond to upper bound values of $\epsilon \in \{1.2, 2.3, 4.5, 10\}$. 
For a given privacy level $\epsilon$, we search the best setting of sensitivity $\Delta$ through grid search in $\{1,3,5,7,9\}$. Then, for each $\epsilon \in \{1.2, 2.3, 4.5, 10\}$, we choose the corresponding $\Delta \in \{3, 3, 5, 5\}$.

	\subsection{Results}

\begin{figure}[t]
	\centering
	\begin{subfigure}{1\textwidth} \centering
		\includegraphics[width=1\textwidth]{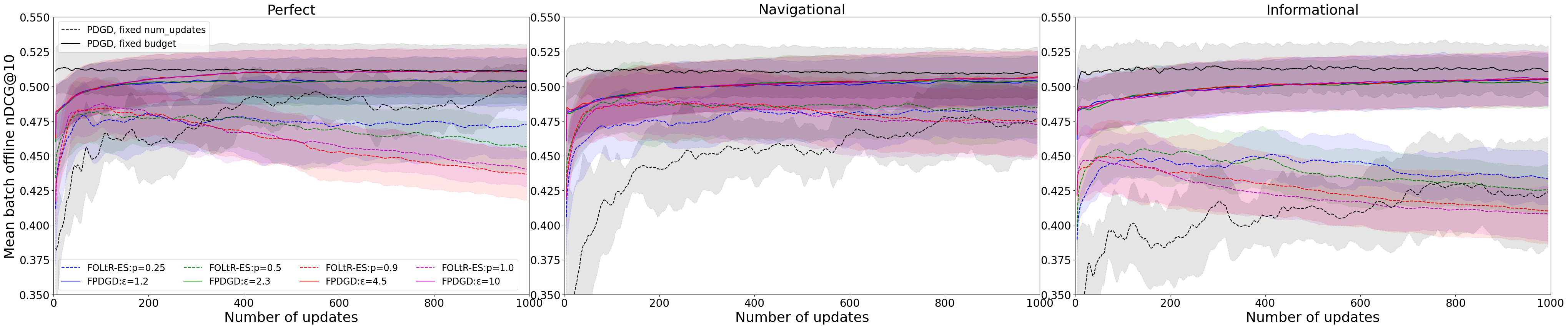}
		\caption{\tiny MQ2007}
		\label{fig:mq2007-ndcg}
	\end{subfigure}
	\begin{subfigure}{1\textwidth} \centering
		\includegraphics[width=1\textwidth]{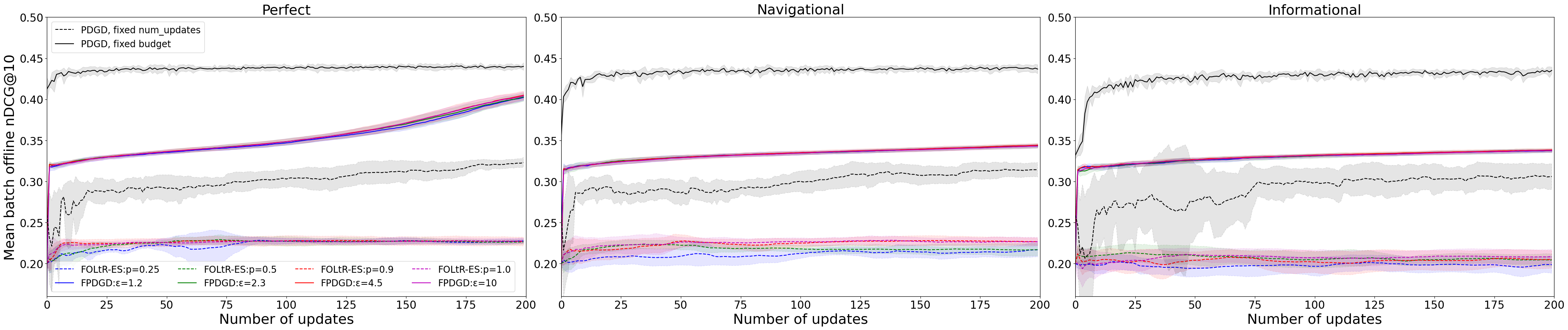}
		\caption{\tiny MSLR-WEB10K}
		\label{fig:mslr-ndcg}
	\end{subfigure}
	\caption{Offline performance (nDCG@10) across datasets, under  different click models, averaged across all dataset splits and experimental runs. Shaded areas indicate the standard deviation. \label{fig:RQ1-ndcg}} 
\end{figure}

\subsubsection{Overall ranking performance}

Figure~\ref{fig:RQ1-ndcg} displays the offline performance (nDCG@10) of the proposed FPDGD method across different settings of differential privacy $\epsilon$, along with the effectiveness of the non-federated PDGD method and the baseline federated FOLtR-ES method (across different settings of differential privacy $p$). The online performance of these methods are reported in Table~\ref{tbl-onlinemetric}.

\begin{table}[t!]
	\centering
	\caption{Online performance (discounted cumulative nDCG@10) for each dataset under different instantiations of CCM. Significant gains and losses of FPDGD over FOLtR-ES (baselines) are indicated by  \enkelop, \enkelneer\ (p $<$ 0.05) and \dubbelop, \dubbelneer\ (p $<$ 0.01), respectively.}\label{tbl-onlinemetric}
	\begin{tabular*}{\textwidth}{@{\extracolsep{\fill} } l  l l l l l  }
		\toprule
		& &{ \small \textbf{$\epsilon = 1.2$}}  & { \small \textbf{$\epsilon = 2.3$}}  & { \small \textbf{$\epsilon = 4.5$}}  & { \small \textbf{$\epsilon = 10$}}\\
		\midrule
		& \multicolumn{5}{c}{\textit{MQ2007}} \\
		\midrule
		\emph{perfect}&FPDGD& 296.03 {\tiny \dubbelneer}& 296.09 {\tiny \dubbelneer}& 313.28 & 313.26 \\
		&FOLtR-ES& 316.66 & 317.93 & 313.80 & 312.29  \\
		\emph{navigational}&FPDGD& 293.29 {\tiny \dubbelneer}& 293.42 {\tiny \dubbelneer}& 303.80 {\tiny \dubbelneer}& 303.82 {\tiny \dubbelneer}\\
		&FOLtR-ES& 319.90 & 325.04  & 324.33 & 323.05 \\
		\emph{informational}&FPDGD& 291.84 & 292.02 & 301.45 {\tiny \dubbelop}& 301.30 {\tiny \dubbelop} \\
		&FOLtR-ES& 292.58 & 294.78 & 288.57 & 285.96 \\
		\midrule
		& \multicolumn{5}{c}{\textit{MSLR-WEB10K}} \\
		\midrule
		\emph{perfect}&FPDGD& 54.62 {\tiny \dubbelop}& 54.61 {\tiny \dubbelop}& 54.64 {\tiny \dubbelop} & 54.61 {\tiny \dubbelop}\\
		&FOLtR-ES& 39.35 & 40.50 & 40.87 & 41.14  \\
		\emph{navigational}&FPDGD& 52.33 {\tiny \dubbelop}& 52.30 {\tiny \dubbelop}& 52.30 {\tiny \dubbelop}& 52.29 {\tiny \dubbelop}\\
		&FOLtR-ES& 38.55& 39.59  & 40.32 & 40.47 \\
		\emph{informational}&FPDGD& 51.11 {\tiny \dubbelop}& 51.16 {\tiny \dubbelop}& 51.14 {\tiny \dubbelop}& 51.18 {\tiny \dubbelop} \\
		&FOLtR-ES& 37.26 & 37.18 & 37.21 & 37.53  \\
		\bottomrule
	\end{tabular*}
\end{table}

\textbf{Impact of federation on PDGD.} We start the analysis of the results by commenting on the impact of federation on the PDGD method, focusing on the offline performance (Figure~\ref{fig:RQ1-ndcg}). When PDGD is given the same number of updates than its federated counterpart, its performance are lower. This is not surprising: the non-federated PDGD only uses the click data from one search interaction (query) for each update, and thus it is updated only 200 times in MSLR-WEB10K (1,000 in MQ2007). While FPDGD is also updated 200 times (1,000), when an update is performed this considers 2 interactions per client and 1,000 clients -- thus the total interaction budget assigned to FPDGD in MSLR-WEB10K is $200 \times 2 \times 1,000 = 400,000$  (4 million in MQ2007) compared to that of 200 for PDGD (1,000 for MQ2007). However if PDGD is given the same interaction budget as its federated counterpart, then PDGD does perform better than FPDGD. This difference is due to PDGD being able to perform a ranker update at each interaction (query), while FPDGD needs to wait the completion of batch size $\times$ clients $ = 2 * 1,000 = 2,000$ interactions in MSLR-WEB10K for doing an update ($4,000$ in MQ2007).

\textbf{Impact of privacy budget.} Next, we analyse the impact of adding differential privacy noise to FPDGD, where lower values of $\epsilon$ correspond to higher privacy guarantees. In theory, with higher privacy guarantee, the performance is worse due to more random noise is added. We find that the privacy budget we select ($\epsilon \in \{1.2, 2.3, 4.5, 10\}$) has little impact on the performance of FPDGD (both online and offline) for MSLR-WEB10K. For MQ2007, lower differential privacy values tend to provide lower performance, though not significantly worse.

\textbf{Comparison with FOLtR-ES.} When comparing the proposed federated approach, FPDGD, with the currently only other federated OLTR method, FOLtR-ES, we note that FPDGD consistently outperforms FOLtR-ES on offline performance across datasets, click models, and differential privacy settings (Figure~\ref{fig:RQ1-ndcg}). Gains over FOLtR-ES are specifically notable for all MSLR-WEB10K experiments and for the informational click model for MQ2007. The fact that the performance of FOLtR-ES are particularly poor in MSLR-WEB10K finds confirmation in previous section, which shows FOLtR-ES has poor performance on large datasets~\ref{sec3.1:results}. In addition, in MQ2007, FOLtR-ES exhibits a surprisingly decreasing trend as global model updates increase, especially for higher values of differential privacy $p$ ($p = 0.9, 1.0$), when the perfect and informational click models are considered. 
Similar observations can be made when considering online performance on MSLR-WEB10K dataset (Table~\ref{tbl-onlinemetric}): FPDGD is significantly better than FOLtR-ES under all three click models within different privacy budgets. However, on MQ2007, our method is not consistently better than FOLtR-ES. When the navigational click model is considered, FPDGD is significantly outperformed by FOLtR-ES. We believe the results are caused by the random noise from the sampling strategy of PDGD that occurs in the early training interactions: results on small-scale datasets like MQ2007 tend to be easily affected by this issue.
%[TODO] In depth analysis of the results on MQ2007\footnote{\textcolor{red}{Not shown here, but will be made available as online appendix.}} reveal that for this dataset FOLtR-ES has better online performance than FPDGD at the very early search interactions, with FPDGD soon catchup with, and then outperforming, FOLtR-ES in later interactions.

\begin{figure*}[t]
	\centering
	\begin{subfigure}{1\textwidth} \centering
		\includegraphics[width=1\textwidth]{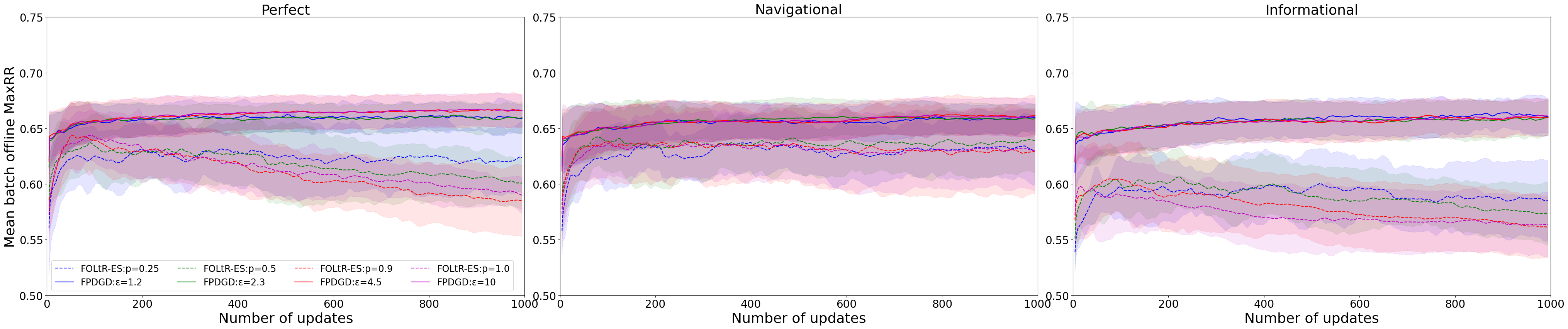}
		\caption{\tiny MQ2007}
		\label{fig:mq2007-mrr}
	\end{subfigure}
	\begin{subfigure}{1\textwidth} \centering
		\includegraphics[width=1\textwidth]{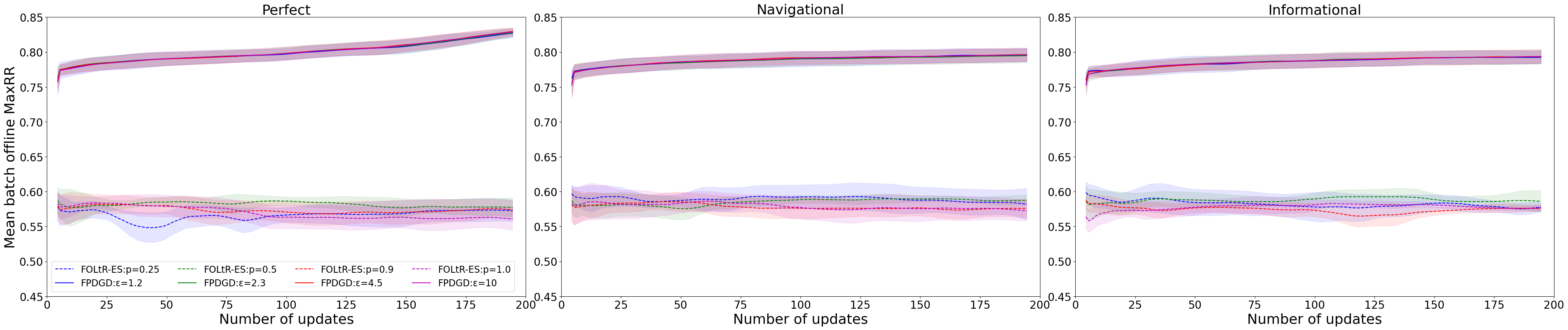}
		\caption{\tiny MSLR-WEB10K}
		\label{fig:mslr-mrr}
	\end{subfigure}
	\caption{Offline performance in terms of MaxRR across datasets, under three different click models,  averaged across all dataset splits and experimental runs. Shaded areas indicate the standard deviation. \label{fig:RQ1-mrr}} 
\end{figure*}

\textbf{Evaluation using MaxRR.} In our experiments, we used nDCG@10 as evaluation measure and optimisation criteria for the OLTR process, in line with previous work on OLTR. FOLtR-ES however was previously evaluated on MaxRR~\cite{kharitonov2019federated}, which is used internally by FOLtR-ES as optimisation criteria. For completeness, in Figure~\ref{fig:RQ1-mrr} we compare FPDGD and FOLtR-ES when MaxRR is used for evaluation. We find that the trends observed when evaluating with nDCG@10 (Figure~\ref{fig:RQ1-ndcg}) are also found when MaxRR is used.

\subsubsection{Influence of number of clients}

\begin{figure*}[t]
	\centering
	\includegraphics[width=1\textwidth]{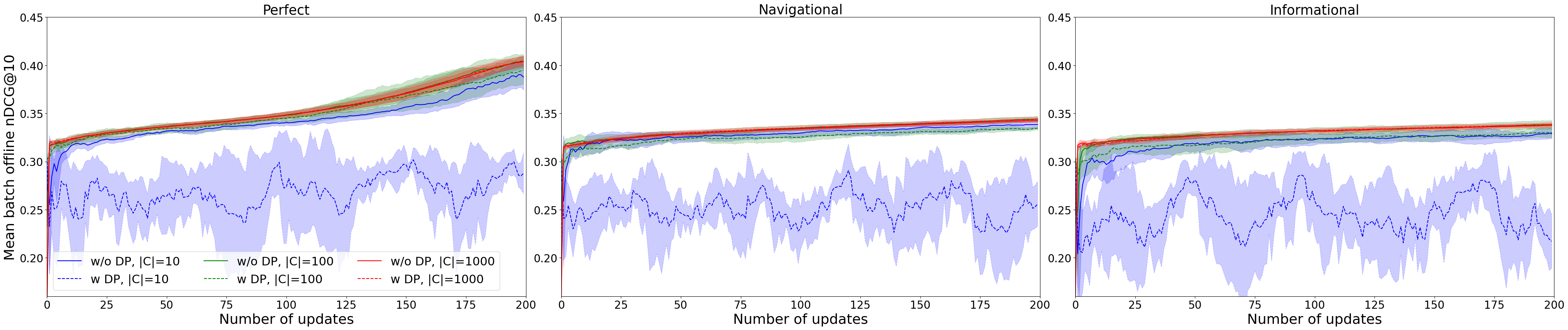}
	\caption{Offline performance on MSLR-WEB10K under three different click models, averaged across all dataset splits and experimental runs, for different number of clients $|C|$. We study FPDGD with differential privacy (\textit{w DP}) and without (\textit{w/o DP}). %Shaded areas indicate the standard deviation. 
	\label{fig:mslr-clients}}
\end{figure*}

To study the influence of the number of clients participating in our federated setup, we vary the number of clients in $|C| \in \{10, 100, 1000\}$. We report the offline nDCG@10 results obtained on MSLR-WEB10K in Figure~\ref{fig:mslr-clients}.
%[TODO](similar conclusions are identified when considering online performance and MQ2007\footnote{\textcolor{red}{Not shown here, but will be made available as online appendix.}}).
We consider two versions of FPDGD: one that uses the introduced differential privacy mechanism with $\Delta = 5$ and $\epsilon = 4.5$ (labelled \textit{w DP} in Figure~\ref{fig:mslr-clients}), and one for which differential privacy is turned off (\textit{w/o DP}).

When FPDGD does not rely on differential privacy, the more clients participate to the federated system the better, although differences are minor and not statistically significant. However, if differential privacy is used, when only a small number of clients participates to the federation, then FPDGD performs poorly.
We believe that this is so because the differential privacy process adds noise to the local gradient updates; this noise hinders the convergence of the global model when relying on very little interaction data for updating (i.e., 10 clients $\times$ 2 batch size $= 20$ queries). In this case, the noise introduced by the differential privacy process has a large impact. 
Conversely, differential privacy has little to no impact on performance if numerous clients participate ($|C| = 100, 1000$). Across all settings, we note that past a certain point, adding more clients to the system has little impact -- this finding is confirmed by prior work in general federated learning, which showed diminishing returns (because of the communication costs) for adding more clients beyond a certain point~\cite{mcmahan2017communication}.

\subsubsection{Influence of batch size}

We further study the influence of the local batch size ($|B|$) on FPDGD by varying $|B| \in \{2, 4, 8\}$. In our federated setting, each client parallelly conducts FPDGD's model update based on the local interactions $B$ (i.e. queries) and sends the updated model to the server. Note that within each batch of $|B|$ interactions, the client updates the local model $|B|$ times. The differential privacy settings are set to $\Delta = 5, \epsilon = 4.5$.

\begin{figure*}[t]
	\centering
	\includegraphics[width=1\textwidth]{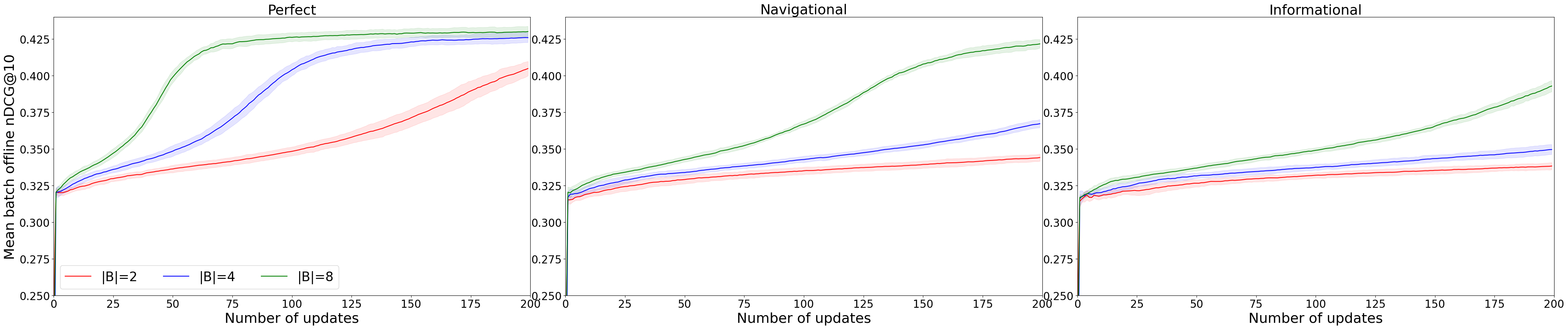}
	\caption{Investigation of the influence of batch size $|B|$ on offline performance for  MSLR-WEB10K under three different click models, averaged across all dataset splits and experimental runs. Shaded areas indicate the standard deviation. \label{fig:mslr-batch}}
\end{figure*}

The offline nDCG@10 performance of FPDGD across different $B$ settings are shown in Figure~\ref{fig:mslr-batch} for MSLR-WEB10K. The results suggest that the larger the batch size, the better the offline performance. This trend is especially significant on the perfect click model, although models trained on different batch sizes tend to ultimately converge after enough updates have been made. 
For the informational click models these differences are less remarked (though still significant), and convergence is long delayed. These results should not be surprising: a larger batch size means that more interactions (queries) have been used during a local update, e.g.,  200 interactions $\times$ 2 batch size $= 400$ queries per client vs. 200 interactions $\times$ 8 batch size $= 1,600$ queries per client.

\begin{figure*}[t!]
	\centering
	\includegraphics[width=1\textwidth]{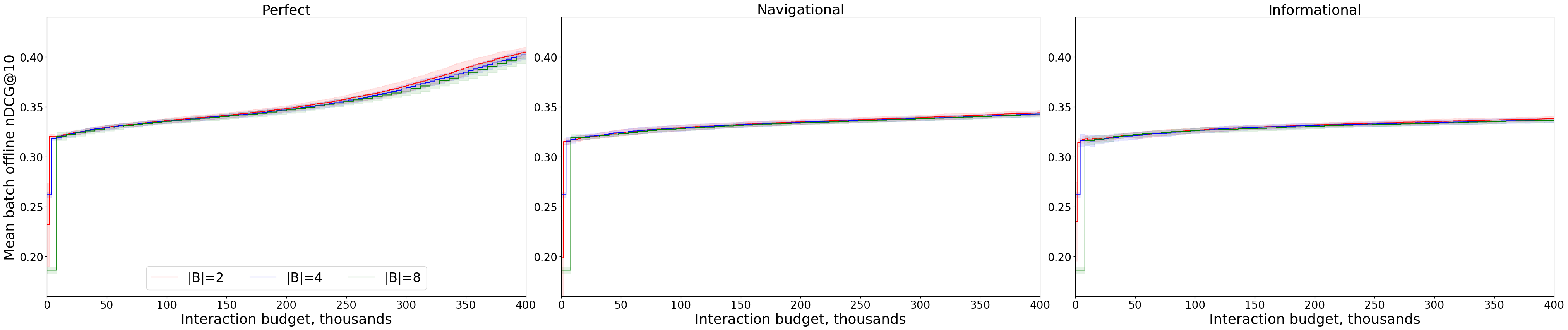}
	\caption{Investigation of the influence of batch size $|B|$ on offline performance for  MSLR-WEB10K under three different click models, averaged across all dataset splits and experimental runs, under \textit{fixed budget}. %Shaded areas indicate the standard deviation. 
	\label{fig:mslr-batch-fixed-budget}}
\end{figure*}

However, when the budget is maintained equal across different batch sizes, and thus smaller batch size means more frequent updates, then the performance obtained across different values of $|B|$ is quite similar (and no significant differences found), although a small batch size guarantees earlier updates, thus resulting in stronger initial performance.  This trend is shown in Figure~\ref{fig:mslr-batch-fixed-budget} for the offline performance on MSLR-WEB10K. The stronger initial performance, which is due to the initial model update occurring earlier when the batch size is small, does result in a slightly higher ranking performance when a small batch size is used. 

	\subsection{Summary}

In this section, we considered the problem of effective online learning to rank within a privacy-preserving environment. To this aim, we cast the Pairwise Differentiable Gradient Descent method (PDGD)~\cite{oosterhuis2018differentiable}, which represents the current state-of-the-art in OLTR, to the Federated Learning framework by using the Federated Averaging algorithm, thus proposing the federated Pairwise Differentiable Gradient Descent method (FPDGD). To further enhance the privacy of the method, we overlaid an $\epsilon$-differential privacy method to the proposed FPDGD, thus enforcing a level of privacy guarantee on the local gradient updates. 

Empirical evaluation shows FPDGD significantly outperforms the only other federated OLTR method (i.e., FOLtR-ES). This resonates with previous section (Section~\ref{sec3.1:results}) that showed the performance of FOLtR-ES to be highly variable across datasets and settings, making the method unstable and of risky use in practice. FPDGD instead represents a reliable, stable, and secured alternative to non-federated state-of-the-art OLTR methods. 

The source code that implements FPDGD, along with experiment code are made available at \url{http://ielab.io/FPDGD}.

%CHAPTER 4
%If you are presenting work which has been previously published, acknowledge this here.
% ***************************************************
% How to introduce a previously published chapter
% ***************************************************
%This is an example of how you might introduce a chapter that has been published previously. 
\cleartoevenpage
\pagestyle{empty}	
%Use this command (above) to suppress the header from the preceding chapter.

\noindent
The following publication has been incorporated as Chapter~\ref{Chap:foltr-noniid}.

\noindent
1.~\cite{wang2022non} \textbf{Shuyi Wang}, and Guido Zuccon, \href{https://doi.org/10.1145/3477495.3531709}{Is Non-IID Data a Threat in Federated Online Learning to Rank?}, \textit{In Proceedings of the 45th International ACM SIGIR Conference on Research and Development in Information Retrieval (SIGIR)}, pages 2801–2813, 2022

%\begin{table}[h]
%	\begin{center}
%	\begin{tabular}{|c|l|l|}
%		\hline
%		Contributor & Statement of contribution & \% \\
%		\hline
%		\textbf{Your Name}				& writing of text 					& 70\\
%															& proof-reading							& 60 \\
%															& theoretical derivations 	& 70\\
%															& numerical calculations 		& 100\\
%															& preparation of figures 		& 80 \\
%															& initial concept						& 10 \\
%		\hline
%		Co-author 1								& writing of text 					& 20\\
%															& proof-reading							& 10 \\
%															& supervision, guidance 		& 20\\
%															& theoretical derivations 	& 10\\
%															& preparation of figures 		& 20 \\
%															& initial concept						& 10 \\
%		\hline
%		Final Author							& writing of text 					& 10\\
%															& proof-reading							& 30 \\
%															& supervision, guidance 		& 80 \\
%															& theoretical derivations 	& 20 \\
%															& preparation of figures 		& 10 \\
%															& initial concept						& 80 \\
%		\hline
%	\end{tabular}
%	\end{center}
%\end{table}
%
%If your task breakdown requires further clarification, do so here. Do not exceed a single page.

% ***************************************************
% Example of an internal chapter
% ***************************************************
%This is an internal chapter of the thesis.
%If you have a long title, you can supply an abbreviated version to print in the Table of Contents using the optional argument to the \chapter command.
\chapter[FOLTR with non-IID Data]{Federated Online Learning to Rank with non-IID Data}
\label{Chap:foltr-noniid}	%CREATE YOUR OWN LABEL.
\pagestyle{headings}

In the previous chapter, we address on the effectiveness of FOLTR paradigm and the factors that affect its performance, with an overview on the proposed and empirically investigated methods~\cite{kharitonov2019federated,wang2021federated,wang2021effective}. In this chapter, we address on an important issue that affects the performance of federated learning systems but has been fully ignored by research on FOLTR: the presence of bias in how the training data is divided across the clients that join the federation. In other words, the fact that clients may hold non-independent and identically distributed (non-IID) data~\cite{zhu2021federated}. 

\begin{figure}[b]
	\centering
	\includegraphics[width=0.9\columnwidth]{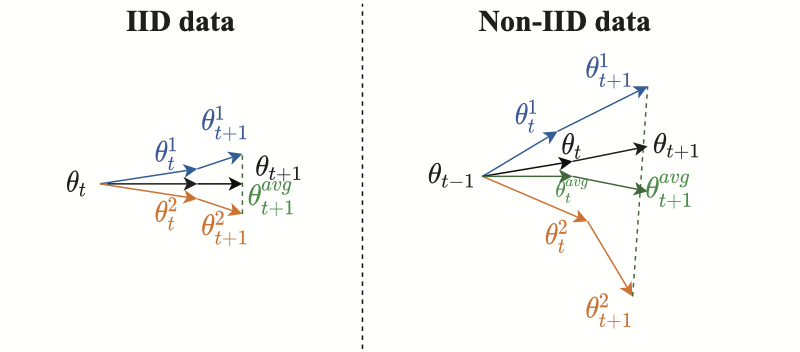}
	\caption{Illustration of the model divergence problem in FL, adapted from Zhu et al.~\cite{zhu2021federated}. $\theta_t$ is the ideal global model under centralised learning, and $\theta^{avg}_t$ is the average model created from the local models of Client 1 ($\theta^1_t$) and Client 2 ($\theta^2_t$) through FedAvg~\cite{mcmahan2017communication}.}
	\label{fig:modeldiver} 
\end{figure}

Non-IID data can pose severe threat to the effectiveness of a federated learning method. Models trained federatively in the presence of non-IID data across the clients that participate in the federation, in fact, display significantly lower effectiveness, and at times experience difficulties for the model to converge~\cite{zhu2021federated}. 
Effectiveness degradation can mainly be attributed to the weight divergence between the local models resulting from the non-IID distribution of the data across the clients~\cite{zhu2021federated}. Local models with the same initial parameters will converge to different models because of the heterogeneity of the local data distributions. This divergence will increase as more communication rounds of the federated learning algorithm are performed. This slows down or even impedes model convergence, worsening the performance of the global model. An illustration of the phenomenon of model divergence for both IID and non-IID data in federated learning is given in Figure~\ref{fig:modeldiver}. The ideal global model ($\theta_t$, under centralised learning) and actual global model ($\theta^{avg}_t$, average model created through FedAvg~\cite{mcmahan2017communication}) coincide when data is IID, but diverge when data is non-IID, showing that this is a sizeable problem when the data is non-IID.
%\textcolor{red}{(TODO: change the figure)}

This chapter provides a systematic understanding of when non-IID data may occur in the FOLTR setting and the impact of non-IID data in such cases. If not specified otherwise, we only consider horizontal FL~\cite{yang2019federated} and we believe our framework can be applied to both cross-device and cross-silo federated learning~\cite{kairouz2021advances}. Specifically, we investigate the following research questions:

\newpage
\begin{enumerate}[label=\textbf{RQ2:}, leftmargin=*]
	\item \bfseries Are FOLTR methods robust to non-IID data among clients? \itshape
	\begin{enumerate}[label=RQ2.\arabic*:, leftmargin=*]
		\item What are the potential types of non-IID data in FOLTR and their impact on model performance?
		\item Do existing methods that deal with non-IID data in FL of general domain generalise to FOLTR?
	\end{enumerate}
\end{enumerate}

To answer RQ2.1, we first enumerate possible data distribution settings that may showcase data bias across clients and thus give rise to the non-IID problem. Then, we study the impact of each setting on the performance of the state-of-the-art FOLTR approach, the Federated Pairwise Differentiable Gradient Descent (FPDGD, introduced in Section~\ref{rq1.2}), and we highlight which data distributions may pose a problem for FOLTR methods. For RQ2.2, we explore how common approaches proposed in the federated learning literature to deal with non-IID data can be cast in the FOLTR framework and their effectiveness in addressing non-IID data in FOLTR.

%Models trained in Federated Learning (FL) usually have worse performance than those trained in the standard centralized learning mode, especially when the training data are not independent and identically distributed (non-IID) on the local devices~\cite{zhu2021federated}. The non-IID issue is one of the challenges arising in FL when it is applied to the real world compared with the centralized learning. The performance degradation can mainly be attributed to weight divergence of the local models resulting from non-IID data~\cite{zhu2021federated}. That is, local models having the same initial parameters will convergence to different models because of the heterogeneity in local data distribution and the divergence will continue to expand together with the increase of communication rounds in FL, slowing down the convergence and worsening the learning performance. An illustration about model divergence for both IID and non-IID data in FL~\cite{zhu2021federated} is shown Figure~\ref{fig:modeldiver}.

%\textcolor{red}{(contribution of this paper)} In this paper, we are dedicated to give a comprehensive analysis of the influence of  non-IID data under different categories to the existing FOLTR system. In the latter part of this paper, we provide corresponding solutions based on existing techniques for handling skewed data distribution~\cite{}.

%	\input{Chapter4_paper2_noniid_sigir2022/sections/related_works.tex}

\begin{figure}[t]
	\centering
	\includegraphics[width=0.8\columnwidth]{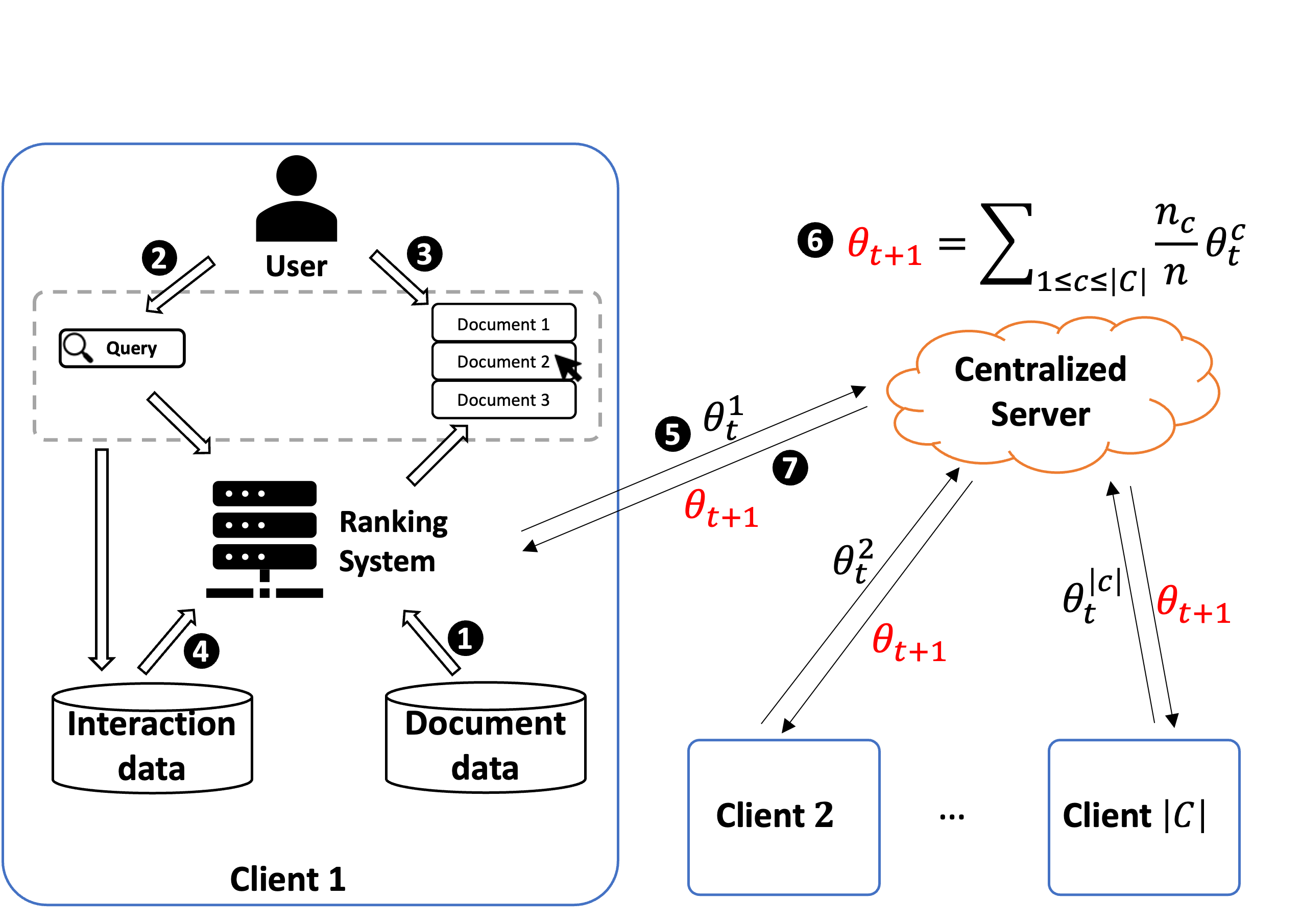}
	\caption{Schematic representation of the FOLTR setting.
	\label{fig:FOLTR}
	}
\end{figure}

\section{FOLTR Framework and FPDGD}

Our study is based on the Federated Pairwise Differentiable Gradient Descent (FPDGD, introduced in Section~\ref{rq1.2}) method, which represents the current state-of-the-art in FOLTR and that we use as a representative method in our experiments to investigate the effect of non-IID data on FOLTR. We did not include the \textit{securing gradient} component in FPDGD (introduced in Section~\ref{dp}) so as not to let the differential privacy noise affect the results of non-IID analysis. We next briefly describe the FOLTR framework used in this chapter.

The federated online learning to rank setting is pictured in Figure~\ref{fig:FOLTR}. Searchable data is stored by each client (\circled{1}) and not shared with the centralised server or other clients. Different clients may hold all, a portion of, none of the same searchable data. Queries and user's clicks occur at a client side (\circled{2} and \circled{3}) and are not communicated to the centralised server or other clients: search is indeed entirely performed on the user device (\circled{2}). Each client exploits search interactions to perform local model updates to the ranker; for FPDGD, the routine executed by the client is shown in Algorithm~\ref{alg:fpdgd:chap4}, and the PDGD update is shown in Algorithm~\ref{alg:pdgd:chap4}. Each client considers $B$ interactions before updating the local ranker using the PDGD gradients. 
These local updates are then shared with the central server (\circled{5}), which in turn combines the ranker updates from the clients to produce a revised ranker (\circled{6}); for FPDGD, this is achieved according to the server routine in Algorithm~\ref{alg:fpdgd:chap4}. The new global model is then distributed to the user's device (\circled{7}).

\begin{algorithm}[t]
	\caption{FederatedAveraging PDGD. \\
		$\bullet$ set of clients participating training: $C$, each client is indexed by $c$; \\
		$\bullet$ number of local interactions for client $c$: $n_c$ ($\sum_{c=1}^{|C|}n_c=n$) \\
		$\bullet$ local interaction set: $B$, model weights: $\theta$.}
	\label{alg:fpdgd:chap4}
	\begin{algorithmic}
		\SUB{Server executes:}
		\STATE initialize $\theta_0$; scoring function: $f$; learning rate: $\eta$
		\FOR{each round $t = 1, \dots \infty$}
		\FOR{each client $c \in C$ \textbf{in parallel}}
		\STATE $\theta_{t+1}^c, n_c \leftarrow \text{ClientUpdate}(c, \theta_t)$
		\ENDFOR
		\STATE $\theta_{t+1} \leftarrow \sum_{c=1}^{|C|} \frac{n_c}{n} \theta_{t+1}^c$
		\ENDFOR
		\STATE
		
		\SUB{ClientUpdate($c, \theta$):}\ \ \ // \emph{Run on client $c$}
		\FOR{each local update $i$ from $1$ to $B$}
		\STATE $\theta \leftarrow \theta + \eta\nabla f_{\theta}$    \hfill \textit{\small //PDGD update shown in Algorithm.~\ref{alg:pdgd:chap4}}
		\ENDFOR
		\STATE return ($\theta, n_c$) to server
	\end{algorithmic}	
\end{algorithm}

\begin{algorithm}[t]
	\caption{Pairwise Differentiable Gradient Descent(PDGD)~\cite{oosterhuis2018differentiable}} 
	\label{alg:pdgd:chap4}
	\begin{algorithmic}[1]
		\STATE \textbf{Input}: initial weights: $\mathbf{\theta}_1$; scoring function: $f$; learning rate $\eta$.  \label{line:novel:initmodel}
		\FOR{$t \leftarrow  1, \ldots B$ }
		\STATE $q_t \leftarrow \mathit{receive\_query}(t)$\hfill \textit{\small // obtain a query from a user} \label{line:novel:query}
		\STATE $D_t \leftarrow \mathit{preselect\_documents}(q_t)$\hfill \textit{\small // preselect documents for query} \label{line:novel:preselect}
		\STATE $\mathbf{R}_t \leftarrow \mathit{sample\_list}(f_{\theta_t}, D_t)$ \hfill \textit{\small // sample list} \label{line:novel:samplelist}
		\STATE $\mathbf{c}_t \leftarrow \mathit{receive\_clicks}(\mathbf{R}_t)$ \hfill \textit{\small // show result list to the user} \label{line:novel:clicks}
		\STATE $\nabla f_{\theta_{t}} \leftarrow \mathbf{0}$ \hfill \textit{\small // initialize gradient} \label{line:novel:initgrad}
		\FOR{$d_k >_{\mathbf{c}} d_l \in \mathbf{c}_t$} \label{line:novel:prefinfer}
		\STATE $w \leftarrow \rho(d_k, d_l, R, D)$  \hfill \textit{\small // initialize pair weight} \label{line:novel:initpair}
		\STATE $w \leftarrow w 
		\frac{ 
			e^{f_{\theta_t}(\mathbf{d}_k)}e^{f_{\theta_t}(\mathbf{d}_l)}
		}{
			(e^{f_{\theta_t}(\mathbf{d}_k)} + e^{f_{\theta_t}(\mathbf{d}_l)}) ^ 2
		}$
		\hfill \textit{\small // pair gradient} \label{line:novel:pairgrad}
		\STATE  $\nabla f_{\theta_{t}} \leftarrow \nabla f_{\theta_{t}} + w (f'_{\theta_t}(\mathbf{d}_k) - f'_{\theta_t}(\mathbf{d}_l))$
		\hfill \textit{\small // model gradient} \label{line:novel:modelgrad}
		\ENDFOR
		\STATE $\theta_{t+1} \leftarrow \theta_{t} + \eta \nabla f_{\theta_{t}}$
		\hfill \textit{\small // update the ranking model} \label{line:novel:update}
		\ENDFOR
	\end{algorithmic}
\end{algorithm}

	\section{Types of non-IID Data in FOLTR}

We consider training a ranker for the OLTR system as a supervised learning task in an FL setup, with each client holding a subset of the data. Each data sample is denoted as $(x,y)$, where $x$ is the feature representation of the data and 
%is the input attribute of feature and 
$y$ is the label. The local distribution of the dataset in client $i$ is denoted as $P_i(x,y)$. The presence of non-IID data can be represented as the difference between local data distributions: that is, for different clients $i$ and $j$, $P_i(x,y) \neq P_j(x,y)$.

In federated learning, data across clients may not be IID due to different reasons: Kairouz et al.~\cite{kairouz2021advances} and Zhu et al.~\cite{zhu2021federated} assert this can be due to how features $x$ and labels $y$ are distributed. However, the translation of these categories to FOLTR is not straightforward. In the following sections, we put forward several situations in which data specific to FOLTR could be distributed in a non-IID manner across clients. Specifically, we consider data in the FOLTR process may not be IID because of biases across clients due to:
\begin{itemize}
	\item \textbf{Type 1:} document preferences skewness (Section~\ref{sec:type1})
	\item \textbf{Type 2:} document label distribution skewness (Section~\ref{sec:type2})
	\item \textbf{Type 3:} click preferences skewness (Section~\ref{sec-type3-4})
	\item \textbf{Type 4:} data quantity skewness (Section~\ref{sec-type3-4})
\end{itemize}
%\todo{@hs: probably you want these to align with the section headers below. You could even ref them. e.g., Document Preferences (Section 5). As a second point, here could be a good place to give an intuition for what each of these types means, maybe with an example.}

\begin{table*}[!t]
	\newcommand{\tabincell}[2]{\begin{tabular}{@{}#1@{}}#2\end{tabular}}
	\centering
	\caption{Summary of non-IID data types in FOLTR. \label{tbl-data-types}}
	\small % Use smaller font size
	\begin{tabularx}{\textwidth}{cXp{10cm}}
		\toprule
		\bfseries Data type & \bfseries Key characteristic & \bfseries When it happens in FOLTR \\ \midrule
		Type 1 & Document Preferences & \tabincell{l}{Different clients have different preferred candidate documents, \\although they are searching for the same query.} \\ \midrule
		Type 2 & Document Label Distribution & \tabincell{l}{Different clients hold candidate documents with different label \\distribution while the conditional feature distribution is the shared.} \\ \midrule
		Type 3 & Click Preferences & \tabincell{l}{Different clients have various preferred click behaviours when \\searching for the same query.} \\ \midrule
		Type 4 & Data Quantity & \tabincell{l}{Different clients have different frequency on issuing queries and \\interacting with the searching system.} \\ \bottomrule
	\end{tabularx}
	
\end{table*}

The last data type, Type 4, i.e., the situation in which different clients hold different quantities of data (and in particular interaction data such as queries and clicks), does not necessarily imply that the data is non-IID. However, we note this case is often studied in the FL literature alongside non-IID data~\cite{zhu2021federated,li2022federated}, and thus we include this situation in our considerations of the non-IID problem. Each data type is defined and investigated in the next sections; in addition we provide a summary overview of the data types in Table~\ref{tbl-data-types}.

We also note that commonly in federated learning, non-IID data occurs because the data is distributed across clients according to its features. In other words, the marginal distribution of the features belonging to the data held by each client may vary, i.e. for different clients $i$ and $j$, $P_i(x) \neq P_j(x)$. This situation may occur in horizontal federated learning settings (also called homogeneous FL)~\cite{yang2019federated}, where each client holds different and overlapping datasets. In this case, the non-IID divergence is usually caused by inconsistent data distributions, e.g., feature imbalance of the training data local to each client. However, this case does not seem applicable to FOLTR (thus is not further studied in this thesis). In FOLTR, each data item is represented by the feature vector of a query-document pair and its relevance label. 
The features often consist of variations of query-dependent features such as TF-IDF scores, BM25 scores, query length, as well as query-independent features such as PageRank, URL lengths, and so on~\cite{DBLP:journals/corr/QinL13}. In this case, bias in the feature distribution across clients would be rare as most features are dependent on the query-document pair. 

Next, we describe the non-IID data types we put forward in this chapter and analyse their impact on FOLTR. We empirically find that only Type 1 and partially also Type 2 data have a strong impact on the FOLTR. We thus predominantly focus our attention on these two data types while providing only a definition and a brief account of the remaining two data types in the thesis. For Type 1 and Type 2, we introduce the simulation of non-IID data, further analysis on impact of non-IID data, and implementation of methods on dealing with non-IID data. For Type 3 and Type 4, we only focus on the simulation and analysis on its impact.

Code, experiment scripts and experimental results are provided at \url{https://github.com/ielab/2022-SIGIR-noniid-foltr}.

	\section{Type 1: Document Preferences} \label{sec:type1}

\emph{Document preference skewness} (Type 1) considers the situation when the conditional distribution $P_i(y|x)$ varies across the clients though $P_i(x)$ remains the same. This happens when different clients have different preferred candidate documents, although they are searching for the same query. As OLTR requires the user's implicit feedback as an optimization objective, which might be highly related to individual preferences, this setting appears to be of very likely occurrence.

\subsection{Simulating Type 1 non-IID Data}

The mechanism we use to simulate non-IID data of Type 1 and IID data to baseline the FOLTR effectiveness relies on a recent work that empirically studied and demonstrated how OLTR methods adapt when user's search intents change overtime~\cite{zhuang2021how}. In particular, Zhuang and Zuccon~\cite{zhuang2021how} created a collection for OLTR with several explicit intent types by adapting an existing TREC collection, as no dataset is available for studying this OLTR problem. Derived from ClueWeb09 and the TREC Web Track 2009 to 2012~\cite{clarke2012overview}, this intent change collection consists of 200 queries with 4 intents each and, on average, 512 candidate documents per query. Furthermore, query-document pairs' relevance judgements are provided per intent. We believe this is an appropriate collection to adapt to study the effect of Type 1 non-IID data on FOLTR. We can regard each intent as a type of user preference. As the average number of relevant documents per intent varies largely across all intent types, the learning difficulty of optimizing a ranker among different intents also varies. To avoid this bias, we follow Zhuang and Zuccon~\cite{zhuang2021how} and we re-label the original intent number for each query through random shuffling: this is possible because all intent types are independent across queries. In our experiments, we repeat this process of re-balancing 5 times, thus giving rise to results averaged across 5 FOLTR experiments. We refer to Zhuang and Zuccon~\cite{zhuang2021how} for further details on the dataset creation, and we further highlight that we have made available an implementation of the dataset creation procedure along with the actual dataset at \url{https://github.com/ielab/2022-SIGIR-noniid-foltr}.

To simulate non-IID data, after randomly shuffling all intents across 4 types, we let each intent represent one client preference. The client preferences differ from each other for the same query-document pair so as the corresponding relevance judgements. The federated setup involves 4 clients (represented by 4 types of intent) and the local updating time $B = 5$ with fixed global communication times $T = 10,000$. These settings are similar to those used in previous work on FOLTR~\cite{kharitonov2019federated,wang2021federated,wang2021effective}\footnote{In Section~\ref{rq1.1}, we have a detailed analysis on the relationships between number of clients, number of local updates $B$, and FOLTR effectiveness.}
For the implicit feedback in FOLTR, same as Chapter~\ref{Chap:fpdgd}, we simulate user clicks based on the popular \emph{Cascade Click Model} (CCM). We limit SERP to 10 documents and use $nDCG@10$ for offline evaluation, cumulative discounted $nDCG@10$~\cite{oosterhuis2018differentiable} for online evaluation, as discussed in Section~\ref{chap2-metric}. We train a linear ranker and a neural ranker on the intent-change dataset. For the linear model, we set the learning rate $\eta = 0.1$ and zero initialization was used. As in the original PDGD studies~\cite{oosterhuis2018differentiable}, the neural ranker is optimized using a single hidden-layer neural network with 64 hidden nodes, along with $\eta = 0.1$.

As in Zhuang and Zuccon~\cite{zhuang2021how}, given that no held-out test set is available, we evaluate both online and offline performance on the original training set across all 4 intent types and average all results. For the IID setting, we merge all intents and mark a document as relevant as long as it is judged relevant for at least one of the intent types. Each client randomly picks a query from the training set and clicks documents based on the same preferences during the federated training with IID data. Other settings remain the same as the non-IID experiments.
%\textcolor{red}{(TODO: modify the definition of IID data.)}

\subsection{Impact of Type 1 non-IID Data}

\begin{figure*}[t]
	\centering
	\begin{subfigure}{1\textwidth}\centering
		\includegraphics[width=1\textwidth]{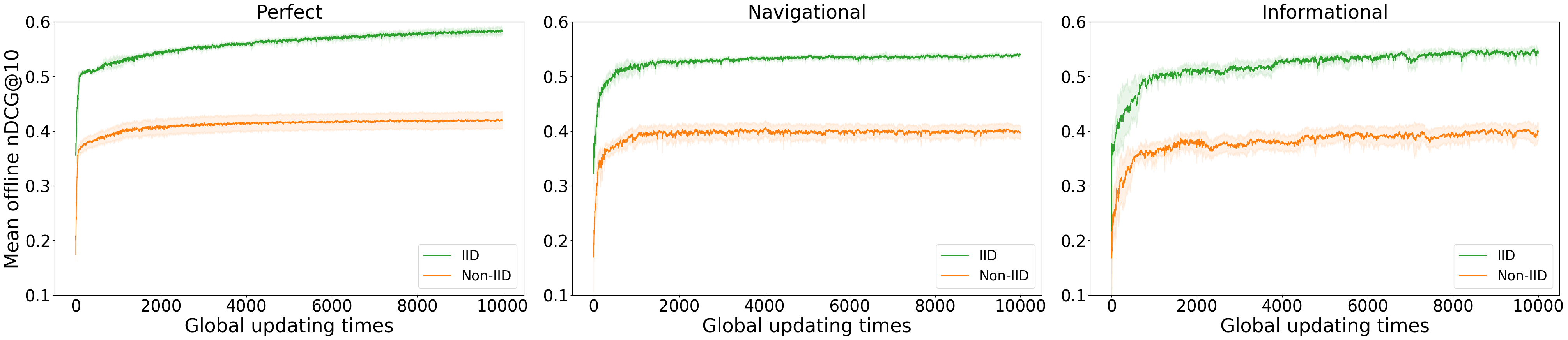}
		\caption{intent-change (linear ranker)}
		\label{fig:intent-linear}
	\end{subfigure}
	\begin{subfigure}{1\textwidth}\centering
		\includegraphics[width=1\textwidth]{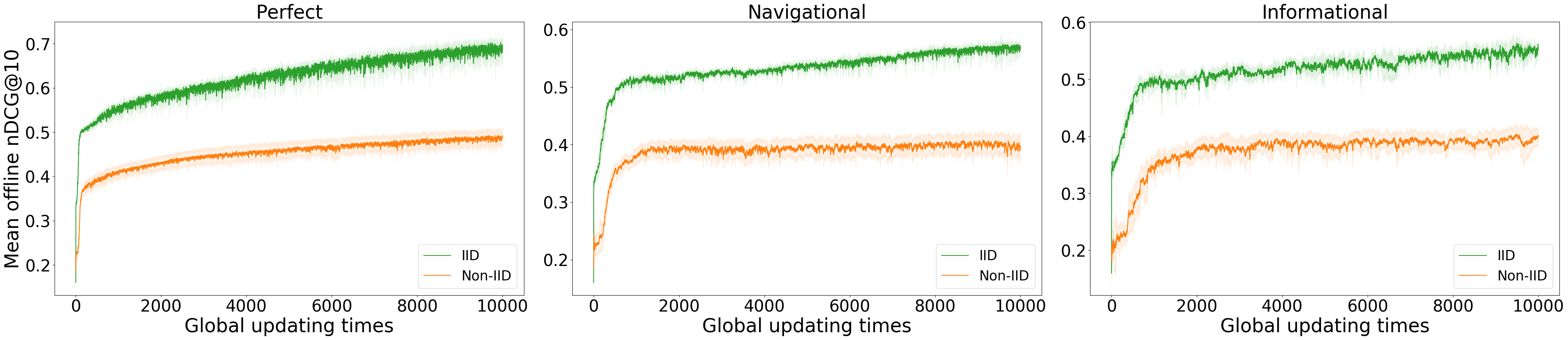}
		\caption{intent-change (neural ranker)}
		\label{fig:intent-neural}
	\end{subfigure}
	\caption{Offline performance (nDCG@10) on Type 1 data; results averaged across dataset splits and experimental runs.
	\label{fig:dps}}
\end{figure*}

The offline performance related to Type 1 data is shown in Figure~\ref{fig:dps}; the corresponding online performance is shown in Table~\ref{tbl:dps_online}.
From the offline performance, it is clear that the presence of non-IID data negatively impacts the performance of the learnt ranker, compared to those obtained when data is IID. In terms of online performance, rankers obtained in the presence of non-IID data are also worse than when trained with IID data. This can be explained as follows. Since each client has its preference (intent), the relevant documents are judged in different ways; this leads to the divergence of each client's local ranker update, as exemplified in Figure~\ref{fig:modeldiver}.

In summary, we find that if data is distributed in a non-IID manner across clients according to Type 1, the effectiveness of FOLTR (and specifically of FPDGD) is seriously affected.

%% merge into one pic
%\begin{figure*}[t]
%	\centering
%	\includegraphics[width=1\textwidth]{Chapter4_paper2_noniid_sigir2022/images/intent-change_FPDGD_clients4_batch5_updates10000_non_iid_intent_change_CCM_linear_and_neural16.png}
%	\caption{Offline performance (nDCG@10) on Type 1 data; results averaged across dataset splits and experimental runs.}
%\label{fig:dps}
%\end{figure*}

%% separate linear and neural into two pics 

\begin{table}[t]
	\centering
	\caption{Online performance (discounted cumulative nDCG@10) on Type 1 data, averaged across dataset splits and experimental runs. Significant differences between IID and non-IID are indicated by \dubbelop \ (p $<$ 0.05).}
	\begin{tabularx}{\textwidth}{XXXXX}
		\toprule
		ranker&data types&\emph{perfect}&\emph{navigational}&\emph{informational} \\
		\midrule
		\emph{linear}&IID& 1002.36 {\tiny \dubbelop} & 872.12 {\tiny \dubbelop} & 894.95 {\tiny \dubbelop} \\
		&non-IID& 648.71 & 546.25 & 566.23 \\
		\midrule
		\emph{neural}&IID& 1061.57 {\tiny \dubbelop} & 834.08 {\tiny \dubbelop} & 842.87 {\tiny \dubbelop} \\
		&non-IID& 668.38 & 505.64 & 490.29 \\
		\bottomrule
	\end{tabularx}
	\label{tbl:dps_online}
\end{table}

\subsection{Dealing with Type 1 non-IID Data} \label{sec:type1dealing}
The employed state-of-the-art FPDGD method is based on the FedAvg algorithm. The fact that FPDGD is affected by non-IID data may be due to the underlying federation algorithm, i.e. FedAvg itself. In federated learning literature, variations of this federation algorithm have been proposed to tackle the non-IID data problem directly. We select two of such methods, FedProx~\cite{li2020federated} and FedPer~\cite{arivazhagan2019federated}, and adapt them to the FPDGD method.

%As shown in the last section, the FedAvg algorithm itself can not handle the performance decrease caused by the non-IID issue. In this section, we analyse the effectiveness of two state-of-the-art, non-IID targeted FL algorithms (FedProx~\cite{li2020federated} and FedPer~\cite{arivazhagan2019federated}) on FOLTR by directly implementing them to our simulated non-IID setting under Type 1.

\begin{figure*}[t]
	\centering
	\begin{subfigure}{1\textwidth}\centering
		\includegraphics[width=1\textwidth]{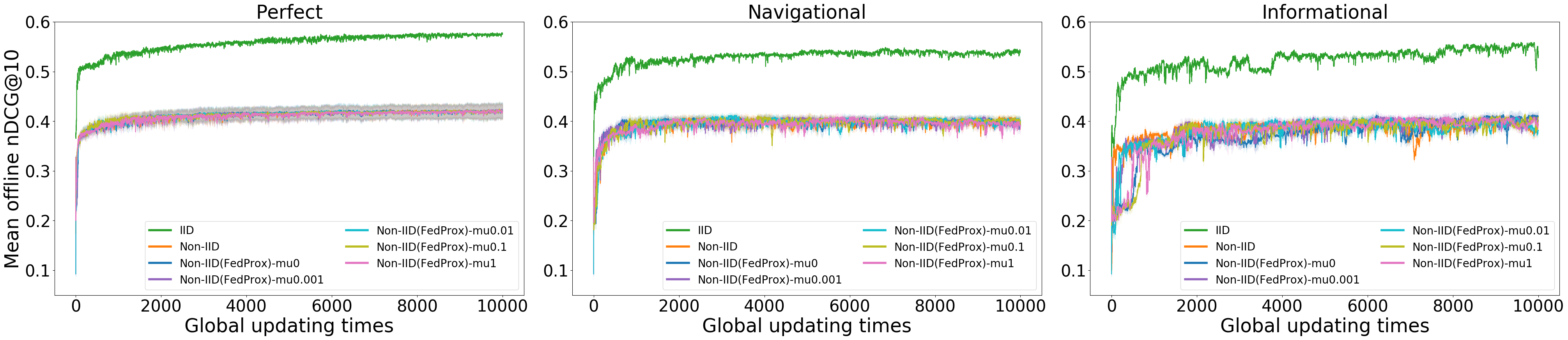}
		\caption{intent-change (linear ranker) - FedProx}
		\label{fig:intent-linear-fedprox} 
	\end{subfigure}
	\begin{subfigure}{1\textwidth}\centering
		\includegraphics[width=1\textwidth]{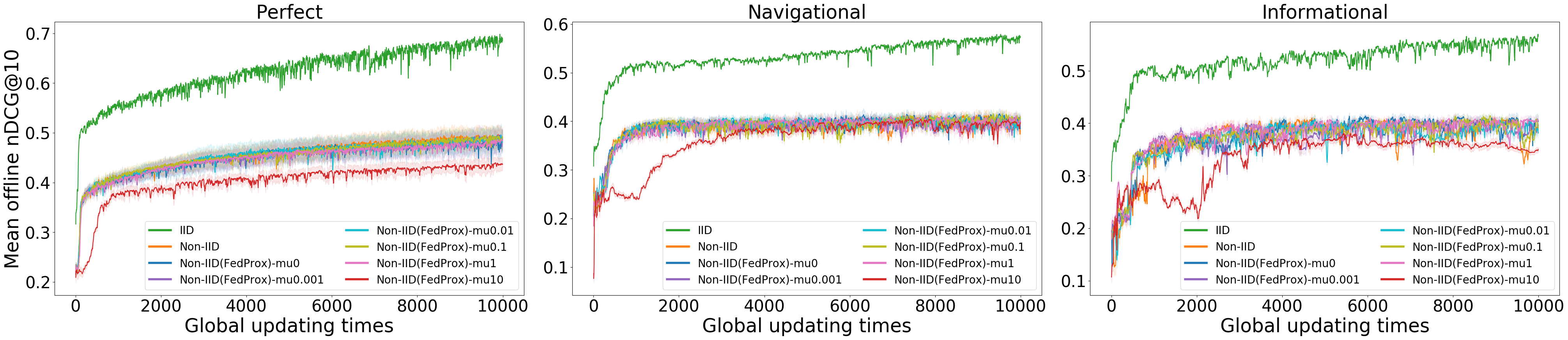}
		\caption{intent-change (neural ranker) - FedProx}
		\label{fig:intent-neural-fedprox}
	\end{subfigure}
	\begin{subfigure}{1\textwidth}\centering
		\includegraphics[width=1\textwidth]{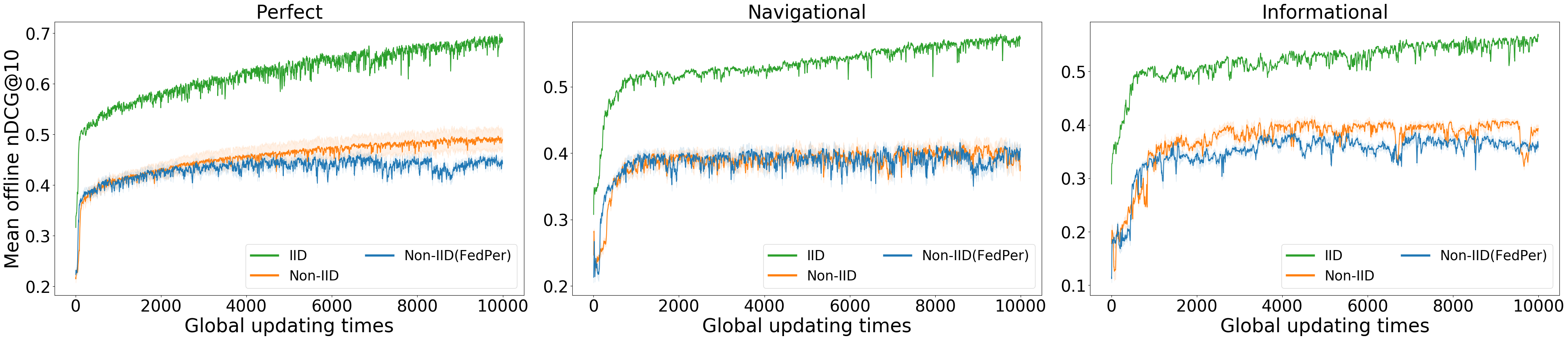}	\caption{intent-change (neural ranker) - FedPer}
		\label{fig:intent-neural-fedper} 
	\end{subfigure}		
	\caption{Offline performance on Type 1 data for FedProx and FedPer; results averaged across dataset splits and experimental runs.
		\label{fig:dps-methods}}
\end{figure*}

FedProx~\cite{li2020federated} improves the local objective of FedAvg. Specifically, it introduces an additional $L_2$ regularisation term (weighted according to a hyper-parameter $\mu$) in the local objective function to limit the distance between the local model and the global model. Thus the use of FedProx adds little computational overhead. However, the main drawback is that the hyper-parameter $\mu$ needs to be carefully tuned: a large $\mu$ may slow the convergence by forcing the updates to get close to the initial point, while a small $\mu$ may not make much difference compared to the use of FedAvg.

FedPer~\cite{arivazhagan2019federated} tackles the presence of non-IID exclusively for deep neural networks by separating them into base layers and personalisation layers.
The base layers are trained collaboratively through FedAvg, where all clients share the same base layers. Instead, the personalisation layers are trained locally using the clients' local data with stochastic gradient descent (SGD). This procedure works as follows: after initialisation, each client merges and updates its base and personalised layers locally using an SGD style algorithm. Each client only sends its base layers to the global server. The server updates the globally-shared base layers using FedAvg and sends back again the updated ones to each client. Intuitively, the base layers are updated globally to learn common high-level representations. In contrast, the distinct personalisation layers never leave the local device and capture the personalisation aspects required by the clients. Except for the training and the maintenance of the local personalisation layers, FedPer is quite similar to FedAvg. FedPer, however, reduces the communication costs as only part of the whole model is transferred and has shown enhanced learning performance under highly skewed non-IID data~\cite{arivazhagan2019federated}.

Our experimental results on FedProx and FedPer are shown in Figure~\ref{fig:dps-methods}; for FedProx we explored $\mu \in \{0.001, 0.01, 0.1, 1, 10\}$. The results clearly show that these federated learning methods, which successfully deal with non-IID data in general machine learning tasks, are not effective in the FOLTR context. In fact, not only do these methods not overcome the gap in effectiveness between IID and non-IID setups, but they even only provide limited improvements, if any, compared to FPDGD with FedAvg. This is an important finding because: (1) it shows a realistic case in which non-IID data largely affects FOLTR effectiveness, and (2) it shows that current methods developed in general FL for non-IID data do not work in FOLTR. Thus, a strong need for new methods specialised in the FOLTR settings emerges from these findings.

%For FedProx under different hyper-parameter, it is clearly shown that neither of $\mu$ from candidate set $\{0.001, 0.01, 0.1, 1, 10\}$ has higher performance than the original non-IID setting for both linear and neural ranker. Similar results are also concluded from FedPer experiments: there is no significant gain in performance under the neural ranker.

\section{TYPE 2: Document Label Distribution} \label{sec:type2}

\emph{Document label distribution skewness} (Type 2) is a widely recognised type of non-IID data type in federated learning. In this setting, the label distributions $P_i(y)$ in each client are different while the conditional feature distribution $P_i(x|y)$ is shared across the clients. In terms of FOLTR, this is equivalent to the following situation. Assume a document is evaluated across the $r$-level relevance grades, from \emph{not relevant} (0) to \emph{perfectly relevant} ($r-1$); then the label distribution on each client is such that, for client $i$, the probability of holding documents with relevance label $k$ is $P_i(R=k) = p_{k}$, where $\sum_{k=0}^{r-1}p_k = 1$, $\forall k, p_k \in [0,1]$. For different clients $i$ and $j$, $P_i(R=k) \neq P_j(R=k)$.

%\begin{align}
%	\label{eq:reldist}
%	P_i(R=k) = p_{k}  \notag  \\
%	where \sum_{k=0}^{r-1}p_k = 1, \\
%	\forall k, p_k \in [0,1] \notag
%\end{align}

In practice, this may be represented by a situation like the following. Several hospitals are collaboratively creating a FOLTR ranker for clinical-decision-support~\cite{roberts2015overview,roberts2016state}. Certain hospitals hold a significantly larger portion of highly relevant health records for a certain disease, while some only a small fraction. In this circumstance, the document label distribution is skewed. Under the context of email search~\cite{narang2017large}, different clients might have unique strategies for managing personal emails~\cite{whittaker1996email}. Some clients frequently clean up their inboxes and use folders to organise emails. In contrast, some hardly use folders or delete irrelevant messages, resulting in different label distribution when following a learning-to-rank approach.

%\begin{table}[t]
%	\centering
%	\caption[centre]{Document label distribution on Client $i$}
%	\begin{tabular}{ p{2cm}<{\centering} p{0.7cm}<{\centering} p{0.7cm}<{\centering} p{0.7cm}<{\centering} p{0.7cm}<{\centering} p{1.2cm}<{\centering}}
%		\hline
%		& \multicolumn{5}{c}{Client $i$}\\
%		%\hline
%		\cmidrule(r){2-6}
%		\emph{Relevance label} &$R = 0$&$R = 1$&$R = 2$&$...$&$R = r-1$\\
%		\hline
%		\emph{$P_i(R=k) = p_{k}$} &  $p_{0}$ & $p_{1}$ &  $p_{2}$ &  ... & $p_{r-1}$\\
%		\hline
%	\end{tabular}
%	\label{labels}
%\end{table}

\subsection{Simulating Type 2 non-IID Data}

In this section, we discuss how we synthetically simulate Type 2 non-IID data and IID data to baseline in the FOLTR effectiveness on the MSLR-WEB10K dataset~\cite{DBLP:journals/corr/QinL13}. %[TODO] For these experiments, we use the popular datasets MSLR-WEB10k~\cite{qin2013introducing} (10,000 queries), Yahoo~\cite{chapelle2011yahoo} (29,900 queries) and Istella-S~\cite{lucchese2016post} (33,018 queries). We report the results for MSLR-WEB10k in the paper; results on the other datasets are similar and are provided in the online appendix.
We simulate $|C|$ clients with each client performing $B$ interactions (queries) locally to contribute to each global model update and restrict the global communication times $T = 10,000$.
For simulating querying behaviour, for each client participating in the federated OLTR, we sample $B$ queries randomly, in line with Chapter~\ref{Chap:fpdgd}. For each query, we use the local ranking model (i.e. that held by the client) to rank documents; we limit SERP to 10 documents. 
For the click behaviour, we rely on the same CCM click models as Section~\ref{sec:type1}. We train both a linear ranker and a neural ranker with same configuration as Section~\ref{sec:type1}, following previous OLTR work~\cite{oosterhuis2018differentiable}.

We specifically consider two types of non-IID data for Type 2: non-IID subtype 1 and non-IID subtype 2. The main difference between the two types is the number of different labels (i.e. the graded relevance assessments) in each client's local dataset. Following similar partitioning strategies by Li et al.~\cite{li2022federated}, suppose each client only has data samples for $k$ different labels. We first generate all possible $k$-combinations of the relevance set $R$ and randomly assign to $\binom{R}{k}$ clients. Then, for the query-document pairs of each label, we randomly and equally divide them into the clients who own the label. In this way, the number of labels in each client is fixed, and there is no overlap between the samples of different clients.

In non-IID subtype 1, each client only holds query-document pairs from one specific value of relevance label. We use $\#R = 1$ to denote this partitioning strategy. This federated setup involves $|C|=5$ clients, which is in line with the 5-level relevance judgements of MSLR-WEB10K. We also vary the local updating time $B \in \{2, 5, 10\}$ to investigate the impact of local updating with a fixed global communication time $T = 10,000$. For non-IID subtype 2, each client holds data samples from two relevance labels -- we denote this as $\#R = 2$. We simulate $|C|=10$ clients and for fair comparison between the two non-IID subtypes, we also simulate  $|C|=10$ clients for $\#R = 1$ (with each label distributed on two different clients).
%For the non-IID setting, the data is sorted and partitioned into 5 portions according to relevance labels. Each client is assigned one subset of the training set: that is, a client only holds query-document pairs with one specific value of relevance label. 
%This federated setup involves 5 clients. We also vary the local updating time $B$(batch) in \{2, 5, 10\} to investigate the impact of local updating with fixed global communication time $T = 10,000$. 
%For the implicit feedback in OLTR, we simulate user clicks based the click models illustrated in Section~\ref{experiment}. We train a linear ranker and a neural ranker on the three LTR datasets considered in this paper and evaluate both online and offline performance on the original held-out unseen test set.

%\textcolor{red}{
%non-IID subtype1: each client has data samples with one single label (5 clients).}
%
%\textcolor{red}{
%non-IID subtype2: each client has data samples with two labels (10 clients).}

The IID experimental setting is the same as the non-IID in terms of federation, ranker parameters and evaluation procedure, except that each client now randomly picks a query from the whole training set with all graded judgements during the training period.

\begin{figure*}[t]
	\centering
	\begin{subfigure}{1\textwidth}\centering
		\includegraphics[width=1\textwidth]{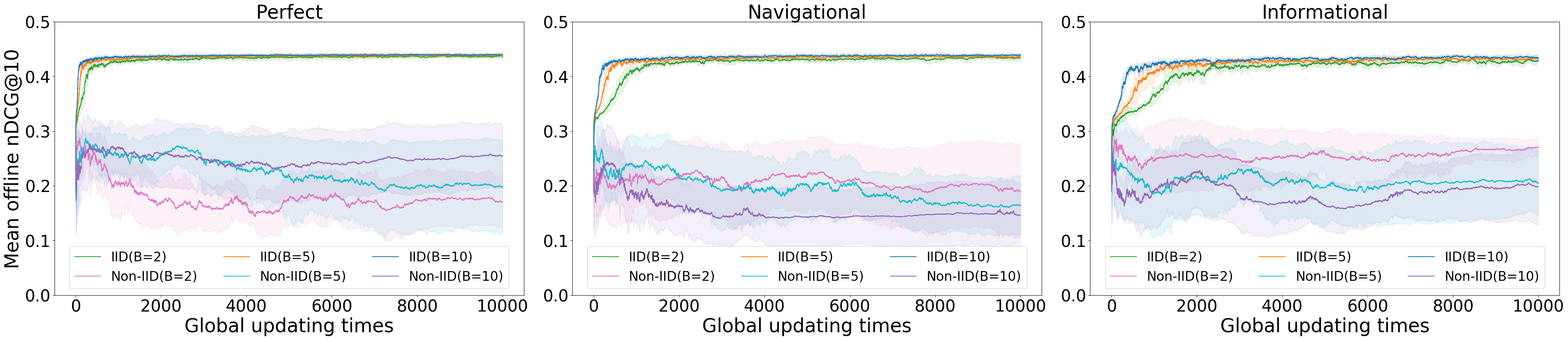} 
		\caption{MSLR-WEB10K (linear ranker)} 
		\label{fig:mslr-r1-linear} 
	\end{subfigure}
	\begin{subfigure}{1\textwidth}\centering
		\includegraphics[width=1\textwidth]{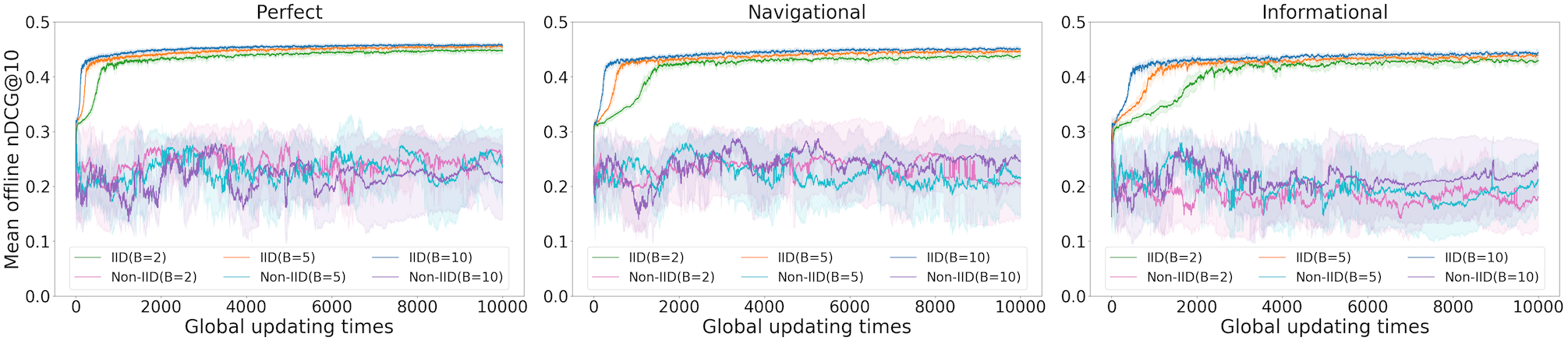}
		\caption{MSLR-WEB10K (neural ranker)}
		\label{fig:mslr-r1-neural} 
	\end{subfigure}
	%	\begin{subfigure}{1\textwidth}\centering
		%		\subfigure[\textbf{Yahoo}] { \label{fig:yahoo-ndcg} \includegraphics[width=12cm]{figures/Yahoo_fndcg.png}} 
		%	\end{subfigure}
	%	\begin{subfigure}{1\textwidth}\centering
		%		\subfigure[\textbf{Istella-S}] { \label{fig:istella-ndcg} \includegraphics[width=12cm]{figures/Istella-s_fndcg.png}} 
		%	\end{subfigure}
	\caption{Offline performance (nDCG@10) on MSLR-WEB10K for Type 2 ($\#R = 1$), under three instantiations of CCM click model and three local updates setting ($B \in \{2, 5, 10\} $); results averaged across all dataset splits and experimental runs.
		\label{fig:lds1}}
\end{figure*}

\begin{figure*}[t]
	\centering
	\begin{subfigure}{1\textwidth}\centering
		\includegraphics[width=1\textwidth]{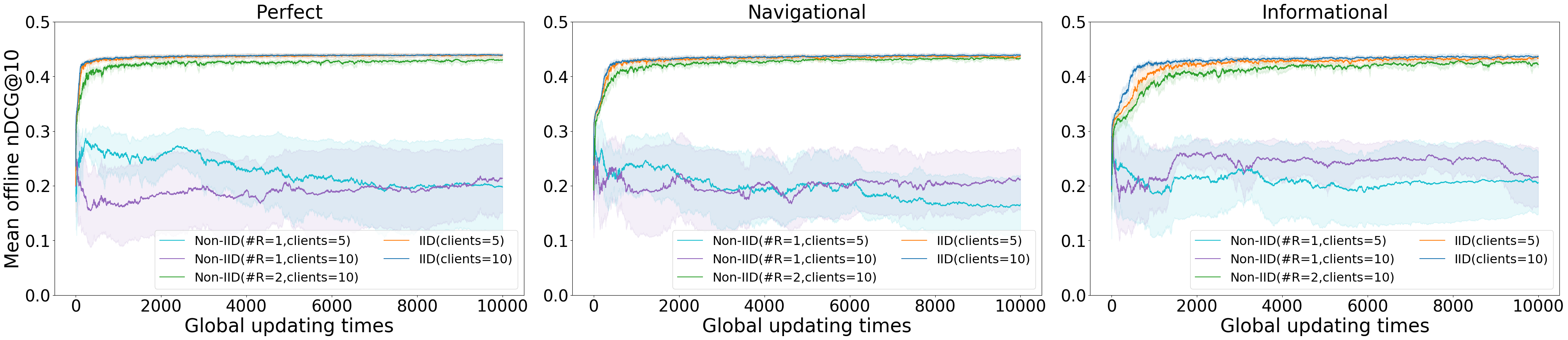}
		\caption{MSLR-WEB10K (linear ranker)}
		\label{fig:mslr-linear}
	\end{subfigure}
	\begin{subfigure}{1\textwidth}\centering
		\includegraphics[width=1\textwidth]{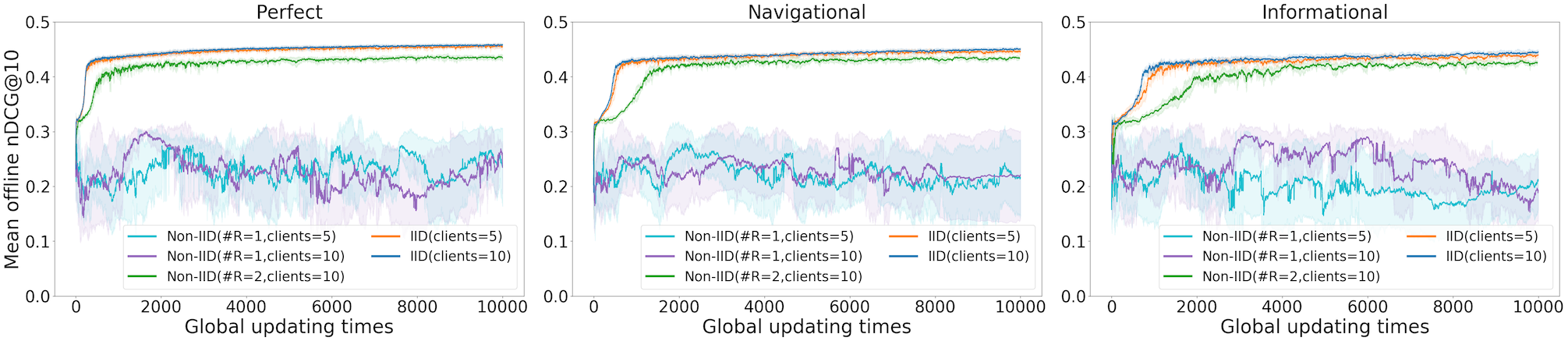}
		\caption{MSLR-WEB10K (neural ranker)}
		\label{fig:mslr-neural}
	\end{subfigure}
	\caption{Offline performance (nDCG@10) on MSLR-WEB10K for Type 2 ($\#R = 2$), under three instantiations of CCM click model with local updates setting ($B = 5$); results averaged across all dataset splits and experimental runs.
		\label{fig:lds2}}
\end{figure*}

\subsection{Impact of Type 2 non-IID Data}

The offline performance for $\#R = 1$ on MSLR-WEB10K is shown in Figure~\ref{fig:lds1} and the corresponding online performance is shown in Table~\ref{tbl:lds_online}. From the offline results, it is clear that the rankers learned with non-IID data under-fit the generalized held-out test set under all three settings of local updating times ($B$). For the linear ranker, a larger number of $B$ achieves better test performance under the \emph{perfect} click model. However, when it comes to noisier clicks (\emph{navigational} or \emph{informational}), the trend is reversed, although differences are minimal and the model performance fluctuates. For the neural ranker, no significant difference is observed from different size of local updates under non-IID settings, which are all outperformed by its IID counterparts.
For the online results, all non-IID settings appear to over-fit the maximum value ($online\_ndcg = 1589.23$) as each client's local data only contains data from one relevance label. 
%(online ndcg = 1589.23 [0.8,0.8,0.8,...])

\begin{table}[t]
	\centering
	\caption{Online performance (discounted cumulative nDCG@10) on MSLR-WEB10K for Type 2 ($\#R=1$), averaged across dataset splits and runs.}
	\begin{tabular}{llcccccc}
		\toprule
		& &\multicolumn{3}{c}{\emph{linear ranker}} & \multicolumn{3}{c}{\emph{neural ranker}}   \\
		\cmidrule(r){2-8}
		&click&$B = 2$&$B = 5$&$B = 10$&$B = 2$&$B = 5$&$B = 10$ \\
		\midrule
		IID&\emph{perfect}& 742.10 & 778.56 & 798.05 & 716.55 & 781.18 & 815.8 \\
		&\emph{navigational}& 698.25 & 743.35 & 771.04 & 649.83 & 728.96 & 775.87\\
		&\emph{informational}& 672.23 & 722.23 & 757.10 & 612.76 & 693.35 & 748.64 \\
		\midrule	
		non-IID&\emph{perfect}& 1589.23 & 1589.23 & 1589.23 & 1589.23 & 1589.23 & 1589.23\\ 
		&\emph{navigational}& 1589.23 & 1589.23 & 1589.23 & 1589.23 & 1589.23 & 1589.23\\
		&\emph{informational}& 1589.23 & 1589.23 & 1589.23 & 1589.23 & 1589.23 & 1589.23\\
		\bottomrule
	\end{tabular}
	\label{tbl:lds_online}
\end{table}

Offline results for $\#R = 2$ are shown in Figure~\ref{fig:lds2}. In this case, the effectiveness of the learnt rankers is much higher than for $\#R = 1$: a diversity in labels held by a client prevents major losses in FOLTR effectiveness. This result is also consistent with previous results in general federated learning with non-IID data: Li et al.~\cite{li2022federated} found that the most challenging setting is when each client only has data samples from a single class (label). We further note that another reason for the performance gap is the pairwise loss used in FPDGD~\cite{wang2021effective}: when each client only has one relevance label, it is hard to infer preferences between document pairs (as they both have the same label). However, given labels from two levels of relevance ($\#R = 2$), pairwise differences can be effectively inferred. This suggests that the results obtained here for Type 2 data may not generalise to other FOLTR methods beyond FPDGD if they do not rely on the pairwise preference mechanism. We further note, however, that FPDGD is the current state-of-the-art method and that the only available alternative~\cite{kharitonov2019federated} displays highly variable and sensibly worse performance compared to FPDGD~\cite{wang2021federated,wang2021efficient}. Therefore, new methods of FOLTR also need to be validated in the presence of Type 2 data.

In summary, we find that if data is distributed in a non-IID manner across clients according to Type 2, the effectiveness of FOLTR (and specifically of FPDGD) is seriously affected in the case of $\#R = 1$; however, if $\#R = 2$ then gaps in effectiveness compared to IID settings are minimal.

\subsection{Dealing with Type 2 non-IID Data}

To mitigate the effect of Type 2 non-IID data, we investigate three existing methods from the federated learning literature: Data-sharing, FedProx and FedPer. FedProx and FedPer have been described in Section~\ref{sec:type1dealing}.
Data-sharing was first proposed by Zhao et al.~\cite{zhao2018federated}. They attribute the performance reduction observed on non-IID data to the weight divergence, which is further affected by the divergence between the local data distribution and the overall distribution. They then introduce a straightforward idea to improve FedAvg: slightly reduce the divergence that causes the global model to underperform. This can be achieved as follows. A globally shared dataset $G$ characterised by the overall data distribution is centralised on the server, and a warm-up global model is trained from $G$. Then, a random $\alpha$ proportion of $G$ is sent to all clients to update the local model by both local training data and the shared data from $G$. Lastly, the server aggregates the local models from the clients and updates the global model with FedAvg. Experimental results on other machine learning tasks show that data sharing can significantly enhance the global model performance in the presence of non-IID data. However, the shortcomings are also pronounced. It is challenging to collect uniformly distributed global datasets in real-world scenarios because either the global server needs some prior knowledge about the local data distributions or each client needs to share parts of the local data (violating the privacy requirement underlying FL).

\begin{figure*}[t]
	\centering
	\begin{subfigure}{1\textwidth}\centering
		\includegraphics[width=1\textwidth]{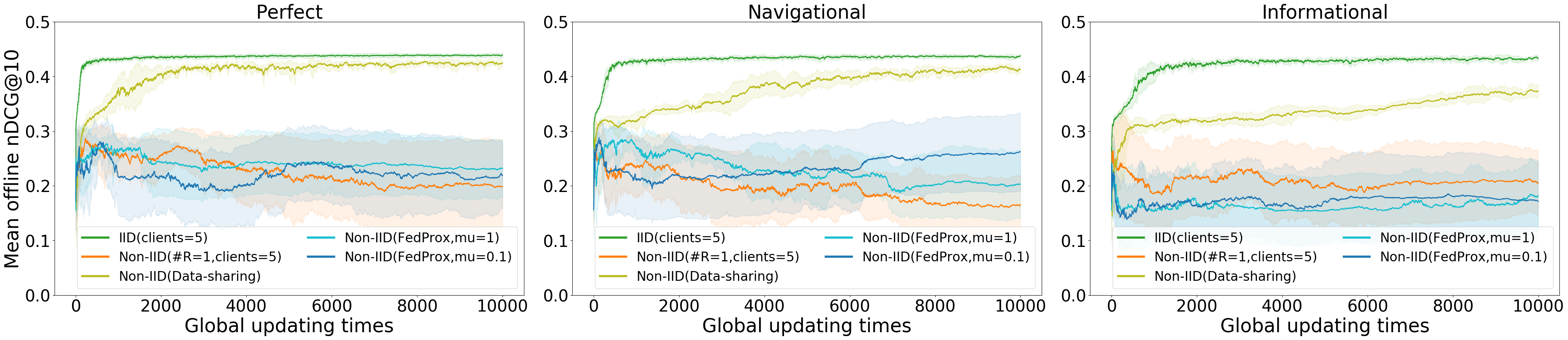}
		\caption{MSLR-WEB10K (linear ranker)}
		\label{fig:mslr-linear-datashare}
	\end{subfigure}
	\begin{subfigure}{1\textwidth}\centering
		\includegraphics[width=1\textwidth]{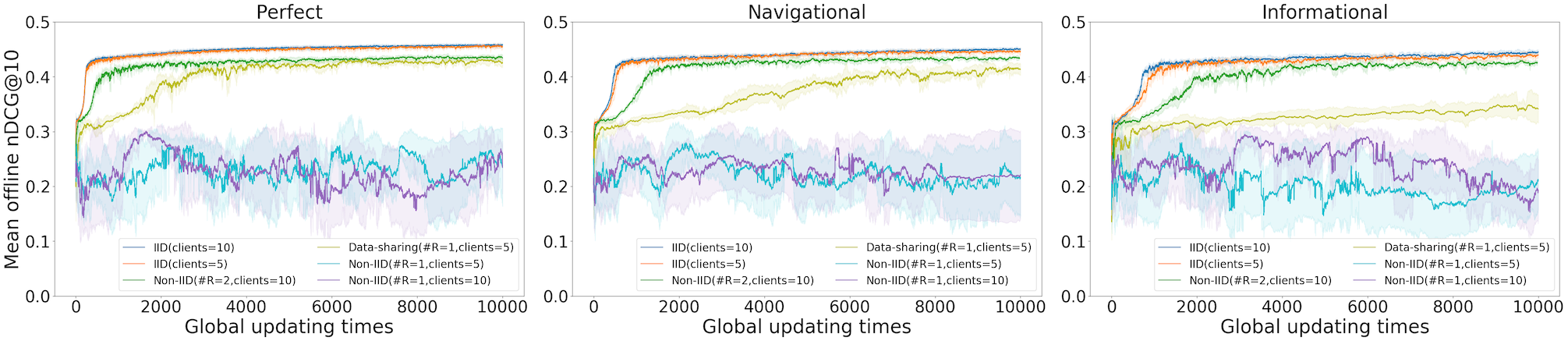}
		\caption{MSLR-WEB10K (neural ranker) - data sharing}
		\label{fig:mslr-neural-datashare}
	\end{subfigure}
	\begin{subfigure}{1\textwidth}\centering
		\includegraphics[width=1\textwidth]{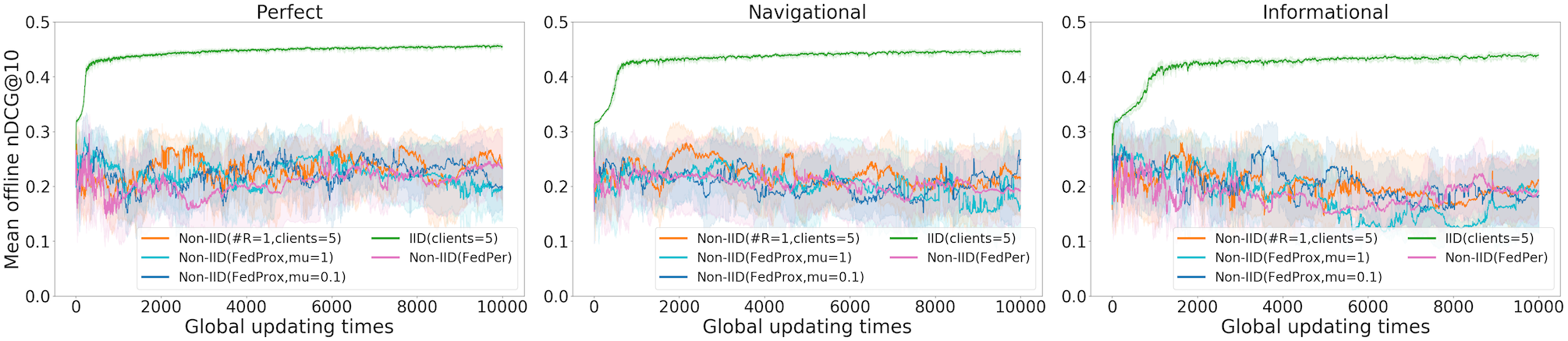}
		\caption{MSLR-WEB10K (neural ranker) - FedProx \& FedPer}
		\label{fig:mslr-neural-fedper}
	\end{subfigure}
	\caption{Offline performance on MSLR-WEB10K when using Data-sharing, FedProx and FedPer on Type 2 non-IID data (with $\#R = 1$); results averaged across dataset splits and experimental runs.
		\label{fig:lds-methods}}
\end{figure*}

Figure~\ref{fig:lds-methods} reports the results for Data-sharing, FedProx and FedPer on MSLR-WEB10K dataset under label distribution skewness $\#R = 1$. We randomly select 10\% of the entire dataset as the globally shared data and simulate $|C|=5$ clients with $B = 5$ local updates before each global update. 
Results show that the global performance can be significantly enhanced with data-sharing for both linear and neural rankers. On the other hand, neither FedProx nor FedPer provides statistically significant gains over the basic FPDGD on Type 2 non-IID data (with $\#R = 1$).

	\section{Other Data Types}
\label{sec-type3-4}
%\todo{@hs: I think you have plenty of room to just use sections for the other two types. I think this may aid in readability.}

%Data type 3, termed as \emph{click preference skewness}, is caused by the noise and biases due to the different click preferences arising from different clients that participate in the FOLTR training

%\subsection{Type 3: Click preferences}
%\label{noniid:cps}

%User click is the implicit feedback which is widely used in OLTR. Though presenting noise and a number of biases~\cite{joachims2017unbiased, pan2007google}, OLTR methods have been learning effectively and overcoming some of those problems~\cite{hofmann2013reusing, schuth2015probabilistic, oosterhuis2016probabilistic, oosterhuis2018differentiable, oosterhuis2021unifying}. Comparing to learning from explicitly labelled data, implicit feedback has the advantage of not requiring costly and lengthy annotated labels. Also it can reveal user's intent change overtime~\cite{zhuang2021how}. 

\begin{figure}[t]
	\centering
	\begin{subfigure}{1\textwidth}\centering
		\includegraphics[width=0.6\textwidth]{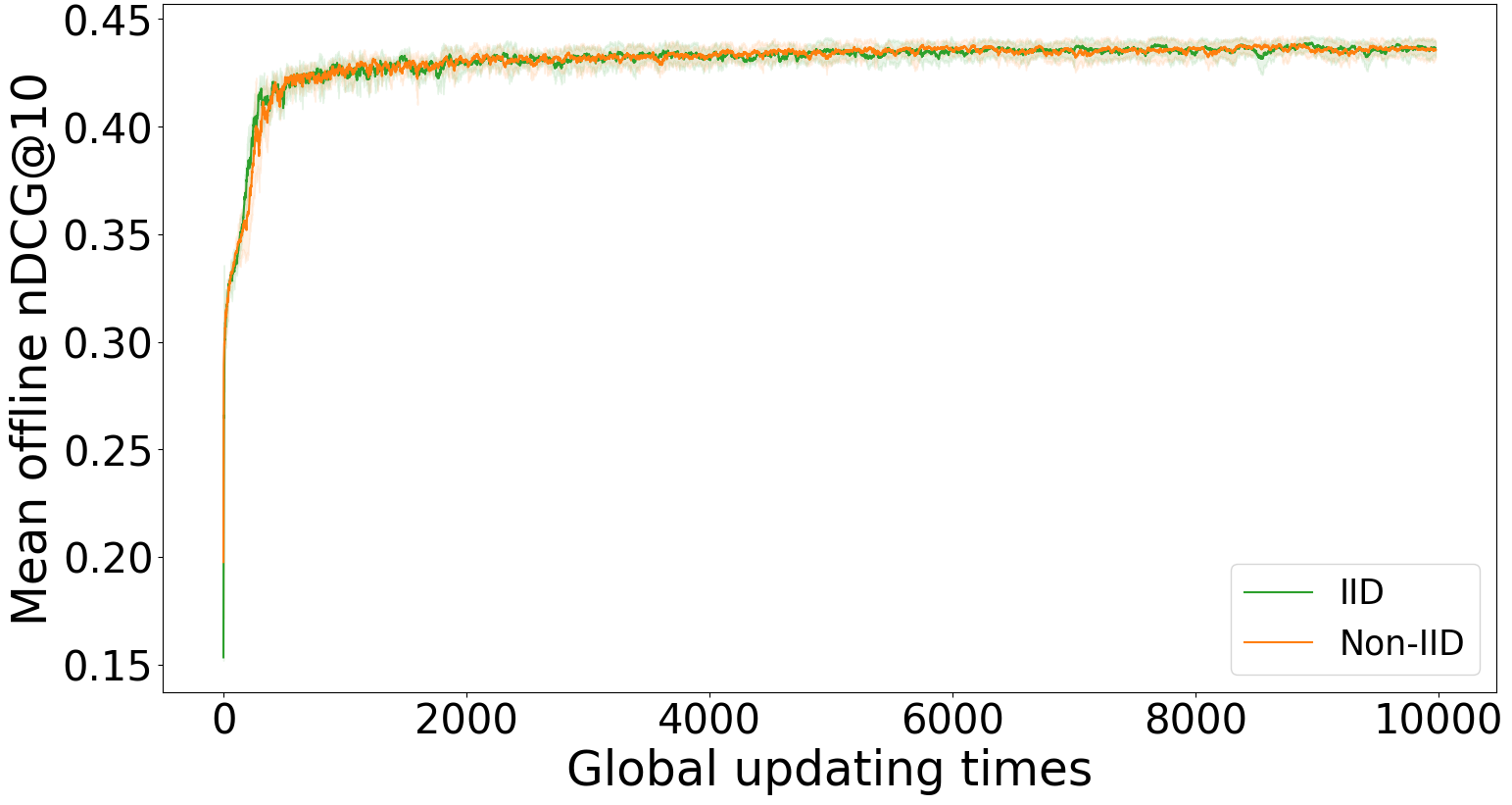}
		\caption{MSLR-WEB10K (linear ranker) under CCM clicks}
		\label{fig:ccm-linear}
	\end{subfigure}
	\begin{subfigure}{1\textwidth}\centering
		\includegraphics[width=0.6\textwidth]{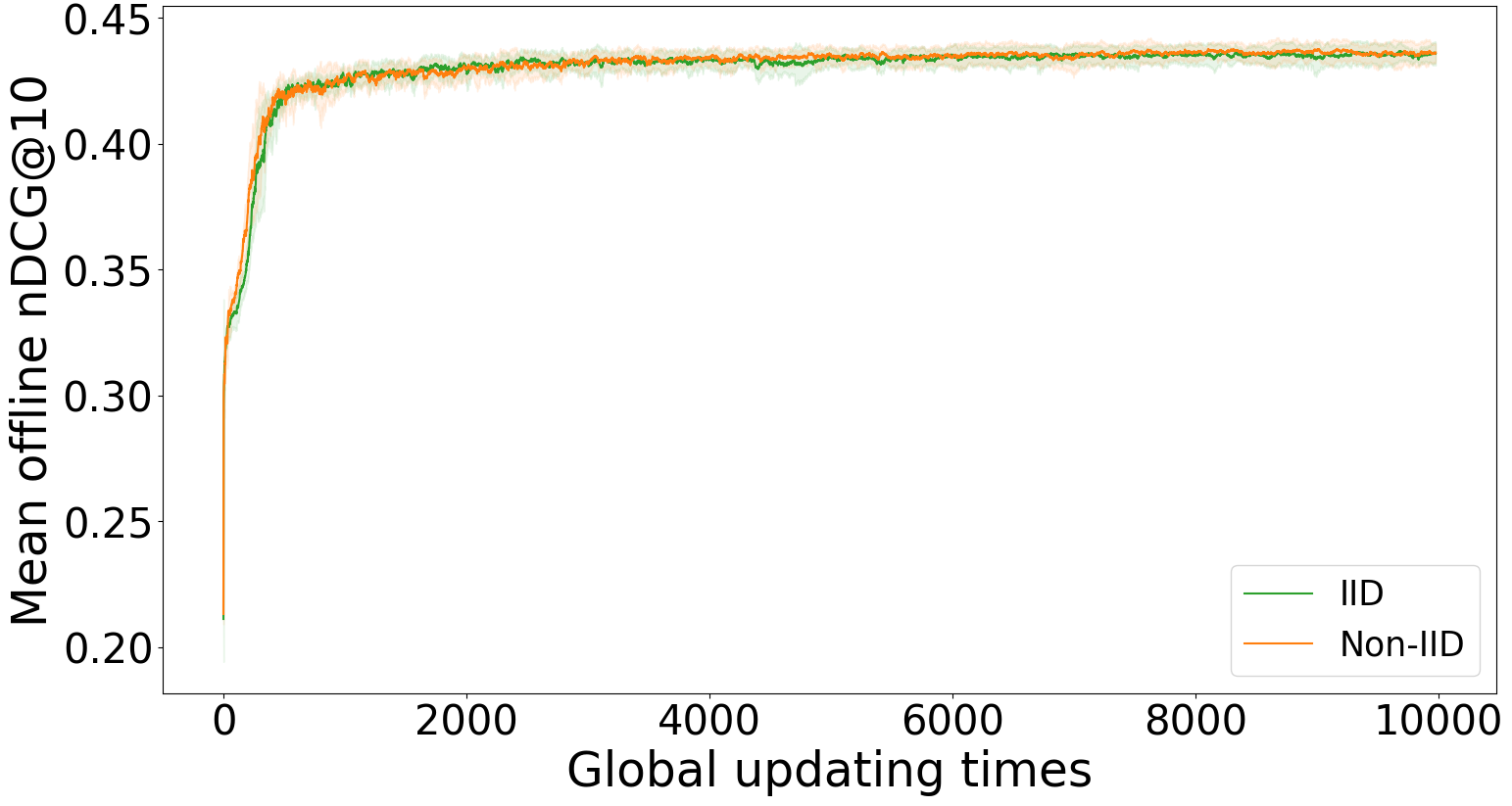}
		\caption{MSLR-WEB10K (linear ranker) under PBM clicks}
		\label{fig:pbm-linear}
	\end{subfigure}
	\caption{Offline performance (nDCG@10) on MSLR-WEB10K for Type 3, separately under CCM and PBM click model; results averaged across all dataset splits and experimental runs.
		\label{fig:cps}}
\end{figure}

\subsection{Type 3: Click Preferences} 
Next, we consider as a source of non-IID data the noise and biases caused by the different click preferences arising from different clients that participate in the FOLTR training; we term this type of non-IID data as \emph{click preference skewness} (Type 3). 

The mechanism to emulate non-IID data of Type 3 and IID data baseline in our FOLTR experiments is as follows.
We study two widely used click models: the \emph{Cascade Click Model} (CCM)~\cite{guo2009efficient}, and the \emph{Position-Based Model} (PBM)~\cite{craswell2008experimental}. For non-IID settings in CCM, each client chooses one of three widely-used instantiations of CCM, namely \emph{perfect, navigational, informational}. For the non-IID settings with PBM, we generate 5 instantiations based on varying the $\eta \in \{0, 0.5, 1, 1.5, 2\}$ parameter: each client is represented by one click type. Thus the federated setup involves 3 clients for CCM clicks and 5 for PBM. We set the local updating time $B = 5$ with fixed global communication times $T = 10,000$. In the IID setting, at every time, each client is simulated based on a click model randomly picked from all click models instantiations detailed above and used in the non-IID setting; this provides a fair comparison between the IID and non-IID settings.
%[TODO ]We experiment on MSLR-WEB10k, Yahoo and Istella-S.
Our experiment on MSLR-WEB10K is shown in Figure~\ref{fig:cps}: the difference between non-IID and IID data for Type 3 is not significant. This might be because different click instantiations only differ from the level of noise and biases they bring, which impact the speed of convergence of the ranker rather than the actual converging direction in FOLTR.                                    

In summary, we find that if data is distributed in a non-IID manner across clients according to Type 3, the effectiveness of FOLTR (and specifically of FPDGD) is not impacted.

\begin{figure}[t]
	\centering
	\begin{subfigure}{1\textwidth}\centering
		\includegraphics[width=1\textwidth]{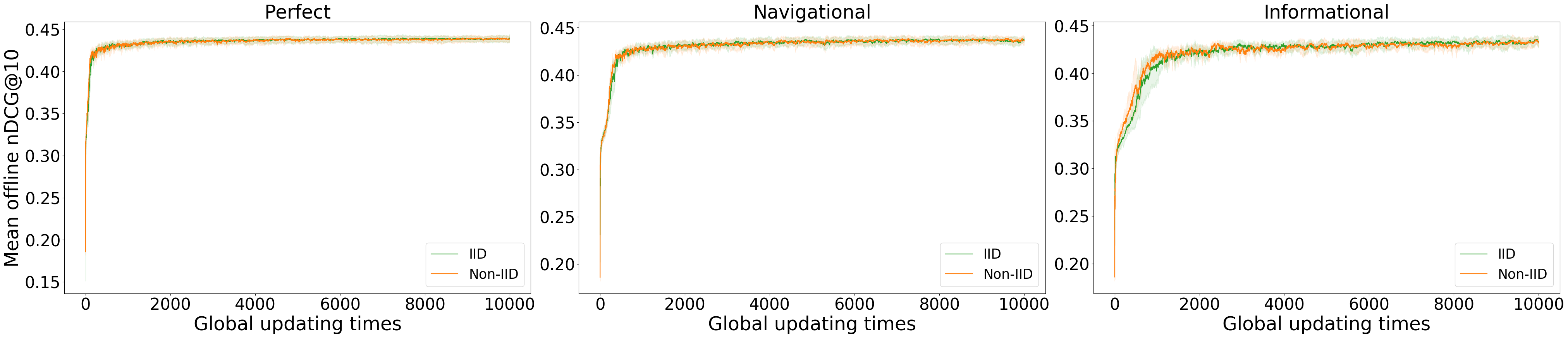}
		\caption{MSLR-WEB10K (linear ranker)}
		\label{fig:mslr-linear}
	\end{subfigure}
	\begin{subfigure}{1\textwidth}\centering
		\includegraphics[width=1\textwidth]{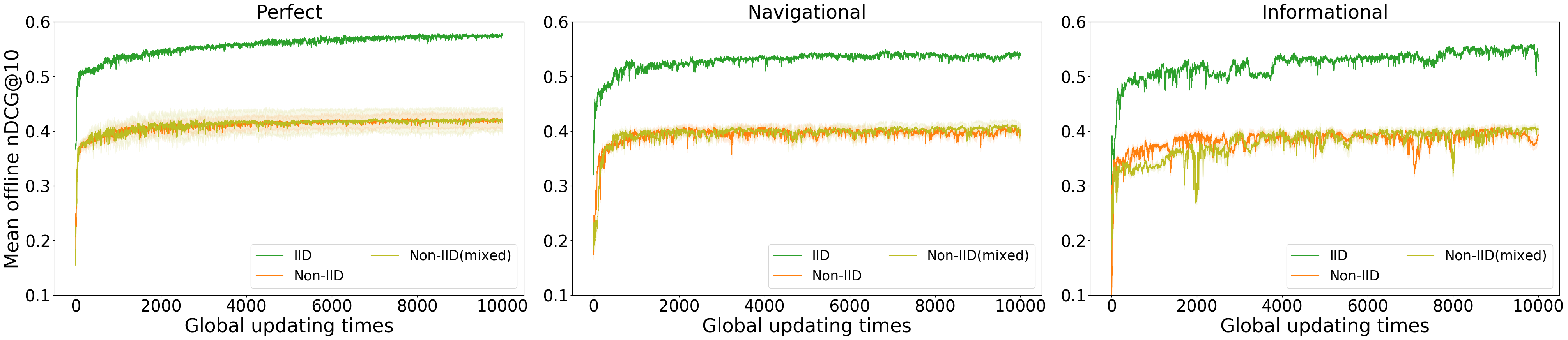}
		\caption{intent-change (linear ranker) mixed with type 1}
		\label{fig:intent-mixed-linear}
	\end{subfigure}
	\begin{subfigure}{1\textwidth}\centering
		\includegraphics[width=1\textwidth]{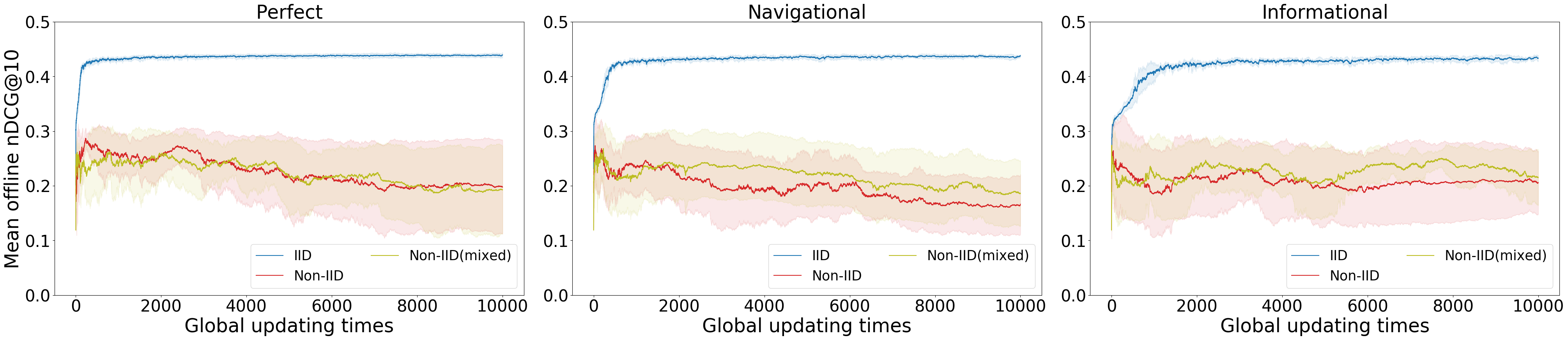}
		\caption{MSLR-WEB10K (linear ranker) mixed with type 2 ($\# R = 1$)}
		\label{fig:mslr-mixed-linear}
	\end{subfigure}
	\caption{Offline performance (nDCG@10) on MSLR-WEB10K and intent-change for Type 4, under three instantiations of CCM click model; results averaged across all dataset splits and experimental runs.
		\label{fig:dqs}}
\end{figure}

\subsection{Type 4: Data Quantity} 

Finally, we consider the case of \emph{data quantity skewness} (Type 4); this occurs when the number of training data varies across different clients. It is a common scenario in real-world applications. For example, in FOLTR, some clients tend to issue more queries and interact more with the searching system than others. Thus, they have more data for training than others. 
The situation represented by Type 4 may occur in combination with the other data types. In our empirical experiments, we have studied Type 4 data both on its own and combined with the document preferences skew (Type 1) and the document label distribution skew with $\#R = 1$ (Type 2).

%Data quantity skewness represents the scenario where the number of training data varies across different clients. It is a common scenario in real-world applications. For example, in FOLTR, some clients tend to issues more queries and perform more interactions with the searching system than other clients. Thus, they have more data for training than others.

%\subsubsection{\textbf{Mechanisms to emulate non-IID data}}
%\subsubsection{Experimental setup for Type 4}
%The mechanism to emulate non-IID data of Type 4 and IID data baseline in our FOLTR experiments is as follows. Except for the normal quantity imbalance, we also consider this type as an adjunctive skew that occur in two aforementioned non-IID types that have demonstrated severe impact: the document label distribution skew with $\#R = 1$ (type 1) and document preferences skew (type 2), to see the effect of the data quantity of each client to the previous  types. 

Type 4 data is simulated by assigning different numbers of queries ($Q$) to each client during the same local updating period, thus leading to different local updating times for each client. The number of queries varies in \{1, 3, 5, 7, 9\} and we simulate $|C|=5$ clients in total with fixed global communication times $T = 10,000$. 
%[TODO]Experiments are carried out on MSLR-WEB10k, Yahoo and Istella-S.

When mixing other non-IID types with Type 4, we follow the same experimental settings of previous non-IID types, and we also assign different numbers of queries to each client (from \{1, 3, 5, 7, 9\}) during the same local updating period. Instead, in the IID simulation, each client has 5 iterations of searching for different queries. For both IID and non-IID, we use CCM click models for click simulation and train a linear ranker using FPDGD on MSLR-WEB10K for Type 1 with $\#R = 1$, and the dataset from Zhuang and Zuccon~\cite{zhuang2021how} (intent-change) for Type 2.

Empirical results on MSLR-WEB10K and intent-change show that if data is distributed in a non-IID manner across clients according to Type 4, the effectiveness of FPDGD is not impacted. We stress that this result may be specific to FPDGD because it uses the FedAvg paradigm and does not generalise to other FOLTR methods.

\section{Chapter Summary}

%The goal of FOLTR is to learn an effective ranker in a federated (without the need for searchable and interaction data to reside on a central server) and online (by exploiting users clicks on SERPs as they occur) manner. In such a FOLTR setup, user data and interactions reside with the user's client, and not in a central server. Clients then do not need to share such data. Instead, they only share ranker updates with a central server whose responsibility is to collect such updates from the clients and aggregate them into a global model. The global model is then pushed back to the clients in an iterative manner as search interactions occur.

Despite federated learning receiving substantial attention, research in FOLTR is still in its early stages, with only two methods available at the time of writing~\cite{kharitonov2019federated,wang2021effective}. Importantly, studies that have proposed FOLTR methods have ignored an important issue that has been shown to affect the performance of federated learning systems~\cite{zhu2021federated}: that of the data not being distributed across the federated clients in an identical and independent manner (non-IID data). This chapter provides the first analysis of the impact of non-IID data on FOLTR thus charts directions for future research. Specifically, we develop a benchmark and introduce four types of non-IID data in FOLTR system. Furthermore, we conduce comprehensive experiments on simulating non-IID data and implementing existing remedy methods. We found that certain types of non-IID data present a threat to the performance of FOLTR and existing solutions employed in general federated learning to mitigate the non-IID data problem do not apply to the FOLTR setting. This study sheds light on future directions in addressing the awareness of non-IID issues and improving robustness to non-IID data for FOLTR.

\cleartoevenpage
\pagestyle{empty}	
%Use this command (above) to suppress the header from the preceding chapter.

\noindent
The following publication has been incorporated as Chapter~\ref{Chap:foltr-poisoning}.

\noindent
1.~\cite{wang2023analysis} \textbf{Shuyi Wang}, and Guido Zuccon, \href{https://doi.org/10.1145/3578337.3605117}{An Analysis of Untargeted Poisoning Attack and Defense Methods for Federated Online Learning to Rank Systems}, \textit{In Proceedings of the 2023 ACM SIGIR International Conference on Theory of Information Retrieval (ICTIR)}, pages 215–224, 2023

%\begin{table}[h]
%	\begin{center}
%	\begin{tabular}{|c|l|l|}
%		\hline
%		Contributor & Statement of contribution & \% \\
%		\hline
%		\textbf{Your Name}				& writing of text 					& 70\\
%															& proof-reading							& 60 \\
%															& theoretical derivations 	& 70\\
%															& numerical calculations 		& 100\\
%															& preparation of figures 		& 80 \\
%															& initial concept						& 10 \\
%		\hline
%		Co-author 1								& writing of text 					& 20\\
%															& proof-reading							& 10 \\
%															& supervision, guidance 		& 20\\
%															& theoretical derivations 	& 10\\
%															& preparation of figures 		& 20 \\
%															& initial concept						& 10 \\
%		\hline
%		Final Author							& writing of text 					& 10\\
%															& proof-reading							& 30 \\
%															& supervision, guidance 		& 80 \\
%															& theoretical derivations 	& 20 \\
%															& preparation of figures 		& 10 \\
%															& initial concept						& 80 \\
%		\hline
%	\end{tabular}
%	\end{center}
%\end{table}
%
%If your task breakdown requires further clarification, do so here. Do not exceed a single page.

% ***************************************************
% Example of an internal chapter
% ***************************************************
%This is an internal chapter of the thesis.
%If you have a long title, you can supply an abbreviated version to print in the Table of Contents using the optional argument to the \chapter command.
\chapter[FOLTR under Poisoning Attacks]{Federated Online Learning to Rank under Poisoning Attacks}
\label{Chap:foltr-poisoning}	%CREATE YOUR OWN LABEL.
\pagestyle{headings}

In the previous chapters, we address the effectiveness and robustness under non-IID data in FOLTR.
Existing FOLTR systems however are not necessarily secure: the federation mechanism provides malicious clients with opportunities for attacking the effectiveness of the global ranker. 
For example, malicious clients can send arbitrary weights to the server so that the convergence of the global ranker can be perturbed after aggregation. This kind of attack is termed as \textit{untargeted poisoning attack} and aim to compromise the integrity of the global model trained federatively~\cite{bagdasaryan2020backdoor}.
% we fill this broad gap
This issue is critical for federated learning systems, but it has not yet been studied for FOLTR. In this chapter, we initiate the investigation of poisoning attacks and corresponding defense methods in the context of FOLTR systems. Specifically, we investigate the following research questions:

\begin{enumerate}[label=\textbf{RQ3:}, leftmargin=*]
	\item \bfseries Are FOLTR methods secure in the face of potential risks brought by malicious participants? \itshape
	\begin{enumerate}[label=RQ3.\arabic*:, leftmargin=*]
		\item Can data poisoning and model poisoning strategies undermine ranking performance on FOLTR?
		\item Can defense strategies mitigate the impact of malicious attacks in FOLTR?
	\end{enumerate}
\end{enumerate}

%In general federate learning domain, attacking and defensing methods have been studied 
%~\cite{bonawitz2019towards, lamport1982byzantine, blanchard2017machine, yin2018byzantine, fang2020local, shejwalkar2021manipulating}. .
%
%A specific class of attacks in this context is that of poisoning attacks, which aim to compromise the integrity of the global model trained federatively~\cite{bagdasaryan2020backdoor}.
%
Outside of FOLTR systems, poisoning attacks on federated learning systems has been shown successful in compromising model integrity across several federated machine learning tasks~\cite{biggio2012poisoning, shafahi2018poison, bhagoji2019analyzing, lyu2020threats, yu2023untargeted}, including in natural language processing and recommender systems. 
% defense
To mitigate or remove the threat posed by poisoning attacks, defense strategies have been designed and optimised~\cite{blanchard2017machine, yin2018byzantine, guerraoui2018hidden}. 
Defense strategies typically act upon the aggregation rules used in the global model updating phase.
The vulnerability of existing FOLTR methods to these attacks and the effectiveness of the related defense mechanism is unknown. 
As shown in the previous chapter (Chapter~\ref{Chap:foltr-noniid}~\footnote{We find that methods for dealing with non identical and independently distributed data in federated learning systems do not generalise to the context of FOLTR. }), findings obtained with respect to federated learning in other areas of Machine Learning or Deep Learning do not directly translate to the online learning to rank context, and therefore the study of these techniques in the context of FOLTR is important. 
%While the effect of poisoning attacks and the corresponding defense strategies have been subject of investigation within machine learning and specific natural language processing and recommender systems tasks~\cite{yoo2022backdoor, zhang2022pipattack}, the vulnerability of existing FOLTR methods to these attacks and the effectiveness of the related defense mechanism is unknown. Previous work in FOLTR has shown that findings obtained with respect to federated learning in other areas of Machine Learning or Deep Learning do not directly translate to the online learning to rank context, and therefore the study of these techniques in the context of FOLTR is important. For example, Wang et al.~\cite{wang2022non} have found that methods for dealing with non identical and independently distributed data in federated learning systems do not generalise to the context of FOLTR. 
%
Specifically, the performance of poisoning attacks and defense methods proposed in general domain cannot be guaranteed when applied to FOLTR: we address this limitation by adapting and investigating these methods to the setting of FOLTR and establish baselines for future studies. Within this scope, to answer RQ3.1, we experiment with and analyse data and model poisoning attack methods to showcase their impact on FOLTR search effectiveness. We further explore the effectiveness of defense methods designed to counteract attacks on FOLTR systems, with the aim of answering RQ3.2. We thus contribute an understanding of the effect of attack and defense methods for FOLTR systems, as well as identifying the key factors influencing their effectiveness.

\begin{figure}[b]
	\centering
	\includegraphics[width=0.8\columnwidth]{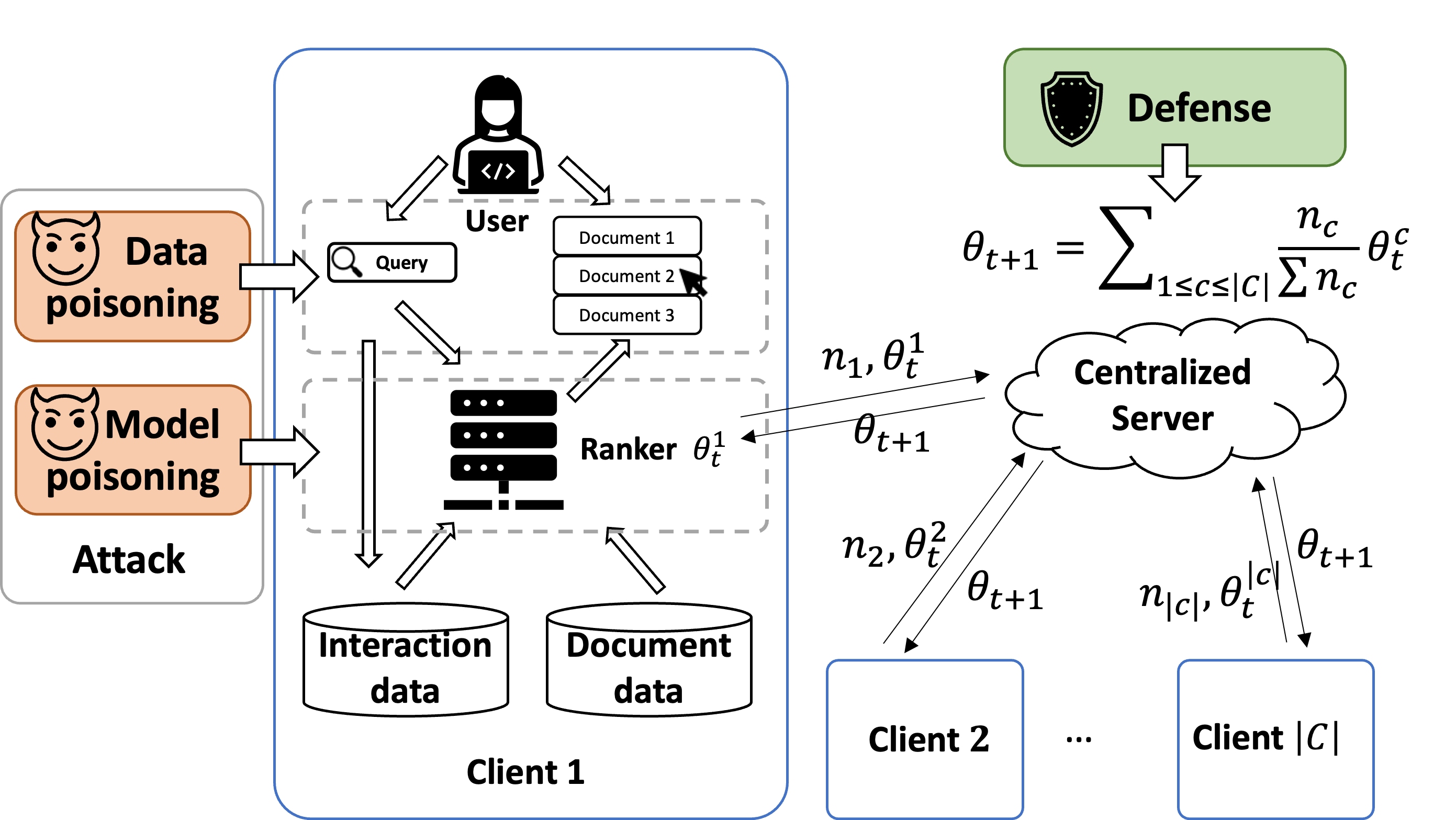}
	\caption{Overview of a FOLTR system with attack and defense modules (the arrows point to where these modules will be applied to).
	\label{fig:foltr:chap5}}
\end{figure}

	\section{PRELIMINARIES}
\label{foltr}

%formally define our research problem
%
%\subsection{FOLTR}

\subsection{Online Learning to Rank (OLTR)}
In OLTR, the ranker is learned directly from user interactions (clicks in our study), rather than editorial labels.
%instead of using manually annotated labels, the ranker learns from direct user interactions (for example, clicks). For simplicity, we use $clicks$ to represent the feedbacks from user interactions. 
In this context, each client performs searches on several queries during each local training phase. For each query $q$, the candidate documents set is $D_q$ and the local training data held by each client is $\{(x_i, y_i),  i=1...|D_q|\}_q$ with feature representation ($x_i$) and user's click signal ($y_i$) for each $(q, d_i)$-pair (where $d_i \in D_q$). The value of the click feedback $y_i$ is either 0 (unclicked) or 1 (clicked). In practice, the $click$ is dependent on the relevance degree of the candidate document $d_i$ to the query $q$, the rank position of $d_i$, and other noise or randomness factors.

\subsection{Federated Pairwise Differentiable Gradient Descent (FPDGD)}

%In our setting for FOLTR, a central server coordinates all participating clients without sharing raw data for the training of online rankers. Instead, each participant updates the local ranker with their local online interactions and then passes on the updated local ranker to the central coordinator. After receiving all updates, the server aggregates them together to produce a global ranker update and then sends the newly-updated global ranker back to each client side so as to perform an update on the local ranker. The training then proceeds iteratively.

%In this section, we briefly describe the state-of-the-art FOLTR method we use in our experiments:

We add our attacking and defense modules to the current state-of-the-art FOLTR system,
the Federated Pairwise Differentiable Gradient Descent (FPDGD)~\cite{wang2021effective}, which is outlined in Algorithm~\ref{alg:fpdgd} and illustrated in Figure~\ref{fig:foltr:chap5}.
%
%This method uses Pairwise Differentiable Gradient Descent (PDGD)~\cite{oosterhuis2018differentiable} algorithm to optimise the local ranking model of each client 
%
%and leverages the widely-used Federated Averaging~\cite{mcmahan2017communication} to aggregate the local model updates. 
%
%The use of this algorithm in FPDGD is shown in Algorithm~\ref{alg:fpdgd}. 
%
Within each iteration $t$, each client $c$ considers $n_c$ interactions and updates the local ranker using Pairwise Differentiable Gradient Descent (PDGD)~\cite{oosterhuis2018differentiable}. 
After the local update is finished, each client sends the trained weights $\theta^c_t$ to the server. 
The server then  leverages the widely-used Federated Averaging~\cite{mcmahan2017communication} to aggregate the local model updates. 
Afterwards, the new global weights $\theta_{t+1}$ are sent back to the clients as their new local rankers. We refer the reader to Section~\ref{sec:PDGD} for more details on FPDGD~\cite{wang2021effective}. 

%
%The local model updates are considered in steps 7-8. 
%
%These updates are obtained by executing the gradient update process of the underlying PDGD algorithm within each local client;
%
% this update is outlined in Algorithm~\ref{alg:pdgd}.

\begin{algorithm}[t]
	\caption{FederatedAveraging PDGD. \\
		$\bullet$ set of clients participating training: $C$, each client is indexed by $c$; \\
		$\bullet$ number of local interactions for client $c$: $n_c$; \\
		$\bullet$ local interaction set: $B$, model weights: $\theta$.}
	\label{alg:fpdgd}
	\begin{algorithmic}[1]
		\SUB{Server executes:}
		\STATE initialize $\theta_0$; scoring function: $f$; learning rate: $\eta$
		\FOR{each round $t = 1, 2, \dots$}
		\FOR{each client $c \in C$ \textbf{in parallel}}
		\STATE $\theta_{t}^c, n_c \leftarrow \text{ClientUpdate}(c, \theta_t)$
		\ENDFOR
		\STATE $\theta_{t+1} \leftarrow \sum_{c=1}^{|C|} \frac{n_c}{\sum_{c=1}^{|C|} n_c} \theta_{t}^c$
		\ENDFOR
	\end{algorithmic}
	\[\]

	\begin{algorithmic}[1]	
	\SUB{ClientUpdate($c, \theta$):} \ \ \ // \emph{Run on client $c$}
	\FOR{each local update $i$ from $1$ to $B$}
	\STATE $\theta \leftarrow \theta + \eta\nabla f_\theta$    \hfill \textit{\small //PDGD update with data from $B$} %shown in Algorithm.~\ref{alg:pdgd}}
	\ENDFOR
	\STATE return ($\theta, n_c$) to server
	\end{algorithmic}	
\end{algorithm}

	\section{Attacks to FOLTR Systems}
\subsection{Problem Definition and Threat Model}

%\textbf{Attacker's goal:} We assume the aim of the attacker is to conduct an untargeted poisoning attack with the goal of reducing the overall effectiveness of the global ranking model. This can be achieved through manipulation of either the training data (data poisoning) or the training process (model poisoning).

We briefly introduce the attacker's capability and background knowledge in our study.

\textbf{Attacker's capability:} Poisoning attacks can come from both members (insiders) and non-members (outsiders) of the FOLTR system. Insiders include both the central server and the clients, while outsiders include eavesdroppers on communication channels and users of the final ranker (this is similar to adversarial attacks during inference).
%Poisoning attacks on FL can be carried out by insiders or outsiders. Insiders include the central server and participants in the FL system. Outsiders include the eavesdroppers on the communication channel between participants and the server, and users of the final FL model when it is deployed as a service (This case is similar as adversarial attack which conducts on the inference stage). Insider attacks are generally stronger than outsider attacks\cite{lyu2020threats}. 
In this study, we focus on insider attacks by malicious participants in the FOLTR system since insider attacks are generally more effective than outsider attacks~\cite{lyu2020threats}. We assume the attacker has control over $m$ collusive clients, which means that the training data and local model updates can be exchanged among the malicious clients. We restrict the percentage of collusive clients to less than 50\%: higher amounts would make it trivial to manipulate the global model.
% as Federated Averaging is the most commonly-used aggregation rule in FL.

\textbf{Attacker's background knowledge:} 
%The attacker can only manipulate the local training process. 
We assume that the attacker has only access to the compromised clients: the training data and local rankers of all remaining clients remain not accessible to the attacker.
Thus, the attacker has limited prior knowledge: the training data and the locally updated models from the poisoned clients, and the shared global model. 
%In our investigation, we principally consider settings in which the attacker has limited prior knowledge. 
The exception of having full prior knowledge\footnote{i.e. the attacker can also access information (training data, model updates) of non-poisoned clients.} will only be for the purpose of analysis and will be clarified in place.

%However, we do also consider the case the attacker has full prior knowledge\footnote{i.e. the attacker can also access information (training data, model updates) of non-poisoned clients.} when we examine a specific model poisoning attack method, the one by Fang's attack which we introduce in Section~\ref{sec:fangattack}.

\textbf{Problem Formulation:} Assume $n$ clients are involved in the FOLTR system. Among them, $m$ clients are malicious.
%, and they can share the respective local training data and model updates among themselves. 
Without loss of generality, we assume the first $m$ participants are compromised. Be $\mathbf{\theta_i}$ the local model that the $i$-th client sends to the central server. The global ranking model is updated through aggregating all $\mathbf{\theta_i}$:

\begin{equation}
	\mathbf{\theta_g} = agg(\mathbf{\theta_1}, ..., \mathbf{\theta_m}, \mathbf{\theta_{m+1}}, ..., \mathbf{\theta_n})
	\label{eq:agg}
\end{equation}

%In Equation~\ref{eq:agg}, the first $m$ models are compromised by the attackers using either data poisoning or model poisoning strategies.

\subsection{Data Poisoning}
\label{data-poison}

Data poisoning methods aim to corrupt the training data in order to degrade the model's effectiveness. This can be done by adding malicious instances or altering existing instances in an adversarial manner.

%In data poisoning methods, the attacker wants to compromise the integrity of the training data so as to compromise performance of the model, such as adding poison instances or adversarially changing existing instances. 

%Recall that in online learning to rank, the ranker is learned directly from user interactions (clicks in our study), rather than from editorial labels.
%%instead of using manually annotated labels, the ranker learns from direct user interactions (for example, clicks). For simplicity, we use $clicks$ to represent the feedbacks from user interactions. 
%In this context, each client performs searches on $B$ queries during each local training phase. For each query $q$, the candidate documents set is $D_q$ and the local training data held by each client is $D_{tr} = \{x_i, c_i\}_q$ with $d$-dimension feature representation ($x_i$) and user's click signal ($c_i$) for each $(q, d_i)$-pair (where $d_i \in D_q$). The value of the click feedback $c_i$ is either 0 or 1: $c_i = 1$ represents that the client clicks the document $d_i$ while $c_i = 0$ represents no click on $d_i$. In practice, the $click$ is dependent on the relevance degree of the candidate document $d_i$ to the query $q$, the rank position of $d_i$, and other noise or randomness factors. 

%We now consider how to perform data poisoning in an online learning to rank setting.

Our data poisoning attack to FOLTR is inspired by the label flipping strategy~\cite{biggio2012poisoning, tolpegin2020data}, in which the labels of honest training samples from one class are flipped to another class, while the features of the flipped samples are kept unchanged. 
In our case, we want to change the label of irrelevant documents into "high-relevant" and vice versa, without any changes to the feature representation of the corresponding query-document pairs. 
%Specifically, the attacker manipulates the training results through reversing the click feedbacks for every interaction during the local training phase. We denote the training data on all benign clients as $D_{clean} = \{x_i, c_{clean}\}_q$ and the poisoned training data held by the malicious clients as $D_{poison} = \{x_i, c_{poison}\}_q$. The only difference between benign and malicious clients is their clicking behaviour. 
%
To achieve so, the attacker needs to intentionally flip the feedback by clicking on irrelevant documents to bring arbitrary noise thus poison the training.

In our experiments, as no click data is available with the considered datasets, we follow the common practice from previous literature in OLTR and FOLTR of simulating click behaviour based on the extensively-used \emph{Cascade Click Model} (CCM) with three instantiations (\textit{perfect}, \textit{navigational}, \textit{informational}) as described in Section~\ref{chap2-click}. This click model has been shown to produce reasonable predictions of real-world user click behaviour.

Inspired by the three instantiations, we manipulate one \textit{poison} instantiation to simulate malicious clicking behaviour. The click probability of \textit{poison} instantiation is the reverse version of the \textit{perfect} click behaviour: the highest probability of clicking is associated with the least relevance label. All stop probabilities in \textit{poison} instantiation are set to zero as we assume the attacker wants to poison as many clicks as possible. The values we adopt for the four instantiations of CCM are reported in Table~\ref{tbl:ccm:chap5}.

%In our experimental setup, we simulate benign users using three types of CCM (\textit{perfect}, \textit{navigational}, \textit{informational}) separately with attackers using \textit{poison} clicks solely.

\begin{table}[t]
	\centering
	\caption{Instantiations of CCM click model for simulating user behaviour in experiments. $rel(d)$ denotes the relevance label for document $d$. Note that in the MQ2007 dataset, only three-levels of relevance are used. We demonstrate the values for MQ2007 in bracket.}
	\begin{tabularx}{1\textwidth}{XXXXXX}
		\toprule
		& \multicolumn{5}{c}{$P(\mathit{click}=1\mid rel(d))$}  \\
		%\hline
		\cmidrule(r){2-6}
		\emph{rel(d)} & 0 & 1 & 2 & 3 & 4 \\
		\hline
		\emph{perfect} &  0.0 (0.0)&  0.2 (0.5)&  0.4 (1.0)&  0.8 (-)&  1.0 (-)\\
		\emph{navigational} & 0.05 (0.05)& 0.3 (0.5)&  0.5 (0.95)&  0.7 (-)&  0.95 (-)\\
		\emph{informational} &  0.4 (0.4)&  0.6 (0.7)&  0.7 (0.9)&  0.8 (-)&  0.9 (-)\\
		\emph{poison} &  1.0 (1.0)&  0.8 (0.5)&  0.4 (0.0)&  0.2 (-)&  0.0 (-)\\
		\hline\hline
		& \multicolumn{5}{c}{$P(\mathit{stop}=1\mid click=1,  rel(d))$} \\
		%\hline
		\cmidrule(r){2-6}
		\emph{rel(d)} & 0 & 1 & 2 & 3 & 4 \\
		\hline
		\emph{perfect}  & 0.0 (0.0)& 0.0 (0.0)&  0.0 (0.0)&  0.0 (-)&  0.0 (-)\\
		\emph{navigational} & 0.2 (0.2)&  0.3  (0.5)&  0.5 (0.9)&  0.7 (-)&  0.9 (-)\\
		\emph{informational} & 0.1 (0.1)&  0.2 (0.3)&  0.3 (0.5)&  0.4 (-)&  0.5 (-)\\
		\emph{poison}  & 0.0 (0.0)& 0.0 (0.0)&  0.0 (0.0)&  0.0 (-)&  0.0 (-)\\
		\bottomrule
	\end{tabularx}
	\label{tbl:ccm:chap5}
\end{table}

\subsection{Model Poisoning}
\label{model-poison}

Unlike data poisoning, model poisoning directly modifies the local model updates (through poisoning gradients or model parameter updates) before sending them to the server. Some literature shows that model poisoning is more effective than data poisoning~\cite{bagdasaryan2020backdoor, bhagoji2019analyzing} while it also requires sophisticated technical capabilities and high computational resources than solely poisoning data. In this section, we investigate two existing model poisoning methods.

\subsubsection{\textbf{Little Is Enough (LIE)}}

Baruch et al.~\cite{baruch2019little} find that if the variance between local updates is sufficiently high, the attacks can make use of this by adding small amounts of noise to the compromised local models and bypass the detection of defense methods. 
They provide a perturbation range in which the attackers can successfully poison the learning process. To conduct the attack, the adversaries first compute the average $\mu$ and standard deviation $\sigma$ of the before-attack benign local model updates of all collusive attackers ($\mathbf{\theta_1}, ..., \mathbf{\theta_m}$). A coefficient $z$ is used and computed based on the number of benign and malicious clients. Finally, the local model of attackers is manipulated as $\mathbf{\theta^m_i} = \mu - z\sigma$ for $i \in \{1, ..., m\}$ and sent to the central server who aggregates updates from all participants under certain rules in Equation~\ref{eq:agg}. Baruch et al.~\cite{baruch2019little} observe that, for image classification tasks, the small noises sufficiently compromise the global model while being sufficiently small to evade detection from defense strategies.

\subsubsection{\textbf{Fang's Attack}}
\label{sec:fangattack}

Fang et al.~\cite{fang2020local} proposed an optimization-based model poisoning attack tailored to specific robust aggregation rules (Krum, Multi-Krum, Trimmed Mean and Median), as will be explained in Section~\ref{defense}. 

Fang's attack is conducted separately under two assumptions: (1) full knowledge, and (2) partial knowledge. Under full knowledge, the attacker has full access to local model updates of all benign clients. This is a strong and impractical assumption, and it is often not the case in real attacks on federated learning systems.
%as the authors would like to investigate the upper bound of their attacks’ threats for the given setting of federated learning. 
In the partial knowledge scenario, the attacker only knows the local training data and models of the compromised clients.

In their attack to the robust aggregation rules Krum and Multi-Krum, the attacker computes the average $\mu$ of the benign updates in their possession, computes a perturbation $\mathbf{s}= - \text{sign}(\mu - \mathbf{\theta_g})$, and finally computes a malicious update as $\mathbf{\theta^m_i} = (\mathbf{\theta_g} + \lambda\cdot\mathbf{s})$ by solving for the coefficient $\lambda$, where $\mathbf{\theta_g}$ is the before-attack global model during each federated training step. Thus, under the full knowledge assumption, the average $\mu$ and perturbation signal $\mathbf{s}$ are computed based on all benign updates ($\mathbf{\theta_1}, ..., \mathbf{\theta_m}, \mathbf{\theta_{m+1}}, ..., \mathbf{\theta_n}$). For updates of the malicious clients ($\mathbf{\theta_1}, ..., \mathbf{\theta_m}$), the before-attack benign updates are leveraged. Under the partial knowledge scenario, only the before-attack benign updates ($\mathbf{\theta_1}, ..., \mathbf{\theta_m}$) are used to estimate the real values for average $\mu$ and the reversed deviation vector $\mathbf{s}$.

When attacking Trimmed Mean and Median, the goal is to craft the compromised local models based on the maximum $\theta_{max,j}$ or minimum $\theta_{min,j}$ benign parameters for each dimension $j$ of the local model (this is one of the key features used by Trimmed Mean and Median for defending). The choice of $\theta_{max,j}$ or $\theta_{min,j}$ depends on which one deviates the global model towards the inverse of its update direction without attacks. 
Similar to when attacking Krum, the reversed deviation vector $\mathbf{s}$ is computed with full knowledge or estimated under partial knowledge with only before-attack updates from all attackers, so as the estimation of $\theta_{max,j}$ and $\theta_{min,j}$. After getting the $j$-th value of vector $\mathbf{s}$, the $j$-th dimension of the compromised local model is randomly sampled from the range built based on $\theta_{max,j}$ (if $s_j = 1$) or $\theta_{min,j}$ (if $s_j = -1$).

	\section{Defense for FOLTR Systems}
\label{defense}

%Byzantine-robust aggregation rules.
%
%defense aggregation methods: Krum~\cite{blanchard2017machine}, Multi-Krum~\cite{blanchard2017machine}, Trimmed Mean~\cite{yin2018byzantine}, Median~\cite{yin2018byzantine}, Bulyan~\cite{guerraoui2018hidden}, ...
%
%\subsection{Problem Definition and Defense Model}
%\subsection{Methods}

The current state-of-the-art defense methods against untargeted poisoning attacks focus on enhancing the robustness of the aggregation rules (Equation~\ref{eq:agg}) used during the global update phase, to counteract attempts by malicious clients to corrupt the training.
%State-of-the-art defense methods for untargeted poisoning attacks focus on optimising the aggregation rules at the global updating phase, which aim to be robust even if certain clients want to compromise the training.
%Next, we review several effectiveness robust aggregation rules.
% The central server can mitigate the impact of poisoning attacks by leveraging robust aggregation rules in Equation~\ref{eq:agg}.

Next, we describe four robust aggregation rules that have been shown effective in general federated learning, but have not been evaluated for FOLTR.
%In Equation~\ref{eq:agg}, the first $m$ models are compromised by the attackers using either data poisoning or model poisoning strategies.

\subsection{Krum and Multi-Krum}

The intuition behind the Krum method for robust aggregation~\cite{blanchard2017machine} is that the malicious local model updates need to be far from the benign ones in order for the success of poisoning the global model. To evaluate how far a model update $\mathbf{\theta_i}$ is from the others, Krum computes the Euclidean distances between $\mathbf{\theta_i}$ and $\mathbf{\theta_j}$ for $i \neq j$. We denote $i \rightarrow j$ if $\mathbf{\theta_j}$ belongs to the set of $n-m-2$ closest local models of $\mathbf{\theta_i}$. Then the sum of $n-m-2$ shortest distances to $\mathbf{\theta_i}$ is computed and denoted as $s(i) = \sum_{i \rightarrow j}^{}Euc\_dist(\mathbf{\theta_i}, \mathbf{\theta_j})$. After computing the distance score $s(i)$ for all local updates, Krum selects the local model with the smallest $s(i)$ as the global model $\mathbf{\theta_g}$:

\begin{equation}
	\begin{split}
	\mathbf{\theta_g} = \textit{Krum}(\mathbf{\theta_1}, ..., \mathbf{\theta_m}, \mathbf{\theta_{m+1}}, ..., \mathbf{\theta_n})
	= \mathop{\arg\min}_{\mathbf{\theta_i}}\ s(i)
	\end{split}
	\label{eq:krum}
\end{equation}

Multi-Krum is a variation of the Krum method. Multi-Krum, like Krum, calculates the distance score $s(i)$ for each $\mathbf{\theta_i}$. However, instead of choosing the local model with the lowest distance score as the global model (as Krum does), Multi-Krum selects the top $f$ local models with the lowest scores and computes the average of these $f$ models ($\mathbf{\theta'_i}$, where $i \in \{1, ..., f\}$) to be the global model.

%proposed as a variant of Krum. Same as Krum, Multi-Krum computes the distance score $s(i)$ for all $\mathbf{w_i}$. Unlike Krum selecting the local model with the smallest distance score as the global model, Multi-Krum selects $f \in \{1, ..., n\}$ among them with the lowest score and computes the average of the $f$ local models $\mathbf{w'_i}$, $i \in \{1, ..., f\}$ as the global model. We set the Multi-Krum parameter $f =  n-m$ same as the previous work~\cite{blanchard2017machine}.

\begin{equation}
	\begin{split}
		\mathbf{\theta_g} = \textit{Multi-Krum}(\mathbf{\theta_1}, ..., \mathbf{\theta_m}, \mathbf{\theta_{m+1}}, ..., \mathbf{\theta_n})
		=\frac{1}{f} \sum_{i=1}^{f} \mathbf{\theta'_i}
	\end{split}
	\label{eq:multi-krum}
\end{equation}

\noindent In our empirical investigation, we set the Multi-Krum parameter $f =  n-m$, as in previous work~\cite{blanchard2017machine}.

\subsection{Trimmed Mean and Median}

Assume that $\theta_{ij}$ is the $j$-th parameter of the $i$-th local model. For each $j$-th model parameter, the Trimmed Mean method~\cite{yin2018byzantine} aggregates them separately across all local models. After removing the $\beta$ largest and smallest among $\theta_{1j}, ..., \theta_{nj}$, the Trimmed Mean method computes the mean of the remaining $n-2\beta$ parameters as the $j$-th parameter of the global model. We denote $U_j = \{\theta_{1j}, ..., \theta_{(n-2\beta)j}\}$ as the subset of $\{\theta_{1j}, ..., \theta_{nj}\}$ obtained by removing the largest and smallest $\beta$ fraction of its elements. That is, the $j$-th parameter of the global model updated by Trimmed Mean is:

\begin{equation}
	\theta_j = \textit{Trimmed Mean}(\theta_{1j}, ..., \theta_{nj}) = \frac{1}{n-2\beta} \sum_{\theta_{ij}\in U_j} \theta_{ij}
	\label{eq:trimmed-mean}
\end{equation}

\noindent In our implementation, as in previous work on general federated learning~\cite{fang2020local, shejwalkar2021manipulating, shejwalkar2022back}, we set $\beta$ to be the number of compromised clients $m$.

The Median method, like the Trimmed Mean method, sorts the $j$-th parameter of $n$ local models. Instead of discarding the $\beta$ largest and smallest values (as in Trimmed Mean), the Median uses the median of $\theta_{1j}, ..., \theta_{nj}$ as the $j$-th parameter of the global model:

\begin{equation}
	\theta_j = \textit{Median}(\theta_{1j}, ..., \theta_{nj})
	\label{eq:median}
\end{equation}

\noindent In case $n$ is an even number, the median is calculated as the average of the middle two values.

%Similar as trimmed mean, the Median aggregation rule~\cite{yin2018byzantine} sorts the $j$th parameter of $n$ local models. Instead of removing $\beta$ largest and smallest values, Median uses the median of $w_{1j}, ..., w_{nj}$ as the $j$th parameter of the global model. If $n$ is an even number, it computes the mean of the middle two values as the median.

%\subsubsection{\textbf{Bulyan}}

	\section{Experimental setup}

We next describe our experimental setup to evaluate the considered attack and defense mechanism in the context of a FOLTR system.

%\begin{table}[]
%	\caption{Default setting for key parameters in experiments. \textcolor{red}{(can remove this or move to appendix if no enough space)}}
%	\begin{tabular}{|c|l|c|}
%		\hline
%		\multicolumn{1}{|l|}{Parameter} & \multicolumn{1}{c|}{Description} & Value             \\ \hline
%		\textit{n}                      & Number of all participants.      & 10                \\ \hline
%		\textit{m}                      & Number of attackers.             & \{1, 2, 3, 4, 5\} \\ \hline
%		\textit{B}                      & Local updating times.            & 5                 \\ \hline
%		\textit{T}                      & Global updating times.           & 10,000            \\ \hline
%	\end{tabular}
%\label{table:exp}
%\end{table}

\textbf{Datasets.}
Our experiments are performed on four commonly-used LTR datasets: MQ2007~\cite{DBLP:journals/corr/QinL13}, MSLR-WEB10K~\cite{DBLP:journals/corr/QinL13}, Yahoo~\cite{DBLP:journals/jmlr/ChapelleC11}, and Istella-S~\cite{lucchese2016post}. Detailed descriptions about each dataset are discussed in Section~\ref{chap2-dataset}.

\textbf{Federated setup.}
We consider 10 participants ($n=10$) in our experiments, among which $m$ clients are attackers. 
This setup is representative of a cross-silo FOLTR system, typical of a federation of a few institutions or organisations, e.g. hospitals creating a ranker for cohort identification from electronic health records~\cite{hersh2020information}. In this chapter we will not consider the setup of a cross-device FOLTR system, where many clients are involved in the federation: this is representative of a web-scale federation. 

We assume that the malicious clients can collude with each other to exchange their local data and model updates to enhance the impact of attacks. In the federated setting, each client holds a copy of the current ranker and updates the local ranker through issuing $B=5$ queries along with the respective interactions. The attackers can only compromise the local updating phase through poisoning the training data or model updates of the controlled malicious clients. After the local updating finishes, the central server will receive the updated ranker from each client and aggregate all local messages to update the global ranker. In our experiments, we consider the following aggregation rules: (1) FedAvg, (2) other robust aggregation rules introduced in Section~\ref{defense}. Unless otherwise specified, we train the global ranker through $T=10,000$ global updating times.

%Unless otherwise specified, the default setting for each parameter in our federated setup is specified in Table~\ref{table:exp}. We repeat each experiments for X trials and report the average results (and variances).

\textbf{User simulations.}
We follow the standard setup for user simulations in OLTR~~\cite{oosterhuis2018differentiable, zhuang2020counterfactual, wang2021effective, wang2022non}. We randomly sample from the set of queries in the static dataset to determine the query that the user issues each time. After that, the pre-selected documents for the query are ranked by the current local ranking model to generate a ranking result. For every query, we limit the SERP to 10 documents. User interactions (clicks) on the displayed ranking list are simulated through the CCM click models introduced in Sec~\ref{chap2-click} and Sec~\ref{data-poison}. For the user simulation in model poisoning, we simulate three types of users using the three click instantiations: \textit{perfect}, \textit{navigational}, and \textit{informational}. We experiment on the three types of users separately in order to show the impact of attacking on different types of users. For data poisoning, we simulate the poisoned click based on the \textit{poison} click combining with benign users on the aforementioned three types of click models separately to show the impact of our data poisoning strategies on different types of benign clicks. %After receiving user's interactions, the ranking model is thus updated by OLTR algorithm (we adopt PDGD~\cite{oosterhuis2018differentiable} in this paper).

\textbf{Ranking models.}
We experiment on a linear and neural model as the ranking model when training with FPDGD. For the linear model, we set the learning rate $\eta = 0.1$ and zero initialization was used. As in the original PDGD and FPDGD studies~\cite{oosterhuis2018differentiable,wang2021effective}, the neural ranker is optimized using a single hidden-layer neural network with 64 hidden nodes, along with $\eta = 0.1$.

\textbf{Evaluation.}
We evaluate the attack methods by comparing the gap in offline performance obtained when a specific attack is performed and when no attack is performed. The higher the performance degradation, the more effective the attack. As we limit each SERP to 10 documents, we use $nDCG@10$ for offline evaluation as discussed in Section~\ref{chap2-metric}. We record the offline $nDCG@10$ score of the global ranker during each federated training update. 

%As we limit each SERP to 10 documents, we use $nDCG@10$ for offline evaluation, and cumulative discounted $nDCG@10$~\cite{hofmann2013fast} for online evaluation. The offline performance is measured through averaging the $nDCG$ scores of the global ranker over the queries in the held-out test dataset with the actual relevance label. We record the offline $nDCG@10$ score of the global ranker during each federated training update. The online performance is designed to evaluate user's experience during training and is measured through the cumulative discounted $nDCG@10$ score of the SERP that is actually displayed during training. Assume we have a sequence of $T$ queries, with $R^t$ representing the SERP displayed to the user for the $t$-th query, then the online performance is measured as: $online\_performance = \sum_{t=1}^{T}nDCG(R^t)*\gamma^{t-1}$. Here, $\gamma$ is a discount factor which assigns more weight to the interactions that occur during the early phases of training; we set $\gamma = 0.9995$ as in previous work. The final online performance in our experiments is averaged on $n$ participants, including the before-attack online performance on the malicious clients. 

\section{Results for Data Poisoning}
\label{sec:data-poison-result}

We perform data poisoning attack and four defense methods across different settings of user behaviours (i.e. click models) and number of attackers ($\{10\%, 20\%, 30\%, 40\%\}$). 
Results on MSLR-WEB10K with a linear ranker are shown as solid lines in Figure~\ref{fig:data-defense} -- results for other datasets are similar and can be found in chapter appendix Section~\ref{chap5:app}.

\subsection{Attacks}

%We start by reporting the results of the data poisoning attacks. 
%
%We implement these attacks by directly compromising the click feedback given by the attackers using the \textit{poison} instantiation introduced in Section~\ref{data-poison}. 
%
%Note that at the same time as the \textit{poison} instantiation is performed on the malicious clients, the three types of benign user click models (\textit{perfect}, \textit{navigational} and \textit{informational}) are executed on the remaining clients. 
%
%We also simulate an all-benign setting (denoted as "honest" in our experimental results). 
%
%This forms the baseline to demonstrate the attack effectiveness by comparing the performance gap between the baseline and the compromised training. 
%
%To further demonstrate the impact of the number of malicious clients on the attack effectiveness, we simulate with percentages of malicious clients in the system in the set $\{10\%, 20\%, 30\%, 40\%\}$ (corresponding to $\{1, 2, 3, 4\}$ clients).

\begin{figure*}[t]
	\centering
	\begin{subfigure}{1\textwidth}\centering
		\includegraphics[width=1\textwidth]{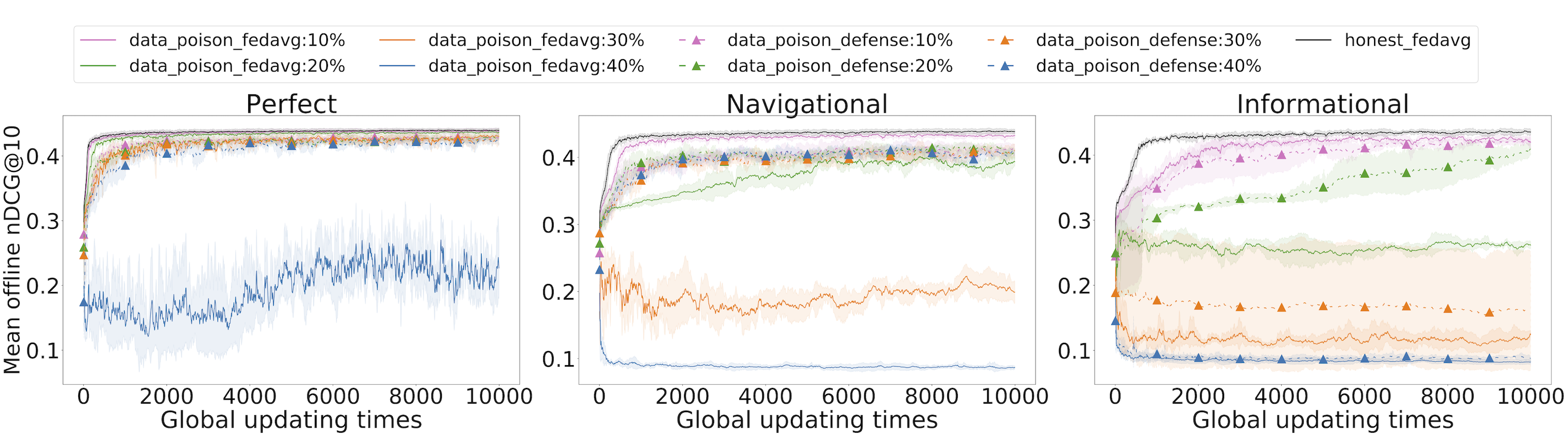}
		\caption{MSLR-WEB10K (Krum)}
		\label{fig:mslr-data-krum}
	\end{subfigure}
	\begin{subfigure}{1\textwidth}\centering
		\includegraphics[width=1\textwidth]{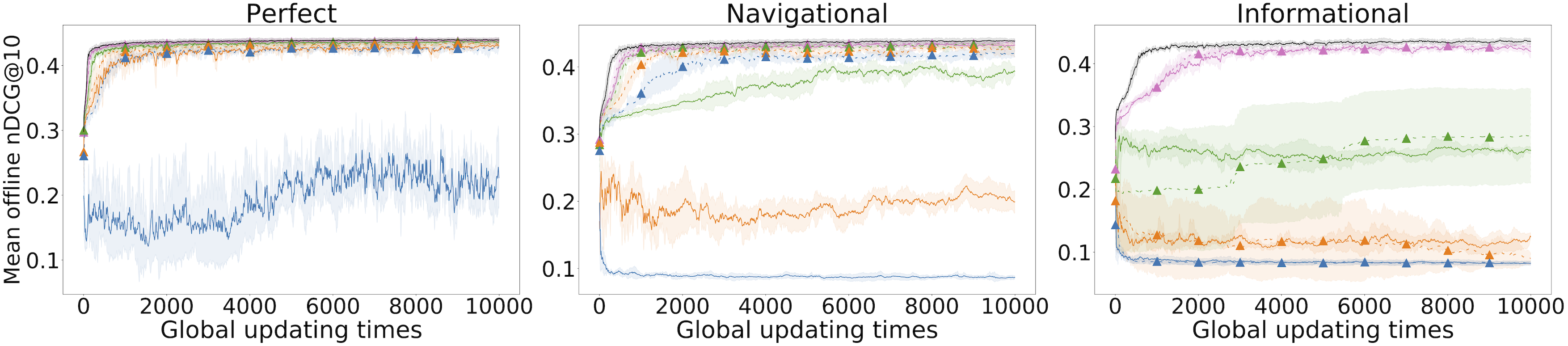}
		\caption{MSLR-WEB10K (Multi-Krum)}
		\label{fig:mslr-data-multikrum} 
	\end{subfigure}
	\begin{subfigure}{1\textwidth}\centering
		\includegraphics[width=1\textwidth]{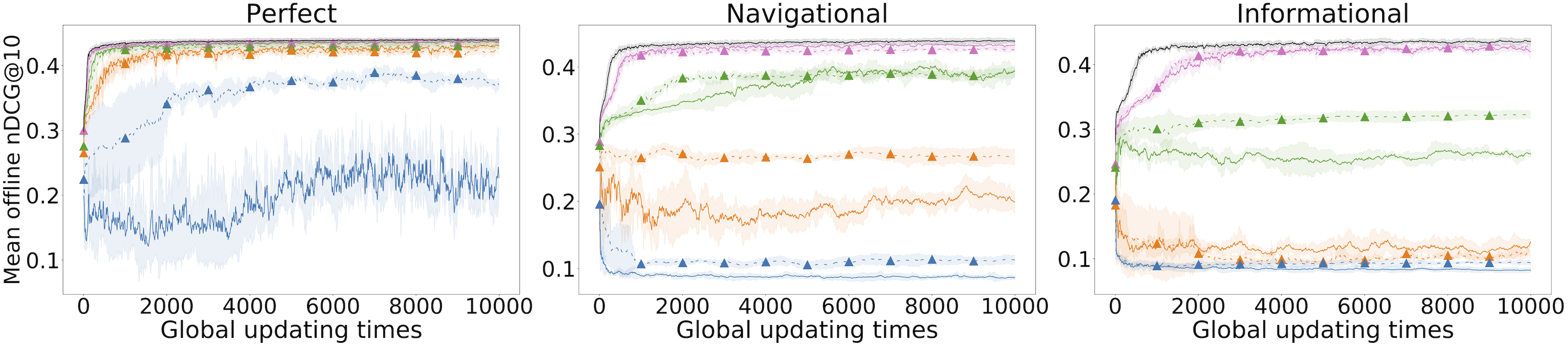}
		\caption{MSLR-WEB10K (Trimmed Mean)}
		\label{fig:mslr-data-trmean}
	\end{subfigure}
	\begin{subfigure}{1\textwidth}\centering
		\includegraphics[width=1\textwidth]{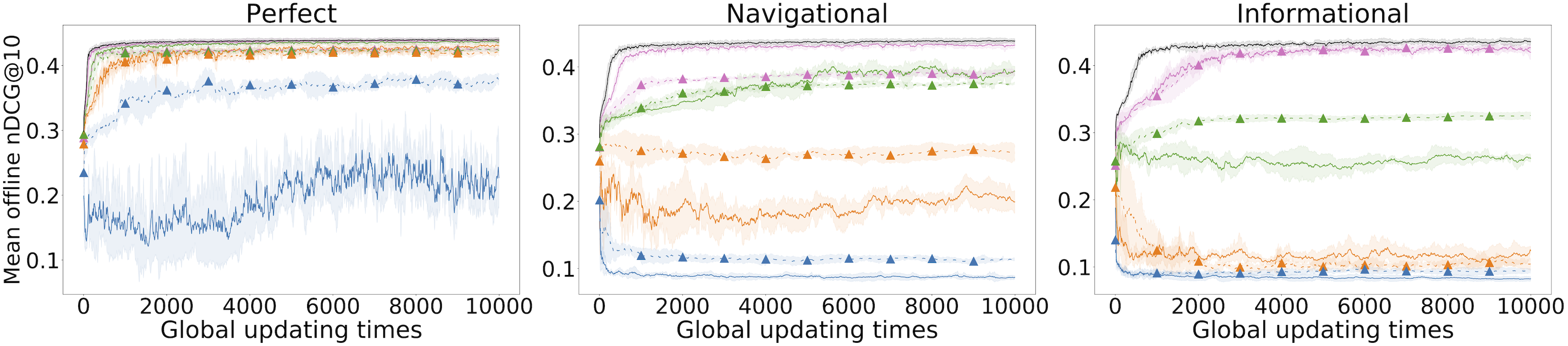}
		\caption{MSLR-WEB10K (Median)}
		\label{fig:mslr-data-median} 
	\end{subfigure}
	\caption{Offline performance (nDCG@10) for MSLR-WEB10K under data poisoning attack and defense strategies, simulated with three benign instantiations of click model and different percentage of attackers equaling to $\{10\%, 20\%, 30\%, 40\%\}$; results averaged across all dataset splits and experimental runs.
		\label{fig:data-defense}}
\end{figure*}

In the plots of Figure~\ref{fig:data-defense}, the solid lines represent the results of data poisoning when no defense method is deployed. Among them, the black line represents no attacking situation ("honest" baseline).
%Results on MSLR10k using the linear ranker are shown in solid line of Figure~\ref{fig:data-defense} -- results for other datasets are similar and are omitted for space constraints. 
%
We can observe that
the effect of data poisoning depends on the settings of user behaviors (i.e. click models) and the number of attackers.

\textbf{Effect of number of attackers.}
By comparing the solid curves in each plot of Figure~\ref{fig:data-defense}, we can observe that the overall performance of the FOLTR system decreases as the number of attackers increases, compared to the ``honest'' baseline.  Thus, the higher the number of attackers, the more degradation on the FOLTR system is experienced.

\textbf{Ease of attack under different user behaviours.}
By comparing the plots within each row, we see the effect of data poisoning is different under different user behaviours.
%The ease of the attack can vary depending on the behaviour of genuine users. 
%
In the navigational and informational settings, attacks carried by as little as 20\% of clients can significantly affect the system. However, to successfully attack the perfect click, a higher number of malicious clients is needed. 
Across all datasets, the informational click model is the most affected by attacks, while the perfect click model only experiences considerable losses when a large number of clients has been compromised.

 %Comparing to the "honest" baseline, the overall performance drops with more attackers engaging. In \textit{navigational} and \textit{informational} settings, 20\% of attackers are enough to sensibly compromise the system. However, to successfully attack the \textit{perfect} click, it is necessary to have a higher number of attackers than what is sufficient for the other two types of clicks. Depending on how genuine users behave, attacks may be easier or more difficult. Across all datasets, the click model that suffers most by attacks is the \textit{informational} click model, while the \textit{perfect} click model displays sensible losses only when many clients have been compromised.

\textbf{Neural ranker vs. linear ranker.}
The findings from results for the neural ranker under data poisoning attack are similar to those for the linear ranker -- and this pattern is valid across all remaining experiments we report. Therefore, we only report experiments using the linear ranker in the following sections. %due to limited space. %and computational costs.

%Results on the neural ranker is shown in Figure~\ref{fig:mslr-data-neural}. We have similar findings on neural ranker with the linear one. Thus, due to space limit and the restricted computational costs associated with experimenting, unless otherwise specified, we implement on the linear ranker when experimenting for the following sections.

%\begin{figure*}[t]
%	\centering
%	\subfigure[\textbf{MSLR-WEB10k (linear ranker)}] {\label{fig:mslr-data-linear} \includegraphics[width=17.5cm]{images/MSLR10K_FPDGD_clients10_batch5_updates10000_data_poison_whole_CCM_linear_offline_performance.pdf}} 
%	\subfigure[\textbf{MSLR-WEB10k (neural ranker)}] {\label{fig:mslr-data-neural} \includegraphics[width=17.5cm]{images/MSLR10K_FPDGD_clients10_batch5_updates10000_data_poison_whole_CCM_neural64_offline_performance.pdf}} 
%	\vspace{-12pt}
%	\caption{Offline performance (nDCG@10) obtained under data poisoning attacks and across three benign instantiations of SDBN click model. The percentages of malicious clients considered are $\{10\%, 20\%, 30\%, 40\%, 50\%\}$. Results are averaged across all dataset splits and experimental runs.
%	\label{fig:data-attack}}
%\end{figure*}

\subsection{Defense}

%Next, we investigate the effectiveness of defense mechanisms under data poisoning attacks; for this, we study the robust aggregation rules introduced in Section~\ref{defense}.
%%implement four robust aggregation rules which are introduced in Section~\ref{defense} to investigate their effectiveness in defending against our data poisoning strategy. W
%We vary the percentage of malicious clients to be in $\{10\%, 20\%, 30\%, 40\%\}$ 
%%-- this is different than the previous experiments 
%because, by design, some defensive strategies do not support systems affected by 50\% or more malicious clients. 
%
%Results on MSLR10k are shown in Figure~\ref{fig:data-defense}; results on other datasets are similar but are omitted due to space limitations.

Next, we demonstrate  the effectiveness of our four defense mechanisms against data poisoning attack.
The results on MSLR-WEB10K are shown by the dashed curves in Figure~\ref{fig:data-defense}. Each row corresponds to one defense method.

\textbf{Krum.} 
Overall, Krum performs well across all datasets and for all three types of click models once the percentage of malicious clients reaches 20\% or higher, with the exception of MQ2007. However, Krum does not work when defending against 10\% of clients, except for Istella-S. 
The accuracy drop from deploying Krum (as shown in Section~\ref{benign-defense}) outweighs its effectiveness in defense, especially when there is a relatively small impact on the effectiveness of the model, as is in the case when 10\% of the clients are malicious. 
Additionally, Krum does not show any improvement in defending certain scenarios under the informational click model, such as for MQ2007 under all percentages of malicious clients, and for MSLR-WEB10K when 40\% of clients are malicious.

%Overall, Krum starts working on all datasets across three click types once the percentage of attackers is equal or more than 20\% with the exception on MQ2007. Meanwhile, Krum to defend on 10\% attackers does not work except for Istella-s. With the relative less impact on compromising the model performance with 10\% attackers, the accuracy drops of deploying Krum (also shown in section~\ref{benign-defense}) washes out its effectiveness in defense. Also, Krum does not demonstrate improvements in defending some scenarios under \textit{informational} click: (1) for MQ2007 under all concerned attacker percent, (2) for MSLR10k with 40\% attackers.

\textbf{Multi-Krum.} 
The results obtained for Multi-Krum show similar effectiveness on the perfect click model as Krum. It is important to note that the perfect click model is the hardest to attack among the three types of click models considered. Multi-Krum provides slightly better defense performance on navigational clicks compared to Krum, especially when there are fewer attackers (30\% or less). However, for the informational click model, Multi-Krum does not perform as well as Krum. This is because the variance of the local model updates is relatively higher in the noisier informational click model. After averaging the selected local models, the advantage of Multi-Krum is reduced, especially when there are more than 30\% malicious clients.

%For results on Multi-Krum, it achieves similar performance on \textit{perfect} clicks with Krum. It is worth to mention that, among three types of clicks, \textit{perfect} type is the hardest for attacking. Multi-Krum demonstrates slightly better defense performance on \textit{navigational} clicks comparing to Krum, especially on systems with fewer attackers (equal or less than 30\%). Under \textit{informational} click: Multi-Krum does not work well comparing to Krum. The reason is that the variance of local model updates is relatively higher under the noisiest \textit{informational} click. After averaging the selected local models, the superiority of Multi-Krum is shrunk especially when attackers account for more than 30\%.

\textbf{Trimmed Mean.}
Across all experiments, Trimmed Mean does not perform well on the noisiest click model (informational) when there are more than 30\% malicious clients involved. When the malicious clients are 20\% or 30\%, Trimmed Mean provides lower performance gains compared to Krum, but it performs similarly to Krum when only 10\% of the clients are malicious. %In general, Trimmed Mean is less effective in defending against data poisoning attacks compared to Krum, except on MQ2007.

%Among all experiments, Trimmed Mean is not effective on the noisiest click model (\textit{informational}) when 30\% or more malicious clients are involved. For 20\% and 30\% attackers, the performance gain by Trimmed Mean is lower than Krum while the performance on 10\% attackers is similar as Krum because of the limited performance drops compromised by fewer attackers. Overall, Trimmed Mean is less effective in defending against data poisoning attack than Krum with the exception on MQ2007.

\textbf{Median.}
Like Trimmed Mean, Median does not provide improved performance on the noisy informational click model when 30\% or 40\% of clients are malicious. Similarly, and like other robust aggregation rules, Median does not show significant improvements when only 10\% of  clients are malicious. In fact, the Median's performance even decreases on the navigational click model for MSLR-WEB10K with 10\% of malicious clients. When the malicious clients are 20\% and 30\% of all clients in the federation, the performance gain provided by Median is similar to that of Trimmed Mean.
%Similar as Trimmed Mean, Median does not show improvements on noisy \textit{informational} click with attackers accounting for 30\% and 40\%. Also, similar as other robust-aggregation rules, no significant gain on 10\% attackers by Median. Performance even drops on \textit{navigational} click in MSLR10k with 10\% attackers. For 20\% and 30\% attackers, the performance gain by Median is similar to Trimmed mean.

\textbf{Summary.} Overall, Krum and Multi-Krum work better than Trimmed Mean or Median when defending against data poisoning attacks, with the exception that Trimmed Mean and Median perform better on the smaller MQ2007 dataset.

\begin{figure}[t]
	\centering
	\begin{subfigure}{1\textwidth}\centering
		\includegraphics[width=1\textwidth]{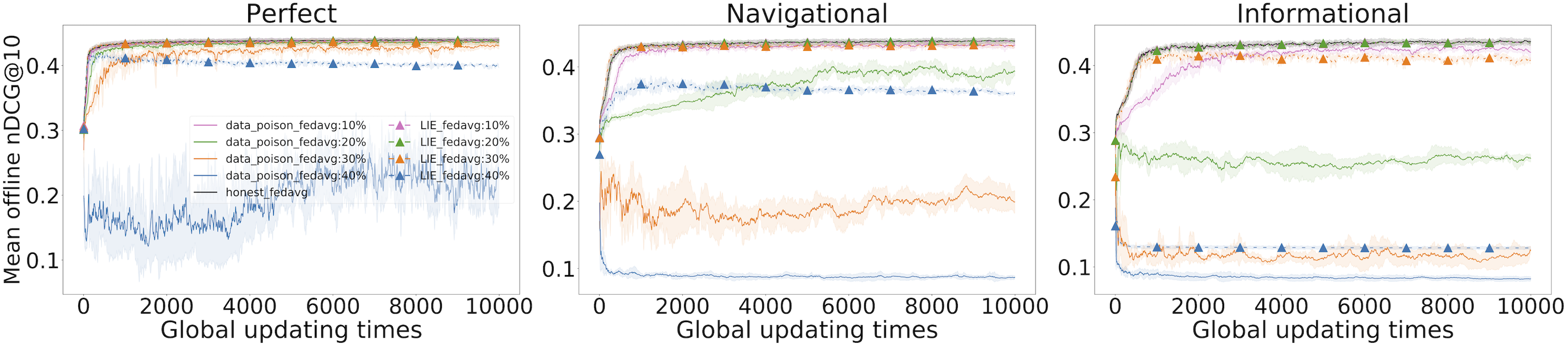}
		\caption{MSLR-WEB10K (LIE)}
		\label{fig:mslr-lie}
	\end{subfigure}
	\begin{subfigure}{1\textwidth}\centering
		\includegraphics[width=1\textwidth]{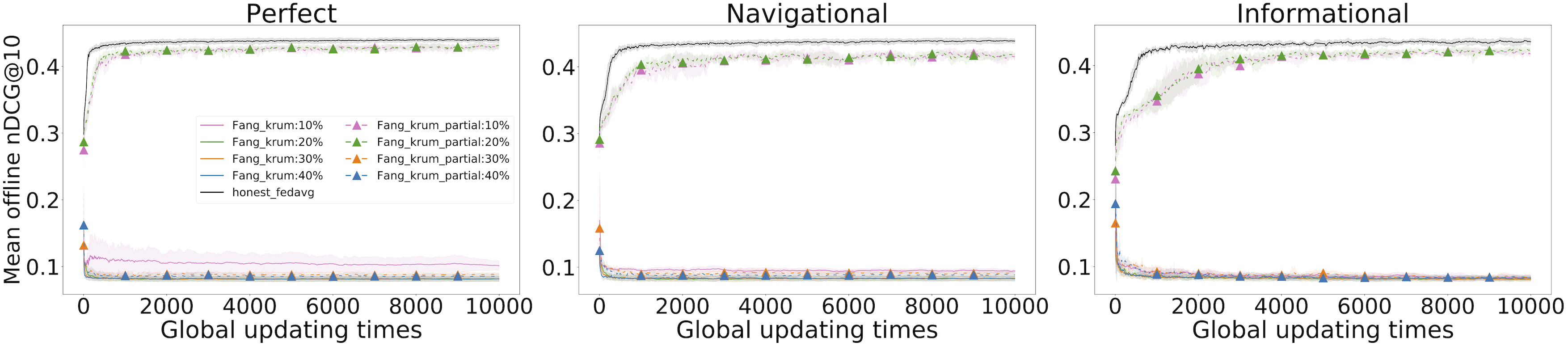}
		\caption{MSLR-WEB10K (Fang's attack on Krum)}
		\label{fig:mslr-fang-krum}
	\end{subfigure}
	\begin{subfigure}{1\textwidth}\centering
		\includegraphics[width=1\textwidth]{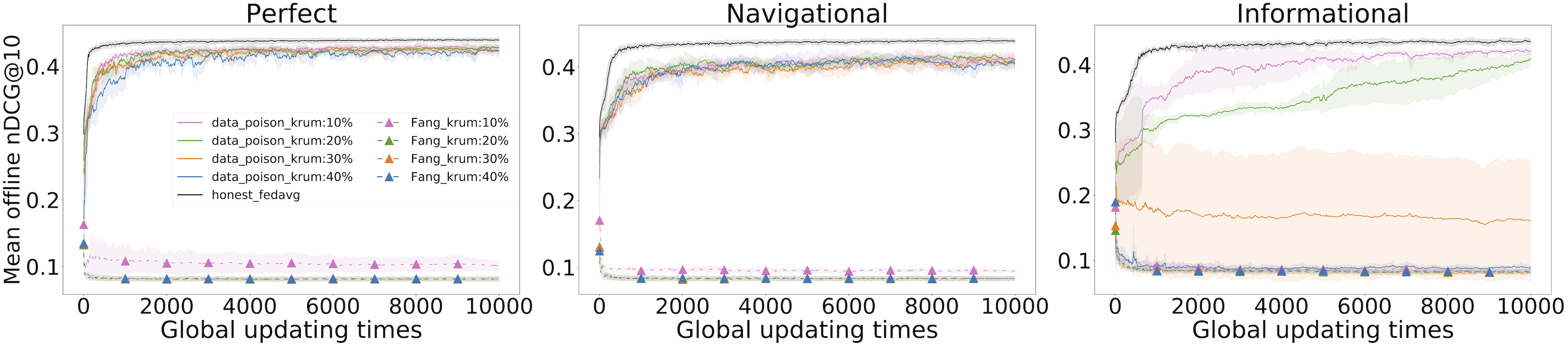}
		\caption{MSLR-WEB10K (Fang's attack on Krum - full knowledge)}
		\label{fig:mslr-fang-krum-data} 
	\end{subfigure}
	\caption{Offline performance (nDCG@10) under model poisoning attacks, simulated using three benign instantiations of click model with percentage of attackers equaling to $\{10\%, 20\%, 30\%, 40\%\}$. The aggregation rule is FedAvg (for Figure~\ref{fig:mslr-lie}) and Krum (for Figure~\ref{fig:mslr-fang-krum}- \ref{fig:mslr-fang-krum-data}).
	\label{fig:model-attack}}
\end{figure}

\section{Results for Model Poisoning}
\label{sec:model-poison-result}

We implement the model poisoning strategies specified in Section~\ref{model-poison} and report results on MSLR-WEB10K~\footnote{Results for other datasets are similar and can be found in chapter appendix Section~\ref{chap5:app}.}, specifically comparing their poisoning effectiveness with that of data poisoning methods.

\subsection{Little Is Enough (LIE)}

The experimental results obtained for LIE are partially shown in Figure~\ref{fig:mslr-lie}, along with a comparison with data poisoning. 

\textbf{Ineffectiveness of LIE.}
The results indicate that LIE is less effective in attacking the performance of the global model compared to data poisoning, with one exception for the perfect click model on the Yahoo dataset when 40\% of the clients are malicious. This shows that adding random noise to compromise the local models is less effective for attacking the global ranker performance than compromising the click signals directly. Because of the poor attacking effectiveness of LIE, we do not investigate how it performs when defense strategies are put in place.

%The experimental results obtained for LIE are shown in Figure~\ref{} and Table~\ref{}, along with a comparison with data poisoning. Based on the empirical results, we find that the attacking performance of LIE is worse than data poisoning with only one exception on \textit{perfect} click with 40\% malicious attackers for Yahoo dataset. Thus, we conclude that with random noise (variance across local model updates) directly added to compromise local model, it is less effective in attacking global performance comparing to our data poisoning strategy which compromises click signals directly. 

\subsection{Fang's Attack}
In our experiments, we implement Fang's attacks on four robust aggregation rules, with each attacking strategy tailoring specific defense strategies except that the same attack method is shared for Trimmed Mean and Median. 

\textbf{Full knowledge vs. partial knowledge.}
First, we compare the attacking performance under both full knowledge and partial knowledge assumptions. 
According to previous findings in general federated learning~\cite{fang2020local}, attacking with full knowledge performs consistently better than with partial knowledge as the tailored attack can be optimised with auxiliary information about benign clients. 
From our results (results on MSLR-WEB10K under Krum are shown in Figure~\ref{fig:mslr-fang-krum}), we observe that full knowledge performs better with fewer malicious clients (10\% and 20\%), but the gap in effectiveness obtained between full and partial knowledge decreases as the number of malicious clients increases (30\% and 40\%), thus leading to differences compared to the general results in federated learning. This is because with more malicious clients, partial knowledge (knowledge of before-attack local model updates for compromised clients) provides enough information to effectively poison the global model while avoiding detection by robust defense strategies.

%From the empirical results, we find that under fewer attackers (i.e. 10\% and 20\%), Fang's attack with full knowledge performs better in manipulating the global ranker. While under some cases with more attackers (i.e. 30\% and 40\%), the performance gap between full knowledge and partial knowledge is minor. Because with more collusive attackers, even under partial knowledge (which means that the attackers can only know the before-attack local model updates on the attacking clients), the poisoning strategy gets sufficient knowledge to effectively poisoning the global model while eluding the detect from robust defensive strategies.

\textbf{Fang's Attack vs. data poisoning.}
Next, we compare Fang's attack under the full knowledge assumption against the data poisoning method under the same robust-aggregation rule (results on MSLR-WEB10K under Krum are shown in Figure~\ref{fig:mslr-fang-krum-data}). We find that Fang's attack can successfully poison FOLTR and mitigate the impact of defense methods compared to data poisoning. This finding aligns with the original results from Fang et al.~\cite{fang2020local}.

	\section{Impact of Defense under No-Attack}
\label{benign-defense}

Robust aggregation rules exhibit improvements in defending against poisoning attacks under some circumstances. But in real-world settings, the administrator of the FOLTR system has no knowledge of whether an attack is taking place. Thus, if the system administrator wishes to ensure protection against attacks, they may be required to deploy defense strategies irrespective of an attack ever taking place, or not. However, is there a price to pay, in terms of search effectiveness, if a defense strategy is deployed on a FOLTR system that is not exposed to an attack? We investigate this next, by comparing the effectiveness of a FOLTR system with no malicious clients and with different defense strategies implemented against the effectiveness of the same system with no defense.

\begin{figure}[t]
	\centering
%	\begin{subfigure}{1\textwidth}\centering
%		\includegraphics[width=1\textwidth]{Chapter5_paper3_poisoning_ictir2023/images/MQ2007_FPDGD_clients10_batch5_updates10000_honest_CCM_linear_offline_ndcg}
%		\caption{MQ2007 (benign clients)}
%		\label{fig:mq2007-benign-defense}
%	\end{subfigure}
	\begin{subfigure}{1\textwidth}\centering
		\includegraphics[width=1\textwidth]{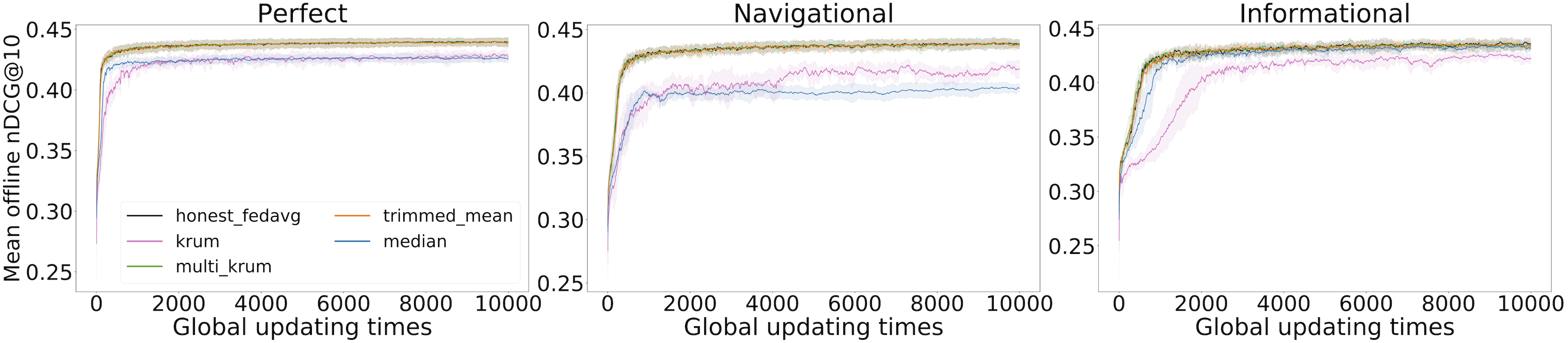}
		\caption{MSLR-WEB10K (benign clients)}
		\label{fig:mslr-benign-defense}
	\end{subfigure}
%	\begin{subfigure}{1\textwidth}\centering
%		\includegraphics[width=1\textwidth]{Chapter5_paper3_poisoning_ictir2023/images/Yahoo_FPDGD_clients10_batch5_updates10000_honest_CCM_linear_offline_ndcg}
%		\caption{Yahoo (benign clients)}
%		\label{fig:yahoo-benign-defense}
%	\end{subfigure}
%	\begin{subfigure}{1\textwidth}\centering
%		\includegraphics[width=1\textwidth]{Chapter5_paper3_poisoning_ictir2023/images/Istella-s_FPDGD_clients10_batch5_updates10000_honest_CCM_linear_offline_ndcg}
%		\caption{Istella-S (benign clients)}
%		\label{fig:istella-benign-defense}
%	\end{subfigure}
	\caption{Offline performance (nDCG@10) of FOLTR system when no attack is present but defense strategies are deployed; results averaged across all dataset splits and experimental runs.
		\label{fig:benign-defense}}
\end{figure}

The experimental results on MSLR-WEB10K reported in Figure~\ref{fig:benign-defense} show that using Krum and Median leads to a decrease in performance compared to the FedAvg baseline when no attacks are present. 
Results for other datasets are similar and are put in chapter appendix Section~\ref{chap5:app} for space reasons. 
This finding has also been reported before in general federated learning literature~\cite{xia2019faba, cao2021fltrust, xu2022byzantine}, especially when each client's local training data is non independent and identically distributed (non-IID). This is because those Byzantine-robust FL methods exclude some local model updates when aggregating them as the global model update~\cite{cao2021fltrust, xia2019faba}. This decrease raises questions about the use of these methods in FOLTR systems when no malicious client is present -- and it suggests that if reliable methods for attack detection were available, then defense mechanisms may better be deployed only once the attack takes place. 

%we investigate the necessity of implementing defensive strategies. In real-world applications, the central server does not have the prior knowledge on the exists of attackers. Thus, we investigate if there is performance drops with all benign clients under the defense strategies comparing to the vanilla FedAvg aggregation rule as we would like to understand the trade-off between the loss on performance and the effectiveness on defending against malicious clients. 

%Our experimental results on MSLR10k is shown on Figure~\ref{fig:benign-defense} (due to space limit, results on other datasets can be found in the online appendix). Our results across all datasets consistently show that Krum and Median strategies bring about inevitable performance drops comparing to the FedAvg baseline, questioning the use of those two methods if no attacker is involved in FOLTR systems.

	\section{Conclusion}
In this chapter we explore attacks and defense mechanisms for federated online learning to rank (FOLTR) systems, focusing on the potential degradation of ranking performance caused by untargeted poisoning attacks. We investigate both data and model poisoning strategies and evaluate the effectiveness of various state-of-the-art robust aggregation rules for federated learning in countering these attacks.
Our findings indicate that sophisticated model poisoning strategies outperform data poisoning methods, even when defense mechanisms are in place.
We also reveal that deploying defense mechanisms without an ongoing attack can lead to ranker performance degradation. This finding recommends care in the deployment of such mechanisms and suggests that future research should explore defense strategies that do not deteriorate FOLTR ranker performance if no attack is underway.

Code, experiment scripts and experimental results are provided at \url{https://github.com/ielab/foltr-attacks}.

%	\begin{acks}
%		Shuyi Wang is the recipient of a Google PhD Fellowship. This research is partially funded by Beijing Baidu Netcom Technology Co, Ltd, for the project "Federated Online Learning of Neural Rankers", under funding schema 2022 CCF-Baidu Pinecone.
%	\end{acks}
	
%	\bibliographystyle{ACM-Reference-Format}
%	\balance
%	\bibliography{ictir2023-foltr-poisoning}
	
	%\appendix
	%
	%\section{Appendix}

%\end{document}
%\endinput
%%
%% End of file `sample-sigconf.tex'.

%\clearpage
\section{Chapter Appendix}
\label{chap5:app}

\subsection{Results for Data Poisoning}

\begin{figure*}[h]
	\centering
	\begin{subfigure}{1\textwidth}\centering
		\includegraphics[width=1\textwidth]{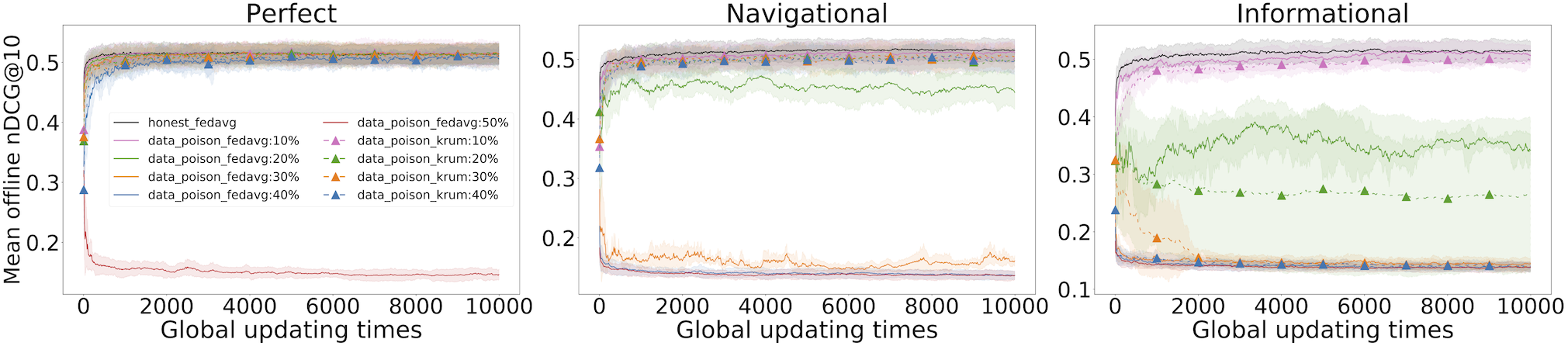}
		\caption{MQ2007 (Krum)}
		\label{fig:mq2007-data-krum}
	\end{subfigure}
	\begin{subfigure}{1\textwidth}\centering
		\includegraphics[width=1\textwidth]{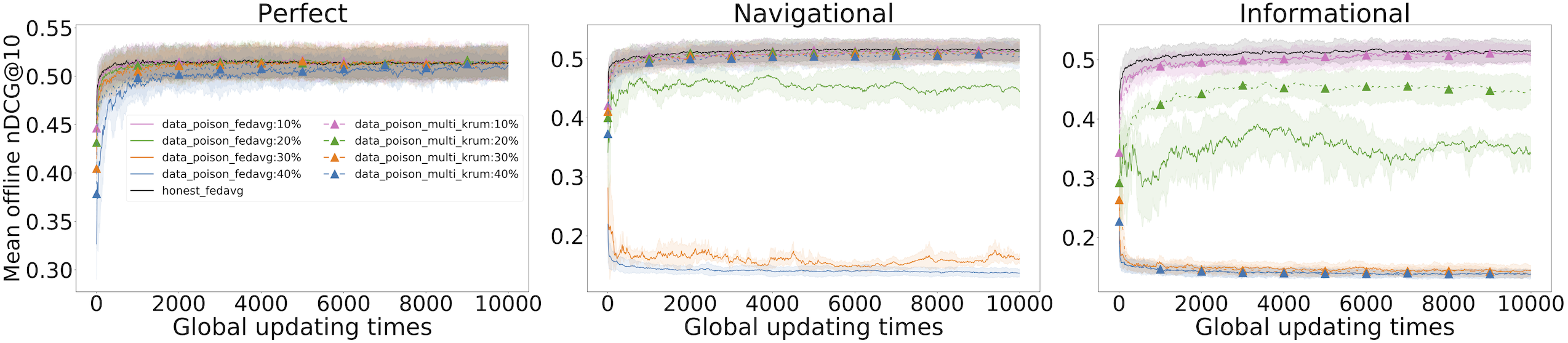}
		\caption{MQ2007 (Multi-Krum)}
		\label{fig:mq2007-data-multikrum} 
	\end{subfigure}
	\begin{subfigure}{1\textwidth}\centering
		\includegraphics[width=1\textwidth]{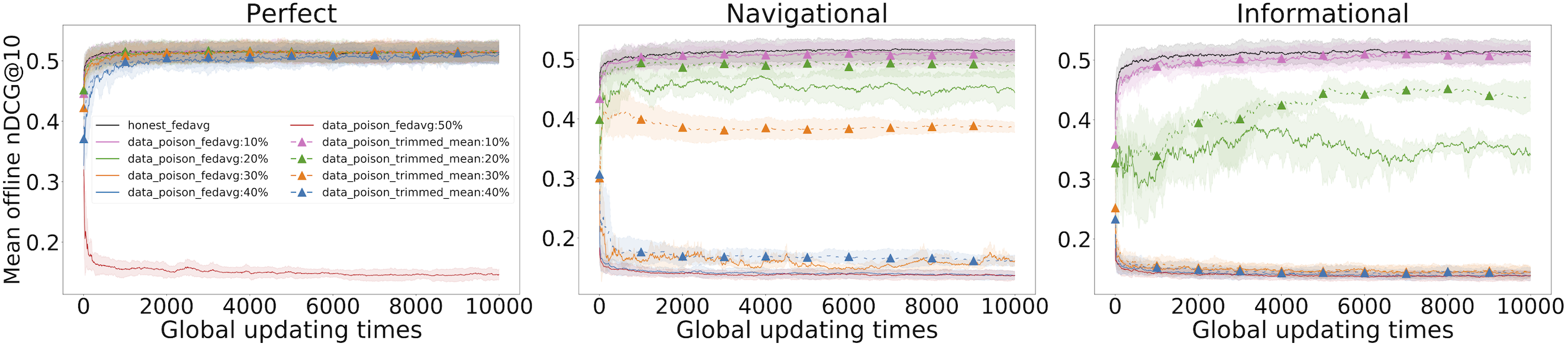}
		\caption{MQ2007 (Trimmed Mean)}
		\label{fig:mq2007-data-trmean}
	\end{subfigure}
	\begin{subfigure}{1\textwidth}\centering
		\includegraphics[width=1\textwidth]{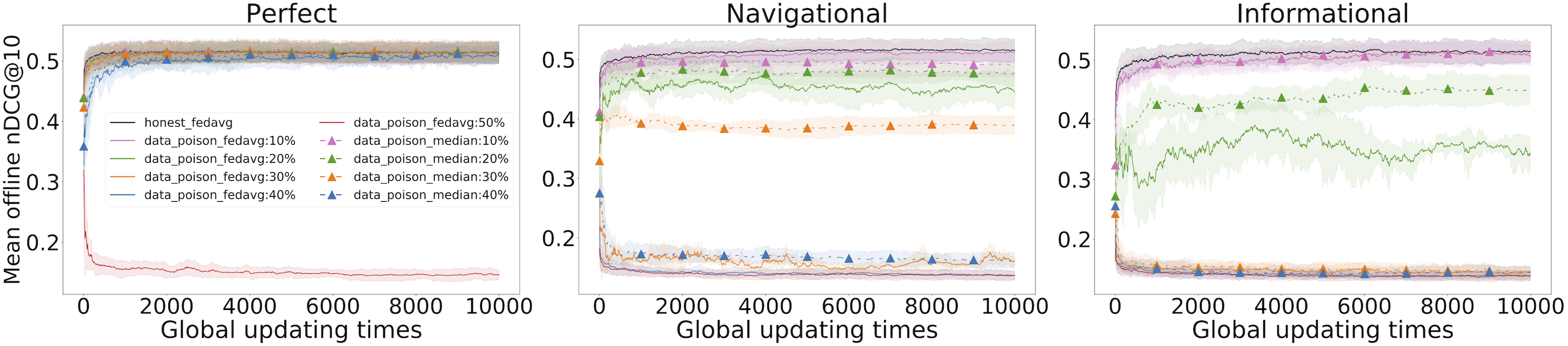}
		\caption{MQ2007 (Median)}
		\label{fig:mq2007-data-median} 
	\end{subfigure}
	\caption{Offline performance (nDCG@10) for MQ2007 under data poisoning attack and defense strategies, simulated with three benign instantiations of click model and different percentage of attackers equaling to $\{10\%, 20\%, 30\%, 40\%\}$; results averaged across all dataset splits and experimental runs.
	\label{fig:mq2007:data-defense}}
\end{figure*}

\begin{figure*}[h]
	\centering
	\begin{subfigure}{1\textwidth}\centering
		\includegraphics[width=1\textwidth]{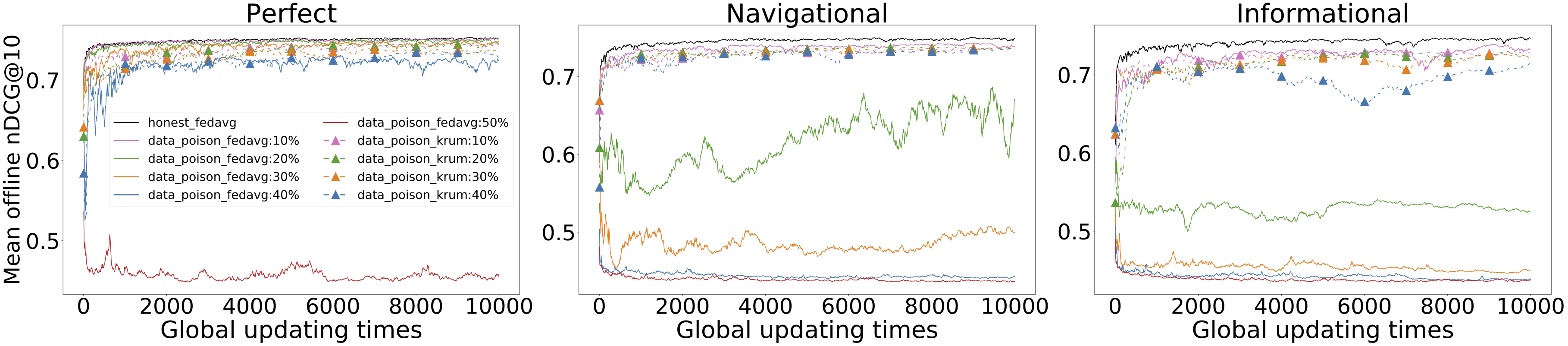}
		\caption{Yahoo (Krum)}
		\label{fig:yahoo-data-krum}
	\end{subfigure}
	\begin{subfigure}{1\textwidth}\centering
		\includegraphics[width=1\textwidth]{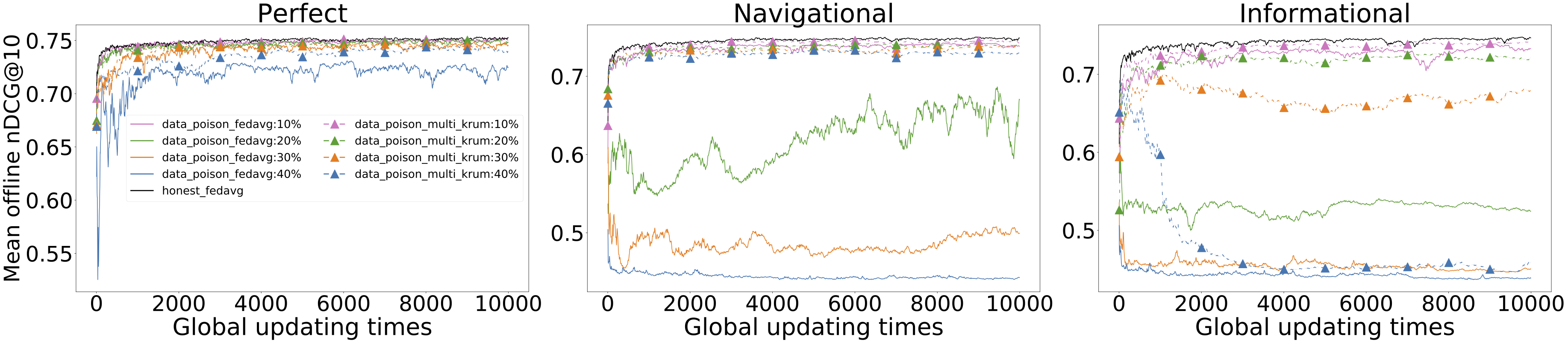}
		\caption{Yahoo (Multi-Krum)}
		\label{fig:yahoo-data-multikrum} 
	\end{subfigure}
	\begin{subfigure}{1\textwidth}\centering
		\includegraphics[width=1\textwidth]{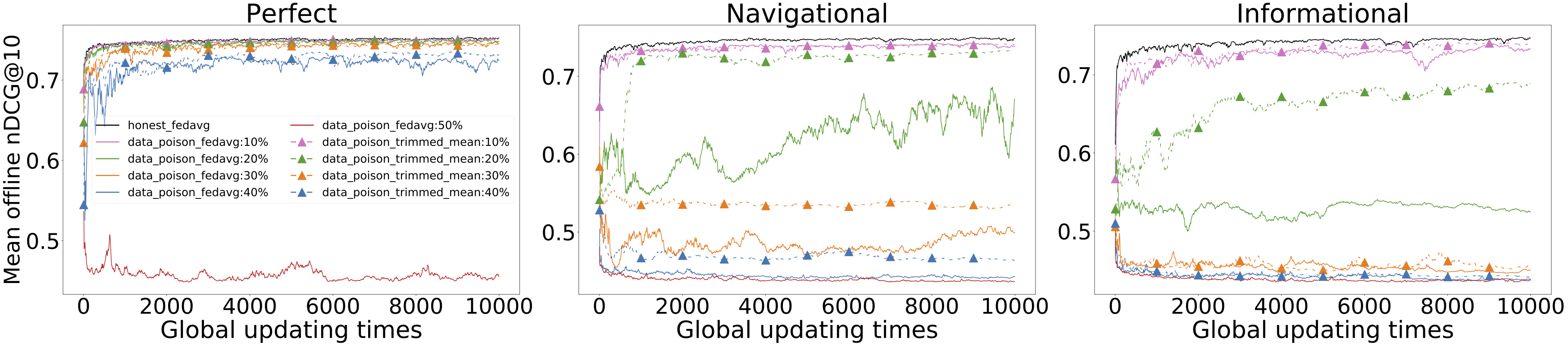}
		\caption{Yahoo (Trimmed Mean)}
		\label{fig:yahoo-data-trmean}
	\end{subfigure}
	\begin{subfigure}{1\textwidth}\centering
		\includegraphics[width=1\textwidth]{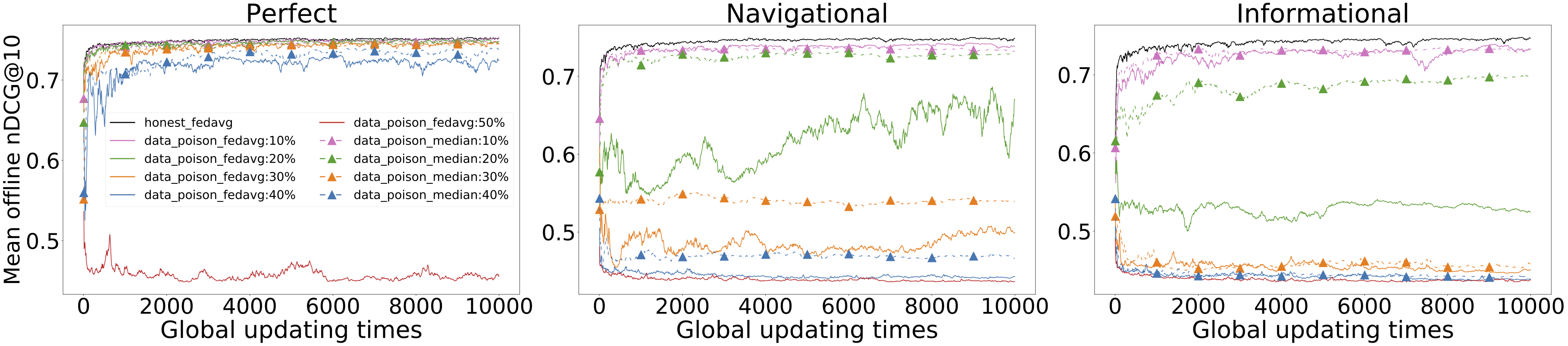}
		\caption{Yahoo (Median)}
		\label{fig:yahoo-data-median} 
	\end{subfigure}
	\caption{Offline performance (nDCG@10) for Yahoo under data poisoning attack and defense strategies, simulated with three benign instantiations of click model and different percentage of attackers equaling to $\{10\%, 20\%, 30\%, 40\%\}$; results averaged across all dataset splits and experimental runs.
		\label{fig:yahoo:data-defense}}
\end{figure*}

\begin{figure*}[h]
	\centering
	\begin{subfigure}{1\textwidth}\centering
		\includegraphics[width=1\textwidth]{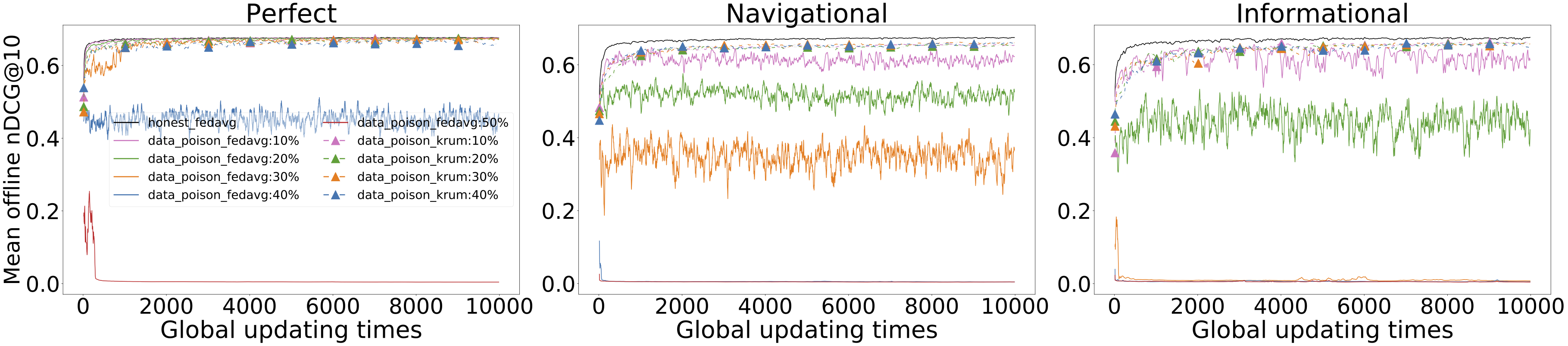}
		\caption{Istella-S (Krum)}
		\label{fig:Istella-s-data-krum}
	\end{subfigure}
	\begin{subfigure}{1\textwidth}\centering
		\includegraphics[width=1\textwidth]{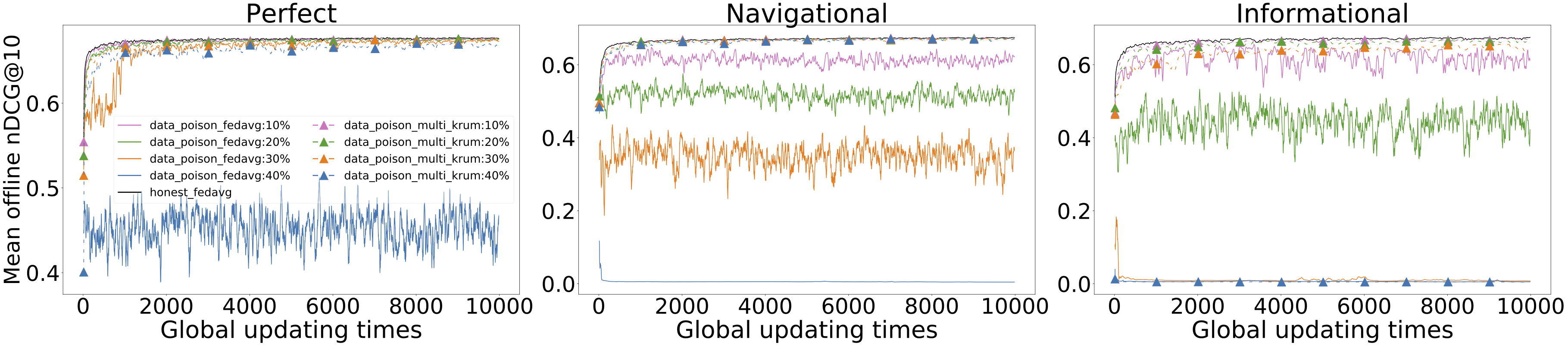}
		\caption{Istella-S (Multi-Krum)}
		\label{fig:Istella-s-data-multikrum} 
	\end{subfigure}
	\begin{subfigure}{1\textwidth}\centering
		\includegraphics[width=1\textwidth]{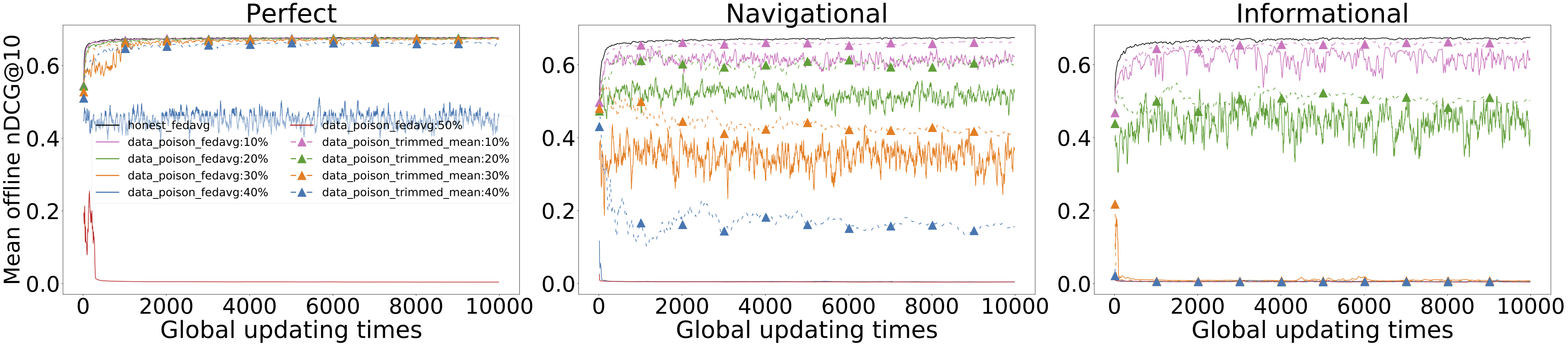}
		\caption{Istella-S (Trimmed Mean)}
		\label{fig:Istella-s-data-trmean}
	\end{subfigure}
	\begin{subfigure}{1\textwidth}\centering
		\includegraphics[width=1\textwidth]{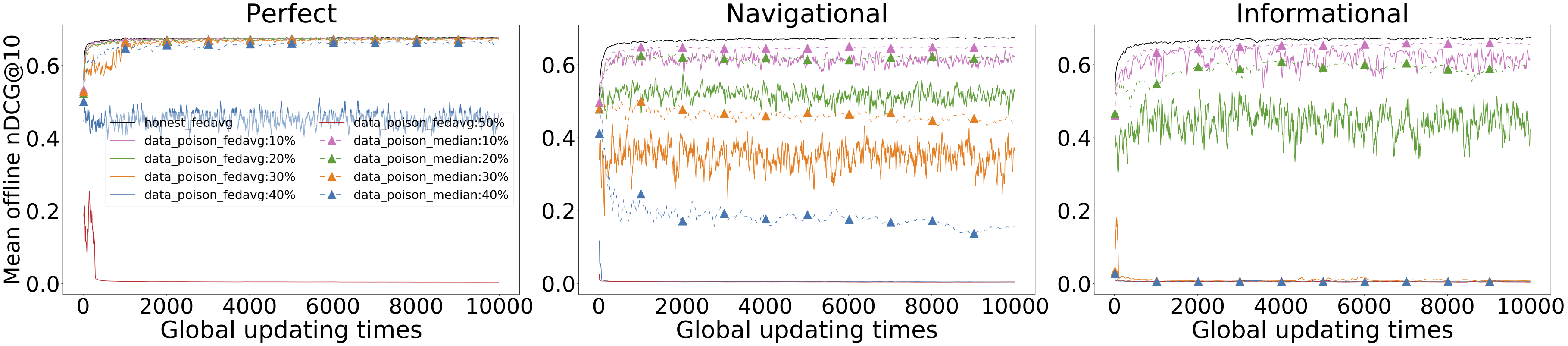}
		\caption{Istella-S (Median)}
		\label{fig:Istella-s-data-median} 
	\end{subfigure}
	\caption{Offline performance (nDCG@10) for Istella-S under data poisoning attack and defense strategies, simulated with three benign instantiations of click model and different percentage of attackers equaling to $\{10\%, 20\%, 30\%, 40\%\}$; results averaged across all dataset splits and experimental runs.
		\label{fig:Istella-s:data-defense}}
\end{figure*}

In this section, we report the results of data poisoning attack and four defense methods across different settings of user behaviours (i.e. click models) and number of attackers ($\{10\%, 20\%, 30\%, 40\%\}$) on the remaining datasets. Results on MQ2007 is shown in Figure~\ref{fig:mq2007:data-defense}. Results on Yahoo is shown in Figure~\ref{fig:yahoo:data-defense}. Results on Istella-S is shown in Figure~\ref{fig:Istella-s:data-defense}. Further analysis on results is detailed in Section~\ref{sec:data-poison-result}

\clearpage
\subsection{Results for Model Poisoning}

\begin{figure}[h]
	\centering
	\begin{subfigure}{1\textwidth}\centering
		\includegraphics[width=1\textwidth]{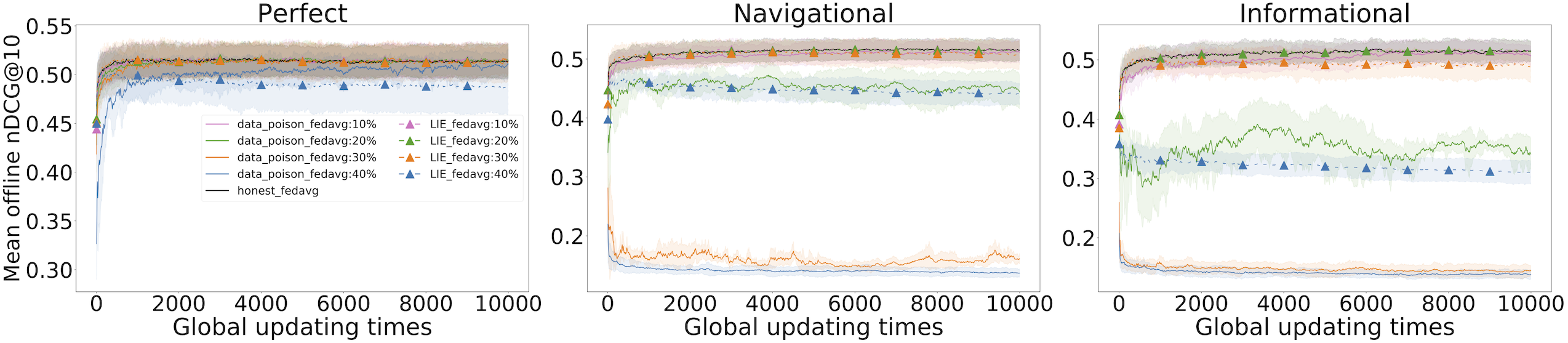}
		\caption{MQ2007 (LIE)}
		\label{fig:mq2007-lie}
	\end{subfigure}
	\begin{subfigure}{1\textwidth}\centering
		\includegraphics[width=1\textwidth]{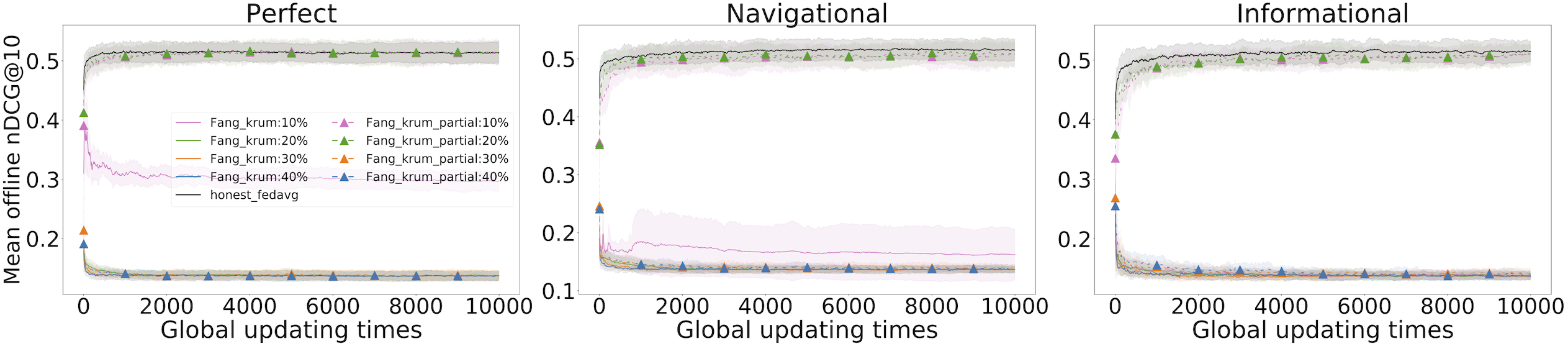}
		\caption{MQ2007 (Fang's attack on Krum)}
		\label{fig:mq2007-fang-krum}
	\end{subfigure}
	\begin{subfigure}{1\textwidth}\centering
		\includegraphics[width=1\textwidth]{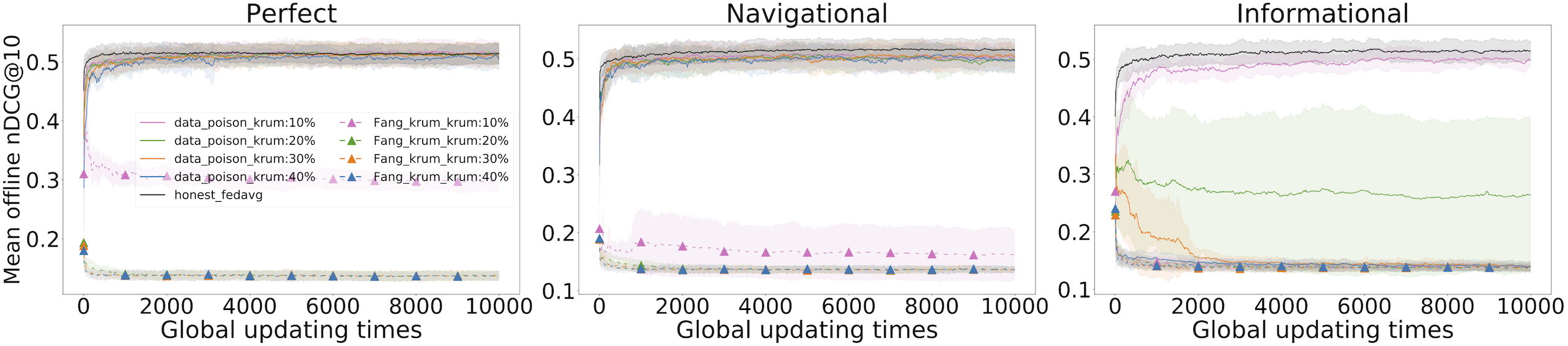}
		\caption{MQ2007 (Fang's attack on Krum - full knowledge)}
		\label{fig:mq2007-fang-krum-data} 
	\end{subfigure}
	\caption{Offline performance (nDCG@10) for MQ2007 under model poisoning attacks, simulated using three benign instantiations of click model with percentage of attackers equaling to $\{10\%, 20\%, 30\%, 40\%\}$. The aggregation rule is FedAvg (for Figure~\ref{fig:mq2007-lie}) and Krum (for Figure~\ref{fig:mq2007-fang-krum}- \ref{fig:mq2007-fang-krum-data}).
		\label{fig:mq2007:model-attack}}
\end{figure}

\begin{figure}[h]
	\centering
	\begin{subfigure}{1\textwidth}\centering
		\includegraphics[width=1\textwidth]{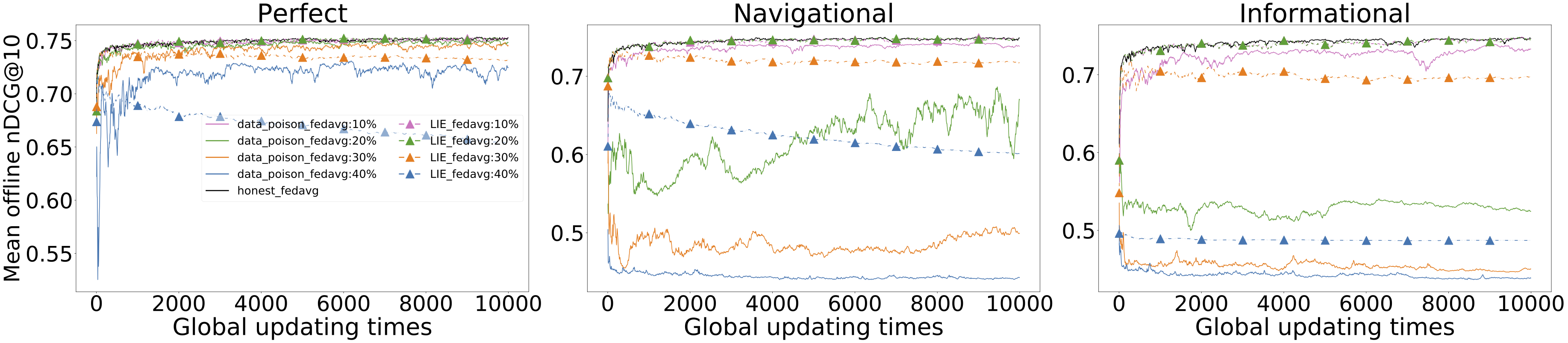}
		\caption{Yahoo (LIE)}
		\label{fig:yahoo-lie}
	\end{subfigure}
	\begin{subfigure}{1\textwidth}\centering
		\includegraphics[width=1\textwidth]{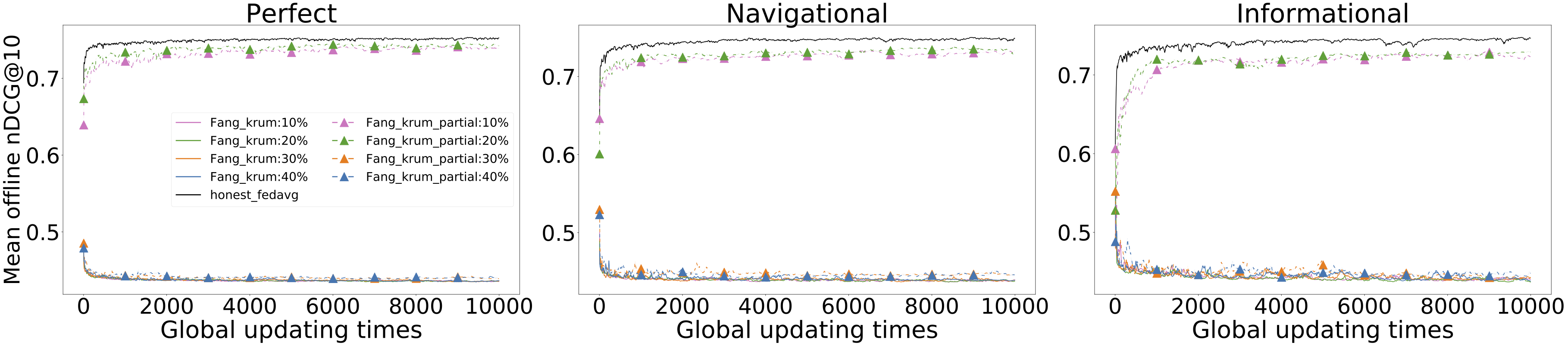}
		\caption{Yahoo (Fang's attack on Krum)}
		\label{fig:yahoo-fang-krum}
	\end{subfigure}
	\begin{subfigure}{1\textwidth}\centering
		\includegraphics[width=1\textwidth]{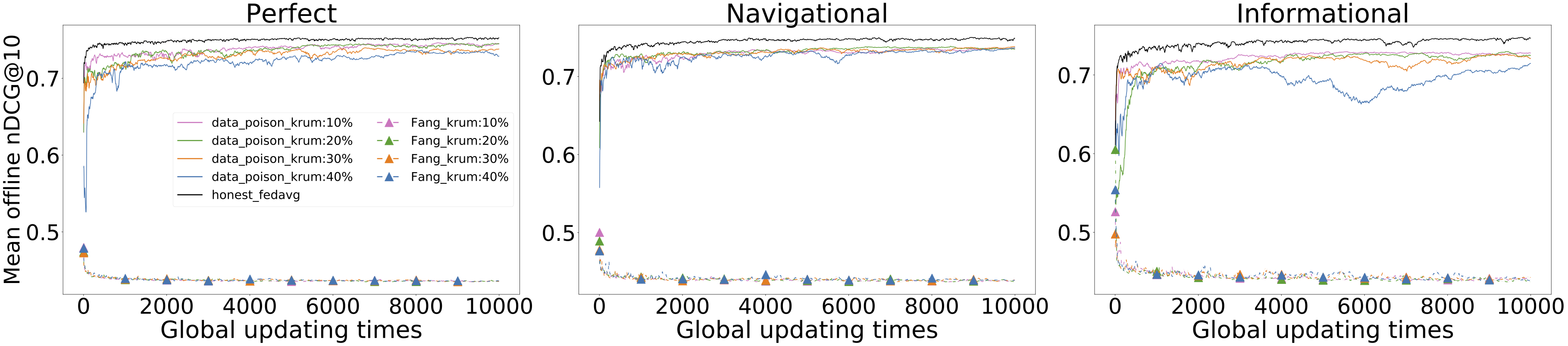}
		\caption{Yahoo (Fang's attack on Krum - full knowledge)}
		\label{fig:yahoo-fang-krum-data} 
	\end{subfigure}
	\caption{Offline performance (nDCG@10) for Yahoo under model poisoning attacks, simulated using three benign instantiations of click model with percentage of attackers equaling to $\{10\%, 20\%, 30\%, 40\%\}$. The aggregation rule is FedAvg (for Figure~\ref{fig:yahoo-lie}) and Krum (for Figure~\ref{fig:yahoo-fang-krum}- \ref{fig:yahoo-fang-krum-data}).
		\label{fig:yahoo:model-attack}}
\end{figure}

\begin{figure}[h]
	\centering
	\begin{subfigure}{1\textwidth}\centering
		\includegraphics[width=1\textwidth]{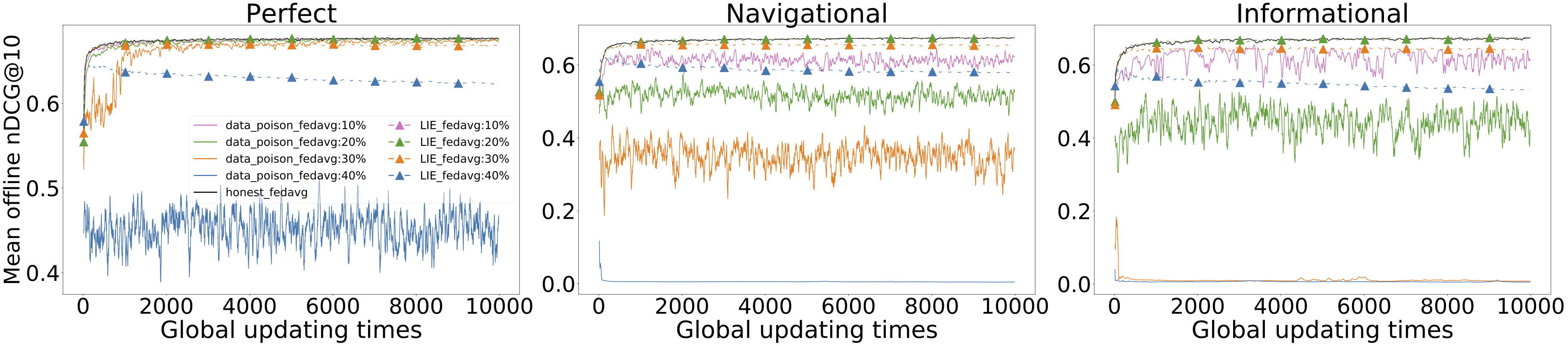}
		\caption{Istella-S (LIE)}
		\label{fig:Istella-s-lie}
	\end{subfigure}
	\begin{subfigure}{1\textwidth}\centering
		\includegraphics[width=1\textwidth]{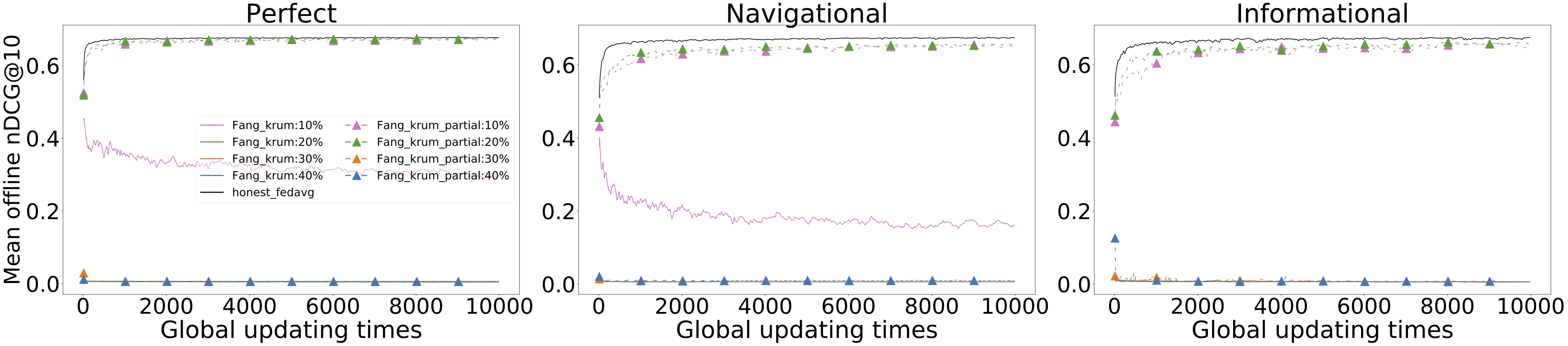}
		\caption{Istella-S (Fang's attack on Krum)}
		\label{fig:Istella-s-fang-krum}
	\end{subfigure}
	\begin{subfigure}{1\textwidth}\centering
		\includegraphics[width=1\textwidth]{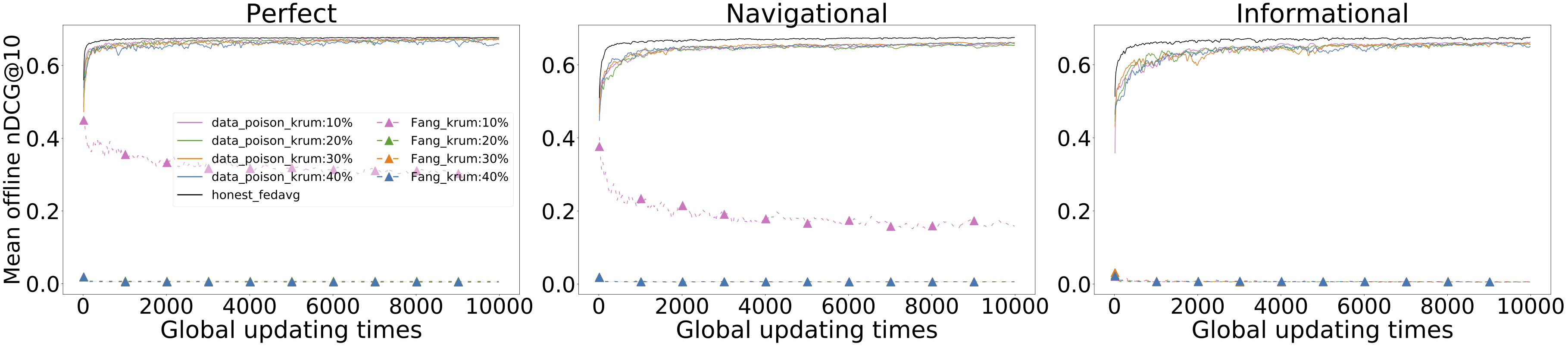}
		\caption{Istella-S (Fang's attack on Krum - full knowledge)}
		\label{fig:Istella-s-fang-krum-data} 
	\end{subfigure}
	\caption{Offline performance (nDCG@10) for Istella-S under model poisoning attacks, simulated using three benign instantiations of click model with percentage of attackers equaling to $\{10\%, 20\%, 30\%, 40\%\}$. The aggregation rule is FedAvg (for Figure~\ref{fig:Istella-s-lie}) and Krum (for Figure~\ref{fig:Istella-s-fang-krum}- \ref{fig:Istella-s-fang-krum-data}).
		\label{fig:Istella-s:model-attack}}
\end{figure}

In this section, we report the results for model poisoning attacks on the remaining dataset. Results on MQ2007 is shown in Figure~\ref{fig:mq2007:model-attack}. Results on Yahoo is shown in Figure~\ref{fig:yahoo:model-attack}. Results on Istella-S is shown in Figure~\ref{fig:Istella-s:model-attack}. Further analysis on results is detailed in Section~\ref{sec:model-poison-result}.

\clearpage
\subsection{Results for Defense under No-Attack}

This section reports the results of four Byzantine aggregation rules with no attacker involving. Further analysis on the results is detailed in Section~\ref{benign-defense}.

\begin{figure}[h]
	\centering
		\begin{subfigure}{1\textwidth}\centering
				\includegraphics[width=1\textwidth]{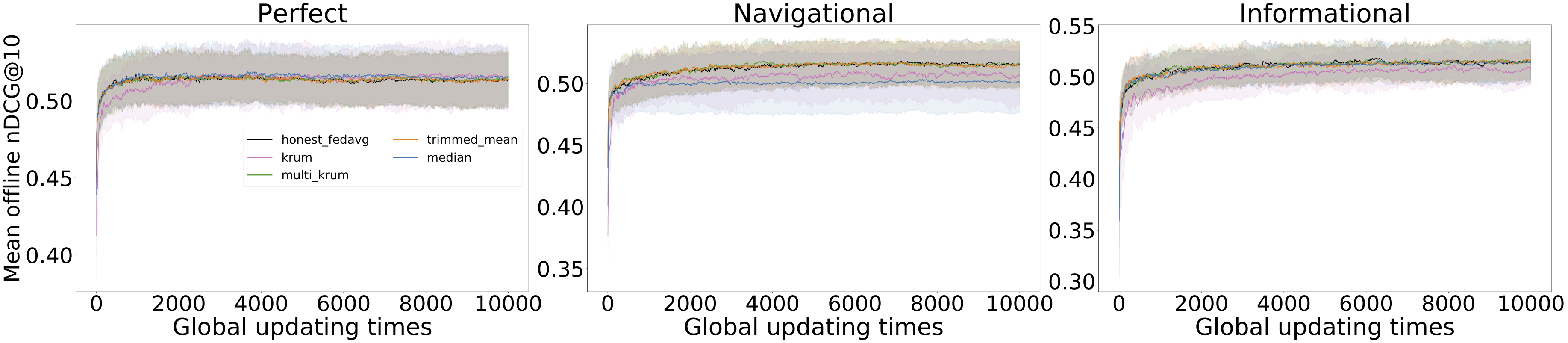}
				\caption{MQ2007 (benign clients)}
				\label{fig:mq2007-benign-defense}
			\end{subfigure}
%	\begin{subfigure}{1\textwidth}\centering
%		\includegraphics[width=1\textwidth]{Chapter5_paper3_poisoning_ictir2023/images/MSLR10K_FPDGD_clients10_batch5_updates10000_honest_CCM_linear_offline_ndcg}
%		\caption{MSLR-WEB10k (benign clients)}
%		\label{fig:mslr-benign-defense}
%	\end{subfigure}
		\begin{subfigure}{1\textwidth}\centering
				\includegraphics[width=1\textwidth]{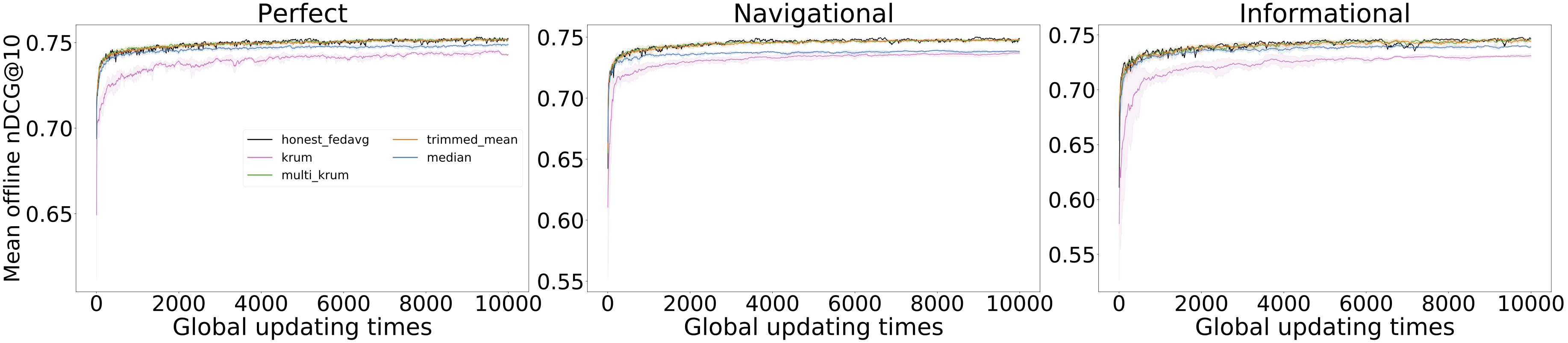}
				\caption{Yahoo (benign clients)}
				\label{fig:yahoo-benign-defense}
			\end{subfigure}
		\begin{subfigure}{1\textwidth}\centering
				\includegraphics[width=1\textwidth]{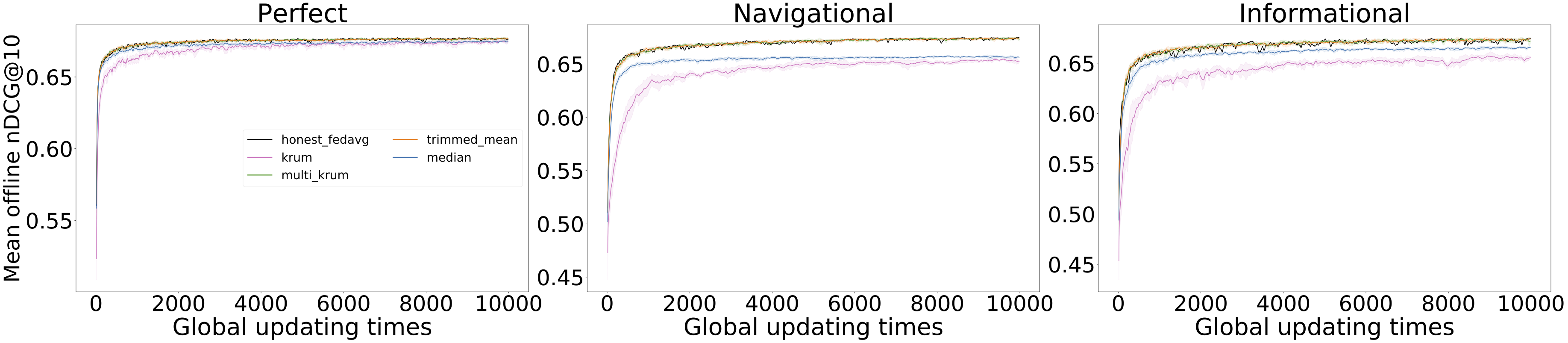}
				\caption{Istella-S (benign clients)}
				\label{fig:istella-benign-defense}
			\end{subfigure}
	\caption{Offline performance (nDCG@10) of FOLTR system when no attack is present but defense strategies are deployed; results averaged across all dataset splits and experimental runs.
		\label{fig:app:benign-defense}}
\end{figure}

%CHAPTER 6
%If you are presenting work which has been previously published, acknowledge this here.
% ***************************************************
% How to introduce a previously published chapter
% ***************************************************
%This is an example of how you might introduce a chapter that has been published previously. 
\cleartoevenpage
\pagestyle{empty}	
%Use this command (above) to suppress the header from the preceding chapter.

\noindent
The following publication has been incorporated as Chapter~\ref{Chap:foltr-unlearning}.

\noindent
1.~\cite{DBLP:conf/ecir/WangLZ24} \textbf{Shuyi Wang}, Bing Liu, and Guido Zuccon, \href{https://doi.org/10.1007/978-3-031-56063-7\_7}{How to Forget Clients in Federated Online Learning to Rank?}, \textit{In European Conference on Information Retrieval (ECIR)}, pages 105--121, 2024

%\begin{table}[h]
%	\begin{center}
%	\begin{tabular}{|c|l|l|}
%		\hline
%		Contributor & Statement of contribution & \% \\
%		\hline
%		\textbf{Your Name}				& writing of text 					& 70\\
%															& proof-reading							& 60 \\
%															& theoretical derivations 	& 70\\
%															& numerical calculations 		& 100\\
%															& preparation of figures 		& 80 \\
%															& initial concept						& 10 \\
%		\hline
%		Co-author 1								& writing of text 					& 20\\
%															& proof-reading							& 10 \\
%															& supervision, guidance 		& 20\\
%															& theoretical derivations 	& 10\\
%															& preparation of figures 		& 20 \\
%															& initial concept						& 10 \\
%		\hline
%		Final Author							& writing of text 					& 10\\
%															& proof-reading							& 30 \\
%															& supervision, guidance 		& 80 \\
%															& theoretical derivations 	& 20 \\
%															& preparation of figures 		& 10 \\
%															& initial concept						& 80 \\
%		\hline
%	\end{tabular}
%	\end{center}
%\end{table}
%
%If your task breakdown requires further clarification, do so here. Do not exceed a single page.

% ***************************************************
% Example of an internal chapter
% ***************************************************
%This is an internal chapter of the thesis.
%If you have a long title, you can supply an abbreviated version to print in the Table of Contents using the optional argument to the \chapter command.
\chapter[Unlearning for FOLTR]{Unlearning for Federated Online Learning to Rank}
\label{Chap:foltr-unlearning}	%CREATE YOUR OWN LABEL.
\pagestyle{headings}

% ********* Enter your text below this line: ********

% ***************************************************

% ********* Enter your text below this line: ********

% ***************************************************

% This is samplepaper.tex, a sample chapter demonstrating the
% LLNCS macro package for Springer Computer Science proceedings;
% Version 2.21 of 2022/01/12
%
%\documentclass[runningheads]{llncs}
%
%\usepackage[T1]{fontenc}
% T1 fonts will be used to generate the final print and online PDFs,
% so please use T1 fonts in your manuscript whenever possible.
% Other font encondings may result in incorrect characters.
%
%\usepackage{graphicx}
% Used for displaying a sample figure. If possible, figure files should
%\usepackage{tabularx}
%\usepackage{booktabs}
%\usepackage{subfigure}
%\usepackage{multirow}
%\usepackage{enumitem}
%\usepackage{wrapfig}
%\usepackage{marvosym}
%\usepackage{xcolor}
%\newcommand{\todo}[1]{{\textcolor{red}{[\em{todo}: #1]}}}
%\newcommand{\modified}[1]{{\textcolor{blue}{[#1]}}}
% If you use the hyperref package, please uncomment the following two lines
% to display URLs in blue roman font according to Springer's eBook style:
%\usepackage{color}
%\renewcommand\UrlFont{\color{blue}\rmfamily}

%
%\begin{document}
	%
	\title{How to Forget Clients in Federated Online Learning to Rank?}

In this chapter, we study another direction for FOLTR, that is, a considerate federated learning system should consider the possibility for clients to leave the federation and request the contributions of their data erased~\cite{liu2021federaser,DBLP:journals/corr/abs-2003-10933,DBLP:conf/infocom/LiuXYWL22}. % cite paper in other domain
This possibility -- dubbed the \textit{right to be forgotten} -- is contemplated in modern data protection legislation, such as in the General Data Protection Regulation (GDPR) emanated by the European Union~\cite{de2020european}. However, the design of existing FOLTR systems is defective as lacking an \textit{unlearning} mechanism to forget certain users' contributions.
An effective and efficient unlearning mechanism is not straightforward to design.
A naive way is to ask all remaining clients to re-execute the training of ranker from scratch. This carries implications in terms of disruption of service and comes with large computational costs, even if the update was done in an offline manner (counterfactually on log data stored in each client), rather than in an online manner (which in turns is impractical as it needs the users to interact again with the search results). 
%would obviously cause huge computation cost. Also, the clients will not accept the burdon of frequent and repeated training locally.
Therefore, a more reasonable unlearning mechanism for FOLTR is necessary, but has not been studied. In this chapter, we aim to fill this gap and provide an initial investigation of unlearning mechanisms for FOLTR. Specifically, we investigate the following research questions:

\begin{enumerate}[label=\textbf{RQ4:}, leftmargin=*]
	\item \bfseries How to effectively and efficiently forget client data from the trained ranker when a client requests to exit the FOLTR system? \itshape
	\begin{enumerate}[label=RQ4.\arabic*:, leftmargin=*]
		\item How to evaluate the effectiveness of unlearning?
		\item How to efficiently forget the client without requiring retraining from scratch?
	\end{enumerate}
\end{enumerate}

The research questions align with two main challenges that we strive to overcome.
The first challenge is how to \textit{efficiently} unlearn without requiring an unreasonable amount of additional computation. Also, the obtained new ranker is expected to have comparable effectiveness to the one retrained from scratch. %without hurting the effectiveness of ranker. 
The second challenge is how to evaluate the \textit{effectiveness} of an unlearning method.
In a FOLTR system with many clients, the effect on ranker's effectiveness of removing a client can be marginal, or even unnoticeable. A natural question from a user leaving the federation is: how can it be proven that the impact of my data on the ranker has been erased? An adequate evaluation method is then required to verify whether an unlearning process is effective in forgetting.

% Thus, we cannot necessarily measure the effectiveness of the unlearning mechanism by observing drops in search effectiveness against an offline search benchmark set -- nor we think that it would be desirable for an effective unlearning mechanism to lead to drops in search effectiveness and thus do not find this evaluation option as appropriate.

% i.e. determine whether the contribution of the client leaving the federation has really been erased from the ranker.

To address these challenges and facilitate future studies, we build the first benchmark for unlearning in FOLTR.
To answer RQ4.2, we adapt an effective unlearning technique emerging from the general federated machine unlearning field (FedEraser~\cite{liu2021federaser}) to the context of FOLTR with adaptation into online training. 
In our method, some historical updates are stored in the local devices and re-used to help retrain a new ranker with much less additional computation cost.
% reusing individual clients' historical updates stored in the local devices. 
To address RQ4.1, for evaluation, we adopt a poisoning attack method~\cite{wang2023analysis} to magnify and control the effect of the client leaving the federation.
%
% (2) an evaluation methodology to measure whether unlearning has happened that is based on the use of a malicious client as the target client for unlearning. 
% In doing so, we also make modifications to the unlearning method to improve its efficiency, by reusing individual clients' historical updates stored in the local devices. 
%
Through extensive empirical experimentation across four learning-to-rank datasets, we study the effectiveness and efficiency of the unlearning method and the factors influencing its performance. 
The utility of our evaluation method is also verified.
	\section{Methodology}
\label{method}

\begin{table}[t]
	\centering
	\caption{Notation used in this chapter. 	\label{tbl_notation} }
	\resizebox{0.9\columnwidth}{!}{		
		\begin{tabular}{ll}
			\toprule
			\bfseries Symbol  & \bfseries Description \\
			\midrule
			$c_i$        & client $i$ in the FOLTR system           \\
			$c^*$        & the unlearned client that requested removal from the FOLTR system           \\
			$n_i$        & number of local updates for client $i$, before unlearning takes place\\ 
			$n'_i$        & number of local updates for client $i$, during the unlearning process \\ 
			$T$        &global update rounds before the unlearning request            \\
			$T_{unlearn}$        &global update rounds for federated unlearning,\\
			&also equals to number of stored local updates for unlearning           \\
			$\Delta t$ & interval of time between stored local updates \\
			$M_i^{local}$        & local ranking model of client $i$ before unlearning            \\
			$\Delta M_i$        & local update of client $i$ in federated learning            \\
			$\Delta M_{i}^{unlearn}$        & local update of client $i$ in federated unlearning            \\
			$\Delta M_{i}^{mal}$        & compromised local update of the client to be unlearned   (client $c^*$)       \\
			$z$        & parameter for poisoning attack (Section~\ref{eval:unlearn})           \\
			\bottomrule
	\end{tabular}}
\end{table}

\subsection{Preliminary}
\label{pre}
Table~\ref{tbl_notation} summarises the  main notations used in this chapter.
In OLTR, for a certain query $q$ and its candidate documents $\mathbf{D}$, a ranking model $M$ is used to compute a relevance score for each candidate document $d \in \mathbf{D}$. The search engine displays the documents according to their scores in descending order, and collects user's interactions with the result page. 
In a centralised setting, $M$ is iteratively trained on the server based on the features of each query-document pair and collected interaction data.

% Unlike in centralised OLTR, FOLTR systems train the ranking model federatively~\cite{wang2021effective,kharitonov2019federated}: 
%methods move the training of the ranking model to a federated setting where

On the contrary, in FOLTR, a global ranking model $M^{global}$ is initialised in a central server and distributed to each client. At training step $t$, each client $c_i$  holds a local ranking model $M_i^{local}$ (received from the central server) and trains the local model with local data (queries, documents, interactions) %using a certain OLTR algorithm (e.g., PDGD~\cite{oosterhuis2018differentiable}) 
for $n_i$ local updating times. After the local training phase\footnote{In our empirical study, we adapt FPDGD in which PDGD algorithm is used in the local training phase. Detailed method is specified in Section~\ref{FPDGD}.}, each client sends its local update $\Delta M_{i,t} = M_i^{local} - M_t^{global}$ to a central server, that aggregates the updates from all clients to generate an updated global model. The most common aggregation rule is FedAvg~\cite{mcmahan2017communication} which updates the global model using the weighted average of all local updates:
\begin{equation}
	M_{t+1}^{global} = M_{t}^{global} + \sum_{i}^{} \frac{n_i}{\sum n_i} \Delta M_{i,t}
	\label{eq_fedagg}
\end{equation}
%The central server broadcasts the newly-updated global model to each client so as to refresh the local models.
The newly-updated global model will be broadcast to each client and replace each local $M^{local}_i$. The whole process is repeated continuously.
%  that then will continue update their local model in a similar fashion.

\subsection{Unlearning in FOLTR}
\label{unlearn}

We next illustrate the unlearning process of FedEraser~\cite{liu2021federaser}, including how we adapted to doing unlearning for a FOLTR system.

Suppose that, during the FOLTR process and after $T$ global update rounds, a client $c^*$ requests to leave the federation and remove all the contributed local updates. Every $\Delta t$ global update rounds (i.e. at rounds $\{1,1+\Delta t,1+2\Delta t, ...\}$ etc.), we instruct each client to store their local updates $\Delta M_{i,t}$.
%We let each client store some local updates $\Delta M_i$ at every $\Delta t$ global update round (i.e. at rounds $\{1,1+\Delta t,1+2\Delta t, ...\}$ etc.). 
The number of local updates stored by each client should then be $T_{unlearn} = [\frac{T}{\Delta t}]$. 

% Borrow the idea from FedEraser~\cite{DBLP:conf/iwqos/LiuMYWL21}, 
The unlearning process takes place as outlined below, in which steps (2)-(4) are performed iteratively and the iterations correspond to the global update rounds $\{1,1+\Delta t,1+2\Delta t, ...\}$ in the original FOLTR process:

\begin{enumerate}
	\item The global ranking model is initialized in the same way as in the original FOLTR process, and is passed to the clients. 
	
	\item Then, each client $c_i$ but $c^*$ (which left the federation), updates the local model by $n_i'$ steps ($n_i' < n_i$, the local step in unlearning $n_i'$ is by design smaller than that before unlearning, i.e. $n_i$). 
	
	\item Each client calibrates the local update $\Delta M_{i,t}'$ using the stored historical local update $\Delta M_{i,t}$ according to:
	\begin{equation}
		\Delta M_{i,t}^{unlearn} = ||\Delta  M_{i,t}|| \frac{\Delta M_{i,t}'}{||\Delta M_{i,t}'||}	
		\label{eq_federaser}
	\end{equation}
	where $||\Delta  M_{i,t}||$ indicates the step size of the global update and $\frac{\Delta M_{i,t}'}{||\Delta M_{i,t}'||}$ indicates the direction of the update ($|| \cdot ||$ is $L_2$-norm). 
	
	\item Each client  $c_i$ sends the calibrated updates $\Delta M_{i,t}^{unlearn}$ to the central server and 
	% After the local updates being calibrated based on Equation~\ref{eq_federaser}, 
	the global model is updated using Equation~\ref{eq_fedagg}. 
\end{enumerate}

%\noindent
%(1) 
%
%\noindent
%(2) 
%
%\noindent
%(3) Before global updating time $t$, each client continue training the ranker locally, and calibrate the local update $\Delta M_{i,t}'$ using the stored historical local update $\Delta M_{i,t}$ as in
%\begin{equation}
%    \Delta M_{i,t}^{unlearn} = ||\Delta  M_{i,t}|| \frac{\Delta M_{i,t}'}{||\Delta M_{i,t}'||}	
%    \label{eq_federaser}
%    \end{equation}
%where $||\Delta  M_{i,t}||$ indicates the step size of global updating and $\frac{\Delta M_{i,t}'}{||\Delta M_{i,t}'||}$ indicates the direction of the update. 
%
%\noindent
%(4) Each remaining client sends calibrated updates to server and 
%% After the local updates being calibrated based on Equation~\ref{eq_federaser}, 
%the global model is updated by Equation~\ref{eq_fedagg}. 

% \noindent
After $T_{unlearn}$ times of global aggregation, the unlearned global model is obtained. In theory, the impact of client $c^*$ is still imposed through $\Delta M_{i,t}$ in Eq.~\ref{eq_federaser}. But this impact is weakened since $\Delta M_{i,t}$ is used in fewer global updating times and the way of using $\Delta M_{i,t}$ is different from the original FOLTR process (Eq.~\ref{eq_fedagg}). The actual impact will be evaluated as in Sec.~\ref{eval:unlearn}.

\subsection{Efficiency Analysis}
In total, the above unlearning approach only requires $n_i' \cdot T_{unlearn}$ local updates for each client $c_i$. Retraining the federated ranker from scratch instead requires $n_i\cdot T$ updates ($n_i'$ is set to be less than $n_i$ to reduce the local computation times). Thus,  in terms of training efficiency compared to the baseline condition of retraining from scratch, the unlearning approach provides a reduction of $(n_i\cdot T)/(n_i' \cdot T_{unlearn}) = \frac{n_i}{n_i'}\cdot \Delta t$ local updates for each client $c_i$. Along with a reduction in local training, the unlearning process also reduces the communications required between  each client and the central server by $\frac{T}{T_{unlearn}} = \Delta t$ times. This reduced number of updates comes at the expense of some extra space required to store the $T_{unlearn}$ updates\footnote{These local updates would ideally be stored within each client, but they could be stored instead in the central server: this though would required extra communication cost to provide the local updates back to the clients when needed.}. For each client, the storage space cost is $[\frac{T}{\Delta t}] \cdot ||\Delta  M_{i,t}||$, where $||\Delta  M_{i,t}||$ indicates the $L_2$-norm of each local updates and it is determined by ranking model structure and the number of features.

\subsection{Evaluating Unlearning}
\label{eval:unlearn}

% To evaluate the effectiveness of unlearning, we need to design methods which can detect if the model removes the contribution of the client to be unlearned. Ideally, remove one client from the federation does not impact the overall performance of ranking model in FOLTR.

The impact of a typical client on the whole FOLTR system can be marginal as later shown in Figure~\ref{fig_origin}. For the purpose of evaluating the unlearning approach, we need to magnify the effect of the client that is leaving the FOLTR system and make this effect relatively controllable. In previous research on machine unlearning and federated unlearning, the effectiveness of unlearning is verified by comparing the effectiveness of the model before and after unlearning~\cite{chundawat2023zero}. This evaluation method comes with a drawback: effective unlearning is associated with a loss in model effectiveness -- a situation that is undesirable and that would penalise methods that can unlearn and do not hurt model effectiveness. 
An alternative direction has been recently proposed: leverage poisoning or backdoor attacks to evaluate unlearning~\cite{yuan2023federated,wu2022federated,halimi2022federated,sommer2020towards,sommer2022athena,hu2022membership}.

Inspired by poisoning attacks methods for federated learning~\cite{baruch2019little} and FOLTR systems~\cite{wang2023analysis} (introduced in Chapter~\ref{Chap:foltr-poisoning}) where malicious clients compromise the effectiveness of the trained models by poisoning the local training data or the model updates, we add noise to the local updates of the client to be unlearned. By doing so, we expect the unlearned clients to be clearly distinguishable from others. Due to the noise being injected, the original (i.e., before unlearning) global model performs worse to some extent compared to the model trained without the poisoned client (i.e., retrained from scratch). However, after unlearning, if the contribution of the poisoned client has been successfully removed, the overall effectiveness of the unlearned global model should improve, achieving similar effectiveness as if it was retrained from scratch. We instantiate this intuition by compromising the unlearned client's ($c^*$) local model after each local updating phase using:

\begin{equation}
M_{c^*}^{mal} = - z \cdot M_{c^*}^{local}	
\label{eq_poison_model}
\end{equation}
where $z>0$ represents how much we compromise the local model, while the negative coefficient is added to change the local model to its opposite direction. Thus, the compromised local update for the client to be unlearned is:

\begin{equation}
\Delta M_{c^*}^{mal} = - z \cdot \Delta M_{c^*}^{local} - (z+1) \cdot M_{t}^{global}	
\label{eq_poison_update}
\end{equation}
In the global updating phase before unlearning (Eq.~\ref{eq_fedagg}), we replace $\Delta M_{c^*}^{local}$ with $\Delta M_{c^*}^{mal}$ for $c^*$. In our experiments, we set $z = 2$ as we found it sufficient in degrading the effectiveness of the global model. 
Note here our poisoning method does not have the burden of hiding from detection as in real poisoning attack scenarios and thus it is simple and its parameters can be tuned as per need.

A key tenet of this evaluation is that the ability to remove such a distinguishable client is equivalent to the ability to remove a much less unique client: we are unsure whether this assumption holds true, and we are not aware of relevant literature that clearly support this.
	\section{Experimental Setup}

We next describe our experimental setup for unlearning in the context of a FOLTR system.

%\begin{table}[t]
%	\centering
%	\caption[centre]{Instantiations of SDBN click model for simulating user behaviour in experiments. $rel(d)$ denotes the relevance label for document $d$. Note that in the MQ2007 dataset, only three-levels of relevance are used. We demonstrate the values for MQ2007 in bracket.}
%	\resizebox{0.8\columnwidth}{!}{		
%	\begin{tabular}{cccccc}
%		\hline
%		& \multicolumn{5}{c}{$P(\mathit{click}=1\mid rel(d))$}  \\
%		%\hline
%		\cmidrule(r){2-6}
%		\emph{rel(d)} & 0 & 1 & 2 & 3 & 4 \\
%		\hline
%		\emph{perfect} &  0.0 (0.0)&  0.2 (0.5)&  0.4 (1.0)&  0.8 (-)&  1.0 (-)\\
%		\emph{navigational} & 0.05 (0.05)& 0.3 (0.5)&  0.5 (0.95)&  0.7 (-)&  0.95 (-)\\
%		\emph{informational} &  0.4 (0.4)&  0.6 (0.7)&  0.7 (0.9)&  0.8 (-)&  0.9 (-)\\
%		\hline\hline
%		& \multicolumn{5}{c}{$P(\mathit{stop}=1\mid click=1,  rel(d))$} \\
%		%\hline
%		\cmidrule(r){2-6}
%		\emph{rel(d)} & 0 & 1 & 2 & 3 & 4 \\
%		\hline
%		\emph{perfect}  & 0.0 (0.0)& 0.0 (0.0)&  0.0 (0.0)&  0.0 (-)&  0.0 (-)\\
%		\emph{navigational} & 0.2 (0.2)&  0.3  (0.5)&  0.5 (0.9)&  0.7 (-)&  0.9 (-)\\
%		\emph{informational} & 0.1 (0.1)&  0.2 (0.3)&  0.3 (0.5)&  0.4 (-)&  0.5 (-)\\
%		\hline
%	\end{tabular}}
%	\label{tbl:click}
%	%	\vspace{-15pt}
%\end{table}

%We next describe our experimental setup to evaluate the considered unlearning process in the context of a FOLTR system.

%\subsubsection*{\textbf{Datasets.}}
\textbf{Datasets.} 
We evaluate using 4 common learning-to-rank (LTR) datasets: MQ2007~\cite{DBLP:journals/corr/QinL13}, MSLR-WEB10K~\cite{DBLP:journals/corr/QinL13}, Yahoo~\cite{DBLP:journals/jmlr/ChapelleC11}, and Istella-S~\cite{lucchese2016post}. Detailed descriptions about each dataset are discussed in Section~\ref{chap2-dataset}. Our experimental results are averaged across all data splits.

%\subsubsection*{\textbf{User simulations.}}
\textbf{User simulations.} 
We follow the standard setup for user simulations in OLTR~\cite{oosterhuis2018differentiable,wang2021effective,wang2022non} by randomly selecting queries for a user and relying on the \textit{Cascade Click Model} (CCM) click model for simulating user's clicks. Specifically, for each query, we limit the search engine result page (SERP) to 10 documents. User clicks on the displayed ranking list are generated based on the implementation of CCM click models detailed in Section~\ref{chap2-click}.

%\subsubsection*{\textbf{Federated setup.}}
\textbf{Federated setup.} We consider 10 clients participating in the FOLTR process; among these 10, one client requests to be unlearned. The original federated setup (before unlearning) involves 5 local updating steps ($n_i = 5$) among all participants and 10,000 global updating steps ($T$). We assume the client requests to leave the federation at global step $T = 10,000$. During the original training in our FOLTR experiments (i.e., before the unlearning request is proposed), each client holds a copy of the current ranker and updates the local ranker by issuing $n_i = 5$ queries along with the respective click responses. After the local updating finishes, the central server will receive the updated ranker from each client and aggregate all local messages to update the global one. At each $\Delta t$ global steps (i.e. at rounds $\{1,1+\Delta t,1+2\Delta t, ...\}$ etc.), each client will also keep a copy of their local ranker update to the local device for the calibration purpose in the unlearning process. In our evaluation of unlearning (results in Figure~\ref{fig_unlearn}), we set $\Delta t = 10$ thus the total number of stored local updates is $[\frac{T}{\Delta t}] = 1000$, equaling to the global steps in the unlearning process ($T_{unlearn}$). For further hyper-parameter analysis (results in Table~\ref{tbl_unlearn}), we set a wider value ranges of $\Delta t \in \{5, 10, 20\}$. The unlearning process follows the same federated setup, with $T_{unlearn}$ global steps, $n'_i \in \{1,2,3,4\}$ local steps for the remaining 9 clients. 
We experiment with a linear model as the ranker with learning rate $\eta = 0.1$ and zero initialization, which we train with and rely on the configuration of the state-of-the-art FOLTR method, FPDGD~\cite{wang2021effective}.

%\subsubsection*{\textbf{Evaluation metric.}}
\textbf{Evaluation metric.} As we limit each SERP to 10 documents, nDCG@10 is used for evaluating the overall offline effectiveness of the global ranker both before and after unlearning. Effectiveness is measured by averaging the nDCG scores of the global ranker on the queries in the held-out test dataset. This is in line with previous work on OLTR and FOLTR. Unlike previous work~\cite{oosterhuis2018differentiable,wang2021effective,jia2022learning}, online evaluation is not considered in this work, as we do not need to monitor user experience during model update. Instead, we focus on measuring the overall impact of the unlearned client (the "attacker" in Figure~\ref{fig_origin}) comparing to other baselines, and the dynamic or final performance gain during the unlearning process (results in Figure~\ref{fig_unlearn} and Table~\ref{tbl_unlearn}) following our evaluation process specifically for unlearning (specified in Section~\ref{eval:unlearn}).

	\section{Results and Analysis}

%\begin{figure}[t]
%	\centering \includegraphics[width=0.6\columnwidth]{images/figure2.pdf}
%	\caption{Relationships between the 9H-1M (green line), 10H-0M (black line), 9H-0M (pink line) FOLTR configurations. Circles are clients.
%		\label{fig_configs}}
%		\vspace{-12pt}
%\end{figure}

\subsection{Validation of Evaluation Methodology}

%\textbf{Before unlearning.} We use the poisoned model update specified in Equations~\ref{eq_poison_model}  and~\ref{eq_poison_update} to instantiate the client to be unlearned. Our experimental results are shown in Figure~\ref{fig_origin}.

\begin{wrapfigure}{r}{0.5\textwidth}
	\vspace{-20pt}
	\begin{center}
		\includegraphics[width=0.5\columnwidth]{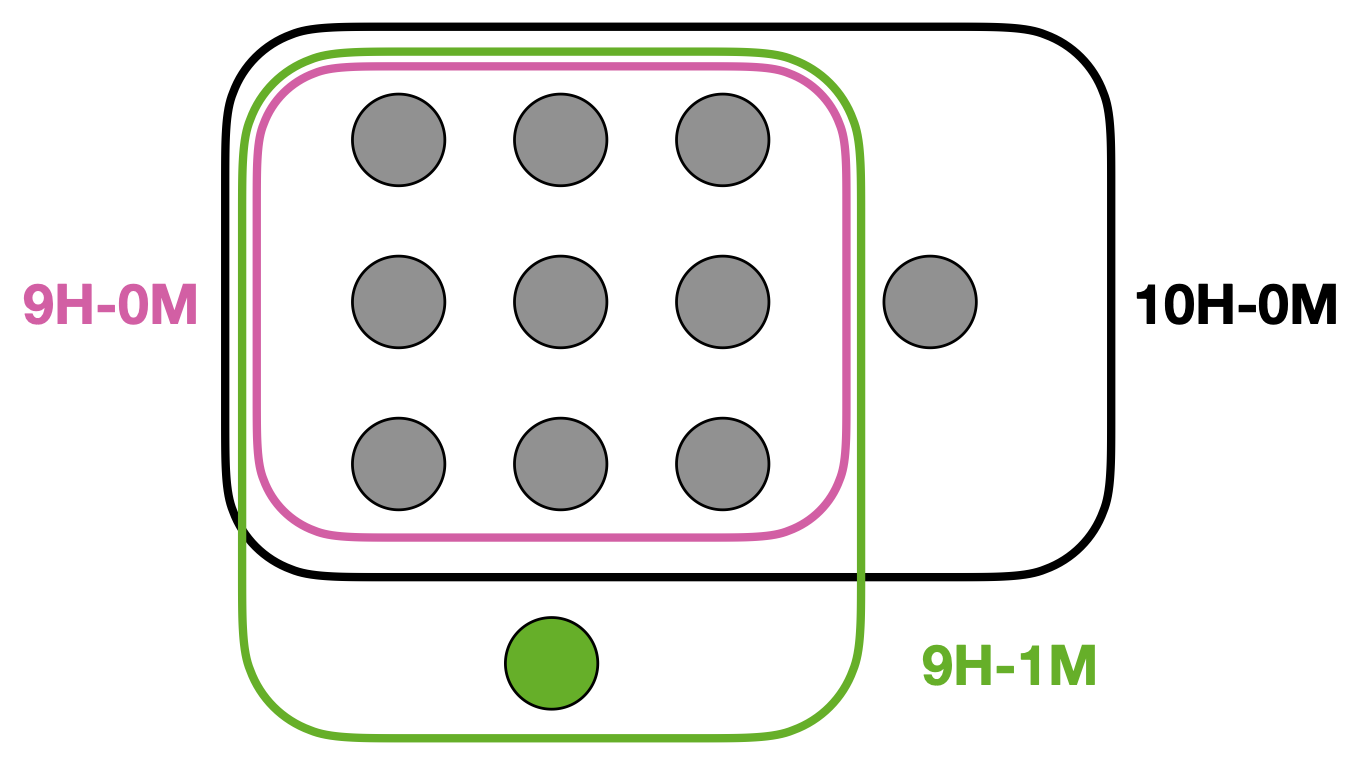}
	\end{center}
	\caption{Relationships between FOLTR configurations: 9H-1M (green line), 10H-0M (black), 9H-0M (pink). Circles are clients.
		\label{fig_configs}}
\end{wrapfigure}

%\textbf{Evaluating unlearning evaluation.} 
The unlearning evaluation methodology is based on instantiating the unlearned client as a malicious actor that injects noise in the learning process. We start by investigating if our evaluation methodology could be effective in identifying whether the unlearning has happened. For this, we consider three configurations (visualized in Figure~\ref{fig_configs}):

\begin{enumerate}
	\item 9H-1M (green line): effectiveness obtained when one of the 10 clients is set to produce noisy updates (i.e. one client behaves maliciously).
	
	\item 10H-0M (black line): effectiveness obtained when all 10 clients behave in an honest way, (i.e. no client is acting maliciously).
	%effectiveness obtained using the unlearning methods outlined in Section~\ref{label}, where the updates made by client that has been unlearned were noisy.
	
	\item 9H-0M (pink line): effectiveness obtained when considering only the 9 honest clients from above, without the client that produces noisy updates (i.e. no malicious client).
\end{enumerate}

Fig.~\ref{fig_configs} clarifies the relationship between these rankers -- all rankers share the same common set of 9 honest clients, but they differ in the 10th client considered: a malicious client for 9H-1M, an honest client for 10H-0M, and no 10th client for 9H-0M.

\begin{figure}[t]
	\centering
	\begin{subfigure}{1\textwidth} \centering
		\includegraphics[width=1\textwidth]{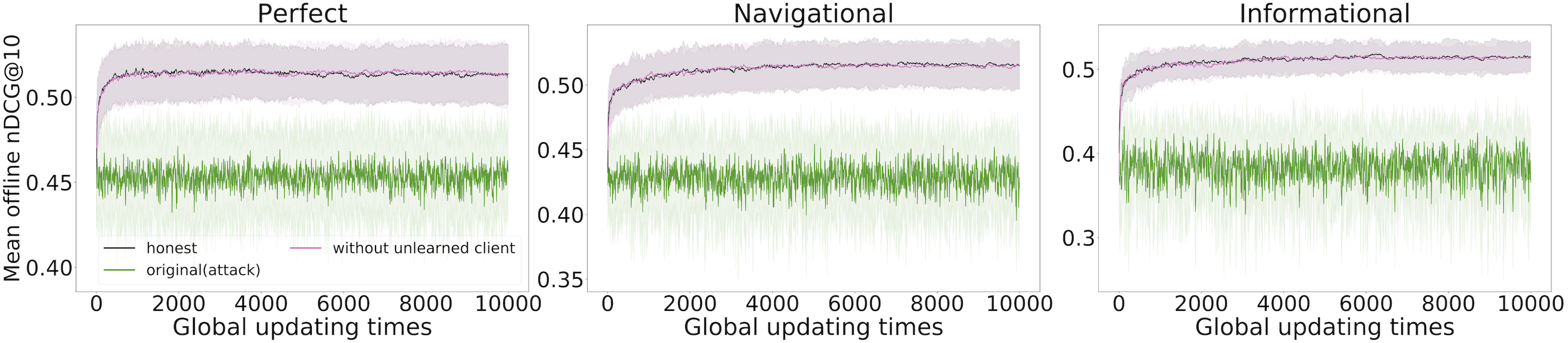}
		\caption{MQ2007}
		\label{fig:mq2007} 
	\end{subfigure}
	\begin{subfigure}{1\textwidth} \centering
		\includegraphics[width=1\textwidth]{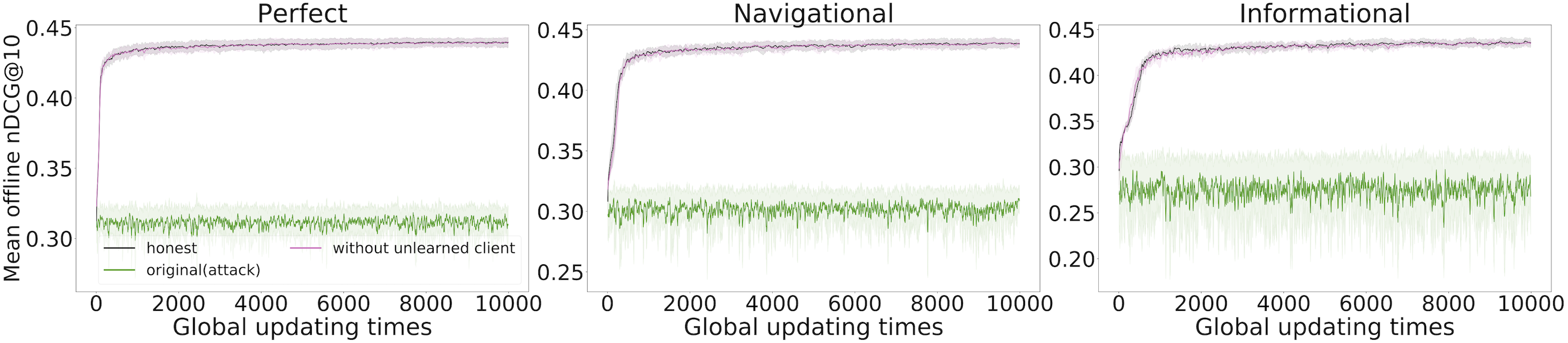}
		\caption{MSLR-WEB10K}
		\label{fig:mslr}
	\end{subfigure}
	\begin{subfigure}{1\textwidth} \centering
		\includegraphics[width=1\textwidth]{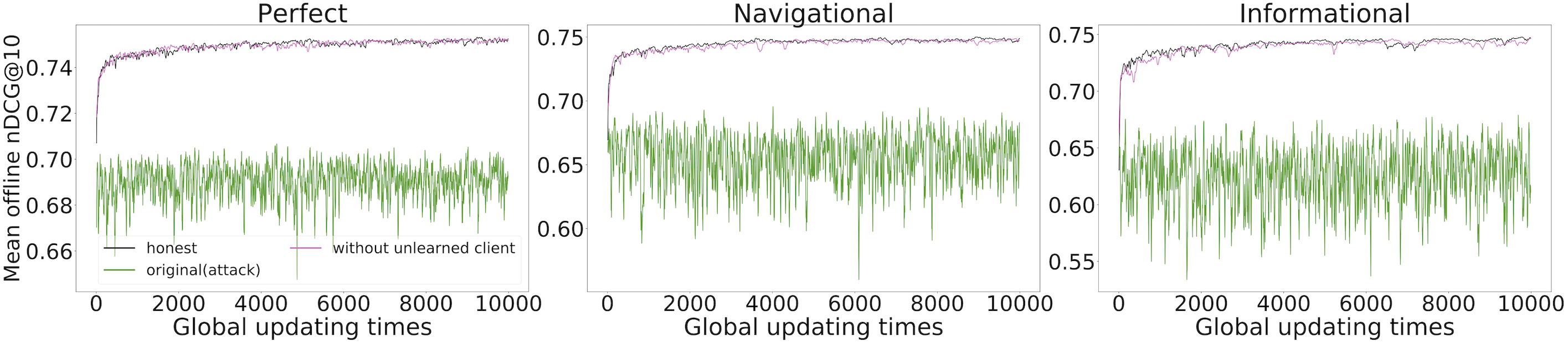}
		\caption{Yahoo}
		\label{fig:yahoo} 
	\end{subfigure}
	\begin{subfigure}{1\textwidth} \centering
		\includegraphics[width=1\textwidth]{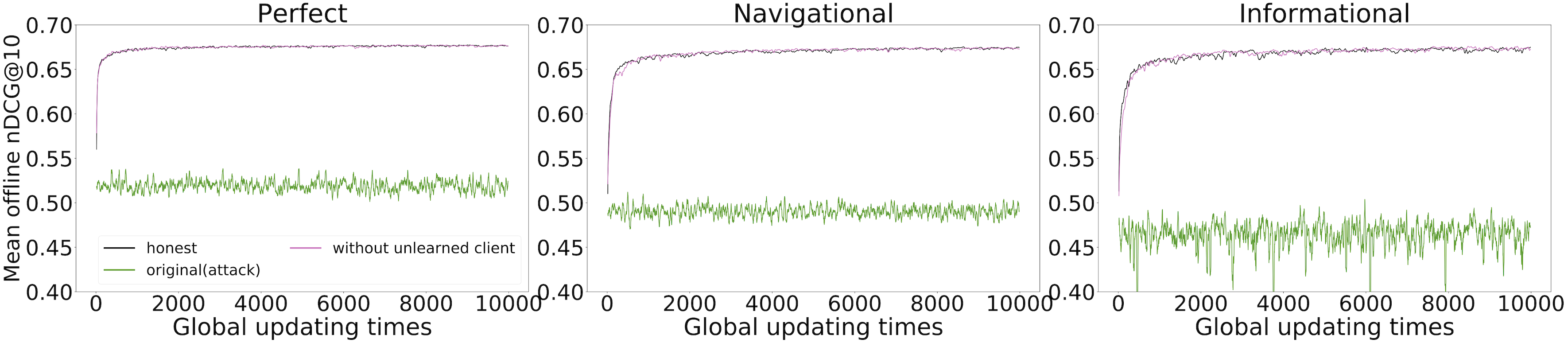}
		\caption{Istella-S}
		\label{fig:istella} 
	\end{subfigure}
	\caption{Offline effectiveness (nDCG@10) obtained under the 9H-1M (green line), 10H-0M (black line), 9H-0M (pink line) FOLTR configurations with three click modes (\textit{Perfect, Navigational, Informational}). Results are averaged across all dataset splits and experimental runs. These results motivate the use of the evaluation methodology based on the malicious client to evaluate the effectiveness of unlearning.
		\label{fig_origin}}
\end{figure}

Fig.~\ref{fig_origin} reports the results obtained by these three conditions across three click modes on the four datasets. The results highlight that it is the addition of the malicious client (i.e. client $c^*$ that will be the target of the unlearning) that sensibly reduces the effectiveness of the ranker. Note that the 9H-0M is the ranker one would obtain if the unlearning process was implemented as a federated re-training of the ranker from scratch by only considering the 9 clients remaining after the removal of client $c^*$. Comparing this ranker with the 10H-0M, we further highlight the need for evaluation based on the malicious client. In fact, 10H-0M and 9H-0M only consider honest clients, but in the 9H-0M one of these clients has been removed -- but the effectiveness of two rankers is indistinguishable.

%; the green line ("original(attack)") denotes the effectiveness throughout the online training process of the original global model before unlearning takes place. We find that compromising one single client's local update consistently degrades the FOLTR effectiveness compared to the remaining two conditions: (1) the "honest" setting (black line), obtained without compromising the local updates, and (2) the "without unlearned client" setting (pink line), which represents the ranker learned from scratch by not considering updates from the client that will eventually leave the federation. 
%The results in Figure~\ref{fig_origin} show that removing a (non-malicious) client from the federation has no or just marginal impact on the overall effectiveness of the FOLTR ranker.

%. Importantly, the "without unlearned client" setting, which represents the ranker learned after the client has left and using our method as unlearning mechanism, performs similarly to the "honest" setting. 
%This result demonstrates that removing an honest client from the federation does not impact the overall performance of the global model significantly.

%% original all results

\begin{figure}[t]
	\centering
	\begin{subfigure}{1\textwidth} \centering
		\includegraphics[width=1\textwidth]{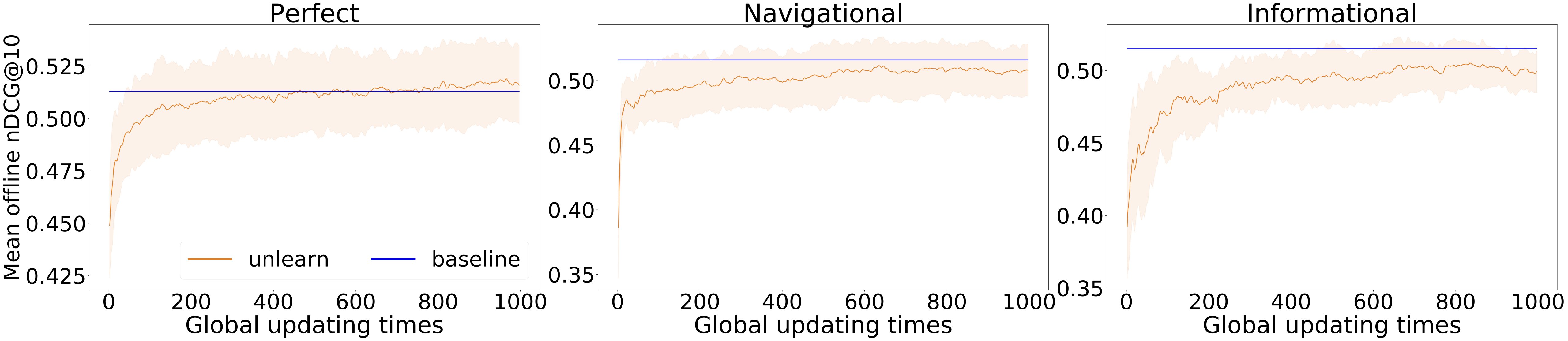}
		\caption{MQ2007}
		\label{fig_mq2007_unlearn}
	\end{subfigure}
	\begin{subfigure}{1\textwidth} \centering
		\includegraphics[width=1\textwidth]{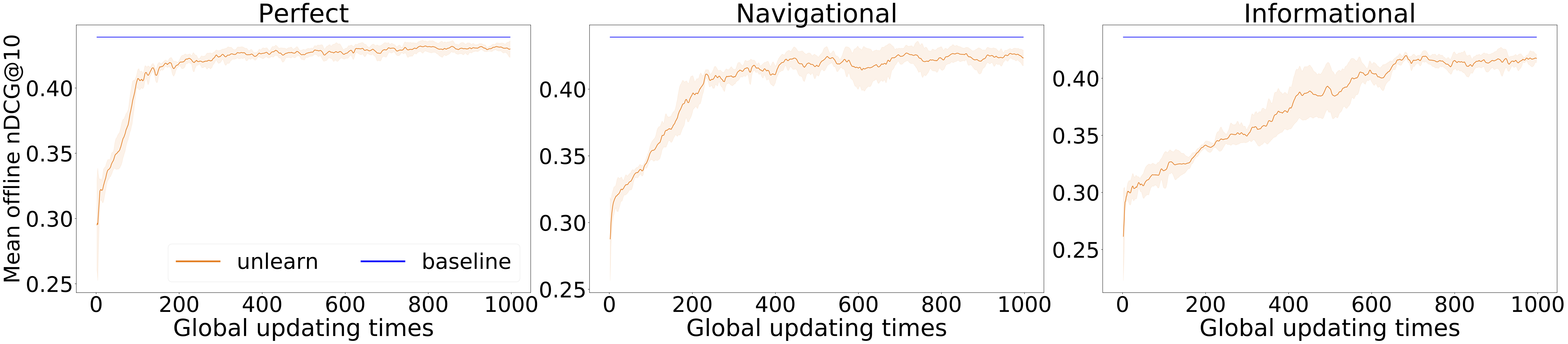}
		\caption{MSLR-WEB10K}
		\label{fig_mslr_unlearn}
	\end{subfigure}
	\begin{subfigure}{1\textwidth} \centering
		\includegraphics[width=1\textwidth]{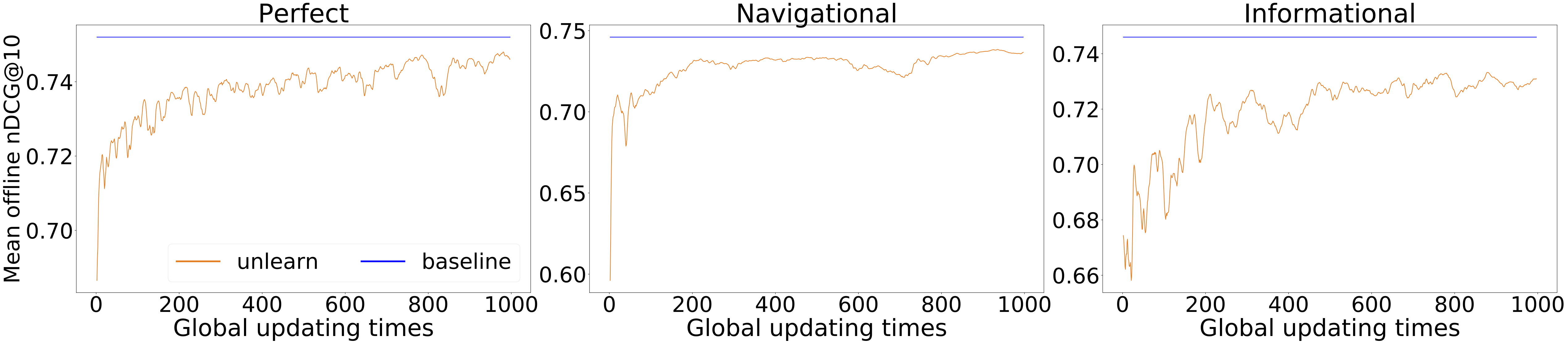}
		\caption{Yahoo}
		\label{fig_yahoo_unlearn}
	\end{subfigure}
	\begin{subfigure}{1\textwidth} \centering
		\includegraphics[width=1\textwidth]{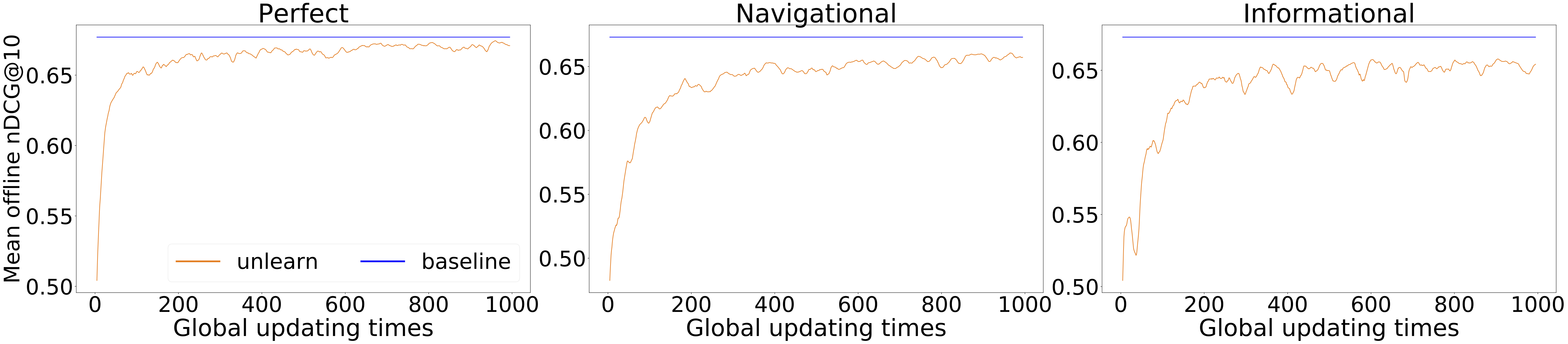}
		\caption{Istella-S}
		\label{fig_istella_unlearn} 
	\end{subfigure}
	\caption{Comparison between the offline effectiveness (nDCG@10) after the unlearning method is applied (ranker $\mathcal{U}$(9H-1M) denoted as "unlearn") and the ranker is retrained from scratch after client $c^*$ is removed (ranker 9H-0M denoted as "baseline"). For the unlearning process, we set $n'_i = 3$ with $\Delta t = 10$ and show the evaluation values across all global steps. For the baseline setup, we only show the final nDCG@10 score after retraining finishes.
		\label{fig_unlearn}}
\end{figure}

\subsection{Effectiveness of Unlearning}
\label{effectiveness_unlearn}
%\noindent\textbf{After unlearning.} 
We now investigate the effectiveness of the unlearning method. For this, we consider ranker 9H-1M and we perform unlearning to remove client $c^*$, which is the malicious client; this leads to ranker $\mathcal{U}$(9H-1M). 
Table~\ref{tbl_unlearn} reports the effectiveness of the global model after unlearning has taken place and under different settings of hyper-parameters $n'_i$ and $\Delta t$. A more detailed analysis of the impact of $n'_i$ and $\Delta t$ is presented in Section~\ref{impact_analysis}.

%To show the effectiveness of the unlearning mechanism, we assume a client asks to leave the FOLTR system at global update time $T = 10,000$. We then consider as baseline the nDCG@10 obtained at $T$ by the global model trained without the unlearned client (pink line in Figure~\ref{fig_origin}) and compare to this for evaluating the effectiveness after unlearning has taken place. Tables~\ref{tbl_analysis_n} \& \ref{tbl_analysis_t} report the global model effectiveness after unlearning under different settings of hyper-parameters $n'$ and $\Delta t$; more detailed analysis of the impact of $n'$ and $\Delta t$ are reported in Section~\ref{impact_analysis}.

In Fig.~\ref{fig_unlearn}, we report the offline effectiveness on four datasets obtained by the unlearning mechanism during the $[\frac{T}{\Delta t}] = 1,000$ global update times, where we set $\Delta t = 10$ and $n'_i = 3$.
Compared to the original model 9H-1M\footnote{The effectiveness of 9H-1M is not shown in Figure~\ref{fig_unlearn} for clarity. The reader can cross reference Figure~\ref{fig_unlearn} with Figure~\ref{fig_origin}, which instead contains the effectiveness of 9H-1M.}, the unlearned model $\mathcal{U}$(9H-1M) achieves better effectiveness and it gradually converges towards the effectiveness of the 9H-0M model, showing that the unlearned model is able to successfully remove the impact of the unlearned client $c^*$.

\subsection{Hyper-parameters Analysis}
\label{impact_analysis}

Next, we study the sensitivity of the unlearning method to its two hyper-parameters: the number of local updates for unlearning $n'_i \in \{1,2,3,4\}$ and interval of time between stored updates $\Delta t \in \{5, 10, 20\}$. We report the results of this analysis in Table~\ref{tbl_unlearn} with the final nDCG@10 score after the unlearning (or retraining). The effectiveness of ranker 9H-0H, i.e. the global model re-trained from scratch for $T = 10,000$ iterations and without client $c^*$, represents the baseline condition. 

%To investigate the impact of two hyper-parameters in the proposed unlearning approach, we set the number of local updates for unlearning to $n' \in \{1,2,3,4\}$ and $\Delta t \in \{5, 10, 20\}$. The experimental results are shown with the performance of unlearned global ranker under the impact of $n'$ with fixed $\Delta t = 10$ (Table~\ref{tbl_analysis_n}) and impact of $\Delta t$ with fixed $n' = 3$ (Table~\ref{tbl_analysis_t}). Baseline represents performance of global model trained in the "without unlearned client" setting in Section~\ref{effectiveness_unlearn} after global updating $T = 10,000$.

%\textbf{Impact of $\mathbf{n'_i$}.}
\textbf{Impact of $n'_i$.} 
By design, lower values of $n'_i$ lead to higher time savings as the time required by each local update is similar.
However, lower values of $n'_i$ may mean there is insufficient training data in each iteration, and this can lead to lower ranking effectiveness. Our experimental findings display this trade-off with relative higher effectiveness obtained when more local updates ($n'_i$) are used during the unlearning process.

\textbf{Impact of $\Delta t$.}
%\textbf{Impact of $\mathbf{\Delta t}$.} 
Larger values of $\Delta t$ correspond to less global updates needed for unlearning. In fact, the required global updating time is$[\frac{T}{\Delta t}]$, where in our experiments $T = 10,000$. Table~\ref{tbl_unlearn} shows that, in most cases, the model for which unlearning has taken place delivers higher ranking effectiveness for small values of $\Delta t$. This higher effectiveness, however, comes at the cost of extra time required by the unlearning process.

\begin{sidewaystable*}[t]
	\centering
	\small
	\caption{Offline effectiveness (nDCG@10) of (1) the ranker trained from scratch after client $c^*$ has been removed (9H-0M), and (2) the ranker for which unlearning is performed with our method ($\mathcal{U}$(9H-1M)). Effectiveness is analysed with respect to the hyper-parameters $n'_i \in \{1,2,3,4\}$ and $\Delta t \in \{5, 10, 20\}$, under three different click models, and averaged across dataset splits. The best results are highlighted in boldface with superscripts denoting results for statistical significance study (paired Student's t-test with $p \leq 0.05$ with Bonferroni correction).
		\label{tbl_unlearn}}
	\begin{tabular}{lcc|lll|lll|lll}
		\toprule
		&&& \multicolumn{3}{c}{Click Model: Perfect} & \multicolumn{3}{c}{Click Model: Navigational} & \multicolumn{3}{c}{Click Model: Informational}\\ 
		\midrule
		Dataset&\textbf{\#}&Ranker &$\Delta t = 5$&$\Delta t = 10$&$\Delta t = 20$ &$\Delta t = 5$&$\Delta t = 10$&$\Delta t = 20$ &$\Delta t = 5$&$\Delta t = 10$&$\Delta t = 20$  \\ 
		\midrule
		\multirow{4}{*}{MQ2007}&&9H-0M&0.513&\textbf{0.513}&\textbf{0.513}&\textbf{0.516}& \textbf{0.516}&\textbf{0.516}& \textbf{0.515}&\textbf{0.515}&\textbf{0.515}\\ 
		&a&$\mathcal{U}$(9H-1M), $n'_i=4$&0.514 &0.513&0.511&0.508&0.507&0.501&0.506&0.502&0.492\\
		&b&$\mathcal{U}$(9H-1M), $n'_i=3$& 0.514&0.513&0.507&0.510&0.504&0.499&0.505&0.498&0.491\\
		&c&$\mathcal{U}$(9H-1M), $n'_i=2$&\textbf{0.515} &0.510&0.507&0.505&0.499&0.499&0.502&0.497&0.483\\
		&d&$\mathcal{U}$(9H-1M), $n'_i=1$& 0.511&0.506&0.501&0.504&0.498&0.496&0.489&0.491&0.484\\
		
		\midrule
		\multirow{4}{*}{MSLR-WEB10K}&&9H-0M&\textbf{0.439}&\textbf{0.439}$^{d}$&\textbf{0.439}$^{cd}$&\textbf{0.439}$^{d}$&\textbf{0.439}$^{cd}$&\textbf{0.439}$^{abcd}$&\textbf{0.436}$^{cd}$&\textbf{0.436}$^{abcd}$&\textbf{0.436}$^{abcd}$\\ 
		&a&$\mathcal{U}$(9H-1M), $n'_i=4$&0.436&0.433&0.429&0.433&0.428&0.422&0.429&0.413&0.389\\
		&b&$\mathcal{U}$(9H-1M), $n'_i=3$&0.435 &0.432&0.427&0.429& 0.428&0.414&0.424&0.412&0.361\\
		&c&$\mathcal{U}$(9H-1M), $n'_i=2$&0.432 &0.431&0.423&0.426&0.421&0.389&0.420&0.386&0.348\\ 
		&d&$\mathcal{U}$(9H-1M), $n'_i=1$&0.430 &0.422 &0.414&0.418&0.382&0.344&0.393&0.343&0.332\\
		
		\midrule
		\multirow{4}{*}{Yahoo}&&9H-0M& \textbf{0.752}&\textbf{0.752}$^{bcd}$&\textbf{0.752}$^{abcd}$&\textbf{0.746}$^{bd}$&\textbf{0.746}$^{bcd}$&\textbf{0.746}$^{abcd}$&\textbf{0.746}$^{abcd}$&\textbf{0.746}$^{abcd}$&\textbf{0.746}$^{abcd}$\\
		&a&$\mathcal{U}$(9H-1M), $n'_i=4$&0.748 &0.746&0.741&0.743&0.743&0.734&0.738&0.725&0.726\\ 
		&b&$\mathcal{U}$(9H-1M), $n'_i=3$&0.748&0.740&0.742&0.738&0.739&0.735&0.736&0.725&0.721\\ 
		&c&$\mathcal{U}$(9H-1M), $n'_i=2$&0.746 &0.746&0.742&0.738&0.735&0.733&0.732&0.725&0.715\\
		&d&$\mathcal{U}$(9H-1M), $n'_i=1$&0.746 &0.739&0.732&0.734&0.726&0.725&0.730&0.720&0.713\\
		
		\midrule
		\multirow{4}{*}{Istella-s}&&9H-0M& \textbf{0.677}$^{cd}$&\textbf{0.677}$^{cd}$&\textbf{0.677}$^{abcd}$&\textbf{0.673}$^{bcd}$ &\textbf{0.673}$^{abcd}$ &\textbf{0.673}$^{abcd}$ &\textbf{0.673}$^{abcd}$&\textbf{0.673}$^{abcd}$&\textbf{0.673}$^{abcd}$\\
		&a&$\mathcal{U}$(9H-1M), $n'_i=4$&0.673 & 0.672 &0.664&0.665& 0.659 &0.650&0.663& 0.655&0.654\\
		&b&$\mathcal{U}$(9H-1M), $n'_i=3$&0.673 & 0.670 &0.663&0.664& 0.662 &0.643&0.664& 0.658 &0.639\\
		&c&$\mathcal{U}$(9H-1M), $n'_i=2$&0.669 & 0.670 &0.663&0.661& 0.652 &0.637&0.659& 0.653&0.635\\
		&d&$\mathcal{U}$(9H-1M), $n'_i=1$&0.669 &0.662 &0.657&0.653& 0.637 &0.604&0.649& 0.641&0.600\\
		
		\bottomrule
	\end{tabular}
\end{sidewaystable*}

	\clearpage
\section{Conclusion}

This chapter is the first study that investigates unlearning in Federated Online Learning to Rank. For this, we adapt the FedEraser method~\cite{liu2021federaser}, developed for general federated learning problems, to the unique context of federated online learning to rank, where rankers are learned in an online manner on implicit interactions (e.g. clicks). 
We further modify the method to leverage stored historic local updates to guide and accelerate the process of unlearning. To evaluate the effectiveness of the unlearning method, we adapt the idea of poisoning attacks to the context of determining whether the contributions of a client to be unlearned made to the ranker are effectively erased from the ranker itself. Experimental results on four popular LTR datasets show both the effectiveness and the efficiency of the unlearning method. In particular: (1) the search effectiveness of the global model once unlearning takes place converges to the effectiveness of the ranker if the removed client did not take part in the FOLTR system from the start, (2) the contributions made to the ranker by the unlearned client are effectively removed, and (3) less local and global steps are required by unlearning compared to retraining the model from scratch.

Code, experiment scripts and experimental results are provided at \url{https://github.com/ielab/2024-ECIR-foltr-unlearning}.
	
	%
	%\subsubsection{Acknowledgements} Please place your acknowledgments at
	%the end of the paper, preceded by an unnumbered run-in heading (i.e.
	%3rd-level heading).
	
%	\bibliographystyle{splncs04}
%	\bibliography{ecir-foltr-unlearning}
%	
%\end{document}

%CONCLUSION CHAPTER
% ***************************************************
% Conclusion
% ***************************************************
\chapter[Conclusion]{Conclusion}
\label{Chap:Conclusion}

% ********* Enter your text below this line: ********
%Conclude your thesis.

% ***************************************************

In this thesis, we explored a new training paradigm for Online Learning to Rank that addresses data privacy: Federated Online Learning to Rank (FOLTR). Our study focused on four critical aspects of FOLTR: effectiveness, robustness, security and unlearning. To guide our investigation, four research questions were proposed:

\begin{enumerate}[label=\textbf{RQ\arabic*:}, leftmargin=*]
	\item How to design an effective and generalisable FOLTR method that can achieve competitive performance compared to state-of-the-art OLTR methods? (\textit{effectiveness}, Chapter~\ref{Chap:fpdgd})
	\item Are FOLTR methods robust to non-IID data among clients? (\textit{robustness}, Chapter~\ref{Chap:foltr-noniid})
	\item Are FOLTR methods secure in the face of potential risks brought by malicious participants? (\textit{security}, Chapter~\ref{Chap:foltr-poisoning})
	\item How to effectively and efficiently forget client data from the trained ranker when a client requests to exit the FOLTR system? (\textit{unlearning}, Chapter~\ref{Chap:foltr-unlearning})
\end{enumerate}

In this final chapter, we will summarise our findings and contributions for each RQ, along with the limitations and future work. Finally, we outline the broader future directions set by this thesis.

\section{RQ1: Effective and Privacy-Preserving FOLTR}

\subsection{Overview}

Federated learning approaches to search have only recently emerged. In the area of OLTR,  FOLtR-ES~\cite{kharitonov2019federated} was the first federated method. FOLtR-ES uses evolutionary strategies similar to those in genetic algorithms to make client rankers explore the feature space, and a parametric privacy preserving mechanism to further anonymise the feedback signal that is shared with the central server by the clients. 
Although showing promising results, the empirical study of FOLtR-ES did not consider many of the datasets and evaluation metrics commonly used in OLTR studies. To further understand this method, we replicated and extended its study to encompass common OLTR evaluation practices. We also compared its ranking effectiveness with the state-of-the-art OLTR method in the centralised setting, with the aim of investigating the limitations of this FOLTR method.

We found the performance of FOLtR-ES vary across datasets, lagging far behind the centralised OLTR method. To address this performance gap, we proposed FPDGD, which leverages the state-of-the-art Pairwise Differentiable Gradient Descent (PDGD) and adapts it to the Federated Averaging framework. For a strong privacy guarantee, we further introduced a noise-adding clipping technique based on the theory of differential privacy to be used in combination with FPDGD. To summarise, the main features of our method are: (1) we use PDGD as the core OLTR optimisation algorithm -- this method is shown to perform well and converge faster than previous OLTR methods and is unbiased with respect to document pair preferences; (2) the FedAvg framework is easy to implement and it has been shown to work efficiently with a large number of clients~\cite{mcmahan2017communication}; (3) we use differential privacy to protect the communication of the local gradient updates between the clients and the server.

\subsection{Findings}

%\subsubsection{Performance of FOLtR-ES}

%\textbf{results on reproducing FOLtR-ES}
\textbf{Our findings question whether FOLtR-ES is a mature method that can be considered in practice:} its effectiveness largely varies across datasets, click types, ranker types and settings. Its performance is also far from that of current state-of-the-art OLTR, questioning whether the privacy guarantees offered by FOLtR-ES are not achieved by seriously undermining search effectiveness and user experience. 
This highlights that more work is needed to improve the performance of FOLTR methods so as to close the gap between privacy-oriented approaches and centralised approaches that do not consider user privacy. Overall, although FOLtR-ES has the merit of being the first FOLTR approach available, our findings suggest that it is far from being a solution that can be considered for use in practice, and more research is required for devising effective, privacy-aware, federated OLTR techniques.

%\subsubsection{Performance of FPDGD}

%\textbf{results on FPDGD}
\noindent
\textbf{Higher performance is observed in FPDGD.} We demonstrate the effectiveness of FPDGD through empirical experiments comparing the method to its non-federated counterpart (PDGD) and the other only federated OLTR method, FOLtR-ES. Compared to the non-federated PDGD, our method converges slower when the same total interaction budget (i.e. number of queries) is considered. In particular, the difference is significant when differential privacy is considered -- this is the price to pay for not sharing the clients' data with the central sever (federated setup) and for protecting the local gradient updates (differential privacy). When compared to FOLtR-ES, the proposed method showcases higher performance. In addition, FPDGD is more robust across different privacy guarantee requirements than FOLtR-ES: FPDGD is therefore more reliable for real-life applications.

\subsection{Contributions}

The results of our empirical investigation of FOLtR-ES help understanding the specific settings in which this technique works, and the trade-offs between user privacy and search performance (in terms of effectiveness and user experience). They also unveil that more work is required to devise effective federated methods for OLTR that can guarantee some degree of user privacy without sensibly compromising search performance. To this aim, we propose FPDGD which leverages a pairwise loss to address the bias from local implicit feedback thus boosts the effectiveness compared to the baseline. We further contribute an understanding on the performance of FPDGD compared to its centralised counterpart, where we compare them under same interactions budgets and same update budgets. Finally, we investigate the impact of differential privacy noise added to the updates sent to the central server.
This privacy preserving mechanism is effective in protecting the gradient updates from privacy attacks~\cite{bhowmick2018protection}. That is, a malicious attacker that is able to observe the ranker model and gradient updates can potentially reconstruct the data that produced such an update~\cite{DBLP:conf/nips/GeipingBD020}, thus finally exposing user data.

\subsection{Limitations and Future Work}

\begin{figure*}[t]
	\includegraphics[width=0.7\linewidth]{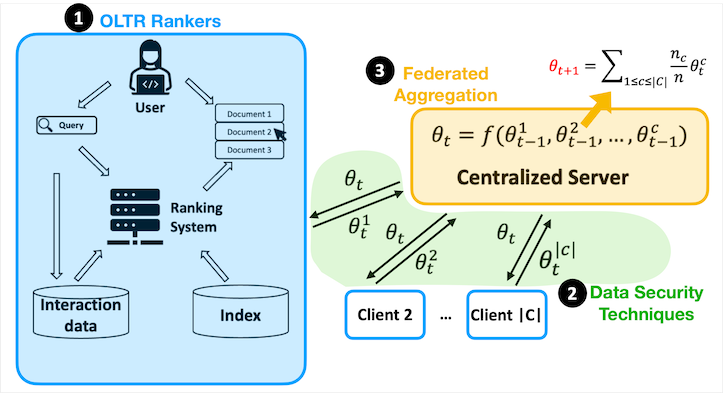} 
	\centering
	\caption{A high-level architecture and key components in FOLTR system.}
	\label{fig:benchmark}
\end{figure*}

%\textcolor{red}{In this study, we only investigate the performance of FPDGD on a linear ranker, though in the following study domenstrated in Chapter~\ref{Chap:foltr-noniid}, a neural ranker is also leveraged and confirmed its utility. [Not a limitation. Think another limitation maybe?]} 
To make FOLTR methods more generalisable, future work can look into creating a FOLTR benchmark where different combination of OLTR methods \circled{1}, data security techniques  \circled{2} (such as differential privacy~\cite{dwork2014algorithmic}, secure multi-party computation~\cite{yao1982protocols} and encryption techniques~\cite{gentry2009fully}), and federated aggregation algorithms \circled{3} (federated averaging~\cite{mcmahan2017communication} and its variants) are studied as shown in a high level architecture in Figure~\ref{fig:benchmark}. In future study, methods for reducing computation and communication load can also be investigated through leveraging the decentralised attributes of federated learning while addressing the bias of implicit online feedback. In addition, there is a growing trend of unifying Online LTR and Counterfactual LTR. For example, Zhuang and Zuccon~\cite{zhuang2020counterfactual} proposed Counterfactual Online Learning to Rank which applies a form of counterfactual evaluation for updating OLTR rankers. Oosterhuis and de Rijke~\cite{oosterhuis2021unifying} linked OLTR to counterfactual LTR by proposing an intervention-aware estimator that has been shown effective for both counterfactual and online LTR. Thus, future work can investigate novel methods for the unification of Online LTR and Counterfactual LTR under the federated setting.

\section{RQ2: FOLTR with non-IID Data}

\subsection{Overview}

A factor influencing the performance of federated learning systems is the presence of bias in the data distribution across clients. In other words, the fact that clients may hold non-independent and identically distributed (non-IID) data. This non-IID issue poses significant challenges, as it can result in suboptimal model performance and reduced generalisability of the trained models~\cite{zhao2018federated}. Addressing this problem is essential for enhancing the robustness and effectiveness of federated learning systems. While FOLTR systems are on their own right a type of federated learning system, the presence and effect of non-IID data in FOLTR has not been studied, so as methods for addressing them. Due to the inherent complexity of users' searching behaviours, the formulation of non-IID data types for FOLTR is not clear. In particular, users may differ in preferences on documents, click behaviours, or frequency on issuing queries, skewing the local learning process towards their local distribution thus hinder the convergence of the global ranker. To this aim, we investigate the effect of non-IID data on FOLTR and explore if the existing methods in handling non-IID can also be adapted to FOLTR, thus chart directions for future work in this new and largely unexplored research area. Specifically, we put forward four different ways data specific to FOLTR could be distributed in a non-IID manner because of biases across clients on: document preferences (Type 1), document label distribution (Type 2), click preferences (Type 3), and data quantity (Type 4). Then, we study the impact of each setting on the performance of current state-of-the-art FOLTR approach.

\subsection{Findings}

\textbf{Impact of non-IID data.} We found that the presence of non-IID characteristics in the distribution of document preferences (Type 1) and specific cases of document labels (Type 2) have severe effects on the effectiveness of FPDGD. Conversely, if data is distributed across clients in a non-IID manner with regard to click preferences (Type 3) or data quantity (Type 4), no significant effects on the quality of FPDGD are observed. These findings contribute an understanding of under which data distributions it is feasible to use FOLTR and when it is not. 
%We believe this paper will encourage researchers to include non-IID data settings when evaluating new FOLTR methods.

\noindent
\textbf{Calling for FOLTR methods to address non-IID issues.} Our findings chart directions to future work on non-IID data in FOLTR concerning the creation of techniques that provide remedies to Type 1 and 2, while deeming solutions for Type 3 and 4 data less critical. Importantly, we show that existing solutions employed in general federated learning to mitigate the non-IID data problem do not apply to the FOLTR setting, despite some of these non-IID cases (and especially Type 1) being likely to occur across many FOLTR systems.

\subsection{Contributions}

This research sheds light on the factors that need to be considered when devising and deploying FOLTR methods. We detail the experimental conditions for simulating non-IID data in FOLTR, paving the way for the development and adaptation to OLTR of existing and new methods for dealing with non-IID data. With this regard, we also show how some of the methods proposed in the federated learning literature to deal with non-IID data could be cast in the FOLTR framework and the gaps that still exist in effectively addressing non-IID data in FOLTR. This allows us to unveil new research gaps that, we argue, future research in FOLTR should consider. This is an important contribution to the current state of FOLTR because, for FOLTR systems to be deployed, the factors affecting their performance, including the impact of non-IID data, need to be thoroughly understood and considered. Our work has also inspired a recent study on investigating non-IID data for federated learning to rank with tree-based rankers~\cite{cecchetti2024learning}.

\subsection{Limitations and Future Work}

\textbf{Privacy should be a high priority when dealing with non-IID data.} Our analysis found that only the data-sharing technique~\cite{zhao2018federated} could address to significant extents Type 2 non-IID data. However, this and similar methods, although performing well, require prior knowledge about the users' local data distributions -- and thus require users to share private data, largely defeating the purpose of federated learning. 
We note that recent work has considered the sharing of synthetic, rather than real, data~\cite{wang2021non}. In such a setting, real data would be used by each client to generate synthetic data, and the synthetic data only would be shared in the federation. However, we could not find evidence of the loss in effectiveness associated with the use of synthetic rather than real data in the data-sharing scheme. Furthermore, it is unclear what the privacy guarantees are in such a synthetic data sharing scheme. Specifically, we wonder whether the use of synthetic data could jeopardise privacy as this synthetic data is generated from the real data: thus analysis of the synthetic data may reveal key aspects of and information contained in the real data.
Thus, how to guarantee user’s privacy needs when designing effective FOLTR algorithms on non-IID data is still an open question.

\noindent
\textbf{Real-world datasets and benchmarks for FOLTR with non-IID data are needed.} The experiments put forward in this study to substantiate our views on the non-IID data problem in FOLTR are based on simulations. While simulations are prevalent in information retrieval and especially in its evaluation~\cite{cooper1973simulation,azzopardi2016simulation,maxwell2016simulating,zhang2017information,balog2021report}, a key aspect we had to simulate was the nature of the non-IID data, including their distributions. On one hand, this allows us to carefully control the experiments; on the other it limits the generalisability of the findings to real non-IID data that may occur in FOLTR settings. Thus, another future direction is building FOLTR benchmark datasets that provide standard simulations on real-world non-IID scenarios as well as standard hyper-parameter settings so that future FOLTR algorithms can be fairly studied.

\noindent
\textbf{Further investigation of online learning to rank algorithm is needed.} The local ranking model of this work is updated based on Pairwise Differentiable Gradient Descent (PDGD) algorithm. A further discussion about PDGD by Oosterhuis and de Rijke~\cite{oosterhuis2021unifying}, which links OLTR to counterfactual LTR, found that PDGD is ineffective in some situation where it is not run in a fully online mode. In other words, PDGD can have detrimental performance if the model updates are based on interactions gathered by a different model than to which they are applied. In this thesis, while the local ranker is updated based on online interactions, the global ranker is updated through federated aggregation and replaces the local ranker. This process dose not follow exactly the same online learning setup as PDGD in a centralised setting~\cite{oosterhuis2018differentiable, oosterhuis2021unifying}. Thus, to fully understand if the cause of this non-IID issue is not running PDGD in an online and centralised setting or from OLTR in general, future work can investigate FOLTR under non-IID setting through considering other OLTR algorithms to enhance the completeness of this study and further explore PDGD running in the federated mode.

\section{RQ3: FOLTR under Poisoning Attacks}

\subsection{Overview}

In this study, we investigated potential risk posed by malicious clients and solutions for defending against them with focus on untargeted poisoning attack. For the untargeted attack, we implement a \textit{data poisoning} method that compromises the local data to impact the training of the model, and two \textit{model poisoning} methods that directly corrupt the local model updates. As for the defense method, we implement four Byzantine-robust aggregation rules (Krum, Multi-Krum, Trimmed Mean and Median) to safeguard against such attacks. These defense mechanisms rely on statistical techniques to identify outliers among the received weights and subsequently exclude them during the aggregation process.

\subsection{Findings}

%From the empirical results presented above, we raise the following key findings:
Based on the empirical results, we identify the following key findings:

\textbf{It is harder to attack FOLTR systems with clients that exhibit low variance between local updates.} Under the \textit{perfect} click type simulation, the users provide the most accurate and lowest noisy feedback. In general, this click type is more difficult to attack compared to the other two click models under both data and model poisoning methods, except in specific instances when employing Fang's attack~\cite{fang2020local} under the full knowledge assumption. To successfully attack a FOLTR system when \textit{perfect} click feedback is present, a larger number of attackers is required due to the relatively low variance between local updates. As a result, more clients must be compromised by the attacker, otherwise the attack is more likely to be mitigated by robust aggregation rules.

\textbf{The attack is more successful when more information about the benign users' local updates is known.} Among all attacking strategies studied, Fang's attack~\cite{fang2020local} with full knowledge emerged as the most successful in diminishing the performance of the global model, though some exceptions were observed in the noisy \textit{informational} click scenario. When there were more malicious clients (i.e. 30\% or 40\% of the total number of clients), Fang's attack with partial knowledge is just as effective as with full knowledge. This indicates that model poisoning is more effective than data poisoning. Furthermore, when defense measures were implemented, Fang's attack demonstrated greater success against Krum and Multi-Krum aggregation rules in comparison to Trimmed Mean and Median.

%	\item Among all attacking strategies studied in this paper, Fang's attack under full knowledge performs consistently better in compromising the effectiveness of global model except for some scenarios under the noisiest \textit{informational} click. With more attackers involving (i.e. 30\%, 40\%), Fang's attack with partial knowledge can achieve similar attacking utility as under full knowledge scenario. This demonstrates the superiority of model poisoning than data poisoning. Under defensive situation, Fang's attack works well on Krum and Multi-Krum aggregation rules than Trimmed Mean and Median.

%	\item Although Krum have shown its effectiveness in defending against data poisoning while Timmed Mean on detecting Fang's attack, more attention should be paid when deploying Krum and Trimmed Mean as the utility drops under no-attacker system brought by those two aggregation rules.
	
\textbf{Defense strategies negatively impact ranking if no attack is in place.} Although Krum has proven effective in countering data poisoning and Trimmed Mean in defending against Fang's attack, deploying these aggregation rules should be exercised with caution as they result in an overall decrease in search performance if the system is not under attack. The selection of these defense mechanisms should thus be carefully considered, taking into account the specific context and risk of potential attacks to strike the right balance between security and search effectiveness.

%It is essential to highlight that while Krum has proven effective in countering data poisoning and Trimmed Mean in defending against Fang's attack, deploying these two aggregation rules should be exercised with caution. It has been observed that they lead to an overall decrease in search performance when the system is not subjected to attacks. Thus, the selection of these defense mechanisms should be carefully considered, taking into account the specific context and risk of potential attacks to strike the right balance between security and search effectiveness.

\subsection{Contributions}

This is the first study that systematically analysed the threats brought by untargeted poisoning attacks to FOLTR systems. It demonstrates the effectiveness and associated drawbacks of existing defense methods on mitigating the impact of malicious adversaries under FOLTR system. Through extensive empirical experiments, we (1) investigate the vulnerability of FOLTR systems to untargeted poisoning attacks, and show under which conditions poisoning attacks can represent a real threat to FOLTR systems; and (2) demonstrate the effectiveness of defense strategies, importantly revealing the presence of issues with defense strategies if applied to FOLTR systems for which an attack is not in place.

\subsection{Limitations and Future Work}

We did not consider the setup of a cross-device FOLTR system, where many clients are involved in the federation: this is representative of a web-scale federation. For example, an email service provider aims to improve its email search functionality to deliver more relevant search results. Due to the private attribute of user's emails and search interactions, an on-device FOLTR system can be leveraged, where the retrieval and ranking is done locally and a ranking model is collaboratively learned with the help of local model updates. As more clients are involved, the portion of users that can be controlled by the malicious attacker is relatively low, making the attack more challenging. Under this circumstance, more sophisticated attacking strategies are required to achieve prominent poisoning impact. Future works need to address the unique challenges under the context of FOLTR: (1) as the federated environment limits the adversary's prior knowledge to its local information (i.e., malicious user's data and local models) and the global model, the randomness of user's queries and clicks might obstruct the success of attacking; and (2) the deployed defense methods should ensure the global ranker maintains high ranking performance, avoiding negative impact on user experience. In addition, to make the study more comprehensive thus strengthening the overall depth and impact of the security evaluation, future works can investigate more critical attacks, such as the DYN-OPT~\cite{shejwalkar2021manipulating}, and key defense mechanisms, including Bulyan~\cite{guerraoui2018hidden} and FLDetector~\cite{zhang2022fldetector}, among others.

\section{RQ4: Unlearning for FOLTR}

\subsection{Overview}

Data protection legislation like the European Union's General Data Protection Regulation (GDPR) establishes the \textit{right to be forgotten}: a user (client) can request contributions made using their data to be removed from learned models. We study how to remove the contributions made by a client participating into a FOLTR system, in other words, how to perform unlearning for FOLTR. We conduct our study along two main directions: (1) how to efficiently unlearn with the guidance of historical local updates; and (2) how to evaluate the effectiveness of unlearning.

\subsection{Findings}

\textbf{Trivial impact on model performance from a single client.}
In previous studies on machine unlearning and federated unlearning, the effectiveness of unlearning strategies has primarily been verified by comparing model performance against baseline models~\cite{bourtoule2021machine,wu2022federated}. Inspired by this, in our empirical simulation, we compare the ranking performance of two models: one trained before the removal of the client wanting to leave the federated system and one trained without this client. Our results indicate that excluding this single client does not significantly affect ranking performance. Although this phenomenon is accentuated due to the synthetic simulation with same configuration of the click model for all participants, it is still reflective of a real-world scenario, particularly in cross-device federated settings. Given the minimal impact on ranking performance from the removal of a single client, a more effective evaluation method for unlearning is required.

\noindent
\textbf{A solution for evaluating unlearning inspired from poisoning attacks.}
For the purpose of evaluating if unlearning takes place, we need to magnify the effect of the client that is leaving the FOLTR system and make this effect relatively controllable. To this aim, we adopt an alternative evaluation which leverages poisoning attacks only on the client requesting to be unlearned but before the unlearning takes place. Thus, mitigating the impact of the poisoned client during unlearning process suggests an effective unlearning strategy.

\noindent
\textbf{The effectiveness of unlearning with the help of historical local updates.}
Under the aforementioned evaluation, we find out that effective and efficient unlearning can be implemented with the help of historical local updates where in particular, we let the historical updates guide the direction for the unlearning process, and set the model update step size based on new interactions gathered during unlearning.

\subsection{Contributions}

This is the first study that investigates unlearning in FOLTR systems, where it is not clear that advances in general federated learning translate to similar improvements. We introduced an unlearning method to effectively and efficiently remove a client's contribution from a global ranking model without the need for complete retraining. The primary challenge tackled is ensuring that the model has genuinely unlearned the specified client's data. To verify this, we employed a novel approach where the client deliberately introduces noise into its updates (a poisoning attack), and the reduction in the impact of this noise post-unlearning is used as a measure of success in unlearning.

%This is because of the significant difference between FOLTR from general classification tasks, e.g. ranking vs. classification, online learning vs. offline learning\footnote{In federated learning, the local model can be trained on the local data repeatedly across several epochs, while in FOLTR, training data is acquired in real time as user interactions occur and it cannot be repeated and reused (e.g., a user cannot be asked to submit the same query they did in the past, and perform the same interactions). }, implicit user feedback vs. ground-truth labels, etc.
%In addition to the impact on FOLTR, our benchmark can enrich the task-level diversity for the evaluation of general federated unlearning methods.

\subsection{Limitations and Future Work}

In this study, we only consider the scenario where only one client requests unlearning to be initiated although we believe the findings might also generalise to the unlearning of multi-clients. However, it is still unclear how to perform multi-unlearning simultaneously and what trade-off between efficiency and effectiveness will be. Additionally, this study utilizes poisoning attacks to evaluate unlearning methods. However, this approach cannot be directly applied in real-world applications due to its disruptive nature. Future research should explore alternative evaluation methods that consider the search experience of remaining clients, ensuring that the unlearning process is both effective and minimally intrusive.

\section{Future Directions: Federated Learning for Rankers based on Pretrained Language Models}

In this thesis, we focused on the learning of feature-based ranking models. Feature-based ranking models are still prevalent in industry~\cite{zou2022large,dato2022istella22}. However, in recent years, there is a growing trend in leveraging Pretrained Language Models (PLMs) based on the Transformer architecture~\cite{vaswani2017attention} for ranking models~\cite{lin2022pretrained}. One of the most common approaches follow a bi-encoder architecture: the query and documents are separately encoded into dense or sparse vectors by PLMs, and relevance scores are estimated based on the similarity between query and document vectors. By doing so, more semantic meanings for queries and documents can be exploited and vocabulary mismatch problem happened in traditional bag-of-words retrieval methods such as BM25 and TF-IDF~\cite{robertson1994some,crestani1998document,robertson2009probabilistic} can be alleviated, demonstrated by the significant effectiveness improvement of the PLMs based rankers~\cite{nogueira2019passage}. 

To our best knowledge, leveraging FL for PLMs-based ranker is still unexplored. Due to the need of large amount of labelled data for fine-tuning PLMs towards ranking tasks, privacy concerns on sharing sensitive training data~\footnote{words in textual documents and queries under this context} and high cost for editorial labelling remain two critical challenges, especially for domains like health, personal and enterprise search. To this aim, this thesis can inspire the study of a federated online PLMs-based ranker across several directions. First of all, future research can look into the design of a framework where PLM rankers are fine-tuned in a federated and online manner using implicit user interactions, such as clicks. Instead of sharing data, clients fine-tune the PLM ranker on their local data and then communicate to a central server the ranker updates they obtained. As the architecture of rankers based on PLMs is largely different from the feature-based rankers, it is unclear how advances in federated learning and our findings on FOLTR apply to PLM rankers. Importantly, as PLM-based rankers are large in size, fine-tuning the PLM ranker is an expensive operation on user devices with low computational power. How to efficiently fine tune them with limited local interactions and limited computation resource remains a challenge. One solution is to use knowledge distillation (KD) rankers~\cite{hofstatter2021efficiently}, where a smaller student ranker is trained by distilling knowledge, such as score distribution and ranking loss, from a larger, more expensive teacher ranker. By reducing the size of the ranking model, both the fine-tuning and the inference costs at the the client side can be reduced. Additionally, PLM rankers typically require a large amount of fine-tuning data to be effective. However, in this FL framework, effective PLM rankers can be trained by using only a small portion of data from each user in the federation. Recent work in information retrieval has explored the promising direction of prompt learning for fine-tuning PLM rankers in the presence of limited labelled data~\cite{hu2022p3} and prompting Large Language Models (LLMs) for zero-shot document ranking~\cite{zhuang2023open}. Future work can investigate how to integrate prompt learning into the FL framework to further improve effectiveness of PLM rankers. Learning from the noisy and biased implicit feedback has been largely investigated in the context of counterfactual learning to rank and online learning to rank. However, aspects still unexplored are how this noisy and biased signal affects the learning of PLM rankers and how effective learning can take place in this context. Despite federated learning not involving the sharing of user data, a malicious agent that is able to observe the gradient updates can potentially reconstruct the data that produced such an update~\cite{DBLP:conf/nips/GeipingBD020}, leaving the potential of exposing user data. In addition, previous work has found that the PLMs underlying the rankers considered in this context can leak private information~\cite{pan2020privacy,morris-etal-2023-text}, and thus this issue plausibly applies to PLM rankers too. Therefore, further studies on privacy issues related to PLM rankers can be conducted, which include (1) the risk caused by the exchange of local ranker updates from the clients, and (2) the inspection of a global or local ranker formed through the aggregation of client models updates; along with guidelines and new methods for controlling such risks.

\section{Summary}

Federated Learning provides a solution for building up effective rankers while respecting user privacy for OLTR. In this thesis, we conducted the first comprehensive study on Federated Online Learning to Rank with focus on the effectiveness, robustness, security and unlearning, tackling a range of unexplored questions for this area. This work will help researchers and practitioners to gain in-depth insights into FOLTR in terms of debiased and effective optimisation, robustness to non-IID data, risks posed by poisoning attacks, and efficient unlearning. As demonstrated in this thesis, there is still much to be done in refining the existing research on FOLTR. We point out the limitations of our work and chart directions for future work.

Our investigation and findings may be valid also in other contexts that consider to create a ranker from feature-based data in a federated manner, e.g., in federated learning to rank~\cite{wang2021efficient} and federated counterfactual leaning to rank~\cite{li2021federated}. Our research questions can inspire the study of federatively learning embedding-based rankers~\cite{lin2022pretrained} which are founded by PLMs and LLMs.

% ***************************************************
% Bibliography
%****************************************************
%CHOOSE YOUR BIB STYLE AND FILE.
%We have included the following two referencing styles for you to use in your thesis. You can add an alternate style if you prefer.

%Style: apalike = this is an (Author, Year) referencing style similar to APA
%Style: elsarticle-num = this is a numbered referencing style that will display the bibliography in citation order

%To use one of the styles provided ensure the % is removed from the start of the line, and the other option is commented out with a % at the start of the line. The style elsarticle-num is active by default.

%\bibliographystyle{apalike}
\bibliographystyle{elsarticle-num}

\bibliography{./References/FOLTR-Bibliography}

%When you have finished your thesis we recommend that you manually fix any errors in your bibliography. 
%To do this, compile, copy the .bbl into a new .tex file and include this here after commenting out the other bibliography commands. Make corrections in that .tex file.

% ***************************************************
% Appendices
%**************************************************** 
%UNCOMMENT THIS SECTION IF YOU ARE USING APPENDICES.
%Simply adapt the same formatting used for other chapters.
\appendix
% If you need appendix in your thesis then consider the following appendix file (you can add more if you need more) otherwise you should not consider it in your main thesis.
%\include{./Appendix/Appendix}

% ***************************************************
% Examples
%**************************************************** 
% The following files are only for examples and you should not include them in your final thesis.
%\include{./Examples/ExampleOfCitations}
%\include{./Examples/ExampleOfCode}
%\include{./Examples/ExampleOfEquations}
%\include{./Examples/ExampleOfFigures}
%\include{./Examples/ExampleOfFlowCharts}
%\include{./Examples/ExampleOfTables}

% ***************************************************
% Back Matter
%**************************************************** 
%COMMENT OUT IF YOU DO NOT WISH TO INCLUDE BACK MATTER.
%\input{./PreliminaryAndBackPages/Back.tex}

\end{document}